\documentclass{article}

\usepackage{microtype}
\usepackage{graphicx}
\usepackage{booktabs} 
\usepackage{hyperref}

\usepackage[preprint]{icml2026}

\usepackage{amsmath}
\usepackage{amssymb}
\usepackage{mathtools}
\usepackage{amsthm}

\usepackage{graphicx}
\usepackage[font=normalsize,labelfont=sf,textfont=sf]{subfig}
\usepackage{textcomp}
\usepackage{stfloats}
\usepackage{verbatim}

\usepackage{multirow}
\usepackage{booktabs}
\usepackage{bbm}
\usepackage{soul}
\usepackage{graphicx}
\usepackage{booktabs}
\usepackage{tabularx}
\usepackage{bm}
\usepackage{adjustbox}
\usepackage{lipsum}
\usepackage{pifont}
\usepackage{amsmath,amsfonts}
\usepackage{tikz,xcolor}

\usepackage[capitalize,noabbrev]{cleveref}

\theoremstyle{plain}

\theoremstyle{definition}

\theoremstyle{remark}

\usepackage[textsize=tiny]{todonotes}

\icmltitlerunning{MLOW: Interpretable Low-Rank Frequency Magnitude Decomposition of Multiple Effects for Time Series Forecasting}

\begin{document}

\twocolumn[
  \icmltitle{MLOW: Interpretable Low-Rank Frequency Magnitude Decomposition \\of Multiple Effects for Time Series Forecasting}



  \icmlsetsymbol{equal}{*}

  \begin{icmlauthorlist}
    \icmlauthor{Runze Yang}{yyy,comp}
    \icmlauthor{Longbing Cao}{yyy}
    \icmlauthor{Xiaoming Wu}{sch}
    \icmlauthor{Xin You}{comp}
    \icmlauthor{Kun Fang}{dd}
    \icmlauthor{Jianxun Li }{comp}
    \icmlauthor{Jie Yang}{comp}\\
    \url{https://github.com/runze1223/MLOW}
  \end{icmlauthorlist}


  \icmlaffiliation{yyy}{Macquarie University}
  \icmlaffiliation{comp}{Shanghai Jiao Tong University}
  \icmlaffiliation{sch}{Nanyang Technological University}
  \icmlaffiliation{dd}{Hong Kong Polytechnic University}
            
  \icmlcorrespondingauthor{Runze Yang}{runze.y@sjtu.edu.cn}

  \icmlkeywords{Machine Learning, ICML}

  \vskip 0.3in
]



\printAffiliationsAndNotice{}  

\begin{abstract}
Separating multiple effects in time series is fundamental yet challenging for time-series forecasting (TSF). However, existing TSF models cannot effectively learn interpretable multi-effect decomposition by their smoothing-based temporal techniques. Here, a new interpretable frequency-based decomposition pipeline MLOW captures the insight: a time series can be represented as a magnitude spectrum multiplied by the corresponding phase-aware basis functions, and the magnitude spectrum distribution of a time series always exhibits observable patterns for different effects. MLOW learns a low-rank representation of the magnitude spectrum to capture dominant trending and seasonal effects. We explore low-rank methods, including PCA, NMF, and Semi-NMF, and find that none can simultaneously achieve interpretable, efficient and generalizable decomposition. Thus, we propose hyperplane-nonnegative matrix factorization (Hyperplane-NMF). Further, to address the frequency (spectral) leakage restricting high-quality low-rank decomposition, MLOW enables a flexible selection of input horizons and frequency levels via a mathematical mechanism. Visual analysis demonstrates that MLOW enables interpretable and hierarchical multiple-effect decomposition, robust to noises. It can also enable plug-and-play in existing TSF backbones with remarkable performance improvement but minimal architectural modifications.

\end{abstract}

\begin{figure}[t]
    \centering
    \includegraphics[width=0.9\linewidth]{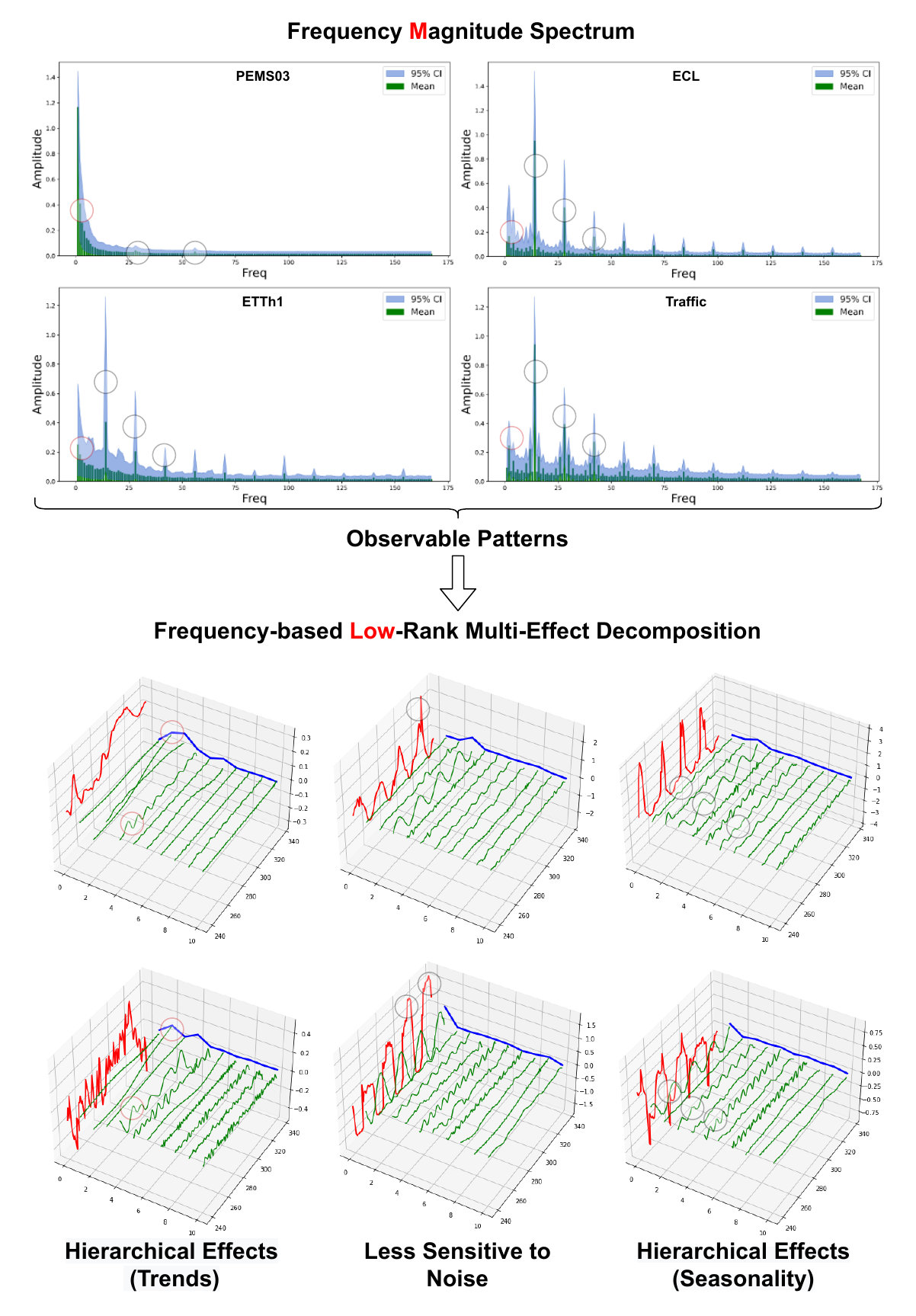}
    \caption{The Motivation and Visualization for MLOW }
    \label{fig:intro}
    \vspace{-10pt}
\end{figure}

\section{Introduction}

Time series forecasting (TSF) is fundamental for a wide range of real-world applications, such as demand prediction, financial risk assessment, climate and environmental monitoring, industrial process control, and healthcare analytics \cite{Cao-AIF20,Cao-AIFt21, hong2016probabilistic, hyndman2018forecasting, huang2025timekan}. 
TSF requires effective time series decomposition of multiple effects in time series, such as seasonality, trends, and residuals, crucial for interpretable TSF and downstream tasks. However, effectively decomposing a time series into meaningful trending, seasonal, and residual effects remains a fundamental challenge in complex time series applications. It requires disentangling intertwined multi-effects, including trend dynamics, multi-scale seasonal patterns, and complex nonstationary fluctuations in real-world time series.

The decomposition module performs initial temporal disentanglement before feeding the input into a TSF backbone. In existing deep TSF backbones, smoothing-based filters such as moving averages serve as the most widely used decomposition techniques, firstly introduced for Autoformer \cite{wu2021autoformer}. Although smoothing methods have been commonly adopted by subsequent methods \cite{zhou2022fedformer, zeng2022transformers, cao2023inparformer, TimeMixer2024, shen2024multi, lin2024sparsetsf, lin2024cyclenet, huang2025timekan}, they are inherently sample-based. As a result, they are sensitive to noises, fail to disentangle different effects, and often produce inaccurate estimates at the boundaries due to padding. Consequently, these methods struggle to consistently enhance deep time series backbones but exhibit poor adaptability when applied to real-world TSF. How to achieve interpretable and effective multi-effect decompositions remains an underexplored problem in TSF.

While frequency-based methods attract increasing recent attention in TSF \cite{zhou2022film, zhou2022fedformer, huang2023crossgnn, yi2023fouriergnn, wu2022timesnet, cao2020spectral, yang2023waveform, wang2024fredf, fei2025amplifier, yi2024filternet, qiu2025duet}, they typically transform signals into frequency complex values and rely on end-to-end supervised hidden state, which lack interpretability. In this work, we pursue interpretable frequency-based multi-effect decomposition beyond transforming with the following insights. We observe that the frequency magnitude spectrum distribution of a time series consistently exhibits observable patterns for different effects within the dataset, as illustrated in Figure \ref{fig:intro}. Trending effects often exhibit an exponentially decreasing pattern from low to high, while distinct seasonal effects typically appear at multiple frequency levels depending on the data granularity. The most prominent patterns usually correspond to the principal effects in a time series. Therefore, relying purely on high-, mid-, and low-frequency levels is insufficient for capturing multi-effects. 

In contrast, we propose MLOW which learns a low-rank representation of the frequency magnitude spectrum to capture underlying patterns of principal trending and seasonal effects in time series. MLOW addresses the above gap and extends interpretability beyond frequency levels to low-rank components. First, a sliding window over the training data obtains the samples. Then, we leverage the Fourier basis expansion \cite{yang2024rethinking} to decompose all sample time series $\mathbf{X}$ into amplitudes $\mathbf{R}$ and phase-aware basis functions $\mathbf{B}$, aiming for learning a low-rank representation on their amplitudes. The amplitude, referring to frequency magnitude spectrum, provides a valuable tool for analyzing time signals \cite{zhang2025frequency, yang2023respiratory}, as it captures the underlying effects in the signal. We observe that all low-rank methods can be expressed as $\mathbf{R} = \mathbf{W}\mathbf{H}$, where we refer $\mathbf{H}$ to the low-rank components and $\mathbf{W}$ as the new coefficients for the low-rank components. However, this triggers two challenges. First, the existing low-rank methods may not be well suited to magnitude spectrum. Second, the limited input horizon restricts the number of frequency levels $\mathbf{R}$ that can be captured in a low-rank representation, leading to frequency (spectrum) leakage issues.



To address the first challenge, we investigate existing decomposition methods for the magnitude spectrum. PCA \cite{pearson1901liii} is one of the most widely used decomposition methods, which suffers from a low interpretability, as it allows both low-rank components $\mathbf{H}$ and the corresponding coefficient $\mathbf{W}$ takes negative values. This results in the decomposed components including negative combinations of phase-aware basis functions, which compromises the original phase effects. On the other hand, the non-negative matrix factorization (NMF) \cite{lee1999learning} addresses the non-negativity issue, but it is computationally inefficient and not well generalizable to unseen data. The interpretability of $W$ is also quite poor as it is based on fitting. The Semi-NMF approach \cite{ding2008convex} faces same issues as PCA and NMF. To overcome these challenges, we propose a new low-rank method, namely Hyperplane-NMF for MLOW, where we force the $\mathbf{W}=\mathbf{R}\mathbf{H}^{T}$ similar as the hyperplane projection by PCA while still following the same principles as NMF with a cosine similarity penalization. Then, Hyperplane-NMF combines the advantages of the aforementioned methods to ensure the interpretability, efficiency and generalization of multi-effect decomposition. To address the second challenge, a common issue in spectral analysis is the frequency (spectral) leakage. Specifically, when the number of available frequency levels is limited, the energy may spread across multiple frequency bins, which increases the difficulty of identifying prominent patterns. This is because the frequency level is constrained to half of the input horizon. To ensure the decomposition independent of the input horizons with flexible selection of frequency levels, we introduce a mathematical mechanism.

As shown in Figure \ref{fig:intro}, we decompose time series into 10 low-rank pieces. The visualization indicates that the MLOW decomposition hierarchically separates distinct effects, with specific pieces more clearly representing trending and seasonal effects while exhibiting greater robustness to noises. The MLOW pipeline also offers a clear interpretability for each piece and can characterize its source of energy. The empirical experiments also demonstrate that MLOW decomposition can remarkably improve existing TSF backbones with minimal changes of their model architectures. To summarize, both empirical results and visual evidence demonstrate the superiority of MLOW decomposition.

The main contributions of this work include:
\begin{itemize}
    \item An interpretable decomposition pipeline to learn low-rank representations of the frequency magnitude distribution capturing multiple effects.
    \item A systematic investigation of strengths and weaknesses of three existing low-rank methods for magnitude spectrum with new Hyperplane-NMF to combine their strengths for an interpretable, efficient, and generalizable decomposition.
    \item A mathematical mechanism that allows flexible selection of frequency levels to compute the low-rank components independently of the input time horizons.
    
\end{itemize}


\section{Related Works}

\subsection{Low-Rank Representation Learning}

PCA \cite{pearson1901liii} can transform high-dimensional data to a smaller set of principal components while retaining most of the variances. It can identify the directions of maximum variances, which can sometimes uncover meaningful patterns. However, it produces negative values in the magnitude spectrum analysis, which are not interpretable as negative energy. In contrast, NMF \cite{lee1999learning} can obtain a low-rank representation by constraining all components and coefficients to be non-negative. However, it is primarily designed for the matrix factorization rather than for efficient and generalizable inference on new data. Semi-NMF \cite{ding2008convex} allows the corresponding coefficients to take negative values while keeping the components positive but faces the same weakness as NMF. Nevertheless, none of these methods can simultaneously satisfy the following three conditions: (i) both the components and corresponding coefficients are non-negative for interpretability; (ii) no optimization is required when new data arrive, ensuring computational efficiency; and (iii) good generalizability and interpretability for $W$ between the training and unseen data.  Therefore, we develop a new low-rank method tailored for magnitude-spectrum study. 

\subsection{TSF Model Backbone}
TSF backbones comprise popular DNN architectures including recurrent neural networks (RNNs) \cite{hochreiter1997long,rangapuram2018deep,jia2023witran,salinas2020deepar, kraus2024xlstm},  convolution neural networks (CNNs) \cite{luo2024moderntcn,liu2022scinet,wang2022micn,franceschi2019unsupervised,sen2019think}, multi-layer perceptron (MLP) based networks \cite{TimeMixer2024,oreshkin2019n, challu2023nhits,yi2023frequency, shi2024time, xia2025timeemb}, and Transformer-based networks \cite{zhou2022fedformer,zhou2021informer,wu2021autoformer,liu2021pyraformer,zhang2022crossformer,cao2023inparformer,fu2024encoder,zhang2023temporal,chen2024pathformer,LiuHZWWML24, ma2025mofo, qiu2025duet}. To validate our decomposition method, we aim to select sophisticated models that require minimal architectural intervention, limited to their initial projection layer. Accordingly, we select two most famous and popular models, iTransformer \cite{LiuHZWWML24} and PatchTST \cite{nie2023time}, which satisfy this requirement. iTransformer introduces an initial projection from the time domain to the hidden state and uses the attention mechanism to capture channel interaction effects. PatchTST introduces an initial projection from the patched time series to the hidden state and uses the attention mechanism to model interactions between patches. Another reason we choose these two models is that they represent the most sophisticated independent and interaction-based architectures.

\section{Methodology}

\subsection{Preliminaries}

MLOW aims for an interpretable multi-effect low-rank frequency magnitude decomposition to improve TSF. All decompositions are initially computed inside the training, validation, and test data loaders, while the low-rank components $\mathbf{H}$ are obtained from the training data loader only. Since the initial projection in the TSF backbone is channel-independent, MLOW is also channel-independent.  Thus, we discuss the decomposition in the univariate case for clarity. We define an input time series $\mathbf{X}^{i}\in \mathbb{R}^T$ paired with target $\mathbf{Y}^{i}\in \mathbb{R}^L$, where $T$ as input horizon and $L$ as forecast horizon.  After applying MLOW on $\mathbf{X}^{i}$ , we obtain the decomposed input $\mathbf{\dot{X}}^{i} =[\mathbf{X}_m^{i}, \mathbf{X}_r^{i}, \mathbf{Z}^{i}]$  $ \mid \mathbf{\dot{X}}^{i} \in \mathbb{R}^{(T \times (V+2) )}$, where $\mathbf{Z}^{i}$ is our $V$ low-rank pieces (MLOW output), $\mathbf{X}_m^{i}$ and $\mathbf{X}_r^{i}$ refer to the mean intercept and the residual, respectively, both having the same shape as $\mathbf{X}^i$.  $\mathbf{X}_m^{i}$ is used for the normalization. Then, the decomposed $V$ pieces and residuals are fed into the downstream mapping network. To minimize intervention in the downstream mapping and demonstrate its effectiveness, we introduce only minimal modifications to the network architecture. Specifically, only the initial projection layer is modified for iTransformer and PatchTST. For iTransformer, its initial projection layer $\texttt{nn.Linear}(T, g)$ is replaced by 
$\texttt{nn.Linear}((V+1)\times T, g)$. For PatchTST, its initial projection layer $\texttt{nn.Linear}(f, g)$ is replaced by $\texttt{nn.Linear}((V+1)\times f, g)$. Here, $T$ denotes the input length, $f$ denotes the patch length, and $g$ denotes the hidden dimension. Thus, the decompose time series is flattened or flattened after patching then fed into the initial projection layer for iTransformer and PatchTST, respectively. We do the same inference to timestamps with same learned $\mathbf{H}$ for iTransformer as they share the same initial projection layer.  



\subsection{MLOW's Decomposition Pipeline and Challenges}

To compute meaningful decompositions, We distinguish different effects (seasonal, and trending, etc.) by rearranging the same effects into a distinct low-rank component. In the frequency domain, the magnitude spectrum provides a global overview of different effects, with distinct effects typically concentrated at separate frequency levels. Some energy at certain frequency levels always tends to co-occur. Thus, we aim to learn a low-rank representation of the magnitude spectrum to capture most dominant effects by identifying the underlying patterns of magnitude spectrum for each dataset. Thus, the decomposition pipeline is as follows:
\begin{align}
\mathbf{X}^{i} &= \mathbf{R}^{i}  \mathbf{B}^{i}+ \mathbf{X}_{m}^{i}, \mathbf{R}^{i} \in \mathbb{R}^{K},  \mathbf{B}^{i} \in \mathbb{R}^{K \times T}  \\  \label{eq:2} 
\mathbf{R}^{i} &\approx  \mathbf{W}^{i} \mathbf{H},  \mathbf{W} \in \mathbb{R}^{V},  \mathbf{H} \in \mathbb{R}^{V \times K},\\ 
\mathbf{P}^{i} &= \mathbf{H} \mathbf{B}^{i},  \mathbf{P} \in \mathbb{R}^{V \times T},\\ 
\mathbf{Z}^{i} &=  \mathbf{W}^{i} \odot \mathbf{P}^{i}, \mathbf{Z}^{i} \in \mathbb{R}^{V \times T},\\ \mathbf{X}^{i}  & \approx  \mathbf{W}^{i} \mathbf{P}^{i}+\mathbf{X}_{m}^{i},  \mathbf{X}_r^{i}=\mathbf{X}^{i}-\mathbf{W}^{i} \mathbf{P}^{i}-\mathbf{X}_{m}^{i}.
\label{eq:4}
\end{align}

Let $\mathbf{X}^{i}$ denote an input time series, with $\mathbf{R}^{i}$ representing its amplitudes and $\mathbf{B}^{i}$ the phase-aware bases from frequency level $1$ to $K$, $\mathbf{X}_{m}^{i}$ is the mean intercept at frequency level $0$. $\mathbf{H}$ denotes low-rank components, with corresponding new coefficients $\mathbf{W}^{i}$, and the reconstructed bases are given by $\mathbf{P}^{i}$. First, we decompose the original time series into amplitude- and phase-aware basis functions. The amplitude,  corresponding to the magnitude spectrum, serves as the coefficient for the phase-aware basis functions. If a low-rank representation can accurately reconstruct the magnitude spectrum, it can also reconstruct the original time series well. Thus, the key is how to find an interpretable, efficient and generalizable decomposition in Eq.~\eqref{eq:2} workable for any seen and unseen samples in a dataset. On the other, if the level of $\mathbf{R}$ is bounded for input horizon for $\mathbf{X}$, but initially too low, it  leads to frequency (spectral) leakage and makes finding quality low-rank components more difficult.

\subsection{Existing Low-Rank Methods vs Our Hyperplane-NMF}

Here, we aim to identify a low-rank method that is suitable for the magnitude spectrum. Thus, we need to use the distribution of $\mathbf{R}^{i}$ in the training set, as $\mathcal{R}\in \mathbb{R}^{N \times K}$, to obtain low rank components $\mathbf{H
}$. $N$ is equal to the total number of sets $i$ multiplied by the multivariate dimension $D$ in the training set, since we mention that we use channel independence for MLOW. We first analyze the strengths and weaknesses of existing low-rank methods. Then, we introduce our Hyperplane-NMF method, following the logic established in the investigation for existing methods. Finally, we explain why  Hyperplane-NMF achieves an interpretable, efficient, and generalizable decomposition, and how it benefits the downstream mapping network.

First, one of the most common dimensionality reduction methods is PCA. PCA usually requires subtracting the mean and dividing by the standard deviation before performing the transformation. However, neither operation is appropriate here. The reason is intuitive: dividing by the standard deviation forces all frequency components to have the same scale, which is clearly not appropriate for making them equally important. Subtracting the mean removes the most important main features and leads to negative energy values that are not interpretable. Thus, the PCA without the standardization is as follows: 
\begin{equation}
\begin{aligned}
\mathcal{R} &= \mathbf{U} \Sigma \mathbf{H}, \\
\mathcal{R} &\approx \mathbf{U}_V \Sigma_V \mathbf{H}_V, \quad \mathbf{H}_V \in \mathcal{R}^{V \times K},\\
\mathbf{W} &= \mathbf{U}_V \Sigma_V= \mathcal{R} \mathbf{H}_{V}^\top, \\
\mathcal{R} &\approx  \mathbf{W} \mathbf{H}_{V}.
\end{aligned}
\end{equation}

The optimization of PCA is based on the SVD decomposition. The strength of PCA is that it generalizes well to new data, as computing the coefficient $ \mathbf{W}$ for new samples is straightforward since it is simply a hyperplane projection $\mathcal{R} \mathbf{H}_V^\top$. However, its weaknesses are also apparent. The decomposition typically introduces negative values in both $ \mathbf{W}$ and $ \mathbf{H}$. Negative values in $ \mathbf{H}$ imply that the reconstructed bases $ \mathbf{P}$ is formed through negative combinations of the original phase-aware bases, while negative values in $ \mathbf{W}$ indicate negative energy for the new reconstructed bases. Both compromise the original phase-aware information, making the decomposition less interpretable for magnitude spectrum.

Second, NMF is also an effective way to reduce dimensionality, especially for all positive matrix like magnitude spectrum. The optimization for NMF is as follows: 
\begin{equation}
\begin{aligned}
\min_{\mathbf{W},\mathbf{H} \ge 0} & J(\mathbf{W},\mathbf{H}) = \frac{1}{2} \| \mathcal{R} - \mathbf{W} \mathbf{H} \|_F^2 , \\
\nabla_{\mathbf{W}} J &= \mathbf{W} \mathbf{H} \mathbf{H}^\top - \mathcal{R} \mathbf{H}^\top, \\
\nabla_{\mathbf{H}} J &= \mathbf{W}^\top \mathbf{W} \mathbf{H} - \mathbf{W}^\top \mathcal{R}, \\
\mathbf{W}_{ij} &\leftarrow \mathbf{W}_{ij} \frac{(\mathcal{R} \mathbf{H}^\top)_{ij}}{(\mathbf{W} \mathbf{H} \mathbf{H}^\top)_{ij}}, \\
\mathbf{H}_{ij} &\leftarrow \mathbf{H}_{ij} \frac{(\mathbf{W}^\top \mathcal{R})_{ij}}{(\mathbf{W}^\top \mathbf{W} \mathbf{H})_{ij}}.
\end{aligned}
\end{equation}

The optimization is based on the EM algorithm, where $\mathbf{W}$ and $\mathbf{H}$ are optimized alternately. Since gradient descent may lead to negative values, multiplicative updates are used instead. However, this approach also faces several issues.  The first issue is that it is not efficient for unseen data: when new data arrive, a new optimization problem must be solved. This is because, while we learn the low-rank components $ \mathbf{H}$ from the training data, the corresponding coefficient $ \mathbf{W}$ for the new data is unknown.  Moreover, since $\mathbf{W}$ is also non-negative, the least squares estimator cannot be applied. The optimization problem for new data is as follows:
\begin{equation}
\begin{aligned}
\min_{\mathbf{W}_\text{new} \ge 0} J(\mathbf{W}_\text{new}) &= \frac{1}{2} \| \mathbf{R}_\text{new} - \mathbf{W}_\text{new} \mathbf{H} \|_F^2.
\end{aligned}
\end{equation}


The second limitation lies in its inferior generalization ability relative to PCA, because adapting to new data requires re-optimizing $\mathbf{W}$, while $\mathbf{W}$ and $\mathbf{H}$ are jointly learned from the training set, thus the components $\mathbf{H}$ are optimized based on the particular $ \mathbf{W}$ learned for the training data. The interpreatability of $\mathbf{W}$ is also quite poor as it is based on fitting. On the other hand, other methods like Semi-NMF face the same negative value issue as PCA and the generalization issue as NMF. More details can be found in Appendix.  

\begin{table}[ht]
\centering
\caption{Comparison for Low-Rank Methods: Hyperplane-NMF vs NMF, Semi-NMF, and PCA}
\begin{adjustbox}{width=\columnwidth,center}
\begin{tabular}{cccccccccccccccc} 
&  Hyperplane-NMF & NMF & Semi-NMF  & PCA    \\ 
\midrule 
Interpretable &$\checkmark$  & \ding{55} &  \ding{55} & \ding{55}\\
Efficient for New Data & $\checkmark$  & \ding{55} &$\checkmark$&$\checkmark$\\
Generalizable for New Data & $\checkmark$  & \ding{55} & \ding{55} &$\checkmark$\\
\midrule 
\end{tabular}
\end{adjustbox}
\label{hyperparameter}
\end{table}

Therefore, we propose the Hyperplane-NMF method to combine the strengths of existing approaches. Our method uses the same constraints as standard NMF decomposition, but enforces the coefficient matrix $\mathbf{W} = \mathcal{R} \mathbf{H}^\top$, which represents a hyperplane projection, similar to PCA. The optimization process for Hyperplane-NMF is as follows:
\begin{equation}
\begin{aligned}
\min_{\mathbf{W} \ge 0, \mathbf{H} \ge 0} 
\mathcal{L}(\mathbf{W},\mathbf{H}) 
&= 
\| \mathcal{R} - \mathbf{W} \mathbf{H} \|_F^2
+
\lambda \sum_{i \neq j} \frac{\mathbf{h}_i^\top \mathbf{h}_j}{\|\mathbf{h}_i\| \|\mathbf{h}_j\|},\\
\frac{\partial}{\partial \mathbf{h}_i} \| \mathcal{R} - \mathcal{R} \mathbf{H}^\top \mathbf{H} \|_F^2 &= (\mathbf{H} \mathcal{R}^\top)(\mathcal{R} \mathbf{H}^\top) \mathbf{H} - (\mathbf{H} \mathcal{R}^\top) \mathcal{R}\\
&=\mathbf{W}^\top \mathbf{W} \mathbf{H} - \mathbf{W}^\top \mathcal{R},\\
\frac{\partial}{\partial \mathbf{h}_i} \sum_{j \neq i} \frac{\mathbf{h}_i^\top \mathbf{h}_j}{\|\mathbf{h}_i\| \, \|\mathbf{h}_j\|}  &=
\sum_{j \neq i} \frac{\mathbf{h}_j}{\|\mathbf{h}_i\| \, \|\mathbf{h}_j\|} 
-
\sum_{j \neq i} \frac{\mathbf{h}_i^\top \mathbf{h}_j}{\|\mathbf{h}_i\|^3 \, \|\mathbf{h}_j\|} \, \mathbf{h}_i\\
&= \nabla_{\mathbf{h}_i}^+ - \nabla_{\mathbf{h}_i}^-,\\
\mathbf{W} = \mathcal{R} \mathbf{H}^\top, \quad \mathbf{H} &\gets  \mathbf{H} \odot \frac{\mathbf{W}^\top \mathcal{R} + \lambda \, \nabla_\mathbf{H}^+}{\mathbf{W}^\top \mathbf{W} \mathbf{H} + \lambda \, \nabla_\mathbf{H}^-}.
\end{aligned}
\end{equation}

There are three main advantages of enforcing $\mathbf{W} = \mathcal{R} \mathbf{H}^\top$. The first advantage is that this formulation can be easily applied to new datasets without new optimization, as required in the standard NMF. The second advantage is that, although the gradient itself does not change from the standard NMF, it now points in the correct direction for the objective, rather than towards a precomputed $\mathbf{W}$ at each iteration. Thus, the optimization constrains $\mathbf{W}$ to be computed via a hyperplane projection as a formalized rule in the objective, which leads to improved generalization on unseen data.  As mentioned before, the $\mathbf{H}$ of NMF are optimized based on particular fitting $ \mathbf{W}$ for the training data. Thirdly, the interpretability of $\mathbf{W}$ is superior to that of NMF, as it can be visualized through hyperplane projection rather than non-interpretable fitting value. This provides a good interpretation of where the source of $\mathbf{W}$ comes from. In addition, to enforce greater diversity among the components of $\mathbf{H}$, we further incorporate a Cosine similarity regularization term with a regularized factor $\lambda$ in our Hyperplane-NMF. This aims to identify the principal effects that are different from each other, with greater diversity being preferable.

As a result, Hyperplane-NMF is efficient and generalizable to unseen data, making it highly effective and interpretable for downstream mapping when applied to the validation or test dataset. In summary, Hyperplane-NMF offers a more interpretable, efficient and generalizable decomposition. In Table~\ref{hyperparameter}, we summarize the strengths of our Hyperplane-NMF compared with NMF, Semi-NMF, and PCA.


\begin{figure*}[t]
    \centering
    \includegraphics[width=\linewidth]{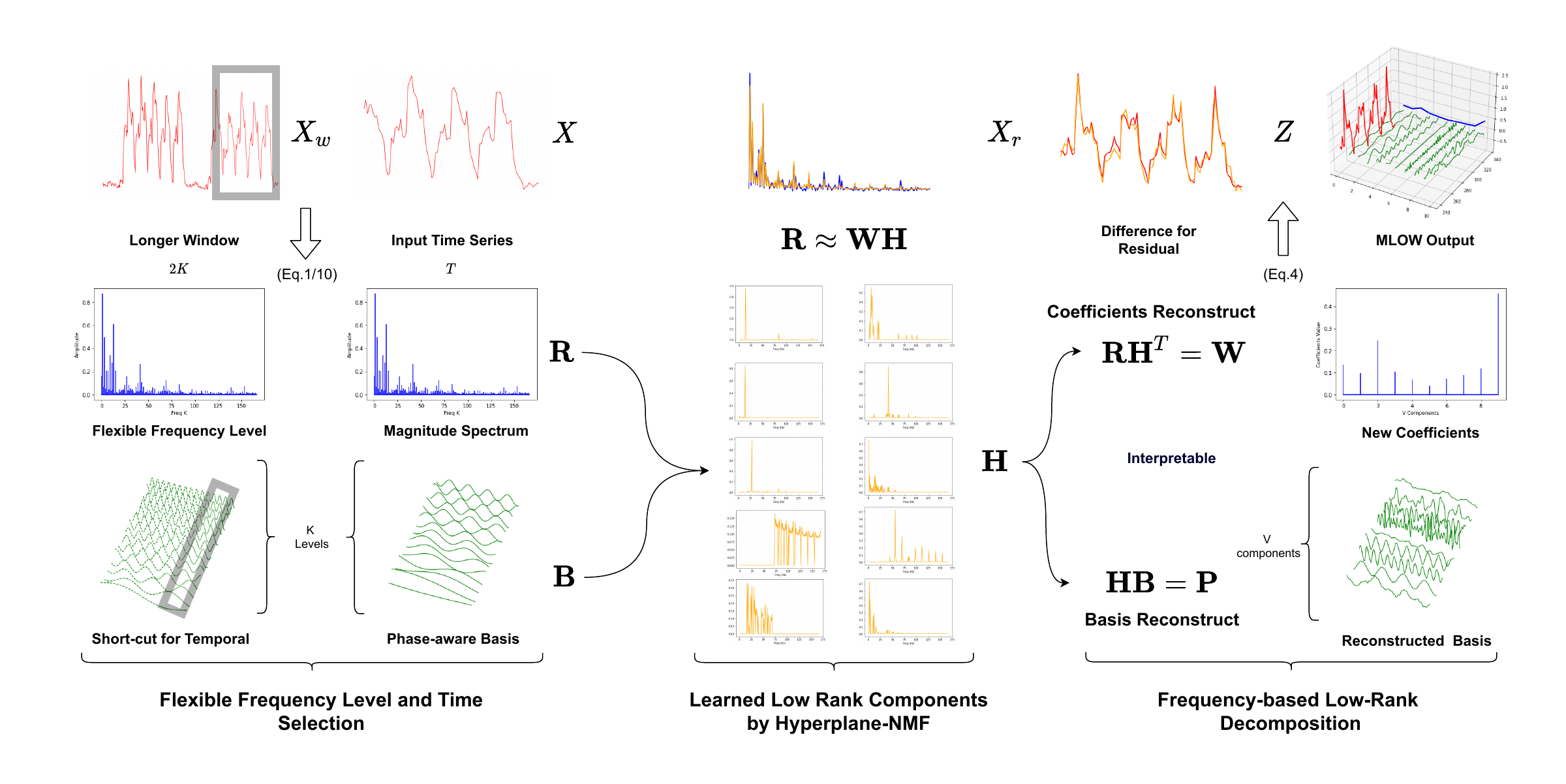}
    \caption{ \textbf{Inference pipeline for MLOW.} The original time series is decomposed into $V$ components and residual. A larger window for $X$ enables more flexible extraction of frequency magnitude levels while preserving the same temporal information.The learned Hyperplane-NMF components provide interpretable representations and serve as interpretable sources for the decomposed pieces.
     }
    \label{Inference}
    \vspace{-10pt}
\end{figure*}

\subsection{Flexible Frequency levels and Input Time Horizons}
\label{flex}
In TSF, the levels of the frequency magnitude are limited to $\frac{T}{2} + 1$ of input horizon. When only a few frequency levels are available, the frequency spectrum is often too coarse to find a good low-rank representation due to frequency (spectral) leakage. To mitigate frequency (spectral) leakage, we want to achieve flexible frequency level selection. Thus, we use a longer historical window with length $2K$ to extract the frequency spectrum, but only retain the most recent $T$ segment of the resulting phase-aware basis functions for modeling. The mathematical formulation is as follows:
\begin{equation}
\begin{aligned}
\mathbf{X}^{i} &= \frac{1}{2K} \sum_{k=0}^{K} 
\mathbf{R}^{i}_{k} \, \cos\Big( \frac{ k \pi n}{K} - \boldsymbol{\varphi}_k \Big)=\mathbf{R}^{i}\mathbf{B}^{i}+ \mathbf{X}_{m}^{i}, \\[2mm]
\mathbf{R}_k &= \mathbf{a_k}\sqrt{\mathbf{r}[k]^2 + \mathbf{i}[k]^2},\quad \boldsymbol{\varphi}_k = \operatorname{atan2}\big( \mathbf{i}[k], \mathbf{r}[k]\big)
, \\[1mm]
\mathbf{a_k}&= \begin{cases} 1,  \quad k=0, K&  \\ 2 ,\quad k=others & \end{cases}   n = 2K-T+1,\ldots,2K.  \\[1mm] 
\end{aligned}
\label{eq:amplitude_phase}
\end{equation}
Here, $r[k]$ and $i[k]$ denote the frequency levels at $k$ of the real and imaginary parts, extracted by a window of length $2K$ from $\mathbf{X}_w^{i}$. This requires additional past information $\mathbf{X}^{i}_{e}$, such that $\mathbf{X}_w^{i} = [\mathbf{X}_e^{i}, \mathbf{X}^{i}]$. Notably, $\mathbf{X}_e^{i}$ is not used in downstream modeling, but only to boost the initial magnitude levels. Since our ultimate goal is to compute a low-rank representation with rank $V$ of the magnitude spectrum, increasing the initial levels for magnitude spectrum at this stage does not affect the final downstream mapping complexity. Thus, this mathematical development helps us build a flexible pipeline to select both time horizon $T$ and frequency level $K$ freely.


\subsection{Interpretable Inference for MLOW Decomposition}



\begin{figure*}[t]
\centering
\begin{minipage}[t]{0.19\linewidth}
\centering
\includegraphics[width=\textwidth,height=0.8\textwidth]{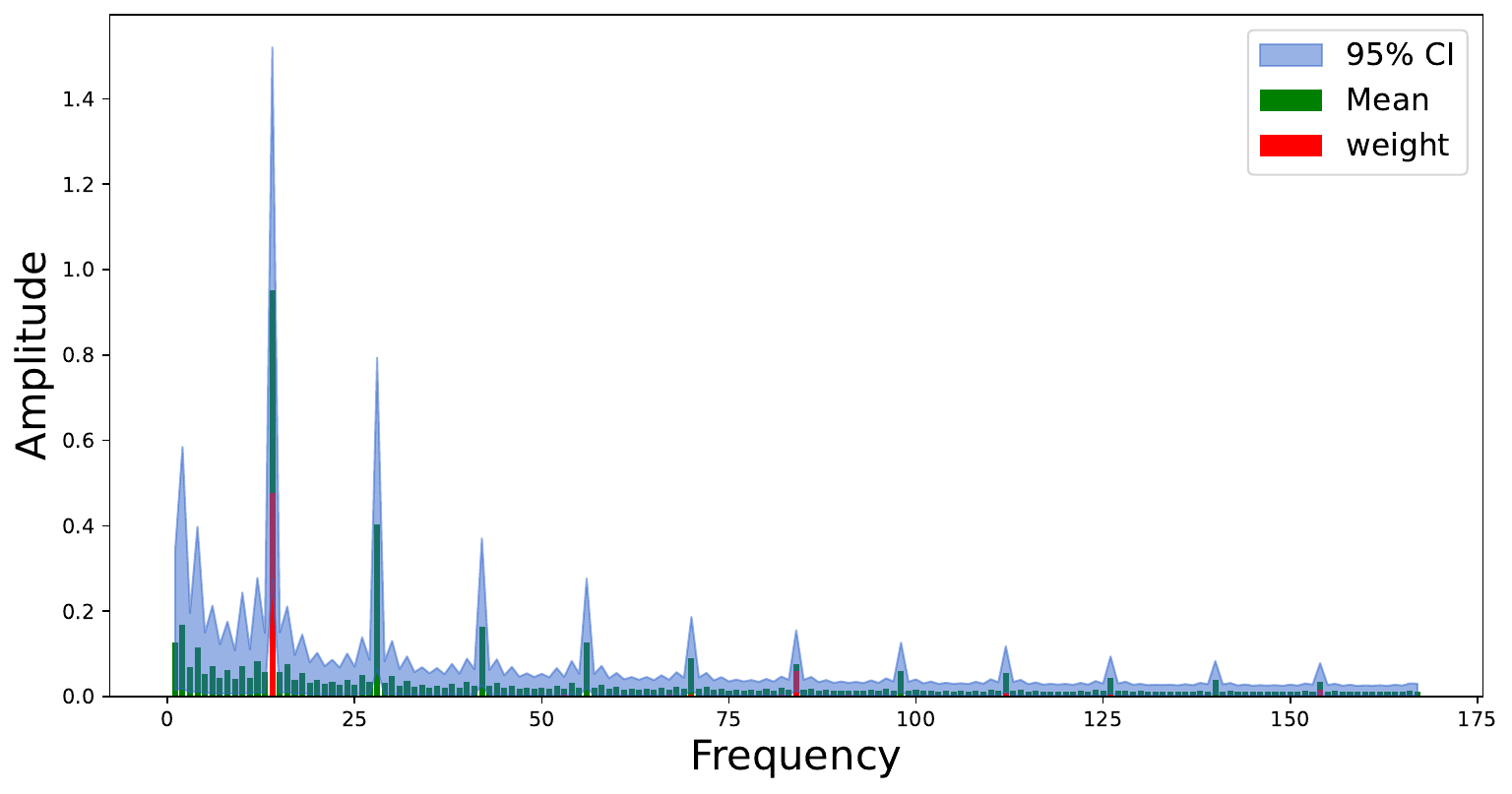}
\end{minipage}%
\begin{minipage}[t]{0.19\linewidth}
\centering
\includegraphics[width=\textwidth,height=0.8\textwidth]{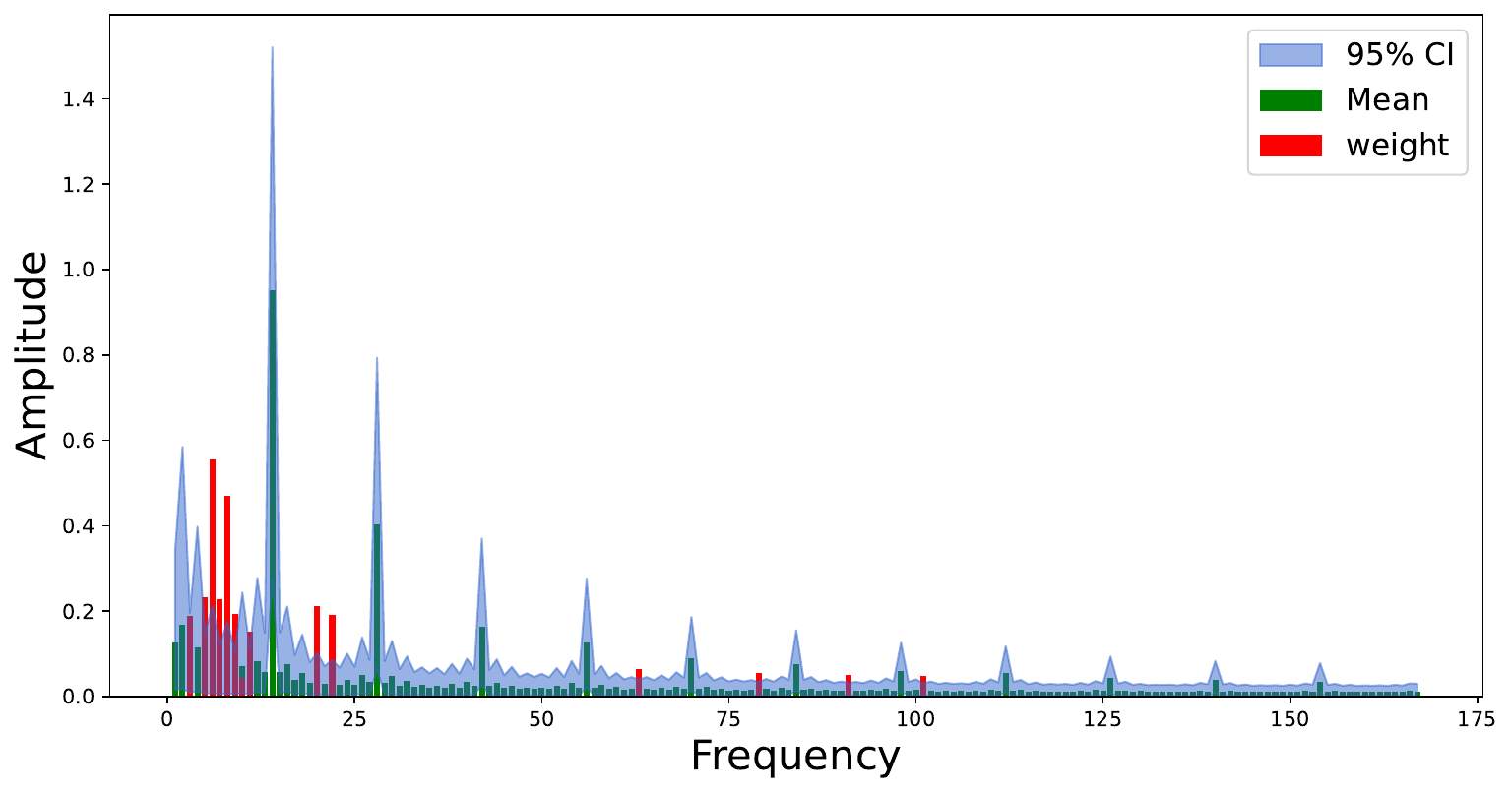}
\end{minipage}%
\begin{minipage}[t]{0.19\linewidth}
\centering
\includegraphics[width=\textwidth,height=0.8\textwidth]{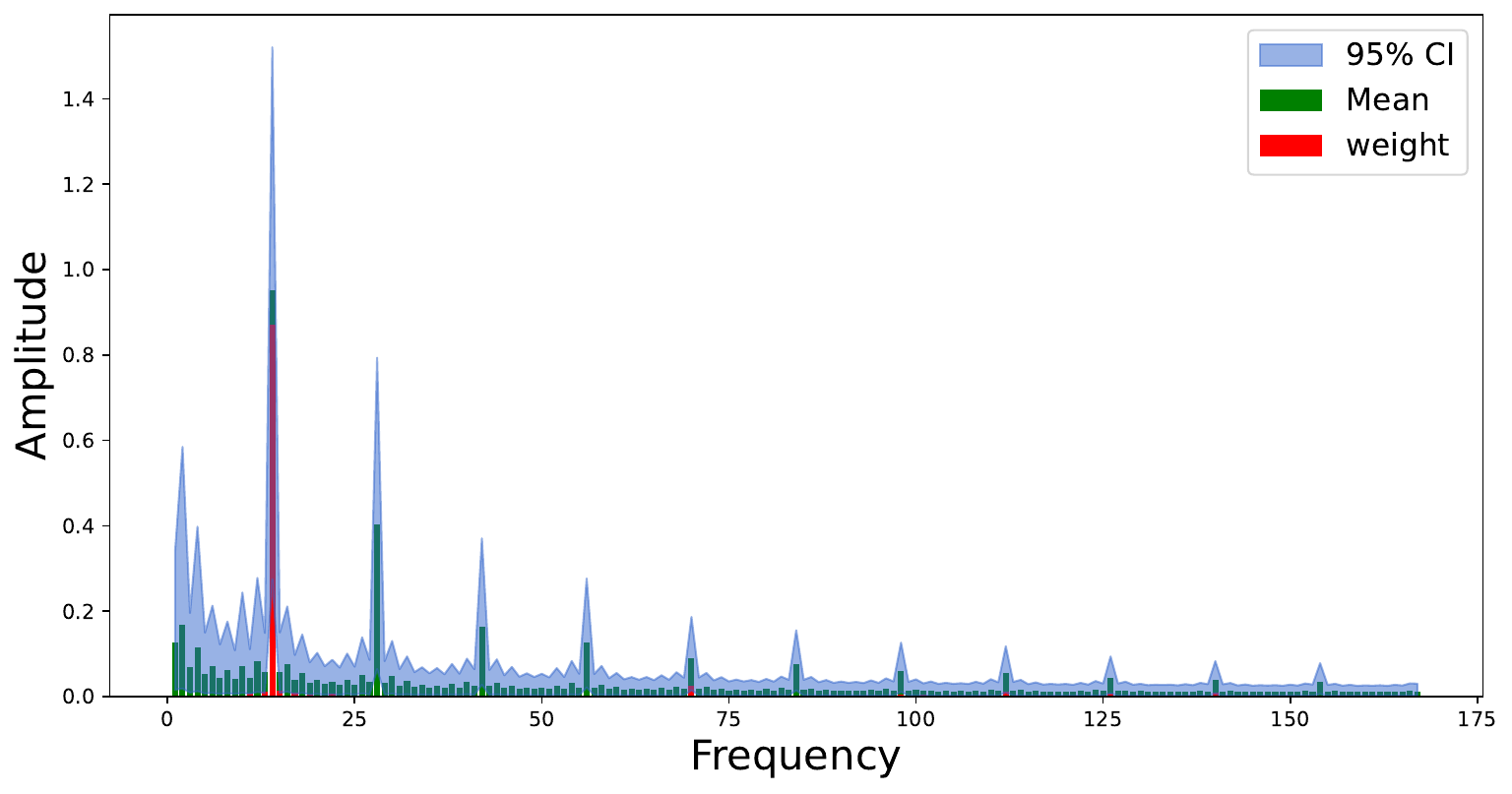}
\end{minipage}
\begin{minipage}[t]{0.19\linewidth}
\centering
\includegraphics[width=\textwidth,height=0.8\textwidth]{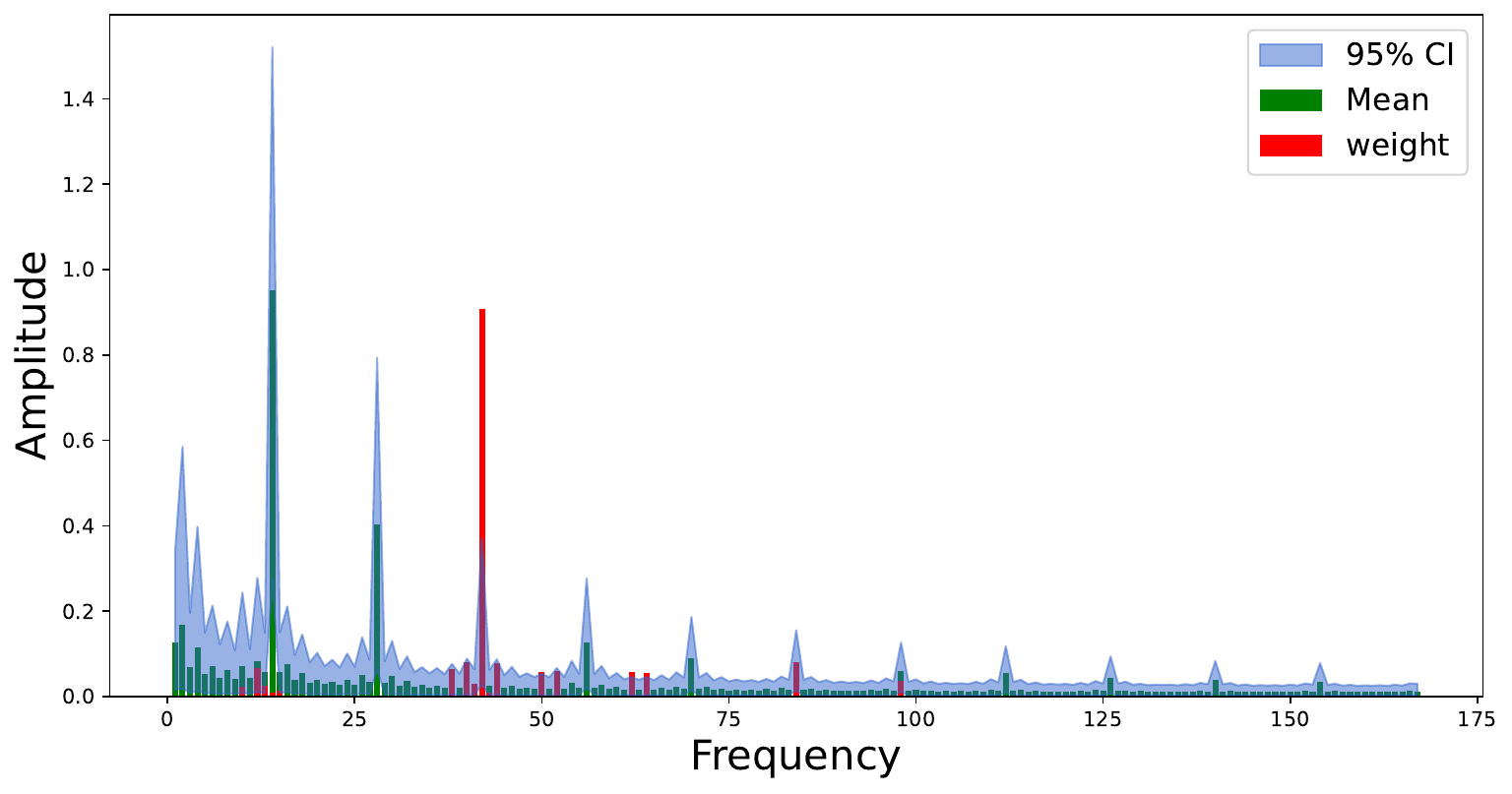}
\end{minipage}
\begin{minipage}[t]{0.19\linewidth}
\centering
\includegraphics[width=\textwidth,height=0.8\textwidth]{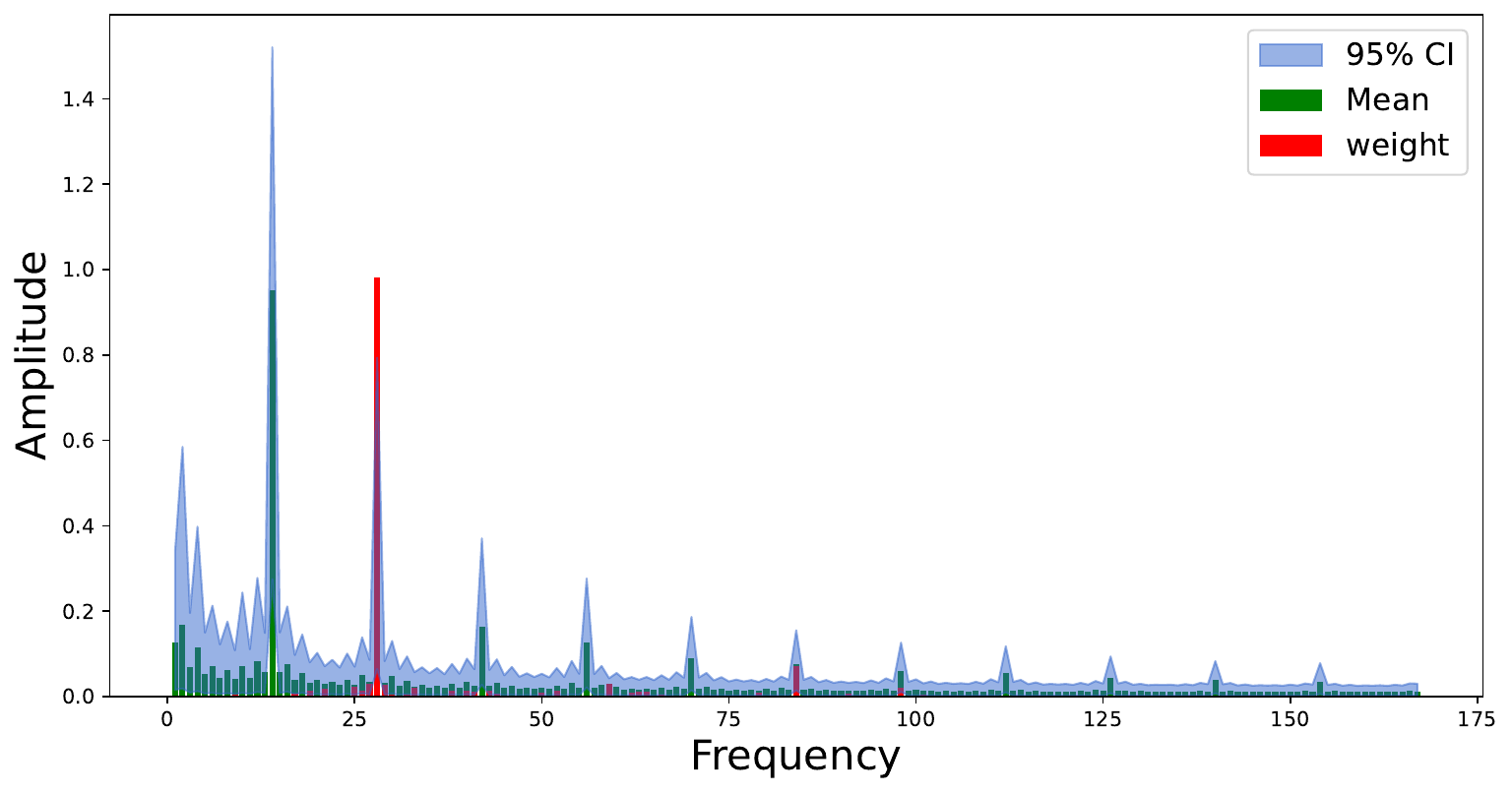}
\end{minipage}
\begin{minipage}[t]{0.19\linewidth}
\centering
\includegraphics[width=\textwidth,height=0.8\textwidth]{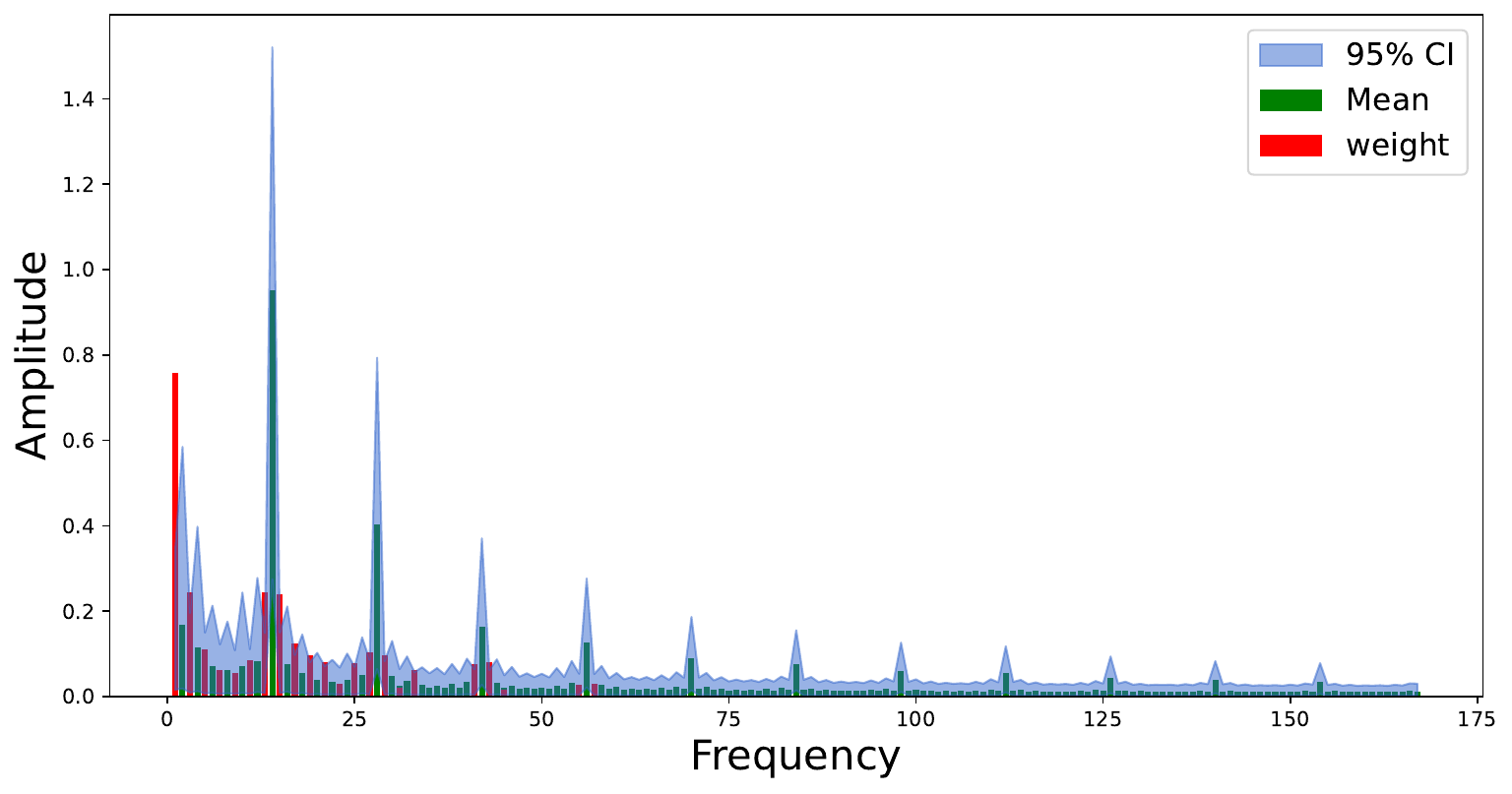}
\end{minipage}%
\begin{minipage}[t]{0.19\linewidth}
\centering
\includegraphics[width=\textwidth,height=0.8\textwidth]{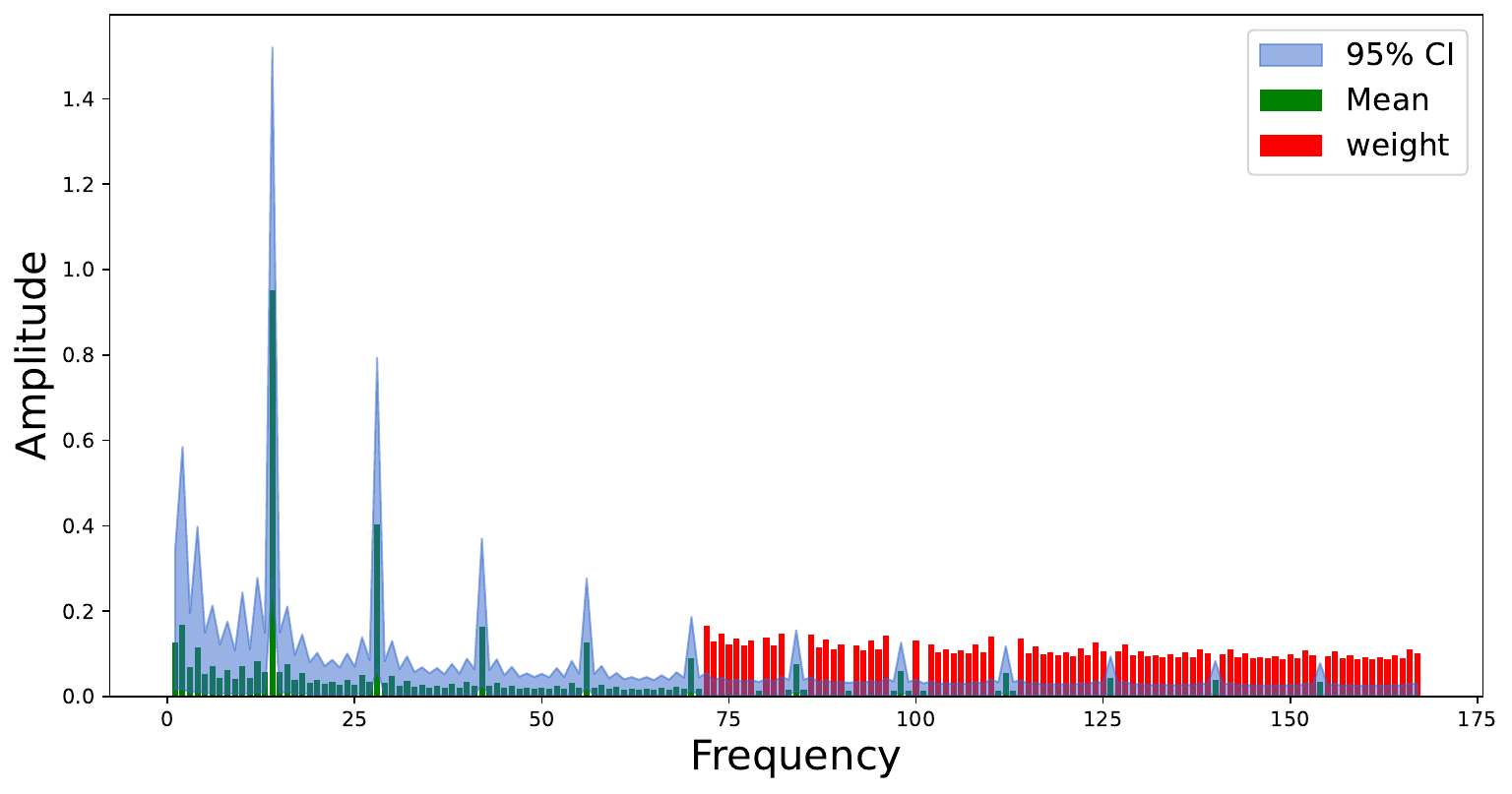}
\end{minipage}%
\begin{minipage}[t]{0.19\linewidth}
\centering
\includegraphics[width=\textwidth,height=0.8\textwidth]{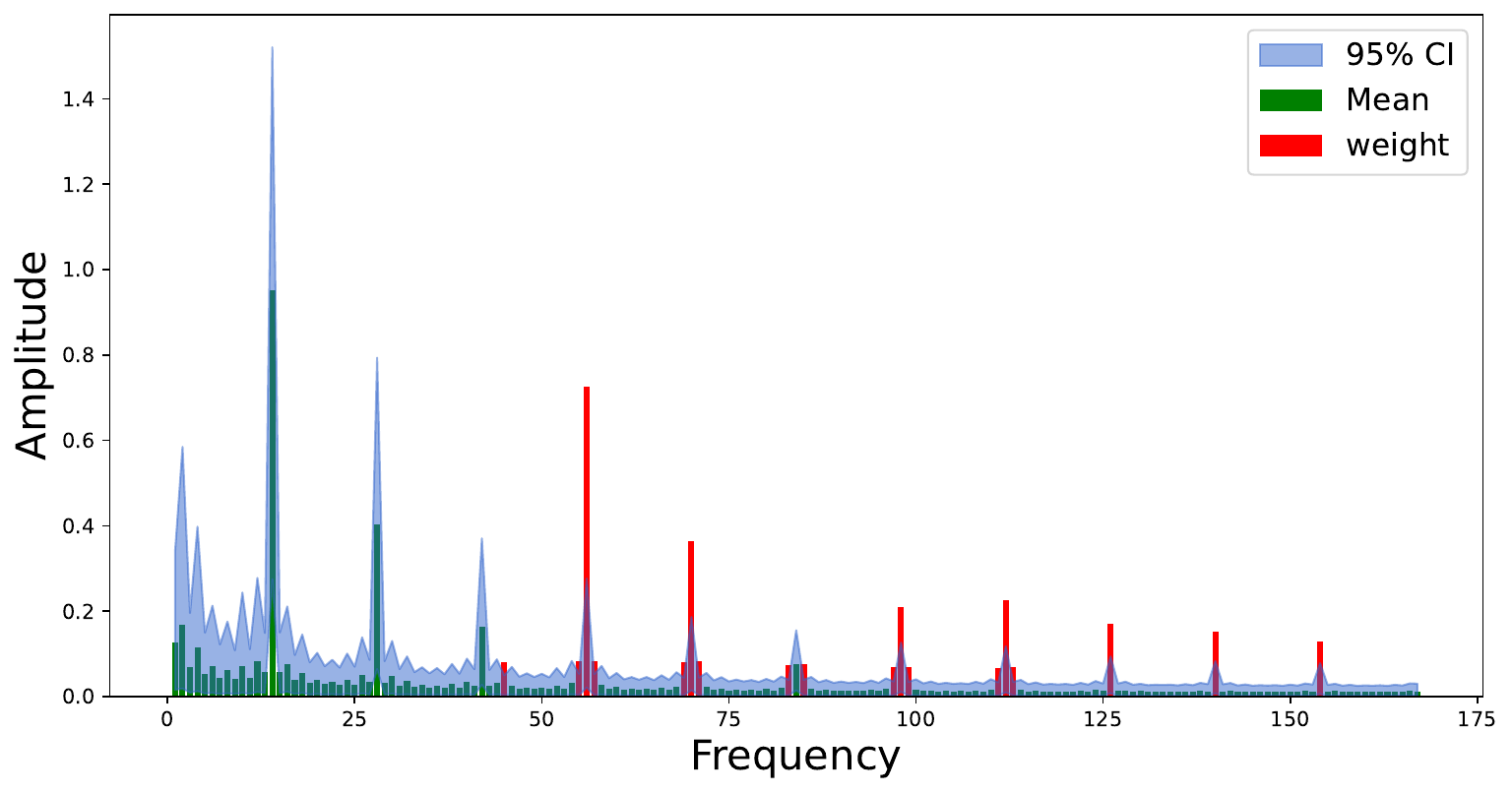}
\end{minipage}
\begin{minipage}[t]{0.19\linewidth}
\centering
\includegraphics[width=\textwidth,height=0.8\textwidth]{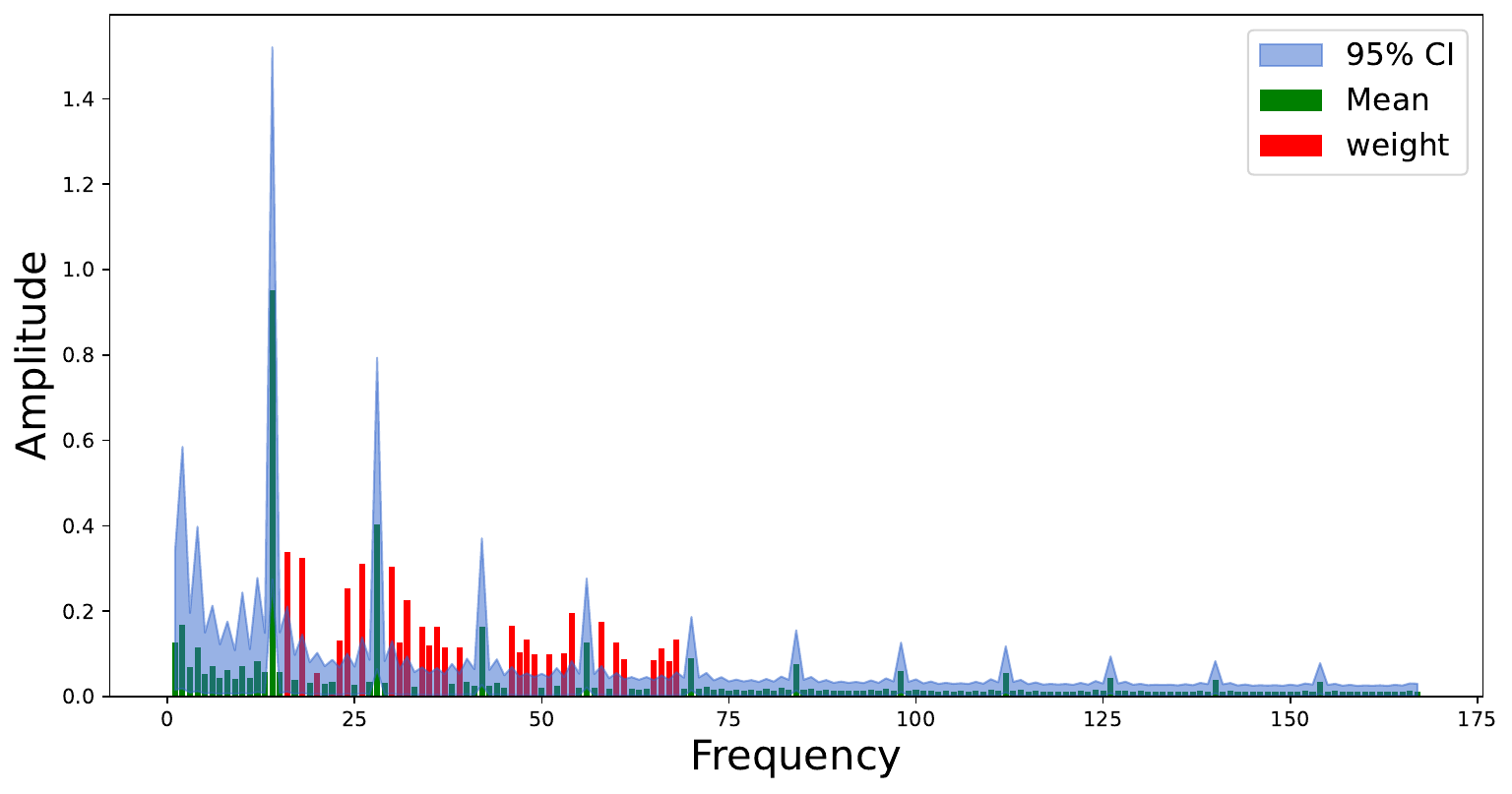}
\end{minipage}
\begin{minipage}[t]{0.19\linewidth}
\centering
\includegraphics[width=\textwidth,height=0.8\textwidth]{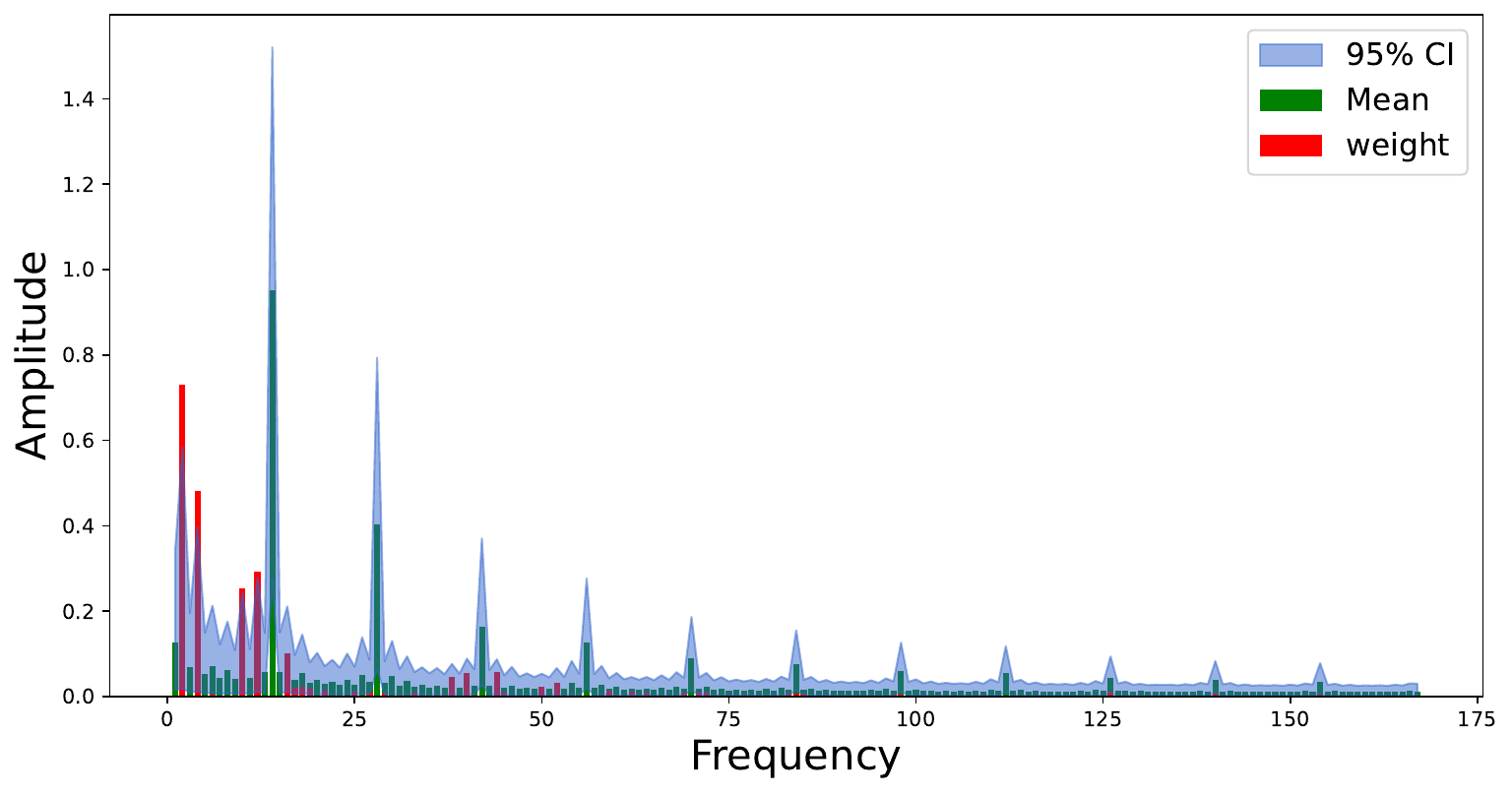}
\end{minipage}
\caption{Learned 10 Components for the ECL Data Compared to Its Magnitude Spectrum Distributions. The blue regions correspond to the 95\% confidence interval of magnitude spectrum, the green columns show the mean magnitude spectrum, and the red columns indicate the weights of the learned components. The larger version is shown in Figure~\ref{components_ECL} in the Appendix.}
\label{components}
\end{figure*}

\begin{figure*}[t]
\centering
\begin{minipage}[t]{0.235\linewidth}
\centering
\includegraphics[width=\textwidth,height=0.8\textwidth]{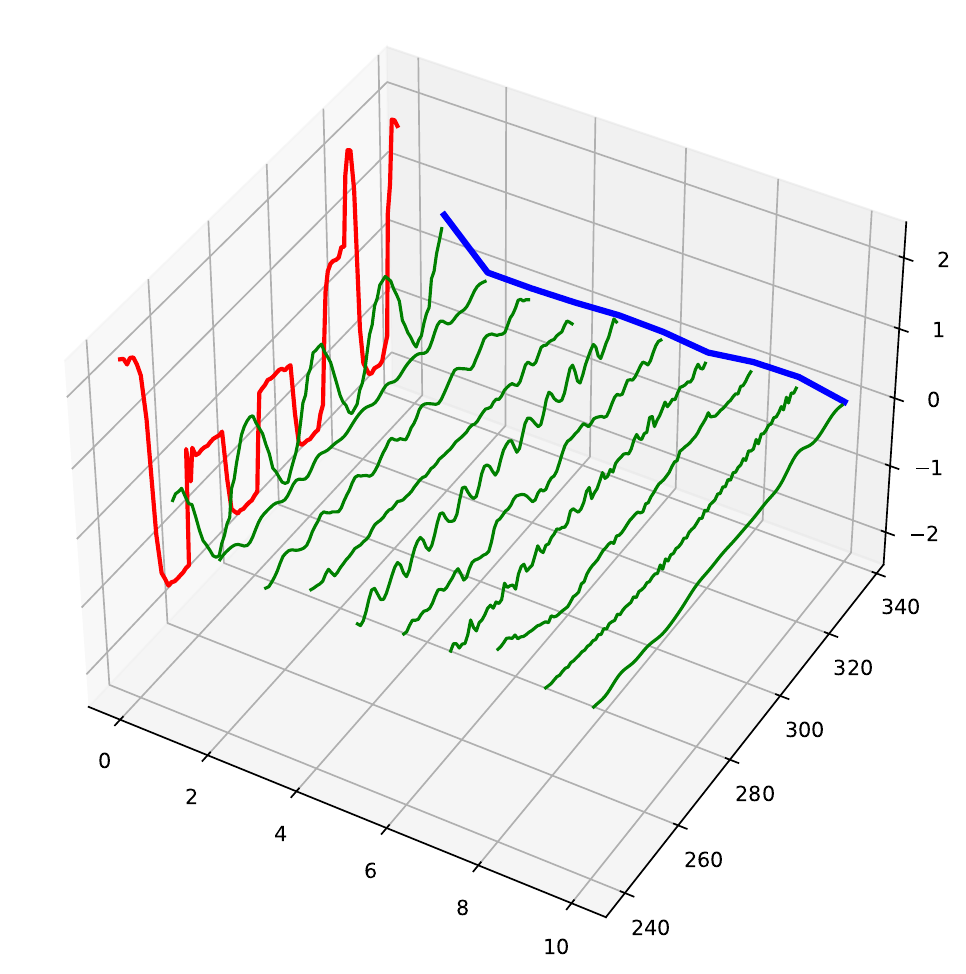}
\caption*{ECL}
\end{minipage}%
\begin{minipage}[t]{0.235\linewidth}
\centering
\includegraphics[width=\textwidth,height=0.8\textwidth]{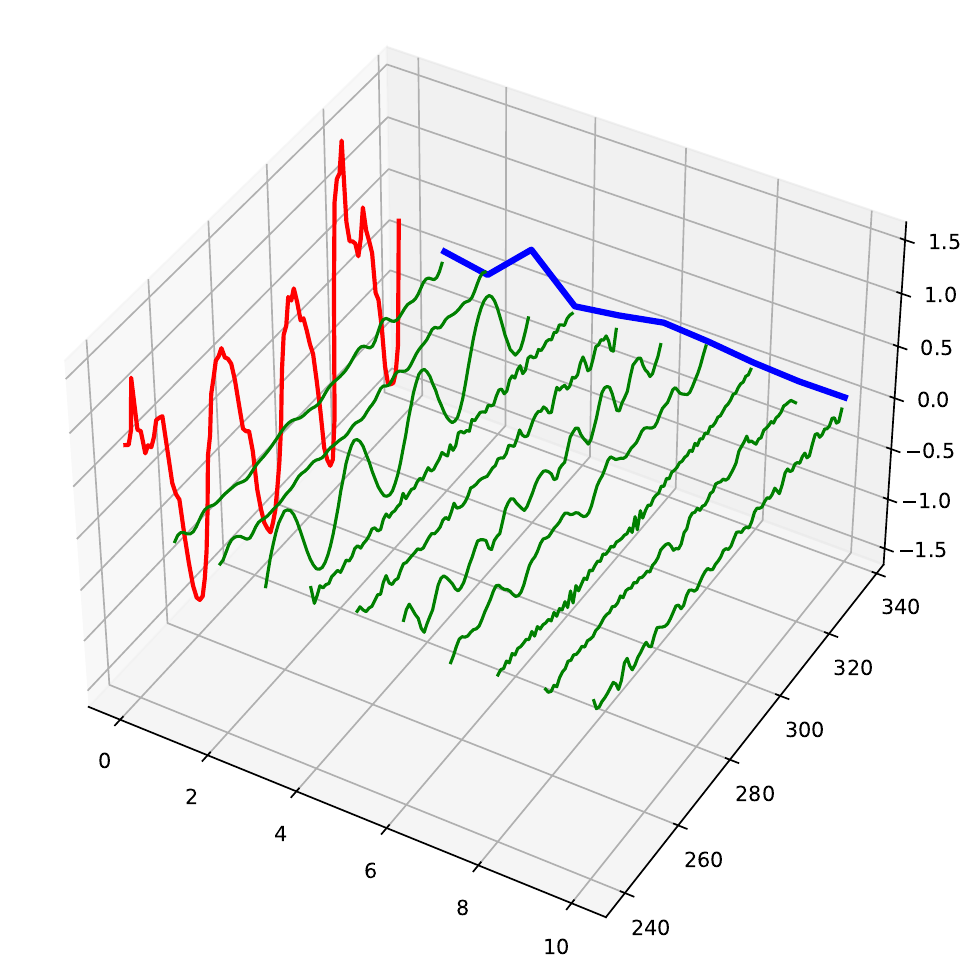}
\caption*{Traffic}
\end{minipage}%
\begin{minipage}[t]{0.235\linewidth}
\centering
\includegraphics[width=\textwidth,height=0.8\textwidth]{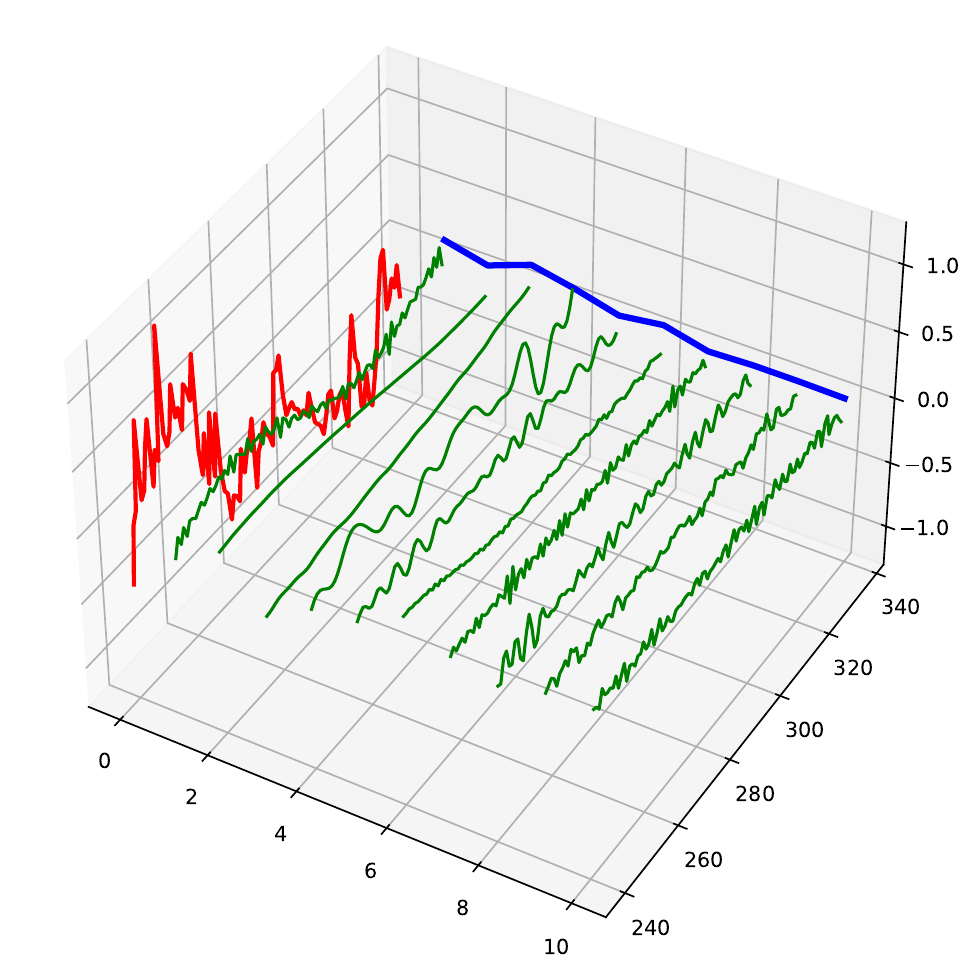}
\caption*{Weather}
\end{minipage}
\begin{minipage}[t]{0.235\linewidth}
\centering
\includegraphics[width=\textwidth,height=0.8\textwidth]{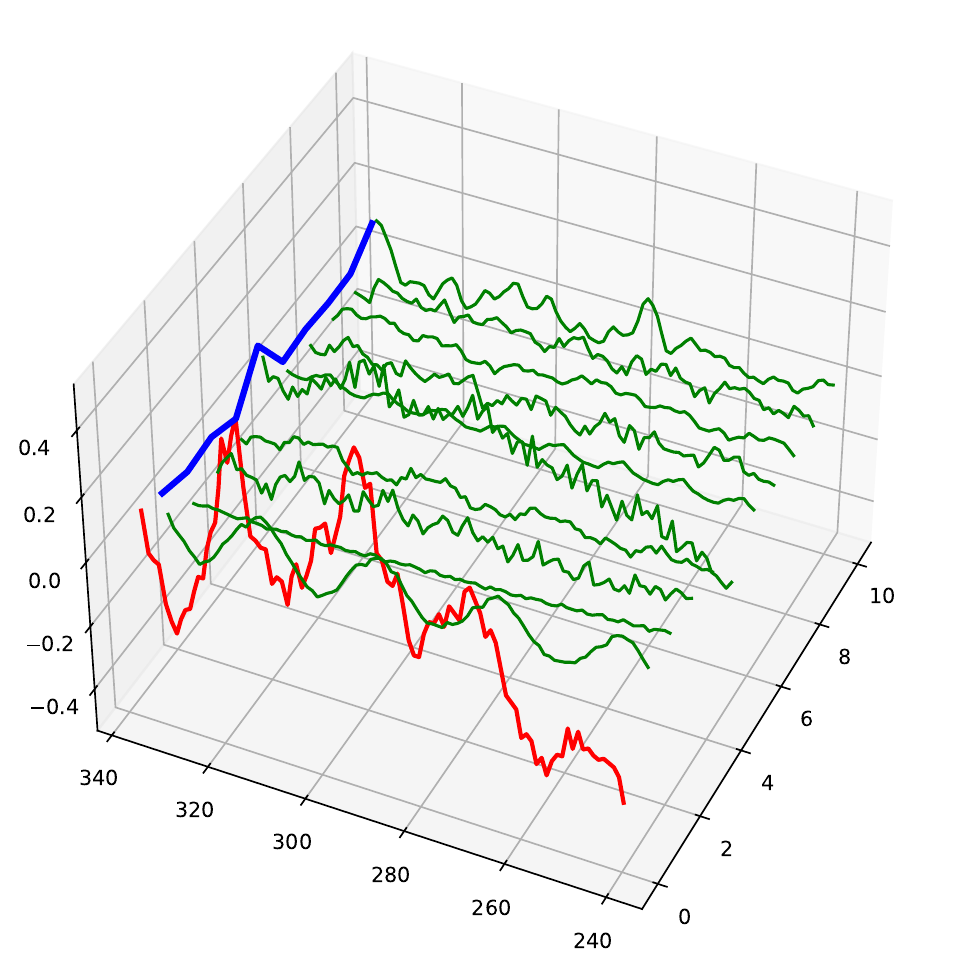}
\caption*{ETTh1}
\end{minipage}
\begin{minipage}[t]{0.235\linewidth}
\centering
\includegraphics[width=\textwidth,height=0.8\textwidth]{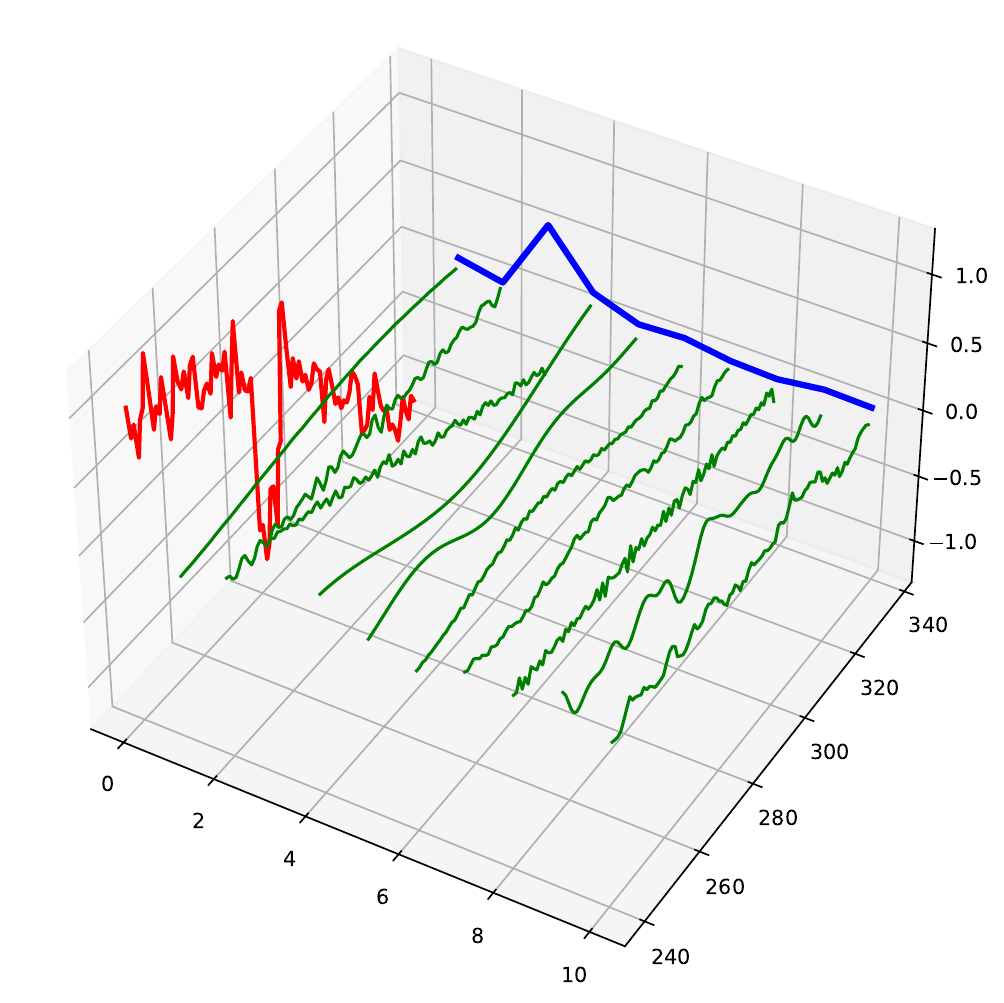}
\caption*{PEMS03}
\end{minipage}%
\begin{minipage}[t]{0.235\linewidth}
\centering
\includegraphics[width=\textwidth,height=0.8\textwidth]{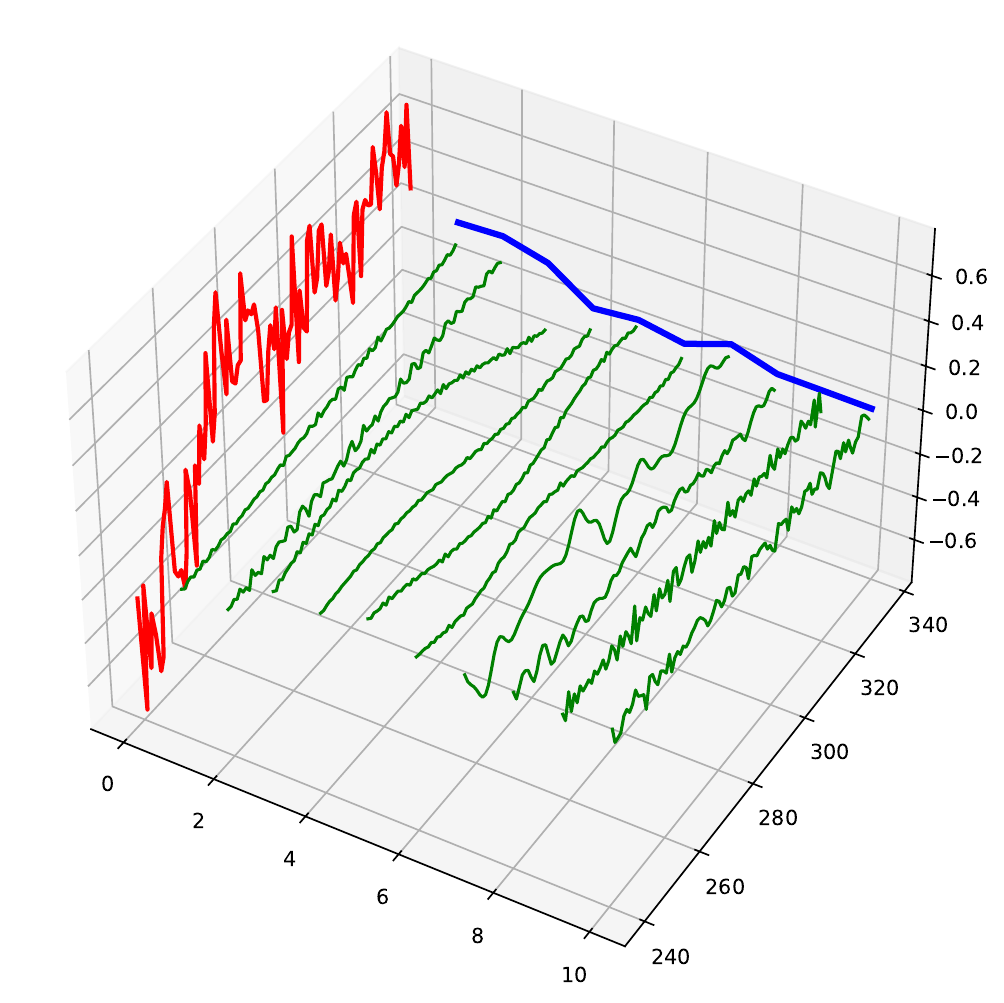}
\caption*{PEMS08}
\end{minipage}%
\begin{minipage}[t]{0.235\linewidth}
\centering
\includegraphics[width=\textwidth,height=0.8\textwidth]{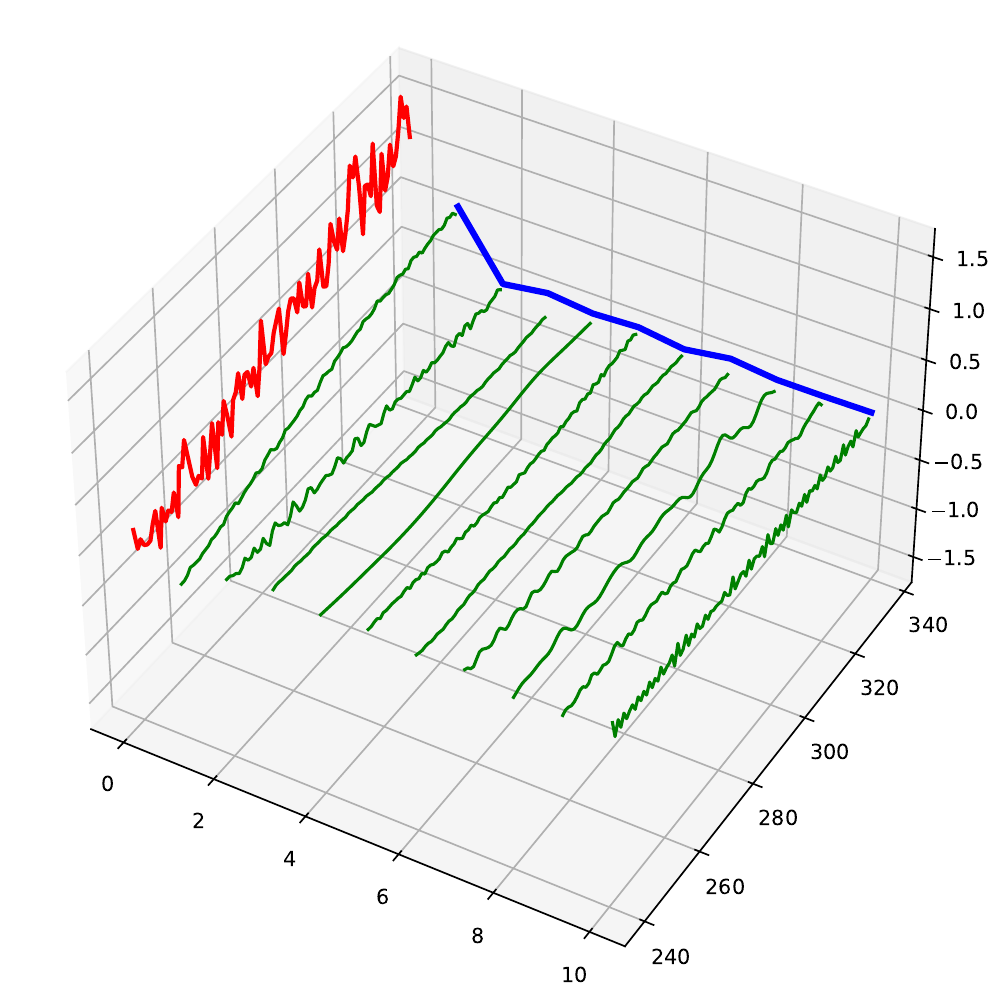}
\caption*{PEMS04}
\end{minipage}
\begin{minipage}[t]{0.235\linewidth}
\centering
\includegraphics[width=\textwidth,height=0.8\textwidth]{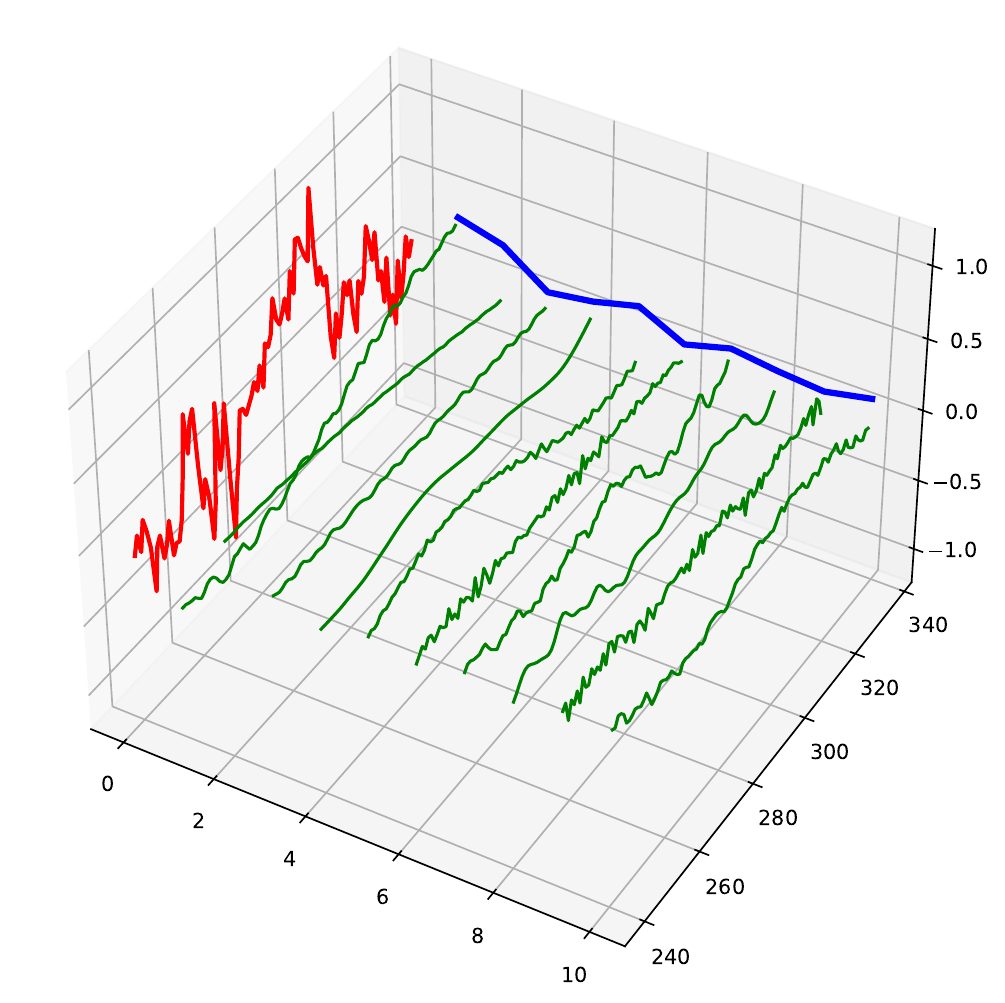}
\caption*{PEMS07}
\end{minipage}
\caption{Visualization of Our Decomposition on Eight Datasets. The results show that our decomposition can capture hierarchical effects. }
\label{plot}
\end{figure*}

More details for training and inference can be found in Appendix Algorithm \ref{alg:freq_decomp}. In Figure~\ref{Inference}, we visualize the inference pipeline of our method for its good interpretability. Here, after decomposition, both $\mathbf{W}$ and $\mathbf{P}$ are computed via a hyperplane projection through leaned $\mathbf{H}$. Our method provides a clear and interpretable meaning for $\mathbf{W}$ and $\mathbf{P}$, especially for $\mathbf{W}$, helping  understand its sources of energy. Commonly co-occurring patterns are more likely to be clustered into the same components, distinguishing them from others. In Figure \ref{components}, we visualize the learned components for the ECL data and observe several interesting patterns, where component weights are grouped by distinct frequency levels. Components with grouped low-level weights, exhibiting an exponential decay pattern,  more likely correspond to trending effects. Components with mid-level weights, which often activate multiples of specific frequency levels,  more likely correspond to seasonal effects. Components with high-level weights that exhibit jumps at certain frequency levels are more likely associated with nonstationary fluctuations. For example, they are visible in components 6, 7, and 8 in Figure~\ref{components}. In particular, component 7 is extremely interesting, as it clusters the high levels into one group while skipping certain levels and assigning them to another component. This demonstrates that our decomposition can distinguish trending, seasonal and nonstationary effects in a more effective and interpretable manner. Thus, this explains why our decomposition looks more promising. In Figure \ref{plot}, the visualization shows that our MLOW decomposition is less sensitive to noises and can successfully separate trending and seasonal components hierarchically. We provide more visualizations for decomposition and learned components for each data in Appendix.

\begin{table*}[t]
\centering
\caption{MLOW vs. MA as Plug-and-Play Decomposition Modules on iTransformer and PatchTST under T = 96 and the Same Hyperparameter. The full results are shown in Table \ref{table_full} in Appendix.}
\begin{adjustbox}{width=2\columnwidth,center}
\begin{tabular}{cccccccc|cccccccccccccc} 
\midrule
\multicolumn{2}{c}{Methods} &\multicolumn{2}{c}{iTransformer+MLOW} & \multicolumn{2}{c}{iTransformer+MA}                 & \multicolumn{2}{c}{iTransformer}                   & \multicolumn{2}{c}{PatchTST+MLOW} & \multicolumn{2}{c}{PatchTST+MA}& \multicolumn{2}{c}{PatchTST} \\

\midrule
\multicolumn{2}{c}{Error} & MSE & MAE  & MSE & MAE  & MSE                    & MAE                    & MSE   & MAE                 & MSE   & MAE                   & MSE   & MAE  \\

\midrule
                     
\multirow{1}{*}{PEMS03} 

                         & avg& \textbf{0.086}	 & \textbf{0.186} & 	0.130 & 	0.232	 & 0.129	 &0.233	 & \textbf{0.108}	 & \textbf{0.222}	 &0.203	 &0.297	 &0.214	 &0.307
\\

\midrule
\multirow{1}{*}{PEMS04} 

                         & avg& \textbf{0.080} &	\textbf{0.177} &	0.157	 &0.250 &	0.157 &	0.251	 &\textbf{0.120} &	\textbf{0.243}	 &0.265 &	0.343	 &0.259	 &0.333

\\

\midrule
\multirow{1}{*}{PEMS08}

                         & avg& \textbf{0.081}&	\textbf{0.173}&	0.142&	0.233&	0.142&	0.232&	\textbf{0.120}&	\textbf{0.232}&	0.221&	0.313&	0.221&	0.308

\\

\midrule
                   \multirow{1}{*}{PEMS07}  

                         & avg& \textbf{0.060}&	\textbf{0.146}&	0.124&	0.218	&0.124&	0.218&	\textbf{0.090}&	\textbf{0.211}&	0.223	&0.310&	0.215&	0.302

\\

\midrule
                   \multirow{1}{*}{ECL}   & avg& \textbf{0.155}&	\textbf{0.248}&	0.180&	0.271	&0.180&	0.271&	\textbf{0.173}&	\textbf{0.281}&	0.199	&0.296&	0.205&	0.295

\\

     \midrule 
                   \multirow{1}{*}{Traffic}  
                         & avg& \textbf{0.393}	& \textbf{0.250}&	0.417&	0.280&	0.428&	0.282&	\textbf{0.405}&\textbf{	0.294}&	0.470&	0.309&	0.467&	0.309

\\

     \midrule
                     
                        \multirow{1}{*}{Weather} 
                         & avg&\textbf{0.231}&	\textbf{0.264}&	0.259&	0.280	&0.261&	0.282&	\textbf{0.232}	&\textbf{0.263}&	0.255&	0.278&	0.253&	0.277\\

     \midrule 

                        \multirow{1}{*}{ETTh1}

                         & avg& \textbf{0.424}&	\textbf{0.438}&	0.462&	0.453&	0.456 &	0.450&	\textbf{0.410}&	\textbf{0.429}&	0.436&	0.440&	0.447&	0.446&
\\

                         
\hline
\end{tabular}
\end{adjustbox}
\label{table_main}
\end{table*}

\begin{table*}[t]
\caption{MLOW + iTransformer / PatchTST vs. Other Deep TSF Methods under T = 96. The full results are in Table \ref{others_full} in Appendix. }
\centering
\begin{adjustbox}{width=2\columnwidth}
\begin{tabular}{cccccccccccccccccc} 
\midrule
 \multicolumn{2}{c}{\multirow{2}{*}{Methods}}  & \multicolumn{2}{c}{iTransformer+MLOW}   & \multicolumn{2}{c}{PatchTST+MLOW} & \multicolumn{2}{c}{DUET} & \multicolumn{2}{c}{CycleNet} & \multicolumn{2}{c}{SparseTSF}  & \multicolumn{2}{c}{TimeKAN}    & \multicolumn{2}{c}{TimesNet} & \multicolumn{2}{c}{TimeMixer} \\ 
 &  &  \multicolumn{2}{c}{\cite{LiuHZWWML24}}  & \multicolumn{2}{c}{\cite{nie2023time}}& \multicolumn{2}{c}{\cite{qiu2025duet}} & \multicolumn{2}{c}{\cite{lin2024cyclenet}} & \multicolumn{2}{c}{ \cite{lin2024sparsetsf}}  & \multicolumn{2}{c}{ \cite{huang2025timekan}}    & \multicolumn{2}{c}{ \cite{wu2022timesnet}} & \multicolumn{2}{c}{ \cite{TimeMixer2024}}\\

\midrule
\multicolumn{2}{c}{Error}   & MSE                    & MAE                    & MSE                    & MAE                    & MSE   & MAE        & MSE   & MAE    & MSE   & MAE   & MSE   & MAE & MSE   & MAE   & MSE   & MAE                          \\ 
\midrule
\multirow{1}{*}{PEMS03}  
                          & avg & \textbf{0.086}& 	\textbf{0.186}& 	0.108	& 0.222	& 0.114	& 0.223& 	0.122	& 0.223& 	0.297	& 0.352	& 0.276	& 0.3465& 	0.149& 	0.257& 	\underline{0.107}& 	\underline{0.212}\\

\midrule
\multirow{1}{*}{PEMS08}  
                       & avg & \textbf{0.081} &	\textbf{0.173} &	0.120 &	0.232 &\underline{	0.104} &	\underline{0.211}&	0.150&	0.237	 &0.298&	0.351	&0.275&	0.346&	0.189&	0.280&	0.120 &	0.225
\\
\midrule
\multirow{1}{*}{ECL}             
                       & avg & \textbf{0.155} &	\textbf{0.248}	&0.173&	0.281	&0.172	&\underline{0.258}	& \underline{0.167}&	0.259	&0.214&	0.288&	0.203&	0.292&	0.192&	0.295&	0.182	& 0.272
\\

\midrule

    \multirow{1}{*}{Traffic}   
                       & avg & \textbf{0.393} &	\textbf{0.250}	&\underline{0.405}&	0.294	&0.451	&\underline{0.269}&	0.471&	0.301	&0.589	&0.339&	0.577&	0.372&	0.619&	0.335&	0.484	&0.297

\\

\midrule

\end{tabular}
\end{adjustbox}
\label{others}
\end{table*}

\vspace{-4pt}

\section{Experiment Results}

\subsection{Datasets and Settings}
\vspace{-3pt}

We conduct our experiments on eight real-world datasets: ETTh1, Electricity (ECL), Traffic, Weather, and PEMS (PEMS04, PEMS08, PEMS03, PEMS07) \cite{LiuHZWWML24}. We split the ETT datasets into 12/4/4 months for training, validation, and test, respectively, while all other datasets are split using a 0.7/0.1/0.2 ratio. We extend the validation and test sets by the past window $2K$ to ensure that their lengths are invariant. The dropout of last batch is disabled for the test set \cite{qiu2024tfb}. The forecasting horizons are ($12$,$24$,$48$,$96$) for short-term TSF on the PEMS datasets and ($96$, $192$, $336$, $720$) for long-term TSF on the other datasets, same as iTransformer \cite{LiuHZWWML24}.  We evaluate the plug-and-play performance on PatchTST and iTransformer, and compare it with moving average (MA) decomposition with kernel size 24. The MA decomposition also modifies only the first initial projection layer for a fair comparison. All downstream mapping networks remain unchanged and use the same hyperparameters. The backbone hyperparameters are provided in Tables \ref{detail1} and \ref{detail2} in  Appendix. For the experiments in Tables~\ref{table_main} and \ref{others}, we use the universal same hyperparameters for MLOW: input sequence length $T=96$, frequency level $K=168$, number of low-rank components $V=10$, and regularization factor $\lambda=20$.

\subsection{Main Results}
As shown in Table~\ref{table_main}, our method significantly improves both PatchTST and iTransformer to a new level, demonstrating the effectiveness of MLOW decomposition in separating different effects for TSF. In contrast, moving-average kernels yield only marginal and often unstable improvements. Our improvement is robust for both long- and short-term TSF across all forecasting horizons. Overall, iTransformer+MLOW outperforms PatchTST+MLOW across most datasets, except ETTh1, as the separation of different effects allows interaction modeling to better capture their relationships, especially in short-term TSF. In addition, we compare our new merged methods with other frequency-based and smoothing-based deep TSF methods on four datasets in Table \ref{others}. The results show that iTransformer+MLOW consistently achieves the best performance and surpasses the second-best method by a large margin.  


\subsection{Ablation and Sensitivity Studies}


Table~\ref{abalation2} also shows that our Hyperplane-NMF outperforms the other low-rank decomposed methods. Firstly, our method outperforms PCA because the proposed decomposition is strictly non-negative, thereby preserving the original phase-aware information without introducing distortion. Secondly, it surpasses NMF  as it exhibits stronger generalization to unseen data. Our learned matrix $W$ is also more interpretable than $W$ obtained through optimization-based fitting in NMF. In addition, incorporating Cosine similarity regularization also help improve NMF, encouraging the model to learn more diverse components. Finally, Semi-NMF performs worse than our method for the mentioned similar  negative and generalizable reasons. Table~\ref{abalation2} also shows that our low-rank method is much more efficient than NMF decomposition, since the optimized for $W$ is omitted. The inference time is measured on an ECL input sample with $T=96$, averaged over 10 runs.

\begin{table}[h]
\caption{Performance of MLOW with Different Low-Rank Methods on iTransformer and Average Inference Time for an Input Time Series on ECL. The full results are in Table \ref{ab2} in Appendix.}
\centering
\begin{adjustbox}{width=\columnwidth}
\begin{tabular}{cccccccccccc} 
\midrule
 \multicolumn{2}{c}{Methods}  & \multicolumn{2}{c}{Hyperplane-NMF}  & \multicolumn{2}{c}{NMF}  & \multicolumn{2}{c}{PCA}& \multicolumn{2}{c}{Semi-NMF} \\ 

\midrule
\multicolumn{2}{c}{Error} & MSE  & MAE                                & MSE & MAE                & MSE   & MAE      & MSE  & MAE                           \\ 
\midrule
\multirow{1}{*}{PEMS08}       
                          & avg  & \textbf{0.081}	 &\textbf{0.173}	 &0.083 &	0.175 &	0.086 &	0.177 &	0.089 &	0.184\\

\midrule
\multirow{1}{*}{ECL}  
                         & avg  & \textbf{0.155}&	\textbf{0.248}&	0.163&	0.258	&0.160&	0.255 &0.166&	0.260\\

\midrule
\multicolumn{2}{c}{Inference Time (s)}    & \multicolumn{2}{c}{\textbf{0.000259}}    & \multicolumn{2}{c}{0.121408}  & \multicolumn{2}{c}{0.000265} &  \multicolumn{2}{c}{0.000288} & \\ 
\midrule
\end{tabular}
\end{adjustbox}
\label{abalation2}
\end{table}

Table~\ref{abalation4} presents the sensitivity studies using different low-rank values of $V$ for the initial frequency levels $K=168$ and $K=48$, respectively. Small values of $V$ make little difference while $V=5$ or $V=10$ usually enables optimal, and continuously increasing $V$ does not necessarily improve performance; in fact, it often degrades it. This is because our ultimate goal is to separate different effects, and the large $V$ components may confuse the model for different effects. However, using $K=168$ consistently yields better results than $K=48$ under the same $V$, even  though model architectures are identical. This shows that our mathematical mechanism in Section \ref{flex} is very important; otherwise,  $K$ can only equal  $\frac{T}{2}$. This is because extra levels for $K$ mitigate the frequency (spectral) leakage effects to help find a better low-rank decomposition. 



\begin{table}[h]
\caption{Sensitivity Analysis for $K$ and $V$ on iTransformer. The full results are provided in Table \ref{ab4} in Appendix.}
\centering
\begin{adjustbox}{width=\columnwidth}
\begin{tabular}{cccccccccccc} 
\midrule
\multicolumn{2}{c}{Error}   & MSE              & MAE                    & MSE                    & MAE                    & MSE   & MAE        & MSE   & MAE                                \\ 

\midrule
 \multicolumn{2}{c}{(K=168)}  & \multicolumn{2}{c}{V=5}  & \multicolumn{2}{c}{V=10}  & \multicolumn{2}{c}{V=15} & \multicolumn{2}{c}{V=20}  \\ 

\midrule
\multirow{1}{*}{PEMS08}  
                             & avg & 0.082	&0.173 &	\textbf{0.081} &	\textbf{0.173} &	0.081 &	0.173	&0.082 &0.174\\

\midrule
\multirow{1}{*}{ECL} 
                         & avg & \textbf{0.154}	&  \textbf{0.246} &	 0.155& 0.248	&0.155	& 0.248& 0.156 & 0.251\\

\midrule
 \multicolumn{2}{c}{(K=48)} & \multicolumn{2}{c}{V=5}  & \multicolumn{2}{c}{V=10}  & \multicolumn{2}{c}{V=15} & \multicolumn{2}{c}{V=20}  \\ 
\midrule
\multirow{1}{*}{PEMS08} 
                          & avg & \textbf{0.086}  &	\textbf{0.182}	 &0.087	 &0.183 &	0.087	 &0.182 &	0.088	& 0.184\\

\midrule
\multirow{1}{*}{ECL}
                         & avg & 0.161 & 0.252&	\textbf{0.160} & \textbf{0.251} & 0.161  & 0.253& 0.162 & 0.253\\

\midrule
\end{tabular}
\end{adjustbox}
\label{abalation4}
\end{table}

\section{Conclusions}

The interpretability of our MLOW decomposition draws the following perspectives: (i) the low-rank pieces visualization shows that our decomposition separates different effects hierarchically and is robust to noise; (ii) the low-rank components provide a meaningful source of $W$ and $P$ by grouping certain frequency levels; (iii) the learned low-rank components are strictly positive and diverse. The empirical results also show that our decomposition improves iTransformer and PatchTST by only adjusting the initial projection layer, which has minimal influence on the model architecture but yields a remarkable improvement on forecasting performance. With MLOW serving as an initial interpretable decomposition, it pushes existing TSF models to a new level. Thus, our method provides a new path for interpretable TSF. Our decomposition is also highly efficient for inference and can be readily applied in real-world scenarios. One limitation is that the optimal rank of $V$ is not universally fixed for each dataset. Thus, our future work will explore developments that enable low-rank methods to assign importance scores to the learned components. We also plan to investigate the application of MLOW in other time series tasks, such as anomaly detection and classification.

\section*{Acknowledgements}

This research is partially supported by NSFC (No. 62376153), and ARC Grants DP240102050, DP260104429, LP230201022 and LE240100131.

\section*{Impact Statement}

MLOW represents a new interpretable low-rank frequency magnitude decomposition approach for time series forecasting. By separating principal trending and seasonal effects, MLOW can substantially enhance both interpretability and predictive performance of any TSF backbones with only minimal changes required to their underlying model architectures. It can transform TSF in applications such as energy demand prediction, traffic forecasting, and environmental monitoring, where characterizing distinct time series effects is essential yet challenging. The MLOW interpretability enables a better analysis of temporal patterns for trustworthy modeling of complex behaviors.


\bibliography{example_paper}
\bibliographystyle{icml2026}


\appendix
\onecolumn

\section{MLOW Training and Inference Process}

We provide a more detail training and inference process in Algorithm \ref{alg:freq_decomp}. We use a universal fixed random seed in the released source codes so that you can reproduce exactly the same components showing in the manuscript. The universal parameters for MLOW are the total number of iterations $F=1000$, input sequence length $T=96$, frequency level $K=168$, number of low-rank components $V=10$, and regularization weight $\lambda=20$. The downstream mapping network uses the same learned $\mathbf{H}$ as presented in the manuscript, averaged over two runs for the results in the table.

\section{Low-Rank Decomposition by Semi-NMF}

We introduce another low-rank method called Semi-NMF, which only requires $H$ to be non-negative. Its optimization formula is as follows:
\begin{equation}
\begin{aligned}
\min_{\mathbf{H} \ge 0} & J(\mathbf{W},\mathbf{H}) = \frac{1}{2} \| \mathcal{R} - \mathbf{W} \mathbf{H} \|_F^2, \\
\mathbf{W} &= \mathcal{R} \mathbf{H}^\top (\mathbf{H} \mathbf{H}^\top)^{-1}, \\
\mathbf{H}_{ij} &\leftarrow \mathbf{H}_{ij} \frac{([\mathbf{W}^\top \mathcal{R}]_{ij})_+ + \mathbf{H}_{ij} ([\mathbf{W}^\top \mathbf{W} \mathbf{H}]_{ij})_-}{([\mathbf{W}^\top \mathcal{R}]_{ij})_- + \mathbf{H}_{ij} ([\mathbf{W}^\top \mathbf{W} \mathbf{H}]_{ij})_+}, \\
\mathbf{A}_+ &= \max(\mathbf{A},0), \quad \mathbf{A}_- = \max(-\mathbf{A},0).
\end{aligned}
\end{equation}

The only strength of this method is that it is relatively efficient for new data, as $\mathbf{W}$ can be directly computed through the least square estimator. However, its weakness is similar to PCA, that negative coefficients compromise the phase-aware temporal information. In addition, although $\mathbf{W}$ can be directly computed using least squares. The optimization would, in principle, be simpler, it fails to converge properly since the gradient does not correctly direct to the objective. This is because $(\mathbf{H} \mathbf{H}^\top)^{-1}$ is unstable and can change quite dramatically, which leads to the optimization for $\mathbf{H}$ being unstable. The gradient is computed with respect to $W$ at the current iteration, rather than the true gradient of the objective.

\begin{algorithm}[t]
\caption{MLOW: Low-Rank Frequency Magnitude Decomposition}
\label{alg:freq_decomp}
\begin{algorithmic}
\REQUIRE Valid time series sample $\mathcal{R} \in \mathbb{R}^{N \times K}$ extracted by $\mathcal{X} \in \mathbb{R}^{N \times 2K}$, frequency levels $K$, low-rank representation $V$, the number of iterations $F$, the length of training set $I$, the number of multivariate $D$, regularized factor $\lambda$,  input sequence length $T$.
\ENSURE The total valid number of frequency spectrum samples $N = (I - 2K)\times D$, covering all valid amplitude extraction instances in the training dataset. This is achieved by a window $2K$ sliding over the training data.

\STATE \textbf{Training Phase:}
\STATE Compute frequency spectrum $\mathcal{R}$ by Eq. (1) in the manuscript for all valid samples.    
\STATE Initialize low-rank factors $\mathbf{H} \in \mathcal{R}^{V \times K}$ with top $v$ eigenvectors of the PCA, but the negative values are replaced by random positive values. This is because our hyperplane projection is the same as PCA, but we don't allow negative values. 
\STATE Force the corresponding coefficient $\mathbf{W}=\mathcal{R} \mathbf{H}^T$.
\FOR{$i = 1$ to $F$}
    \STATE Update $H$ by Eq. (9) in the manuscript.
\ENDFOR 
\STATE Obtained leaned low-rank components $\mathbf{H}$. 
\STATE \textbf{Inference Phase:}
\STATE This is required when $2K$ is larger than $T$. For an input time series $\mathbf{X}^{i}$ with time horizon T, we extend the input horizon to $2K$ to extract the magnitude spectrum by $\mathbf{X}^{i}_{w} = [\mathbf{X}^{i}_{\text{e}}, \mathbf{X}^{i}] \in \mathbb{R}^{2K}$, then we express it as  $\mathbf{X}^{i}_{w} = \mathbf{R}^{i}\mathbf{B}^{i}_{w}$, while only the last $T$ timestamps in $\mathbf{B}^{i}_{w}$ are used as the effective temporal information. Thus, we only take $\mathbf{B}^{i}$ of $\mathbf{B}^{i}_{w}=[\mathbf{B}^{i}_{e}, \mathbf{B}^{i}]$ as the phase-aware bases we need. All the windows here contain strictly past information and do not include any forecast horizon. 
\STATE  Obtain the  $\mathbf{X}^{i}= \mathbf{R}^{i}  \mathbf{B}^{i} +\mathbf{X}^{i}_m$ by Eq. (10) in the manuscript.
\STATE Compute the new coefficient $\mathbf{W}^{i}=\mathbf{R}^{i}\mathbf{H}^T$.
\STATE Compute the new reconstructed basis $\mathbf{P}^{i}=\mathbf{H}\mathbf{B}^{i} $.
\STATE Obtain the low-rank decomposition $\mathbf{Z}^{i}=\mathbf{W}^{i} \odot \mathbf{P}^{i}$. 
\STATE Obtain the residual $\mathbf{X}^{i}_r= \mathbf{X}^{i}-  \mathbf{W}^{i}\mathbf{P}^{i}-\mathbf{X}^{i}_m $.
\STATE Do the same for timestamps decomposition if timestamps are required. This is because, in the initial iTransformer setting, the timestamps share the same initial projection weights.
\end{algorithmic}
\end{algorithm}

\section{Tables}

The full results for Tables \ref{table_main}, \ref{others}, \ref{abalation2} and \ref{abalation4} are provided in Tables \ref{table_full}, \ref{others_full}, \ref{ab2} and \ref{ab4}, respectively. Table \ref{data_detail} provides more details for data description. The training details for iTransformer and PatchTST are provided in Tables \ref{detail1} and \ref{detail2}, and the hyperparameters for MLOW have already been mentioned in the manuscript. 

\section{Visualization}

\begin{figure*}[t]
\centering
\begin{minipage}[t]{0.48\linewidth}
\centering
\includegraphics[width=0.9\textwidth]{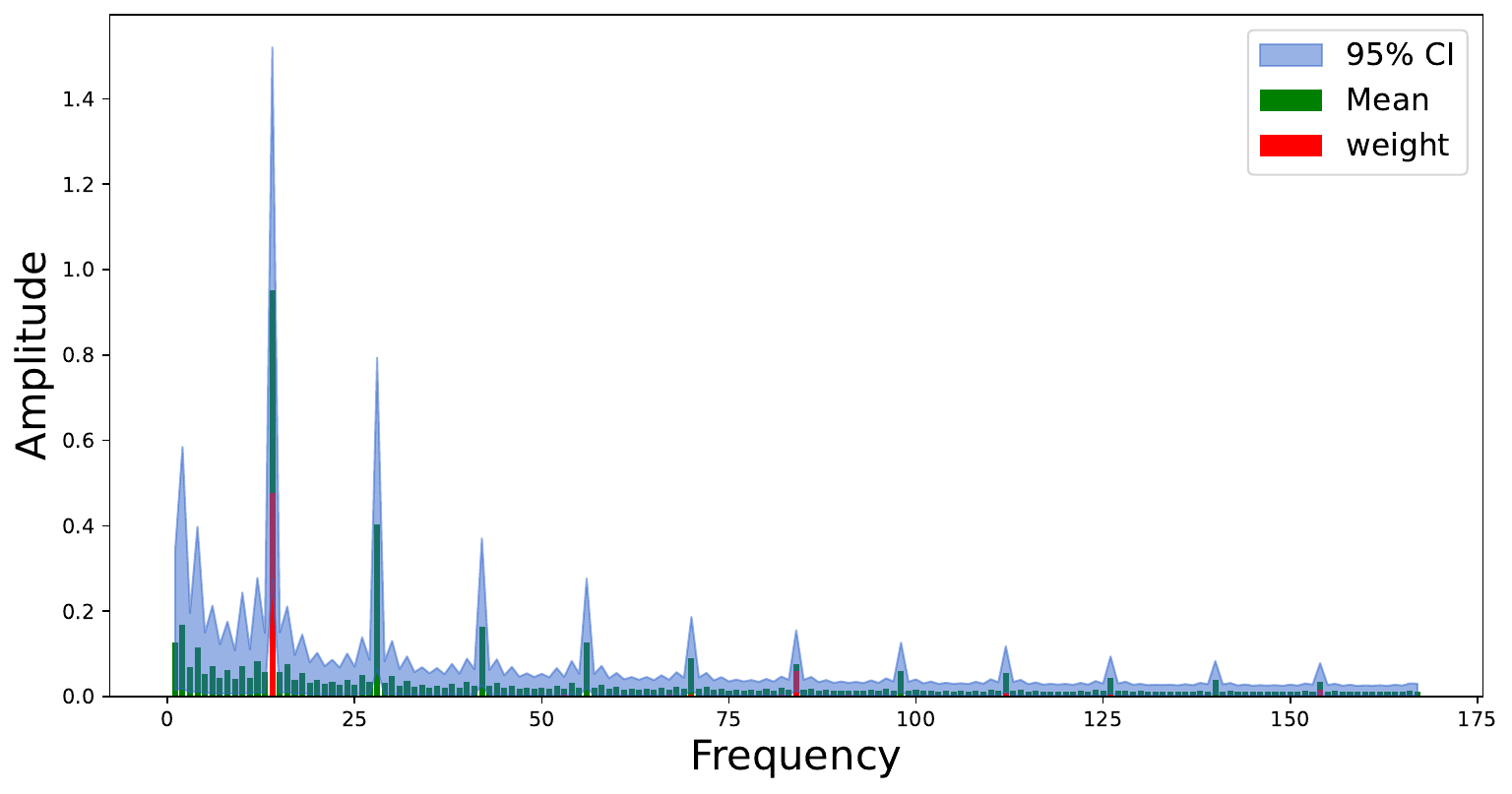}
\end{minipage}%
\begin{minipage}[t]{0.48\linewidth}
\centering
\includegraphics[width=0.9\textwidth]{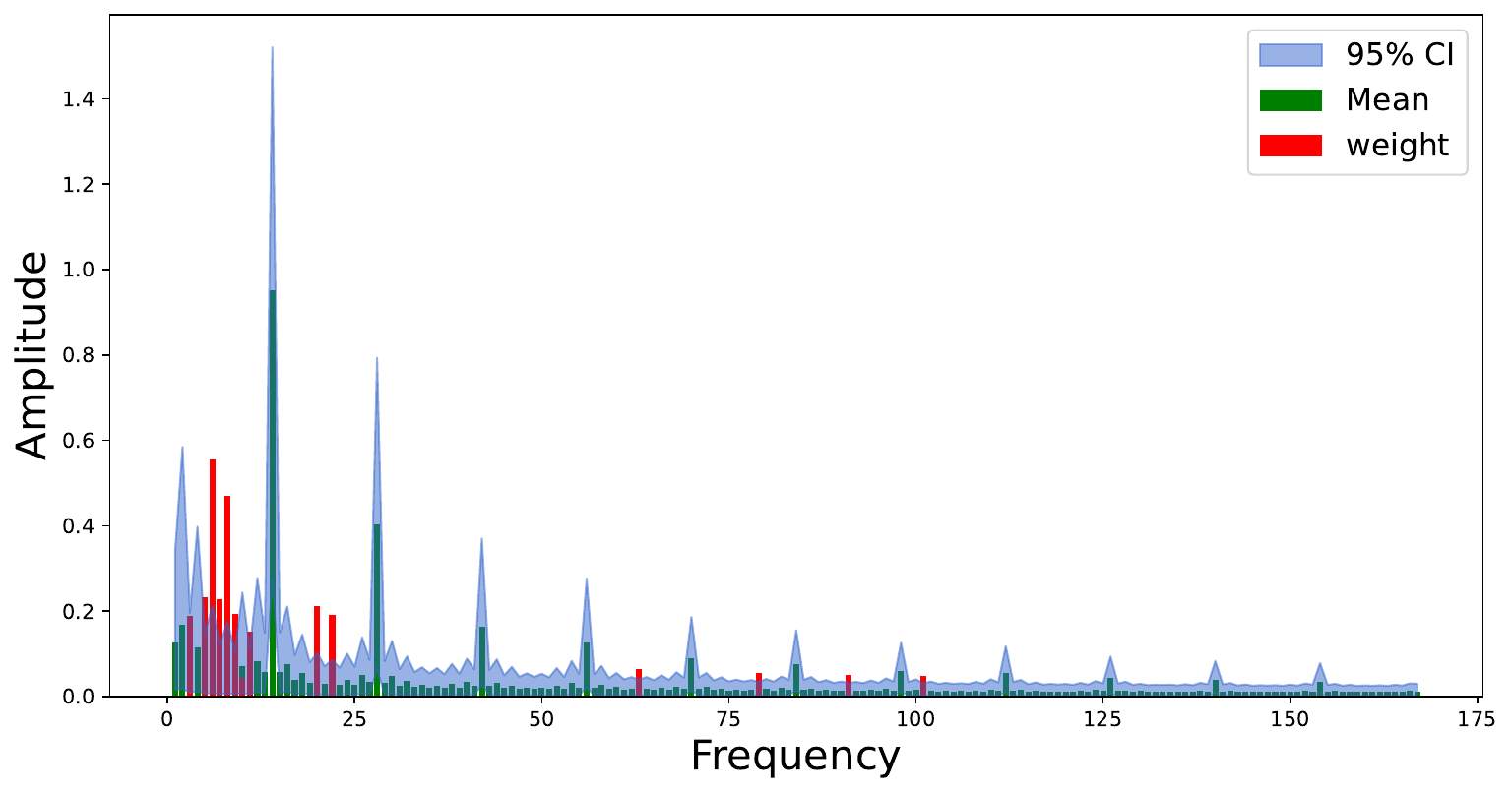}
\end{minipage}%
\vspace{0.5em}
\begin{minipage}[t]{0.48\linewidth}
\centering
\includegraphics[width=0.9\textwidth]{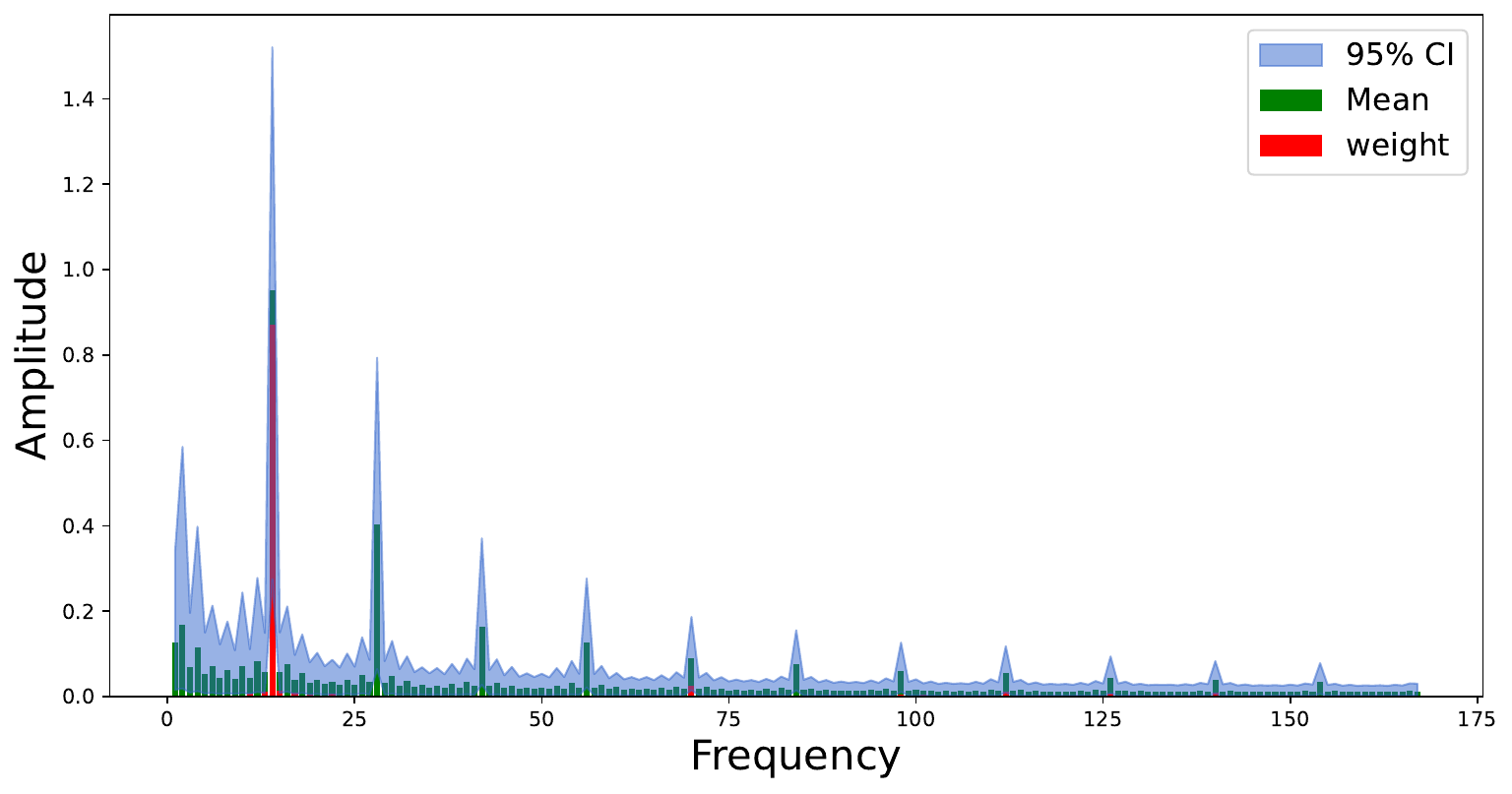}
\end{minipage}
\begin{minipage}[t]{0.48\linewidth}
\centering
\includegraphics[width=0.9\textwidth]{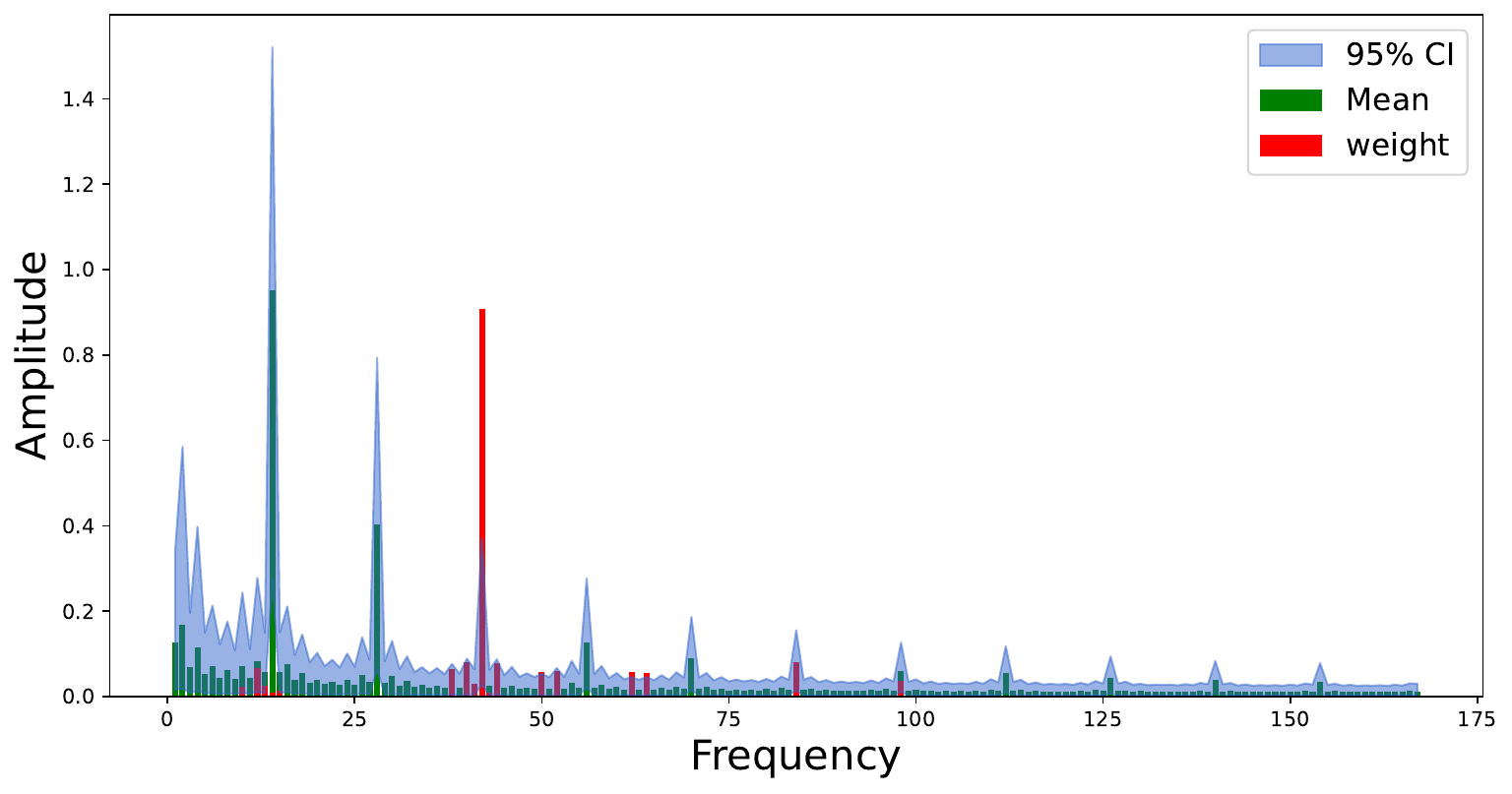}
\end{minipage}
\vspace{0.5em}
\begin{minipage}[t]{0.48\linewidth}
\centering
\includegraphics[width=0.9\textwidth]{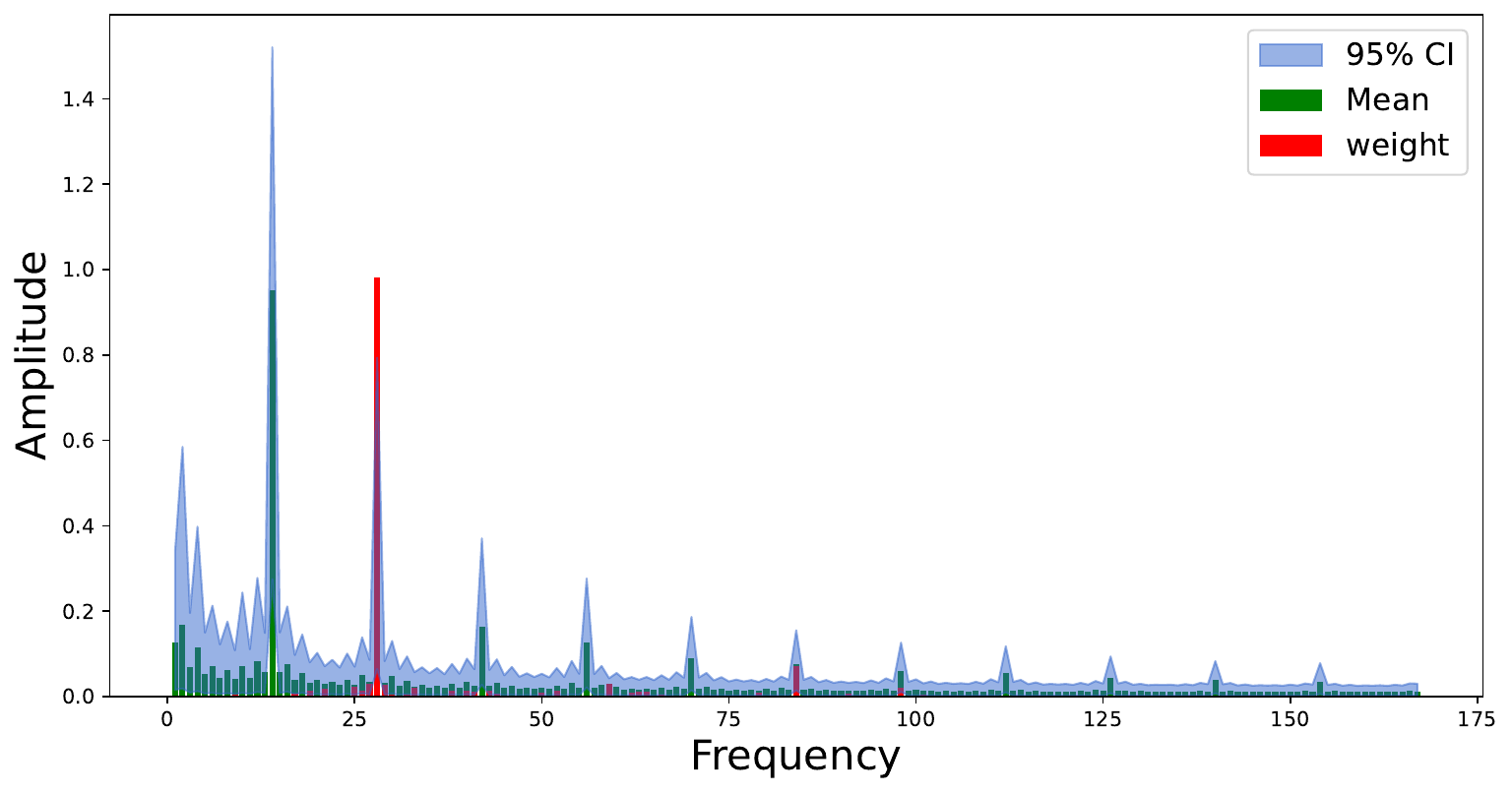}
\end{minipage}
\begin{minipage}[t]{0.48\linewidth}
\centering
\includegraphics[width=0.9\textwidth]{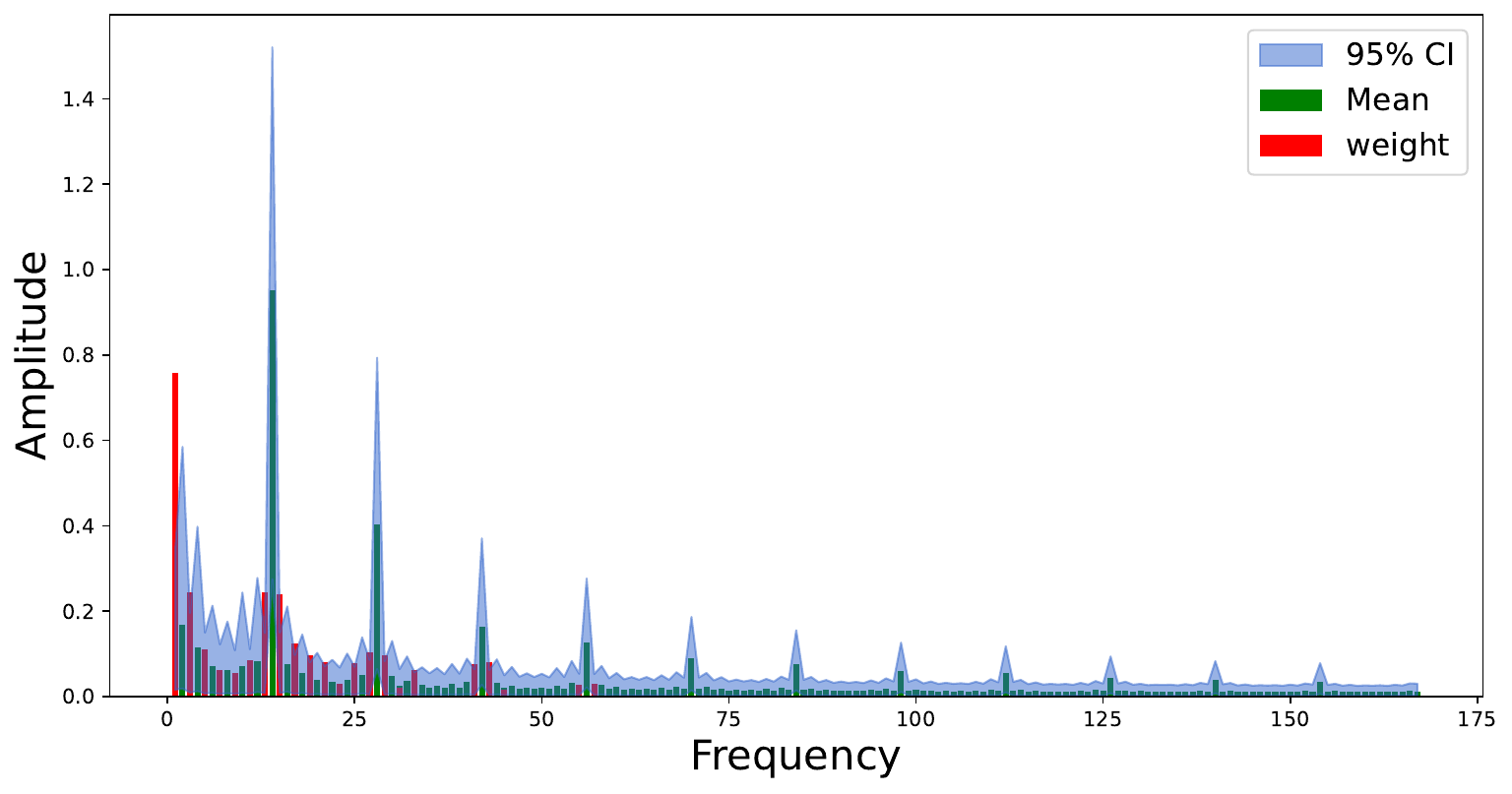}
\end{minipage}%
\vspace{0.5em}
\begin{minipage}[t]{0.48\linewidth}
\centering
\includegraphics[width=0.9\textwidth]{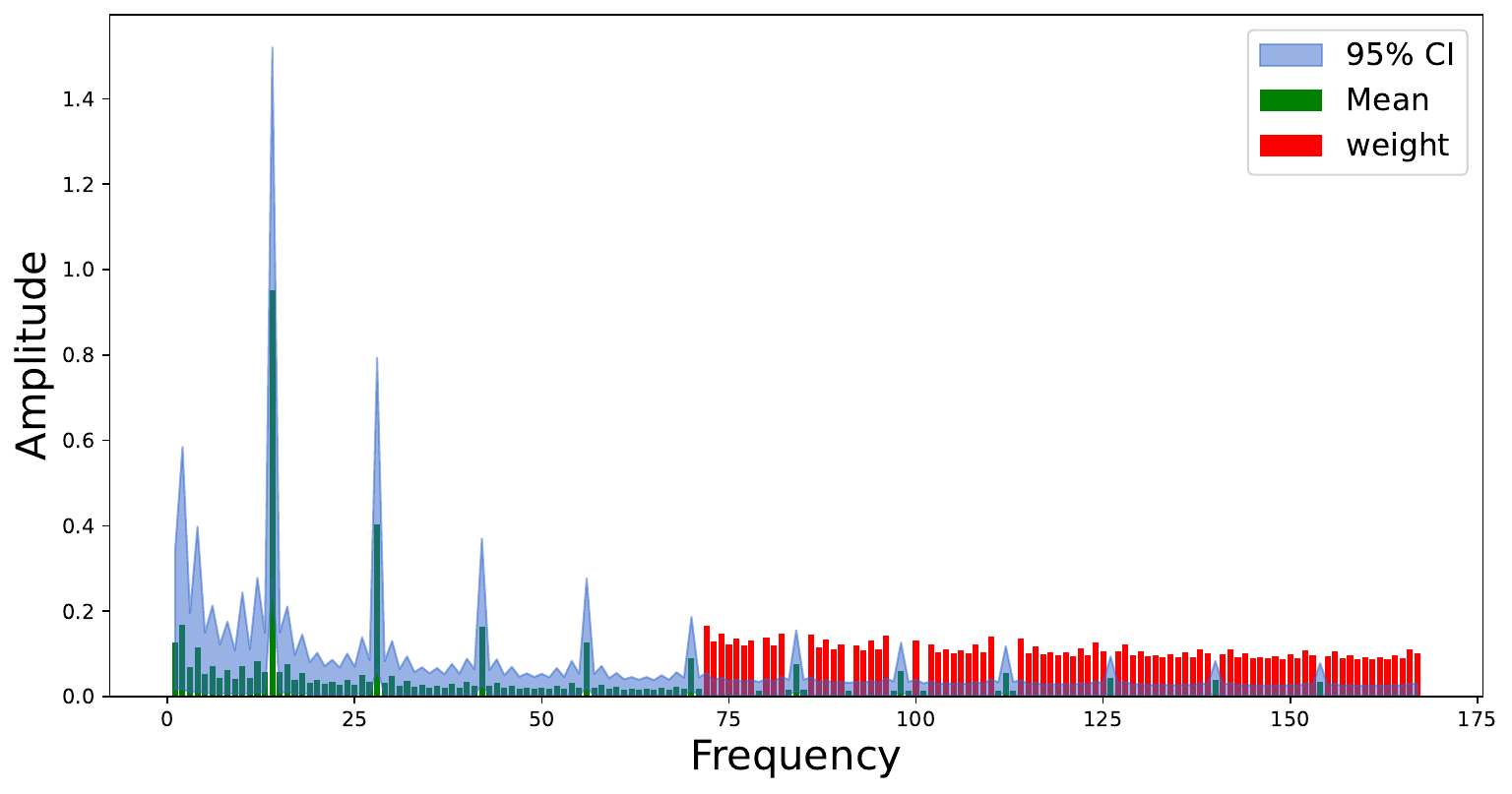}
\end{minipage}%
\begin{minipage}[t]{0.48\linewidth}
\centering
\includegraphics[width=0.9\textwidth]{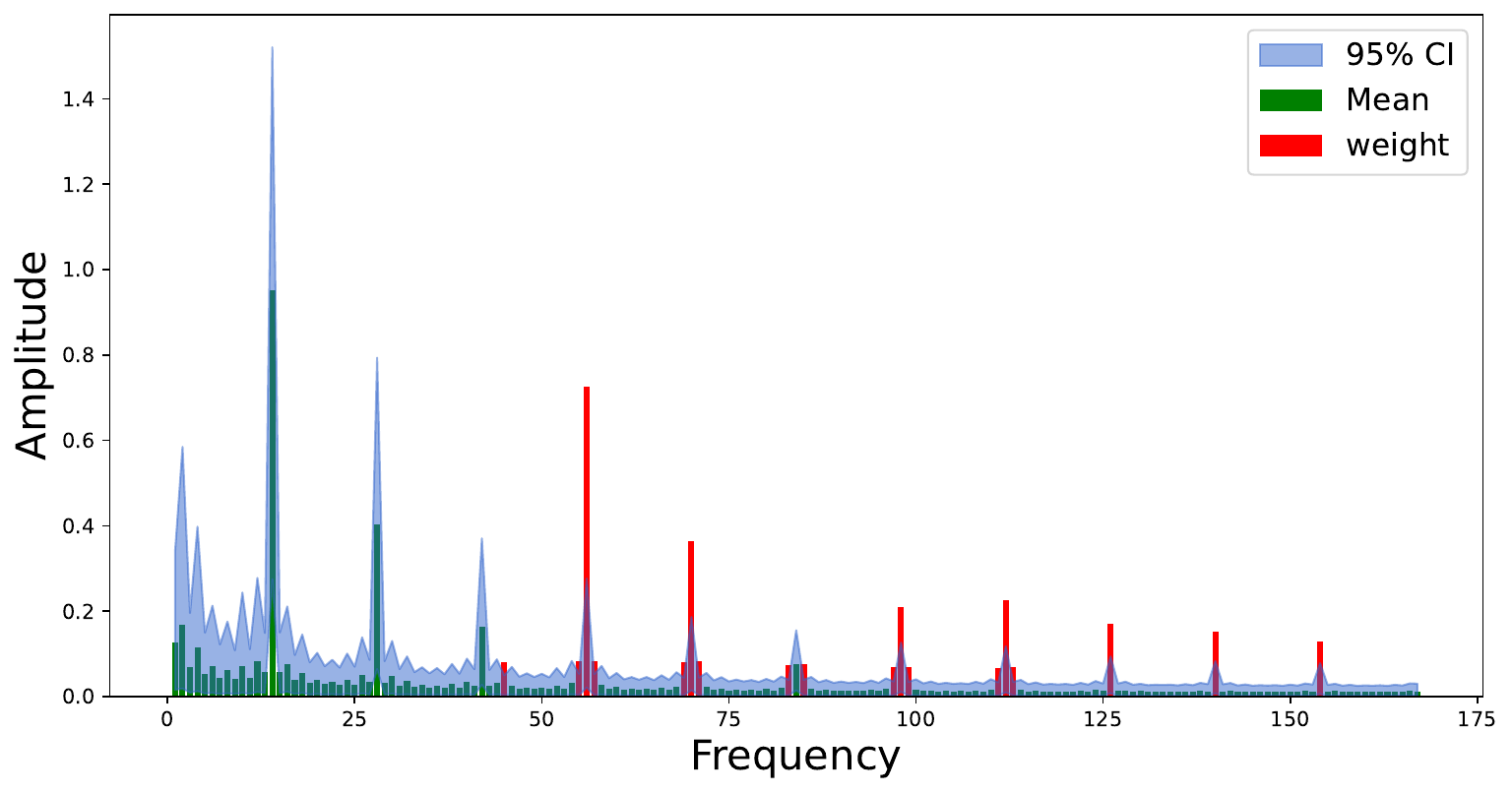}
\end{minipage}
\vspace{0.5em}
\begin{minipage}[t]{0.48\linewidth}
\centering
\includegraphics[width=0.9\textwidth]{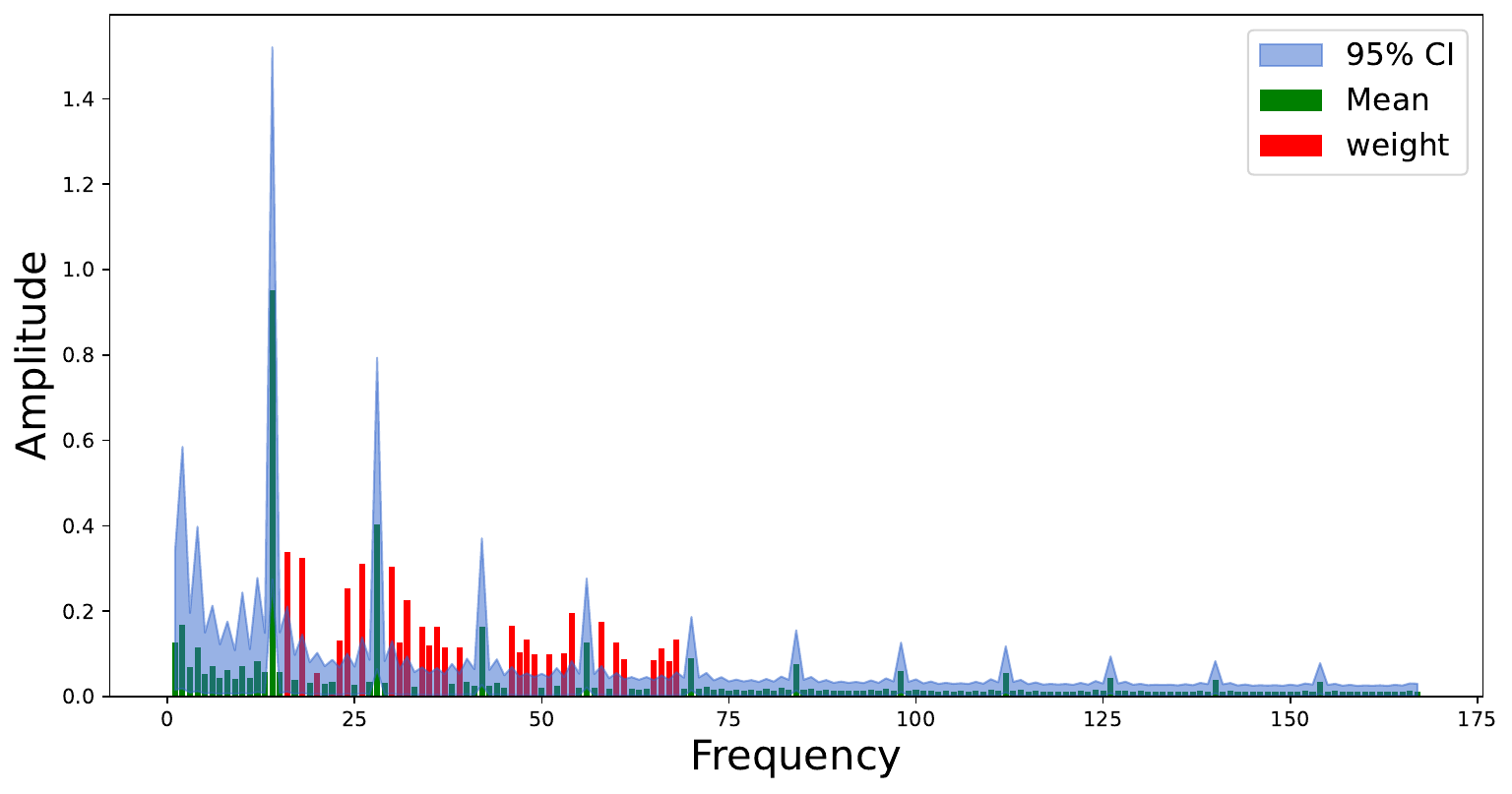}
\end{minipage}
\begin{minipage}[t]{0.48\linewidth}
\centering
\includegraphics[width=0.9\textwidth]{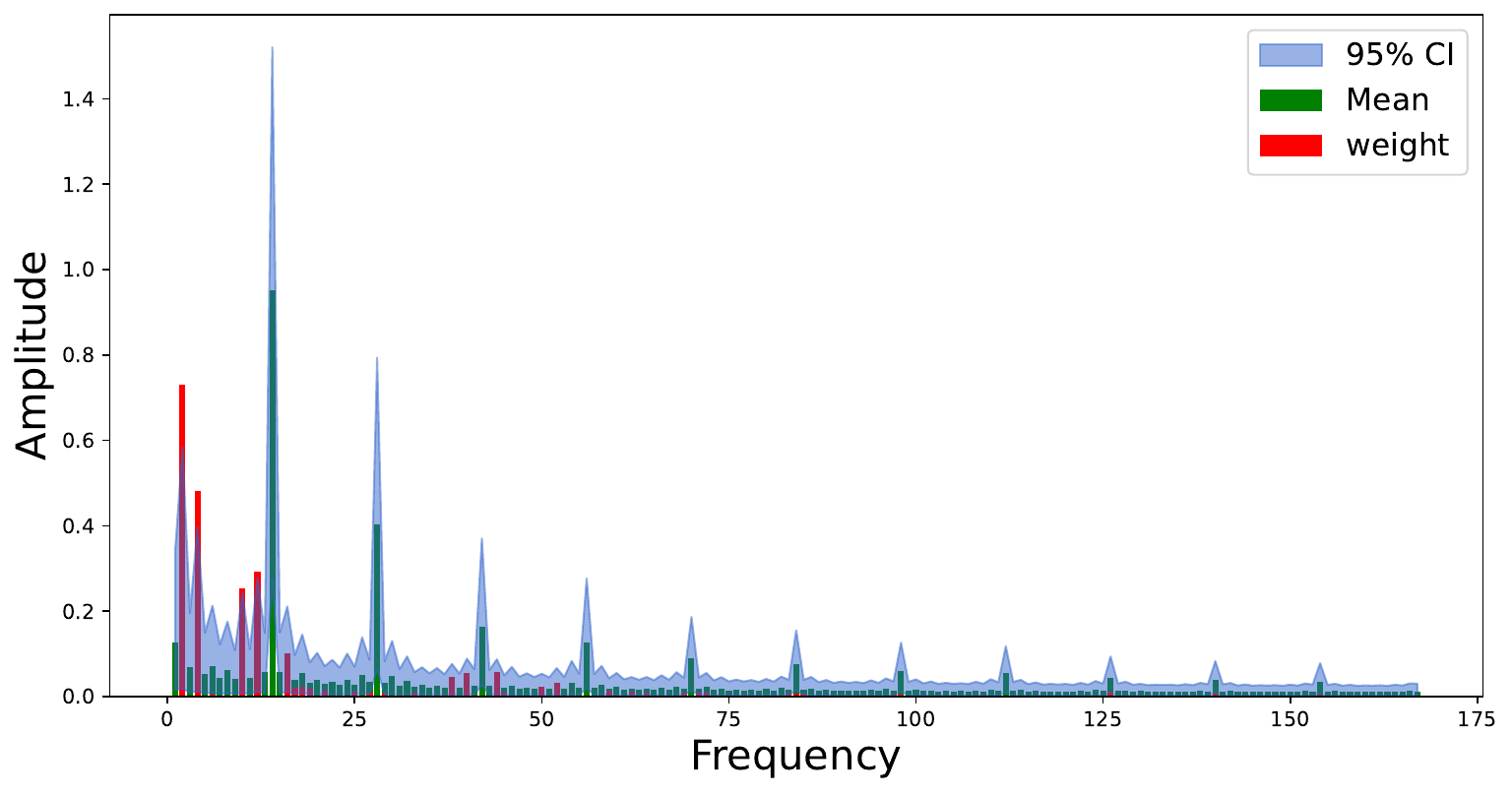}
\end{minipage}
\caption{Learned 10 Components for the ECL data Compared to Its Magnitude Spectrum Distribution. The blue regions correspond to the 95\% confidence interval of magnitude spectrum, the green columns show the mean magnitude spectrum, and the red columns indicate the weights of the learned components.}
\label{components_ECL}
\end{figure*}

\begin{figure*}[t]
\centering
\begin{minipage}[t]{0.48\linewidth}
\centering
\includegraphics[width=0.9\textwidth]{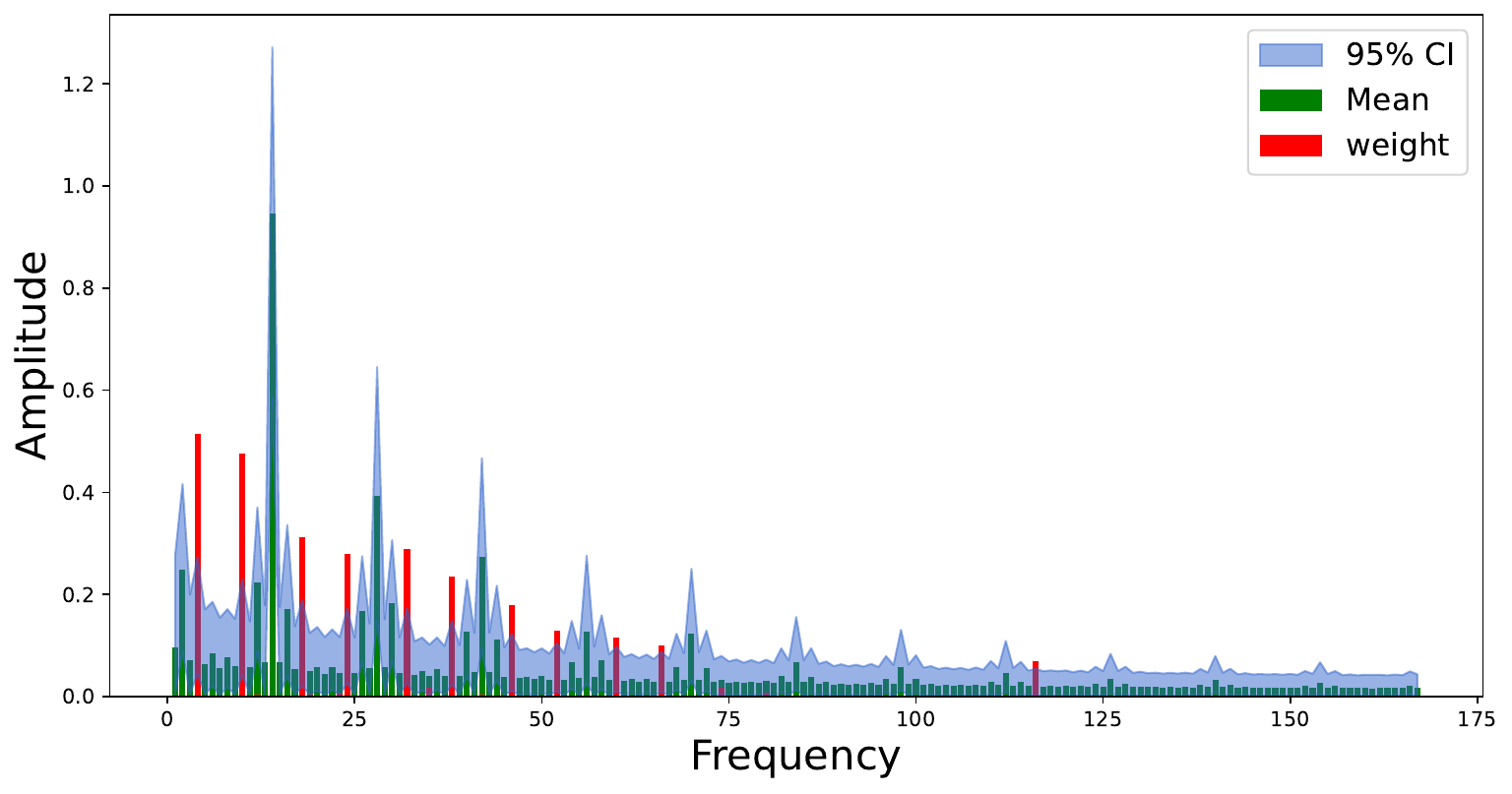}
\end{minipage}%
\begin{minipage}[t]{0.48\linewidth}
\centering
\includegraphics[width=0.9\textwidth]{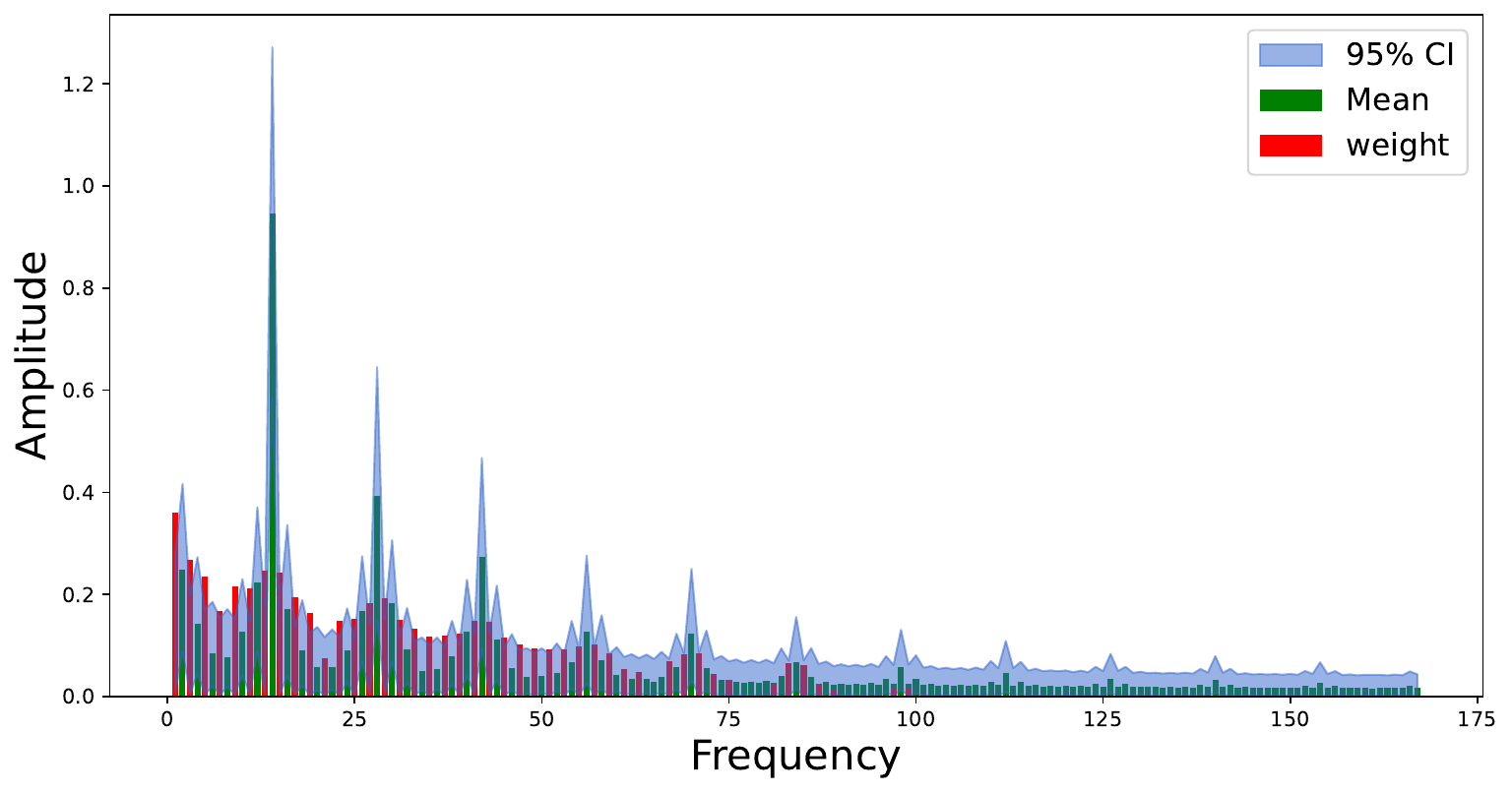}
\end{minipage}%
\vspace{0.5em}
\begin{minipage}[t]{0.48\linewidth}
\centering
\includegraphics[width=0.9\textwidth]{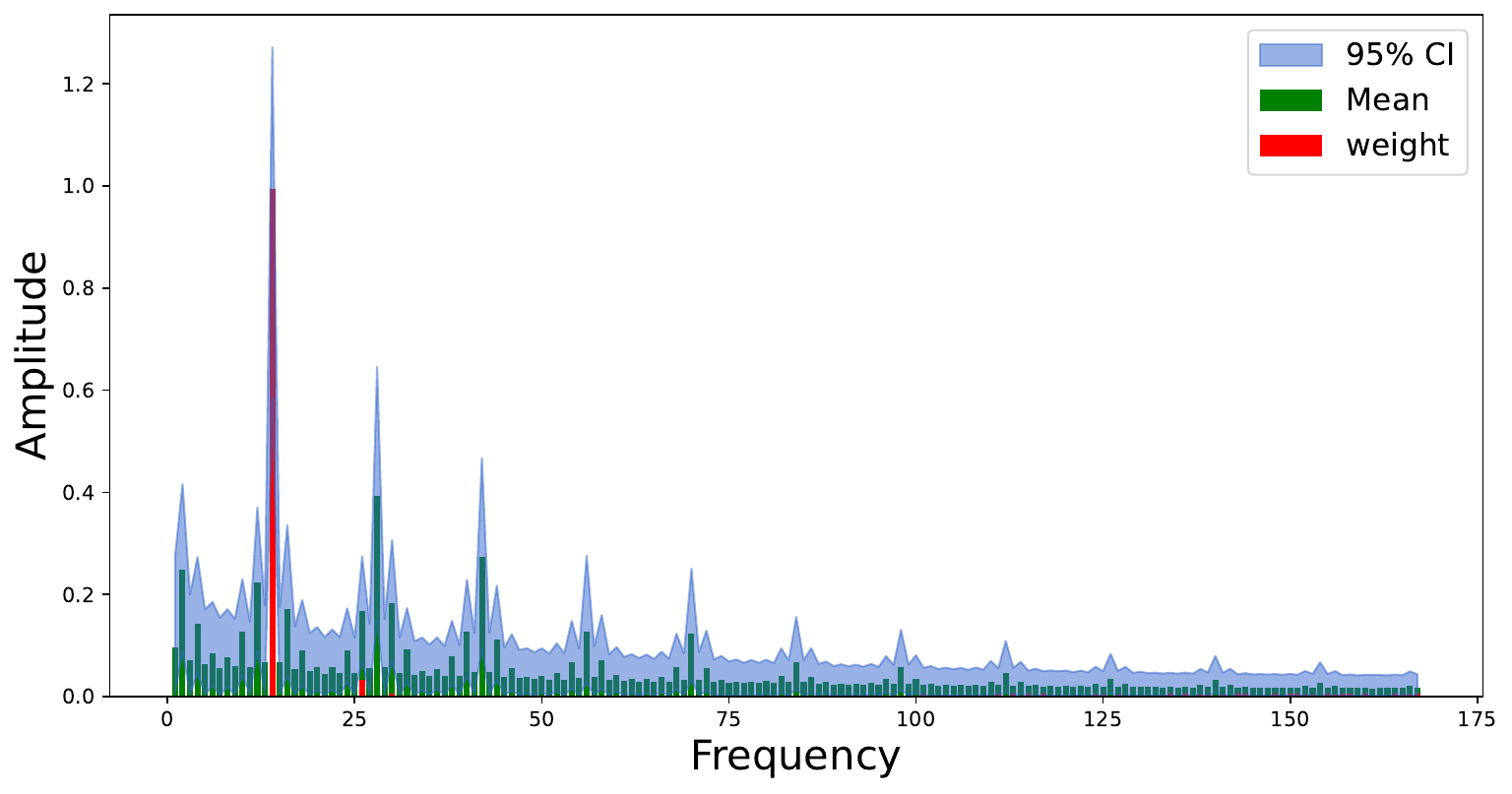}
\end{minipage}
\begin{minipage}[t]{0.48\linewidth}
\centering
\includegraphics[width=0.9\textwidth]{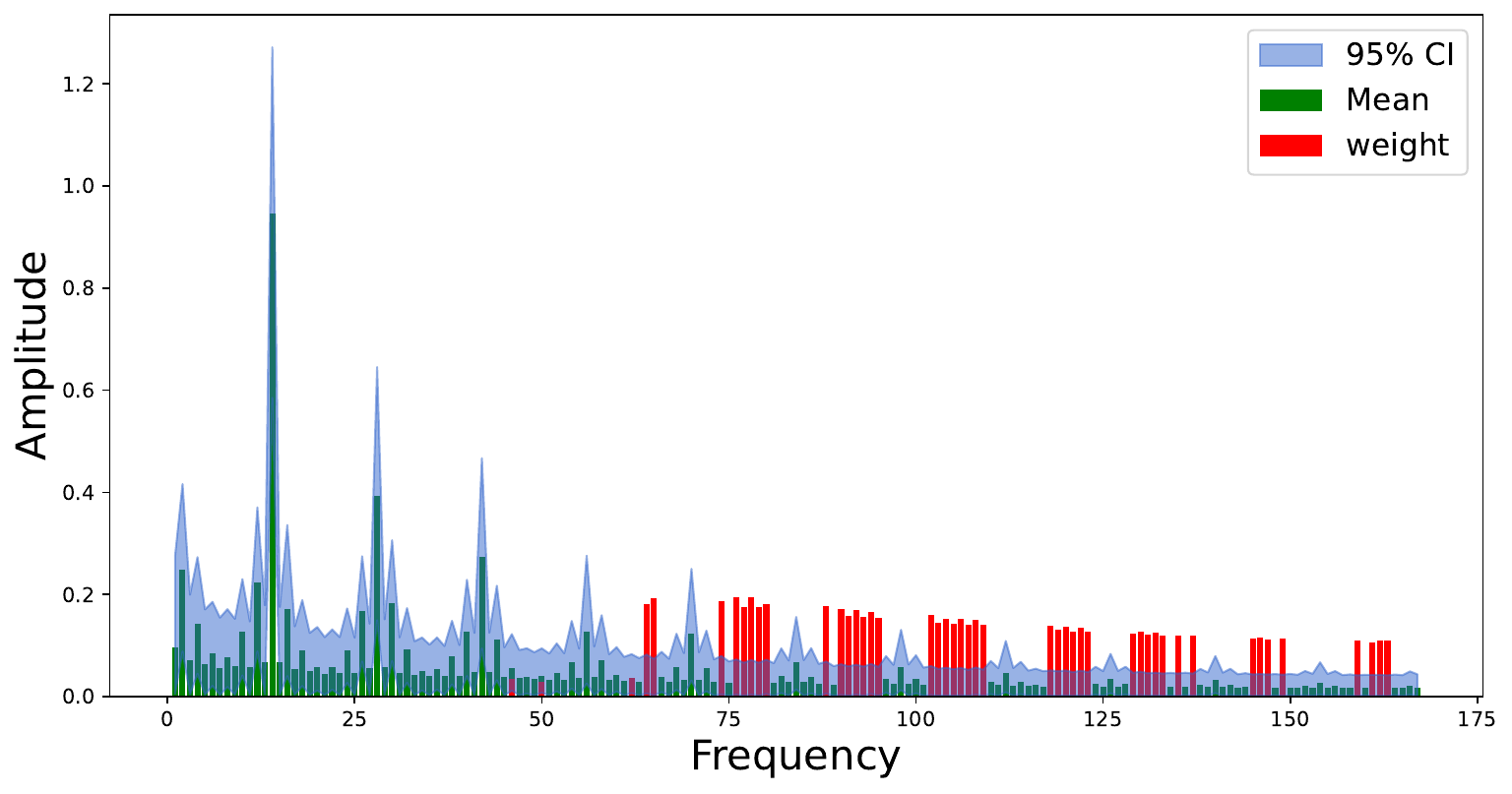}
\end{minipage}
\vspace{0.5em}
\begin{minipage}[t]{0.48\linewidth}
\centering
\includegraphics[width=0.9\textwidth]{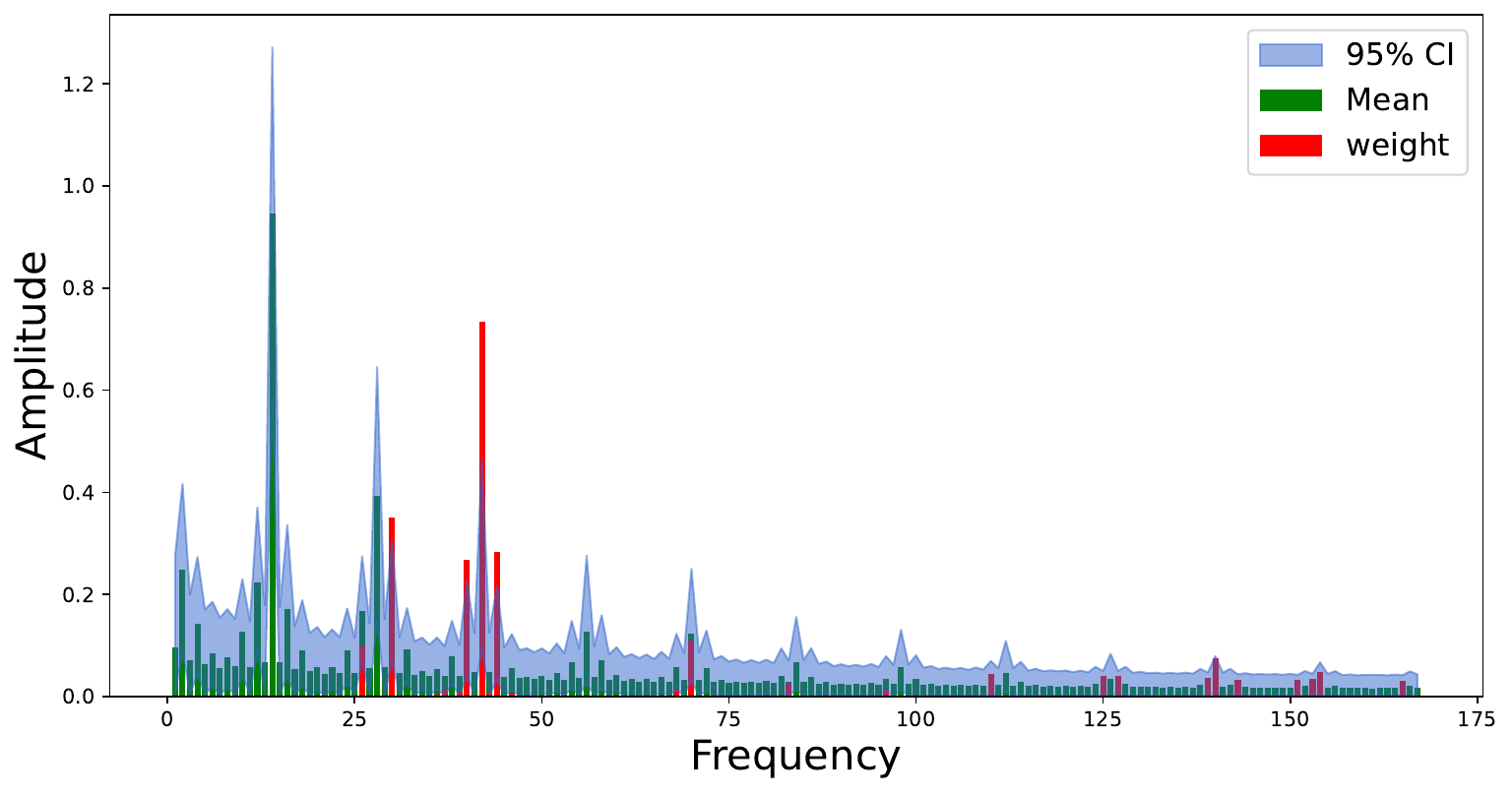}
\end{minipage}
\begin{minipage}[t]{0.48\linewidth}
\centering
\includegraphics[width=0.9\textwidth]{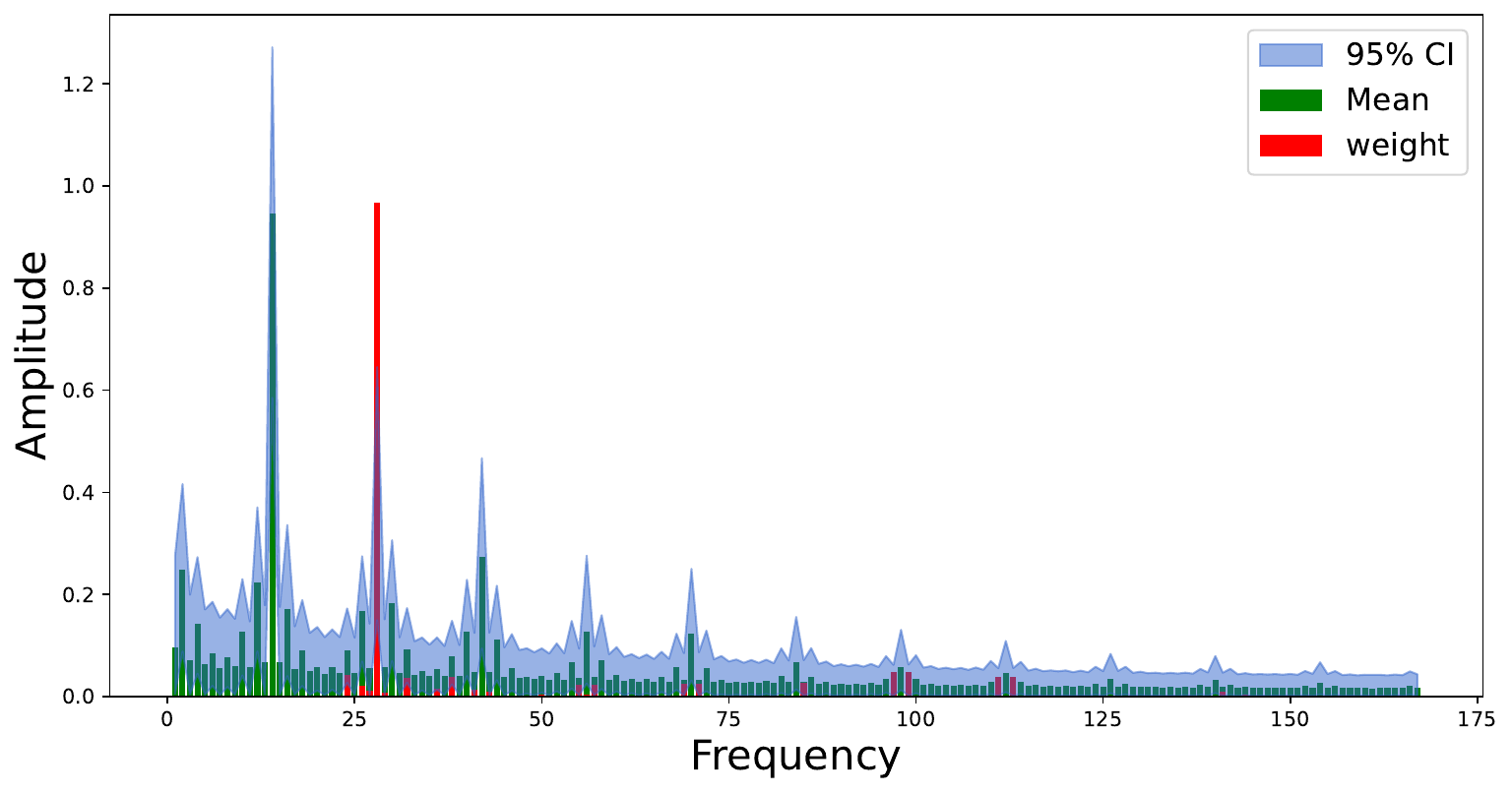}
\end{minipage}%
\vspace{0.5em}
\begin{minipage}[t]{0.48\linewidth}
\centering
\includegraphics[width=0.9\textwidth]{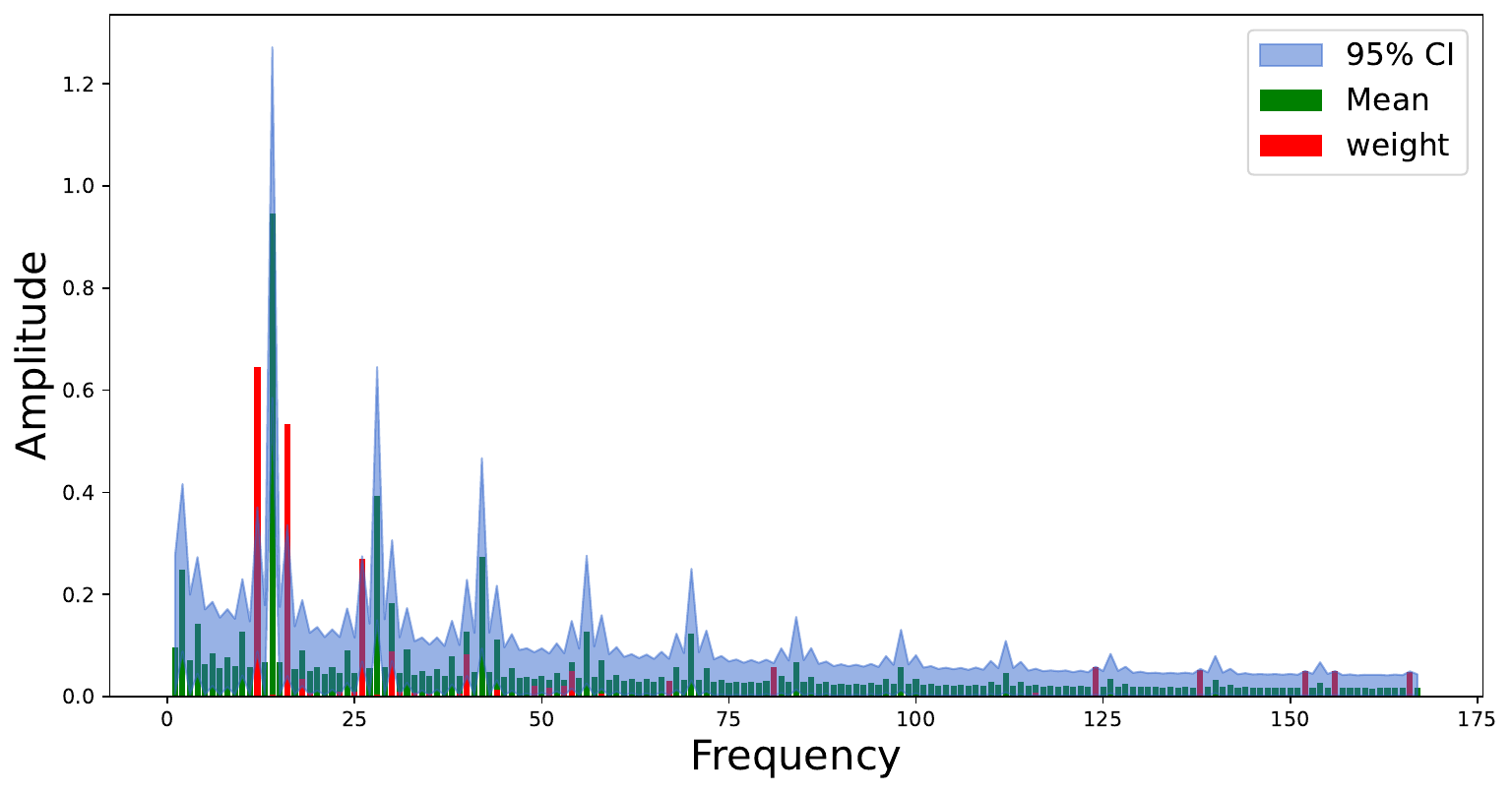}
\end{minipage}%
\begin{minipage}[t]{0.48\linewidth}
\centering
\includegraphics[width=0.9\textwidth]{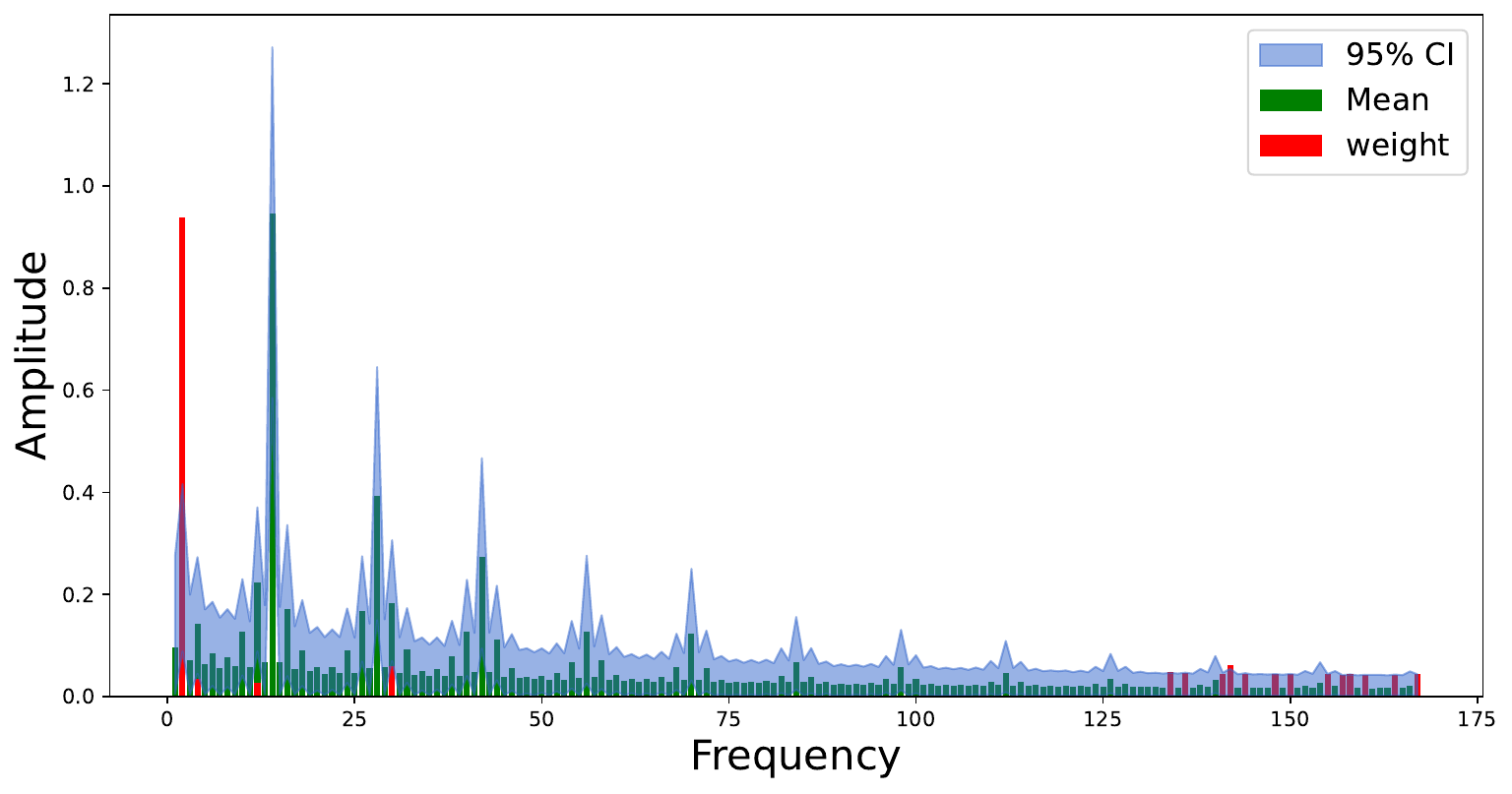}
\end{minipage}
\vspace{0.5em}
\begin{minipage}[t]{0.48\linewidth}
\centering
\includegraphics[width=0.9\textwidth]{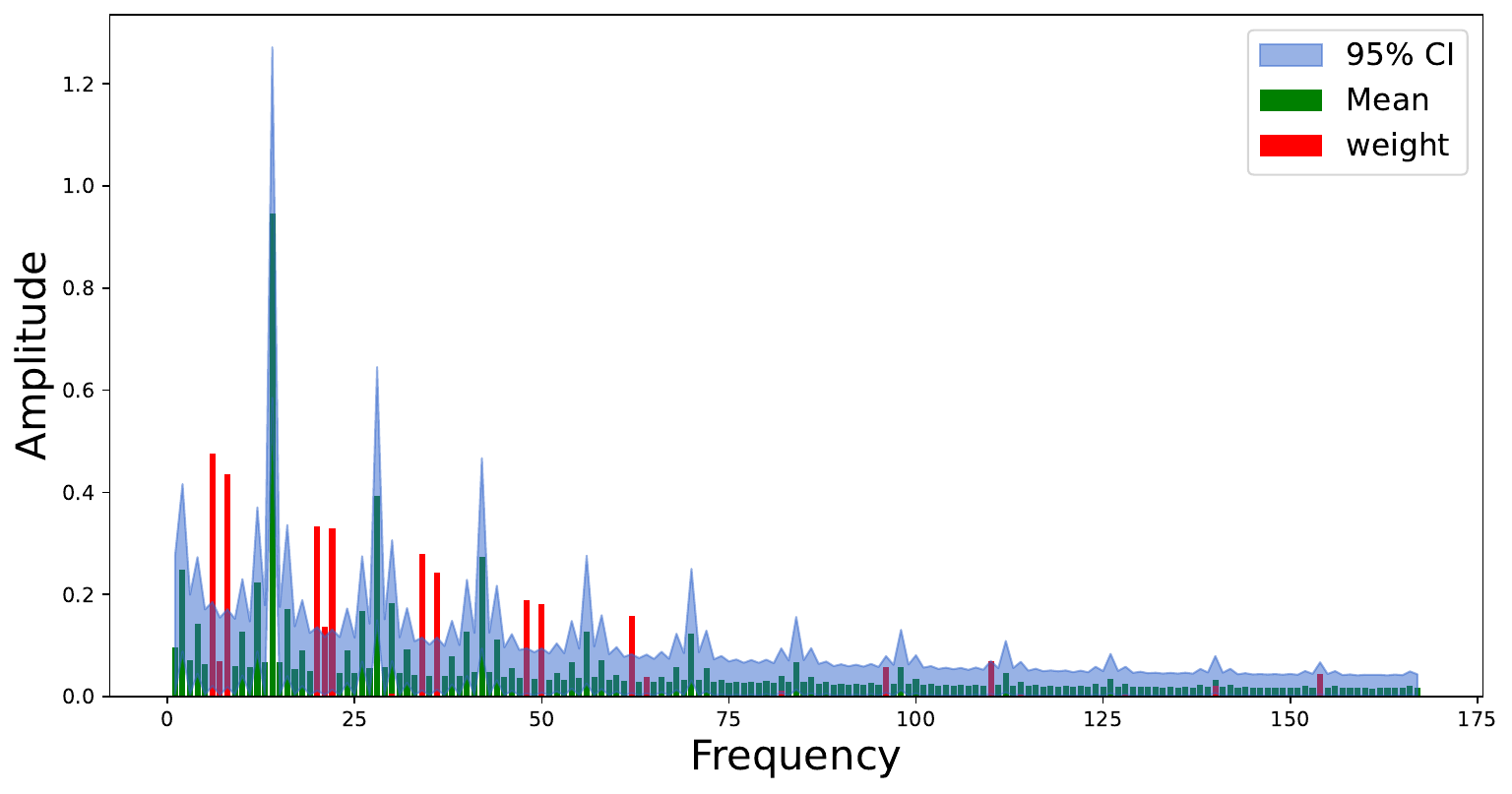}
\end{minipage}
\begin{minipage}[t]{0.48\linewidth}
\centering
\includegraphics[width=0.9\textwidth]{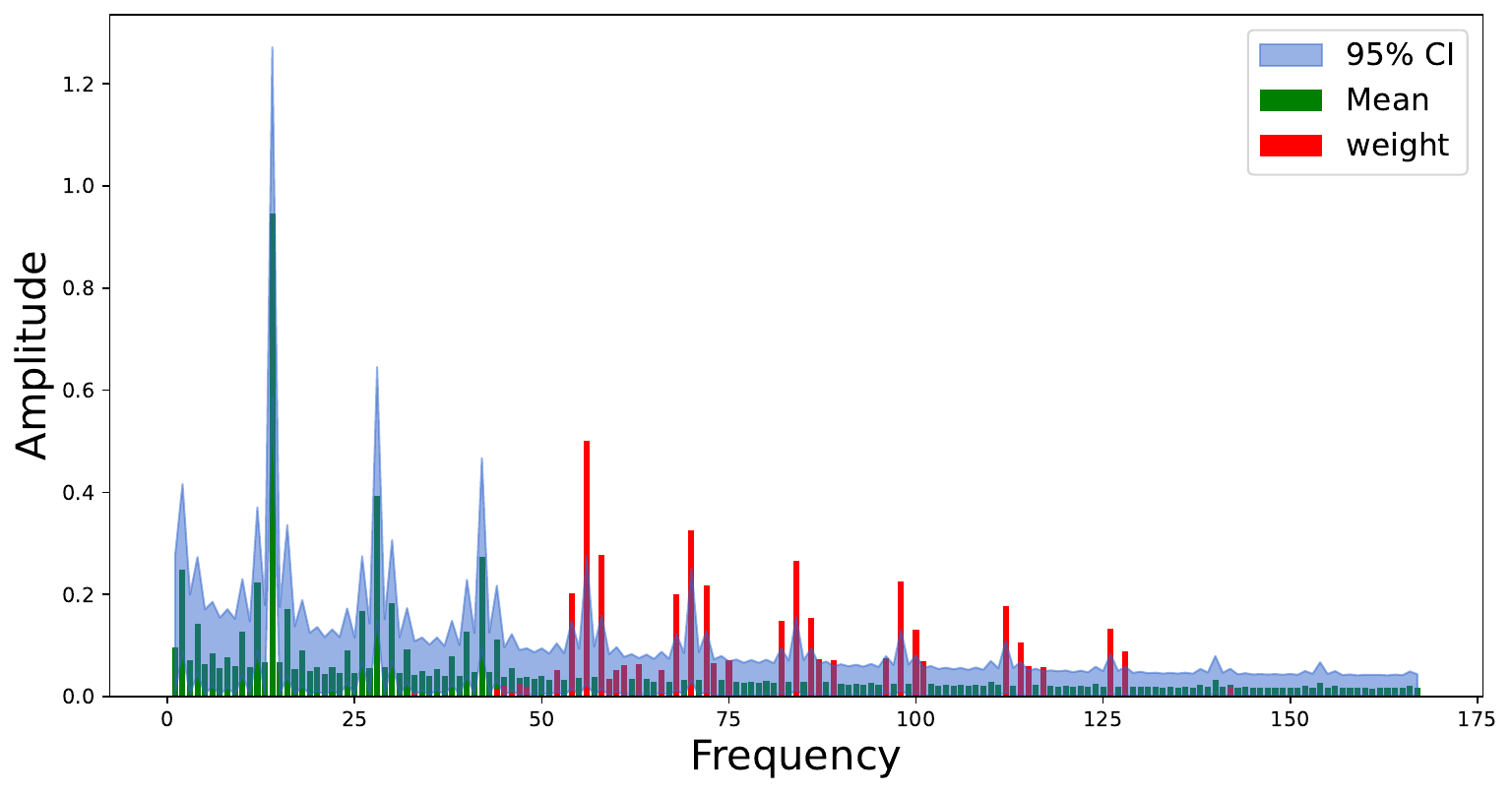}
\end{minipage}
\caption{Learned 10 Components for the Traffic data Compared to Its Magnitude Spectrum Distribution. The blue regions correspond to the 95\% confidence interval of magnitude spectrum, the green columns show the mean magnitude spectrum, and the red columns indicate the weights of the learned components.}
\label{components_traffic}
\end{figure*}

\begin{figure*}[h]
\centering
\begin{minipage}[t]{0.48\linewidth}
\centering
\includegraphics[width=0.9\textwidth]{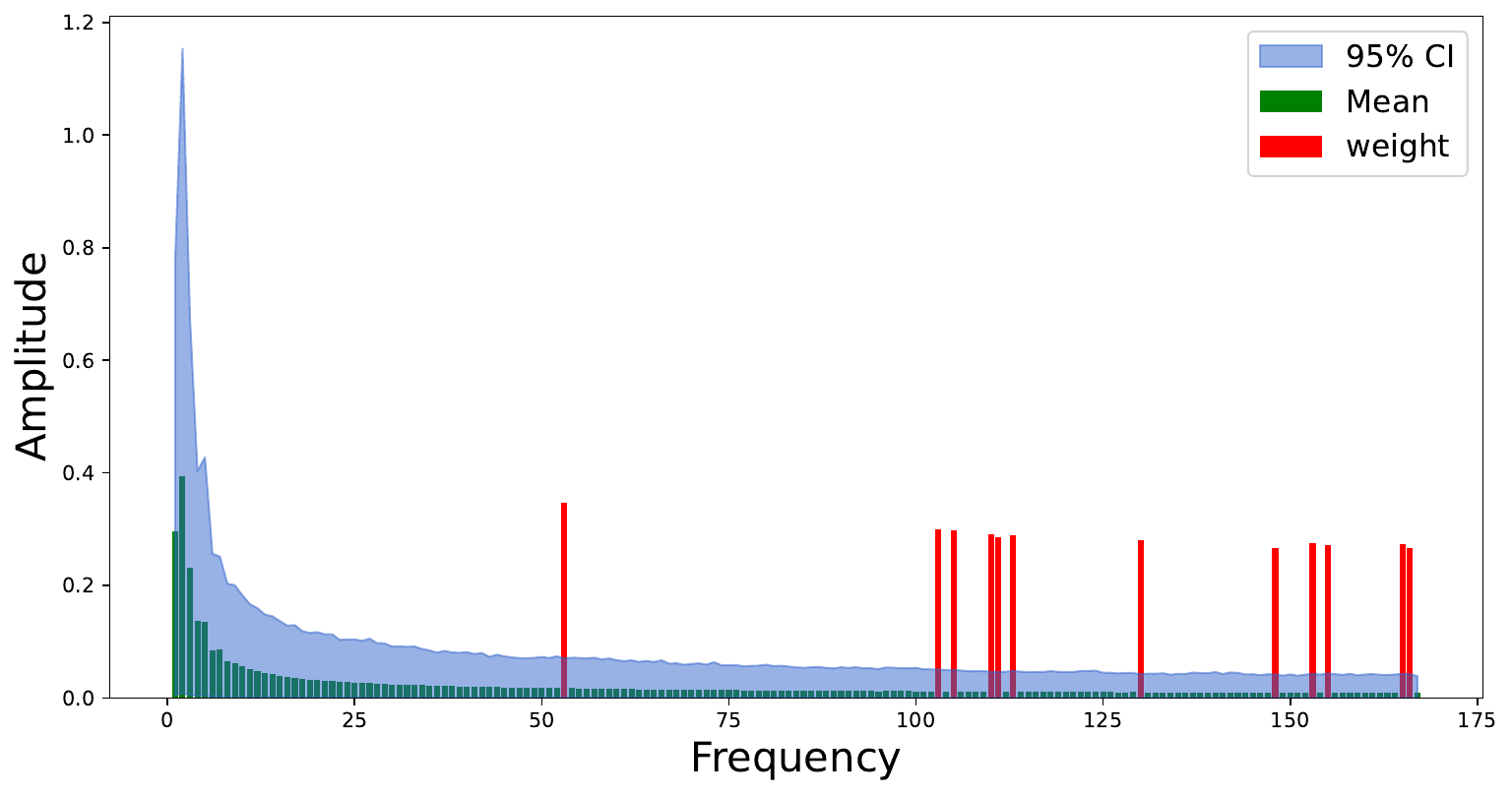}
\end{minipage}%
\begin{minipage}[t]{0.48\linewidth}
\centering
\includegraphics[width=0.9\textwidth]{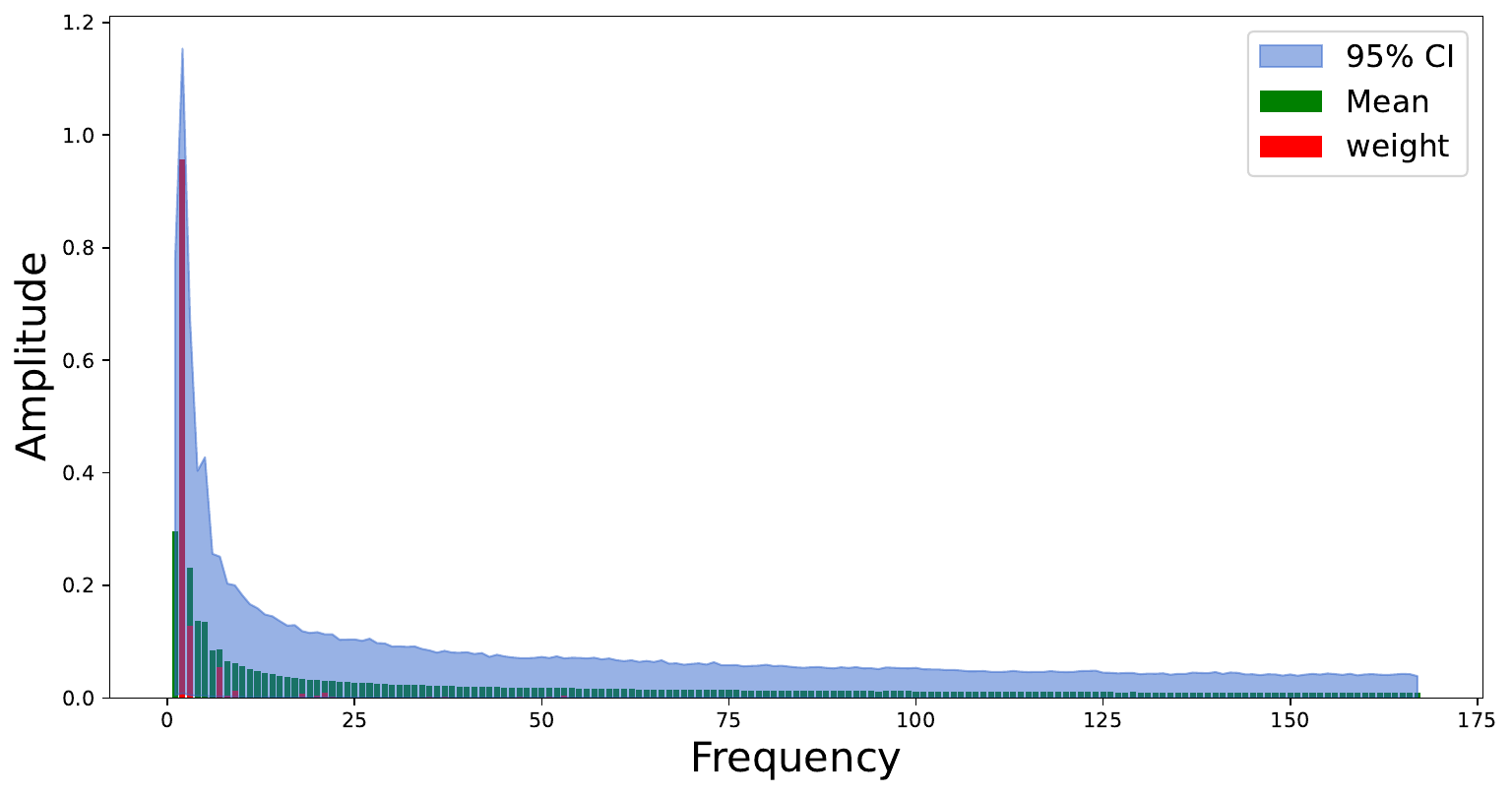}
\end{minipage}%
\vspace{0.5em}
\begin{minipage}[t]{0.48\linewidth}
\centering
\includegraphics[width=0.9\textwidth]{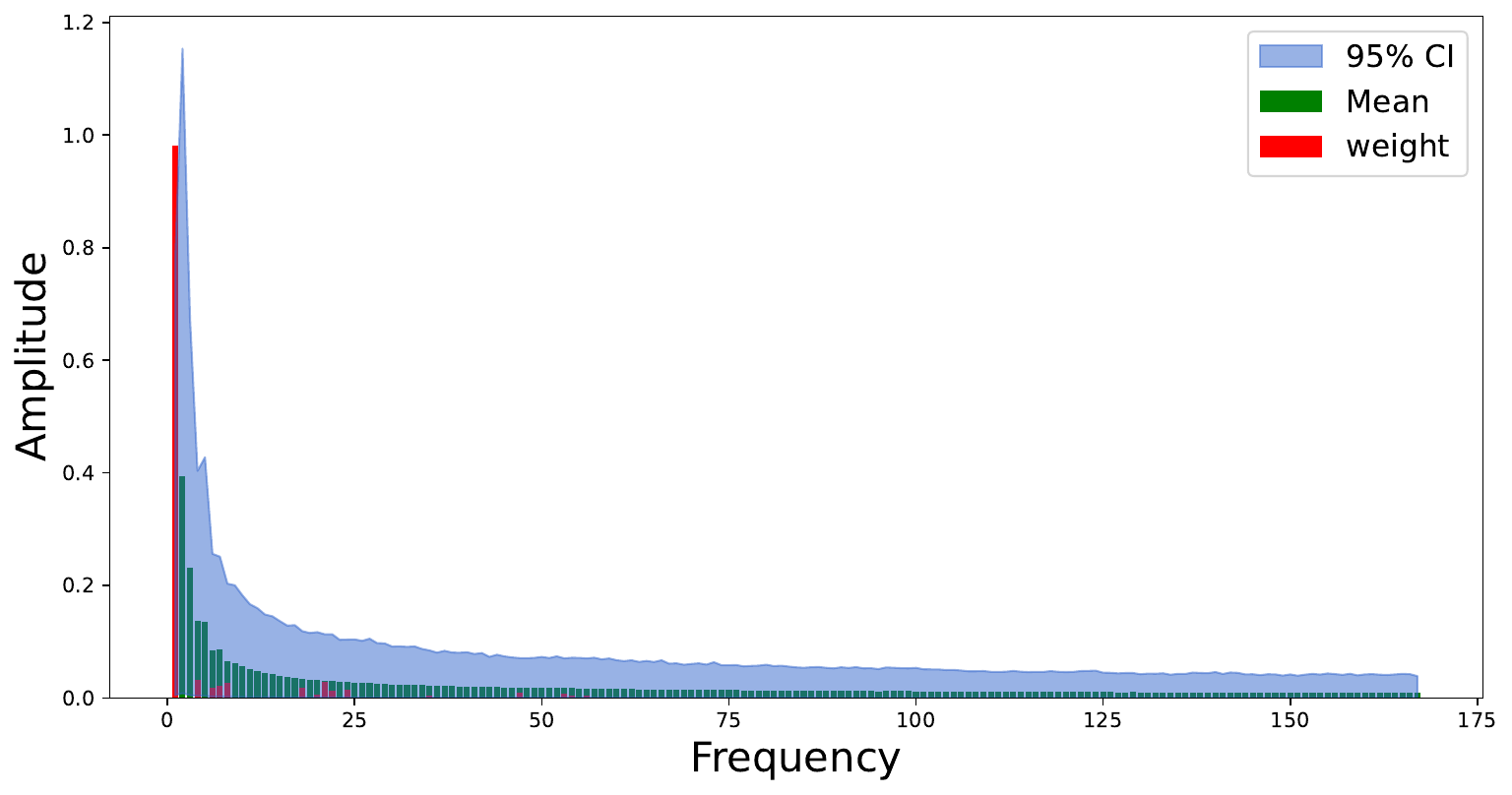}
\end{minipage}
\begin{minipage}[t]{0.48\linewidth}
\centering
\includegraphics[width=0.9\textwidth]{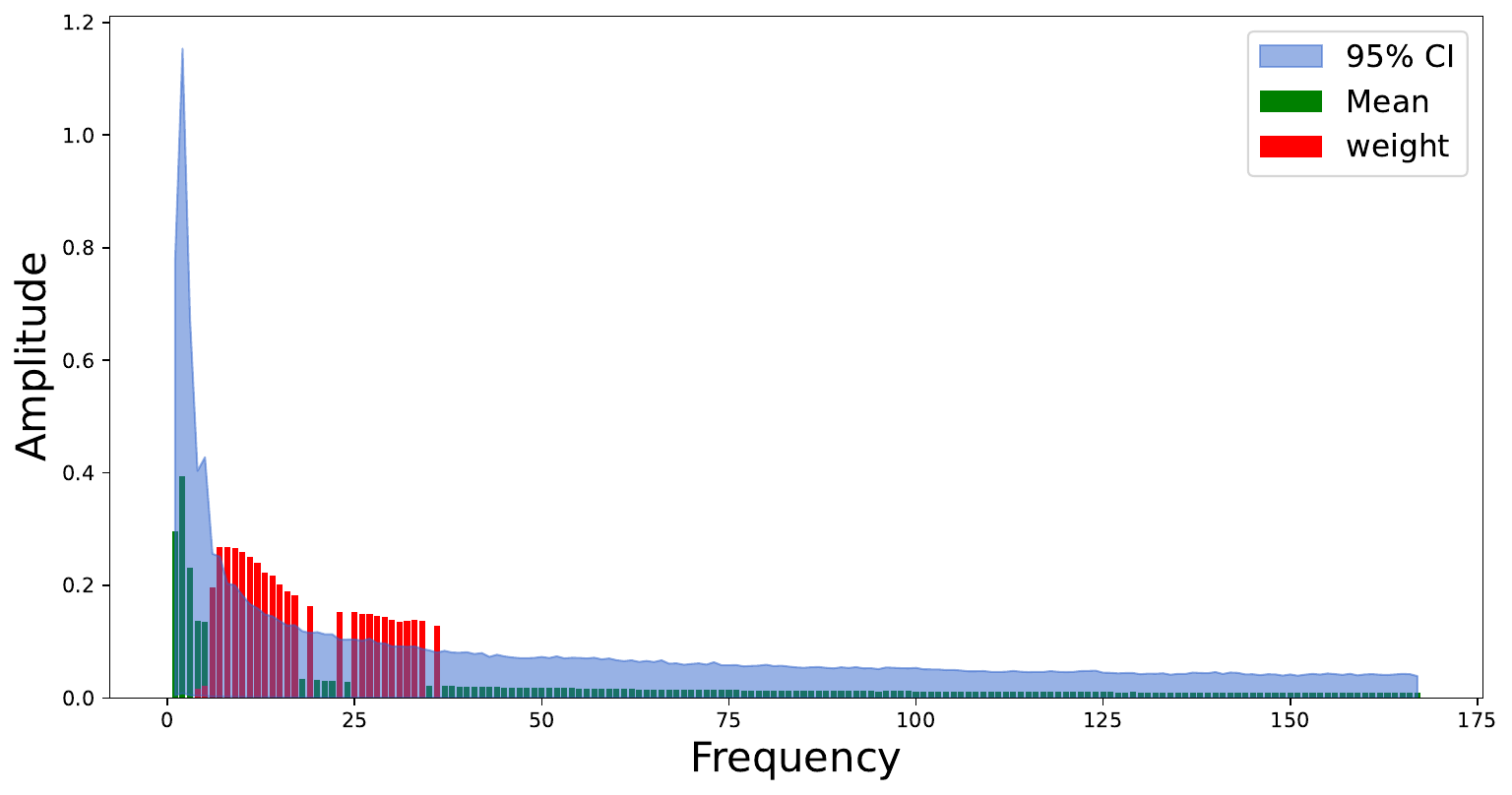}
\end{minipage}
\vspace{0.5em}
\begin{minipage}[t]{0.48\linewidth}
\centering
\includegraphics[width=0.9\textwidth]{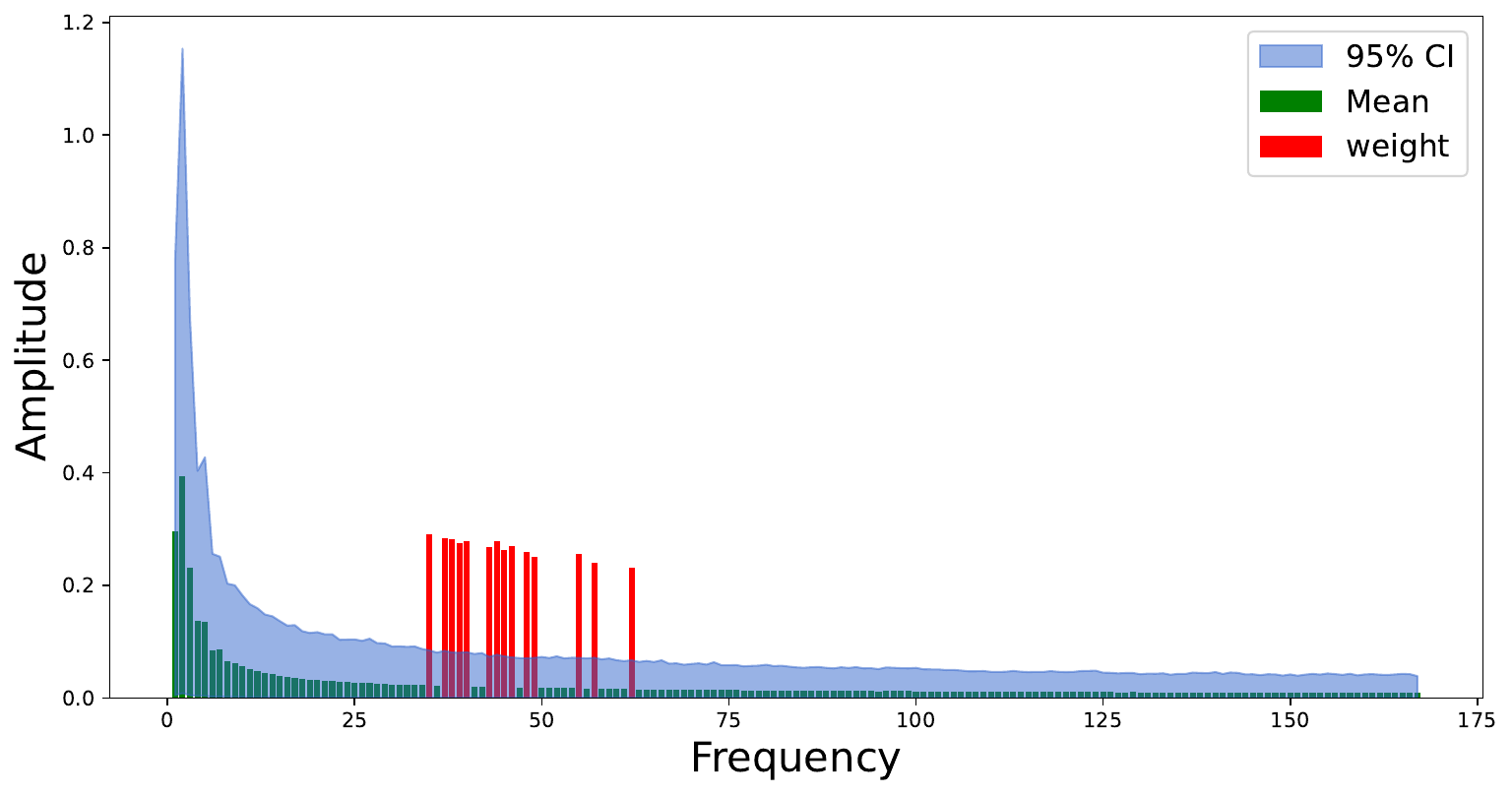}
\end{minipage}
\begin{minipage}[t]{0.48\linewidth}
\centering
\includegraphics[width=0.9\textwidth]{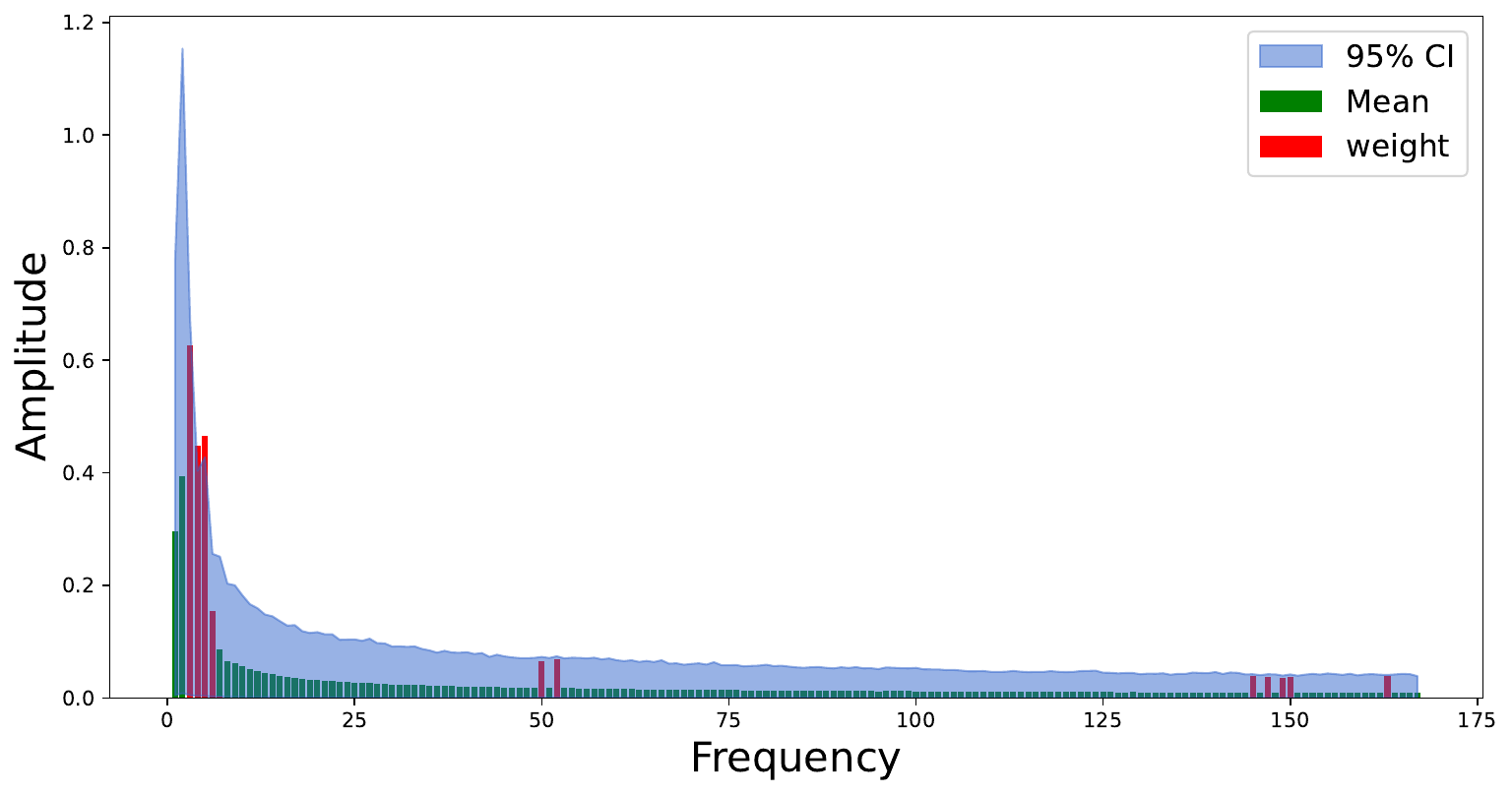}
\end{minipage}%
\vspace{0.5em}
\begin{minipage}[t]{0.48\linewidth}
\centering
\includegraphics[width=0.9\textwidth]{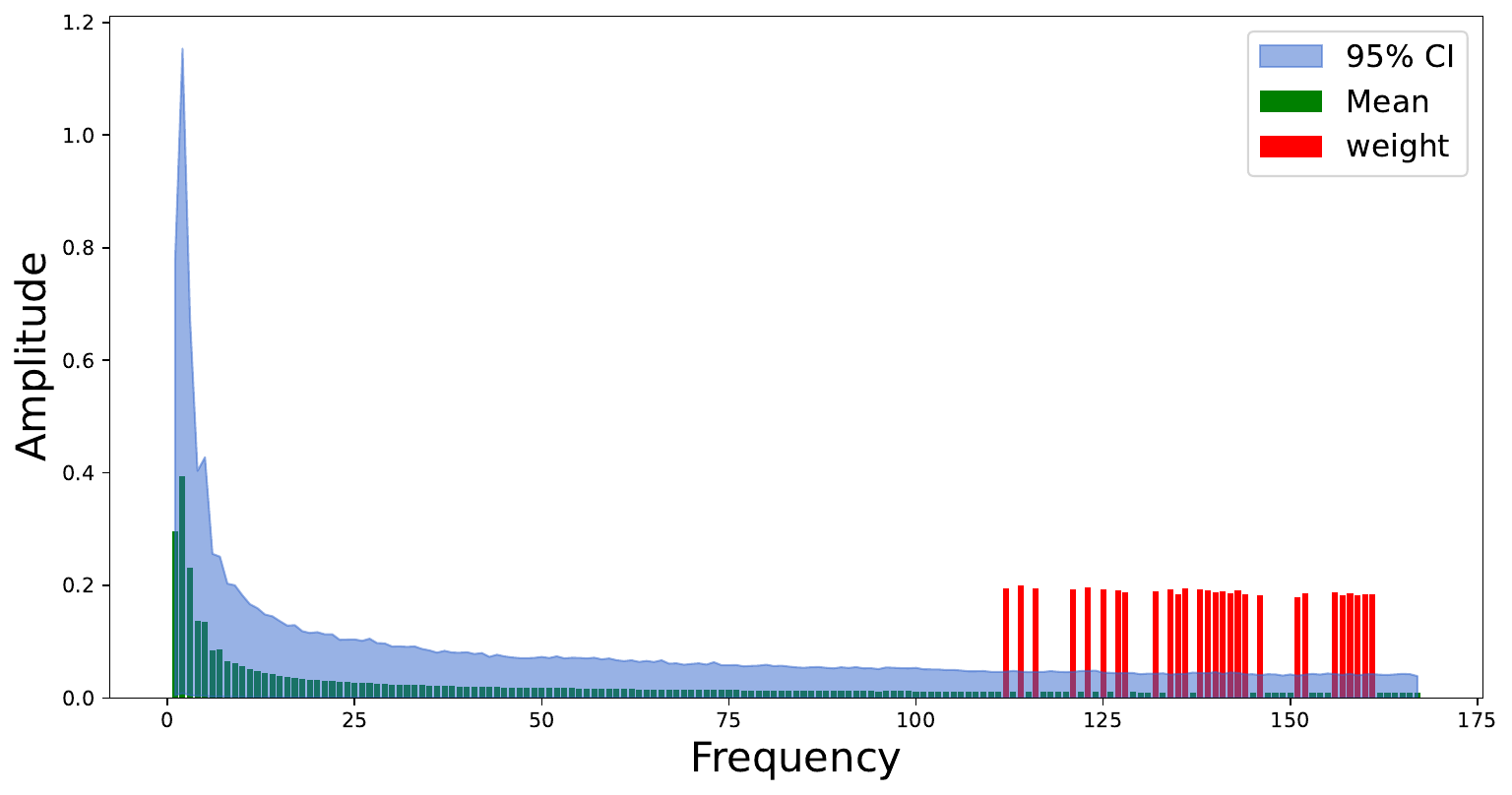}
\end{minipage}%
\begin{minipage}[t]{0.48\linewidth}
\centering
\includegraphics[width=0.9\textwidth]{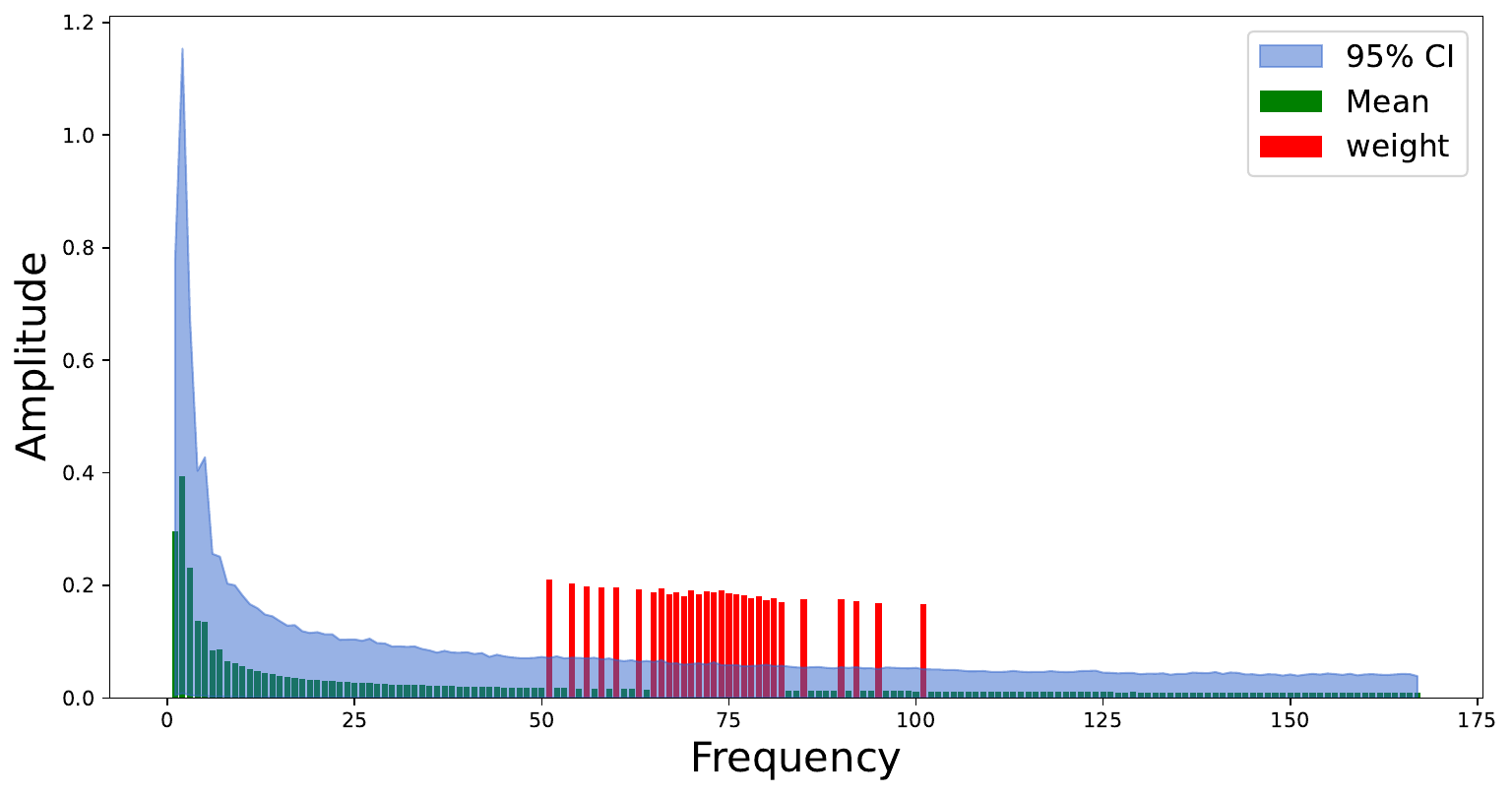}
\end{minipage}
\vspace{0.5em}
\begin{minipage}[t]{0.48\linewidth}
\centering
\includegraphics[width=0.9\textwidth]{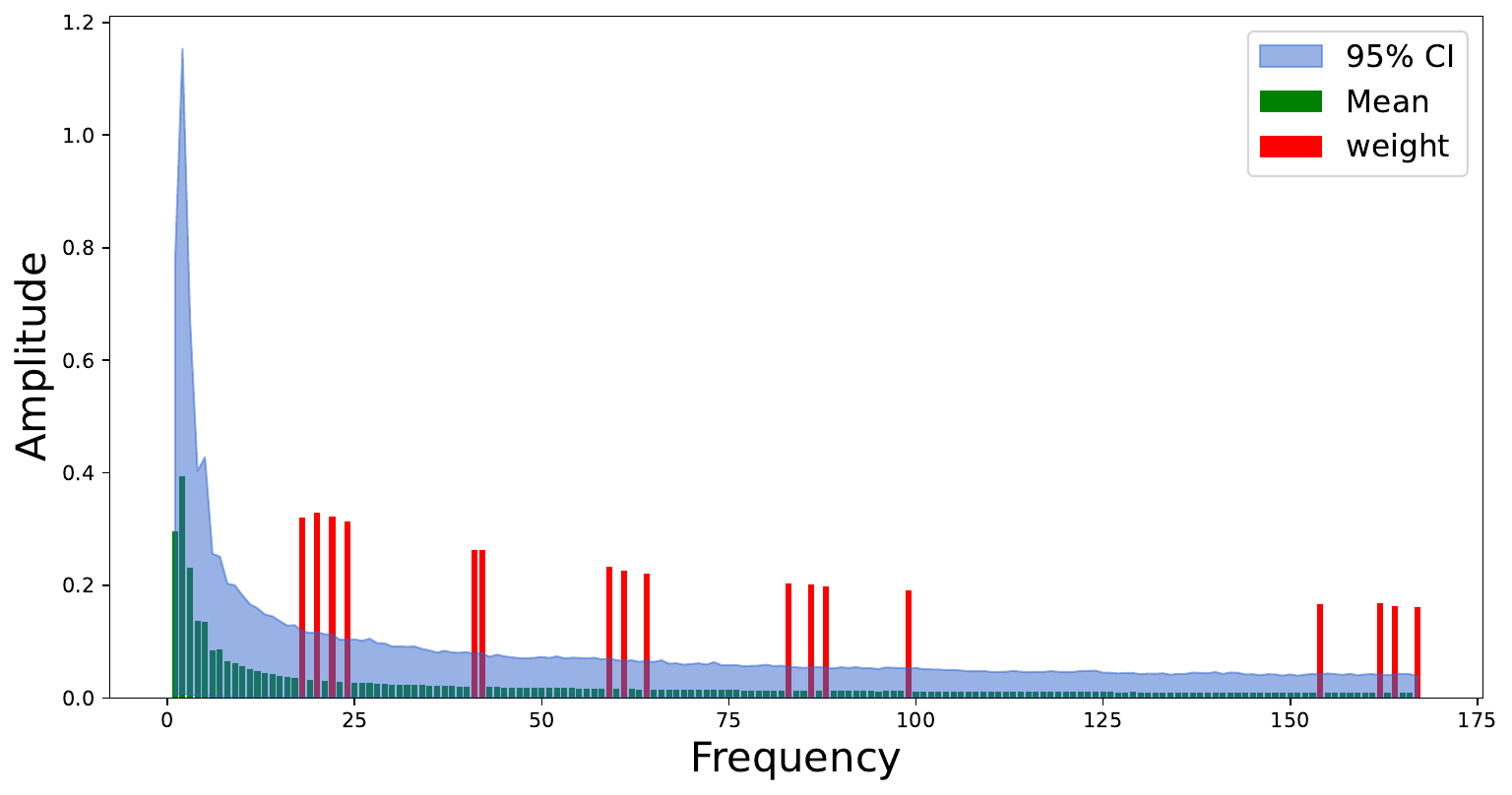}
\end{minipage}
\begin{minipage}[t]{0.48\linewidth}
\centering
\includegraphics[width=0.9\textwidth]{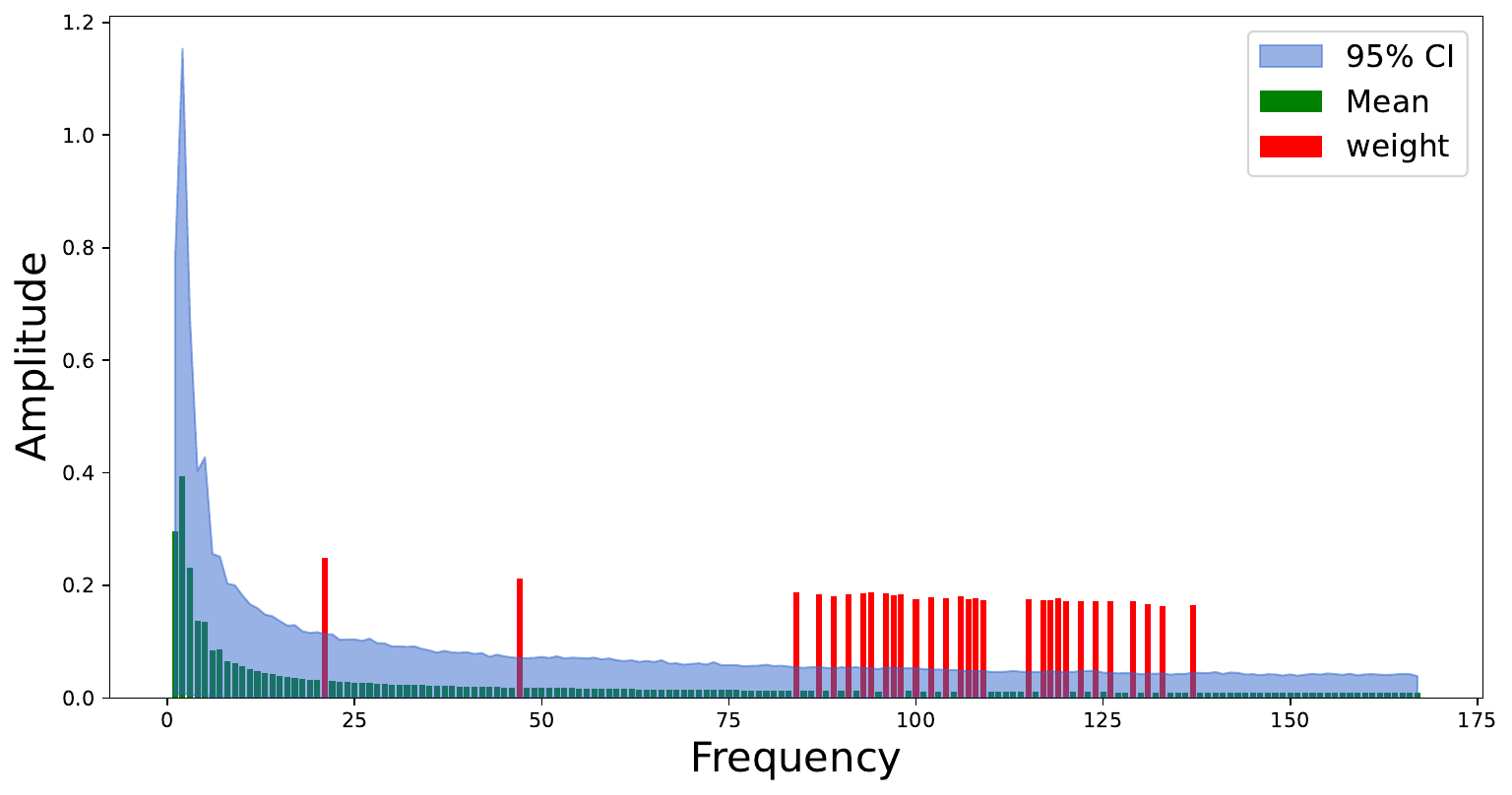}
\end{minipage}
\caption{Learned 10 Components for the Weather data Compared to Its Magnitude Spectrum Distribution. The blue regions correspond to the 95\% confidence interval of magnitude spectrum, the green columns show the mean magnitude spectrum, and the red columns indicate the weights of the learned components.}
\label{components_Weather}
\end{figure*}

\begin{figure*}[t]
\centering
\begin{minipage}[t]{0.48\linewidth}
\centering
\includegraphics[width=0.9\textwidth]{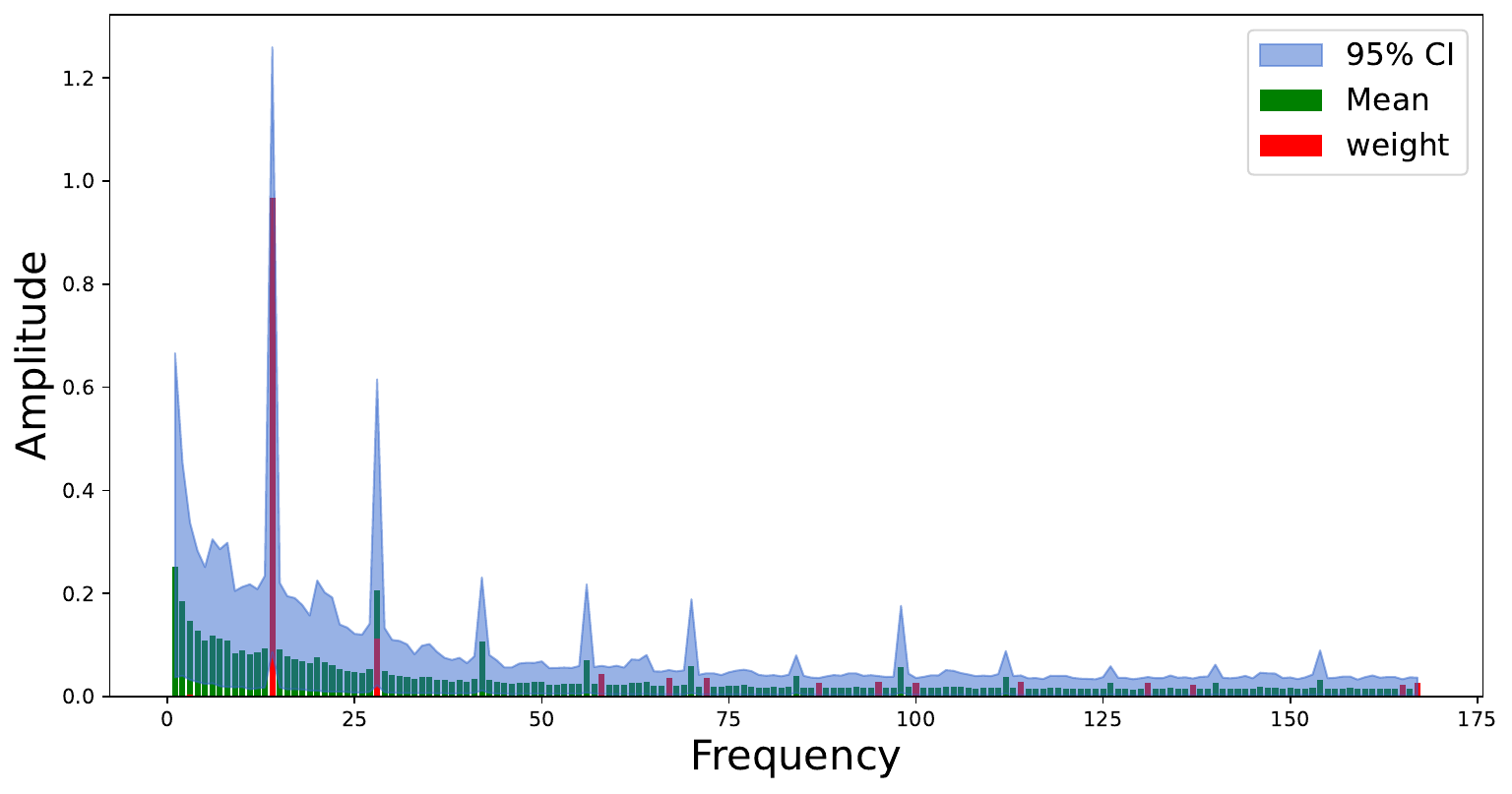}
\end{minipage}%
\begin{minipage}[t]{0.48\linewidth}
\centering
\includegraphics[width=0.9\textwidth]{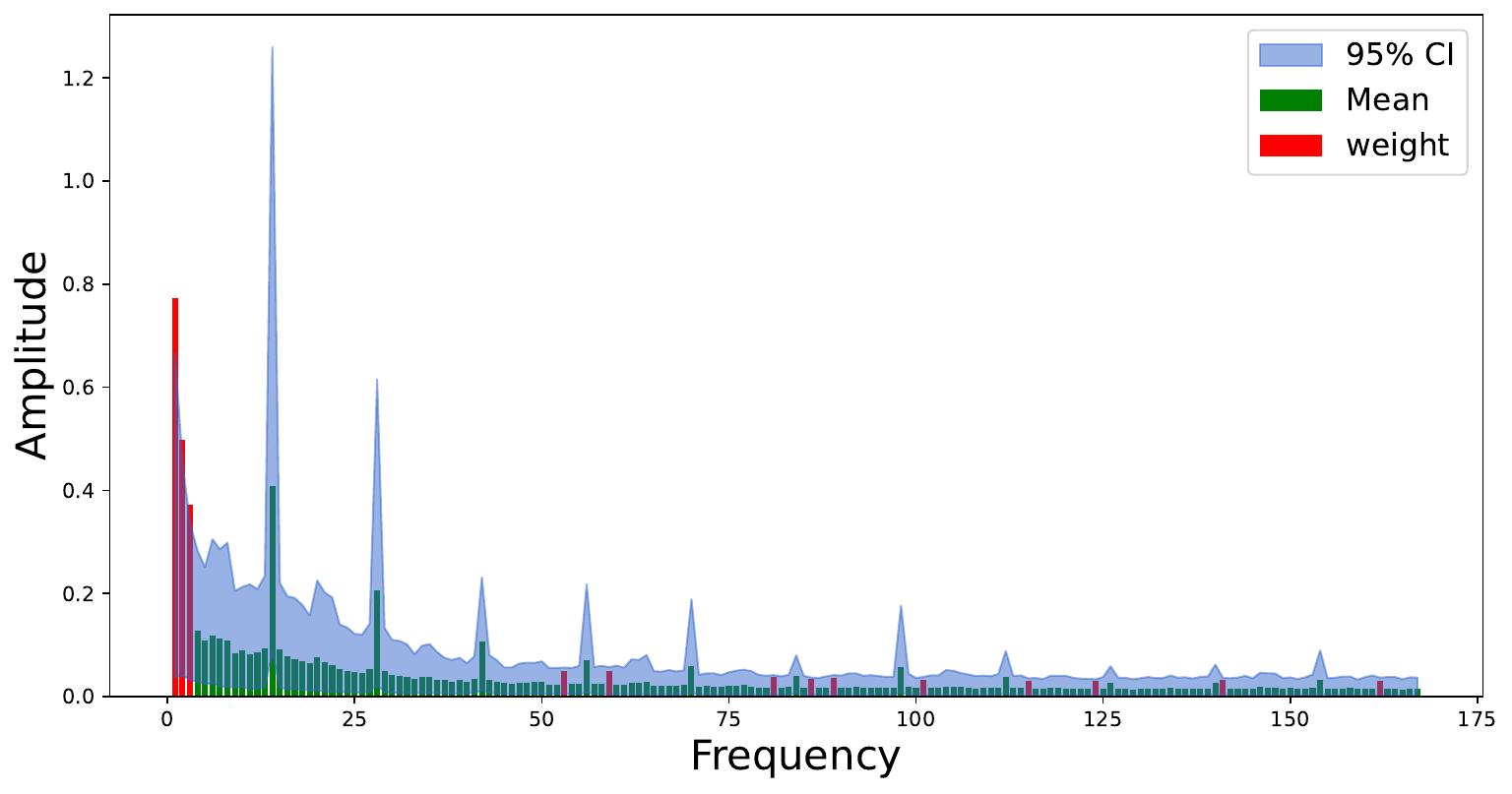}
\end{minipage}%
\vspace{0.5em}
\begin{minipage}[t]{0.48\linewidth}
\centering
\includegraphics[width=0.9\textwidth]{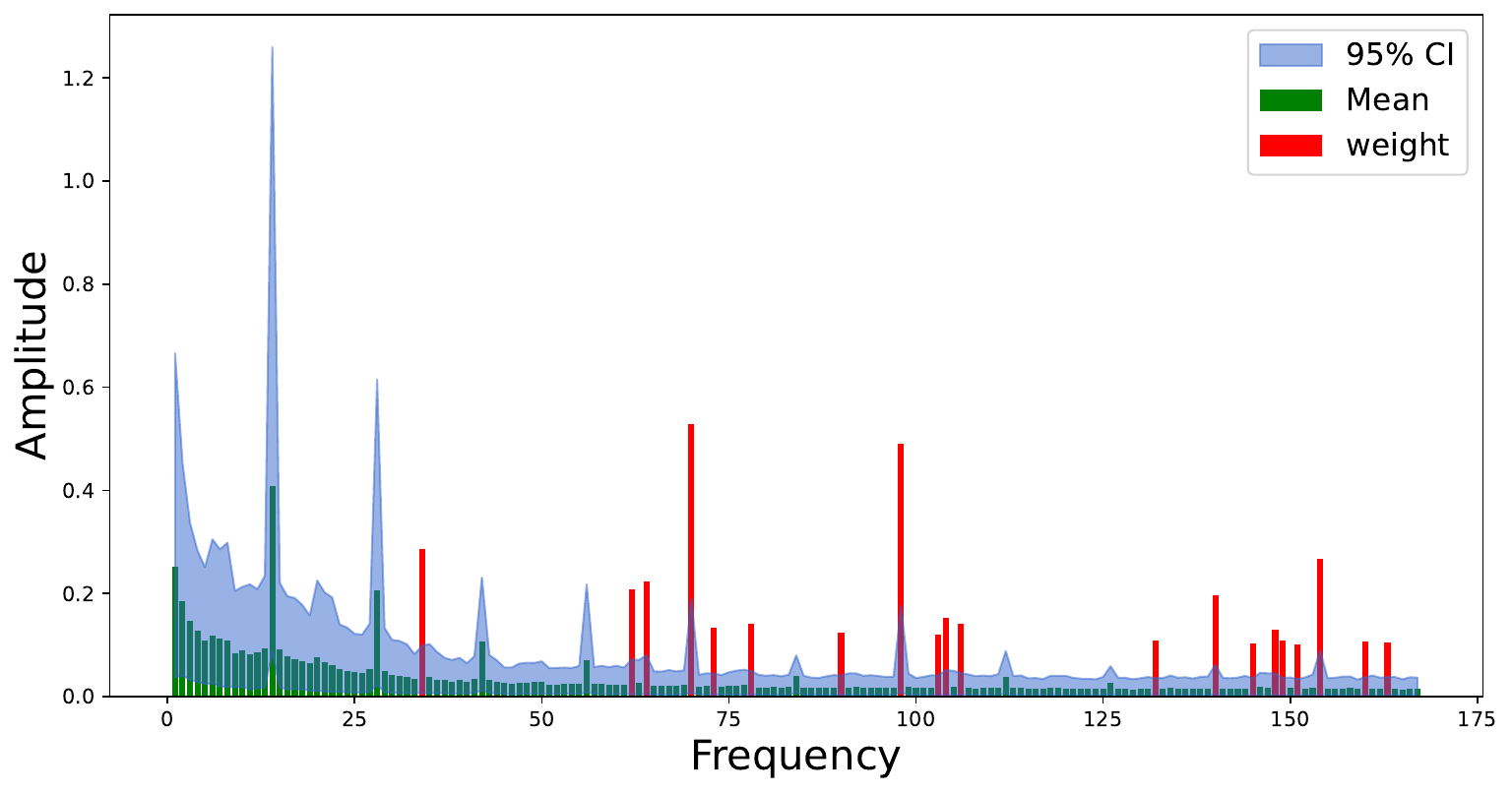}
\end{minipage}
\begin{minipage}[t]{0.48\linewidth}
\centering
\includegraphics[width=0.9\textwidth]{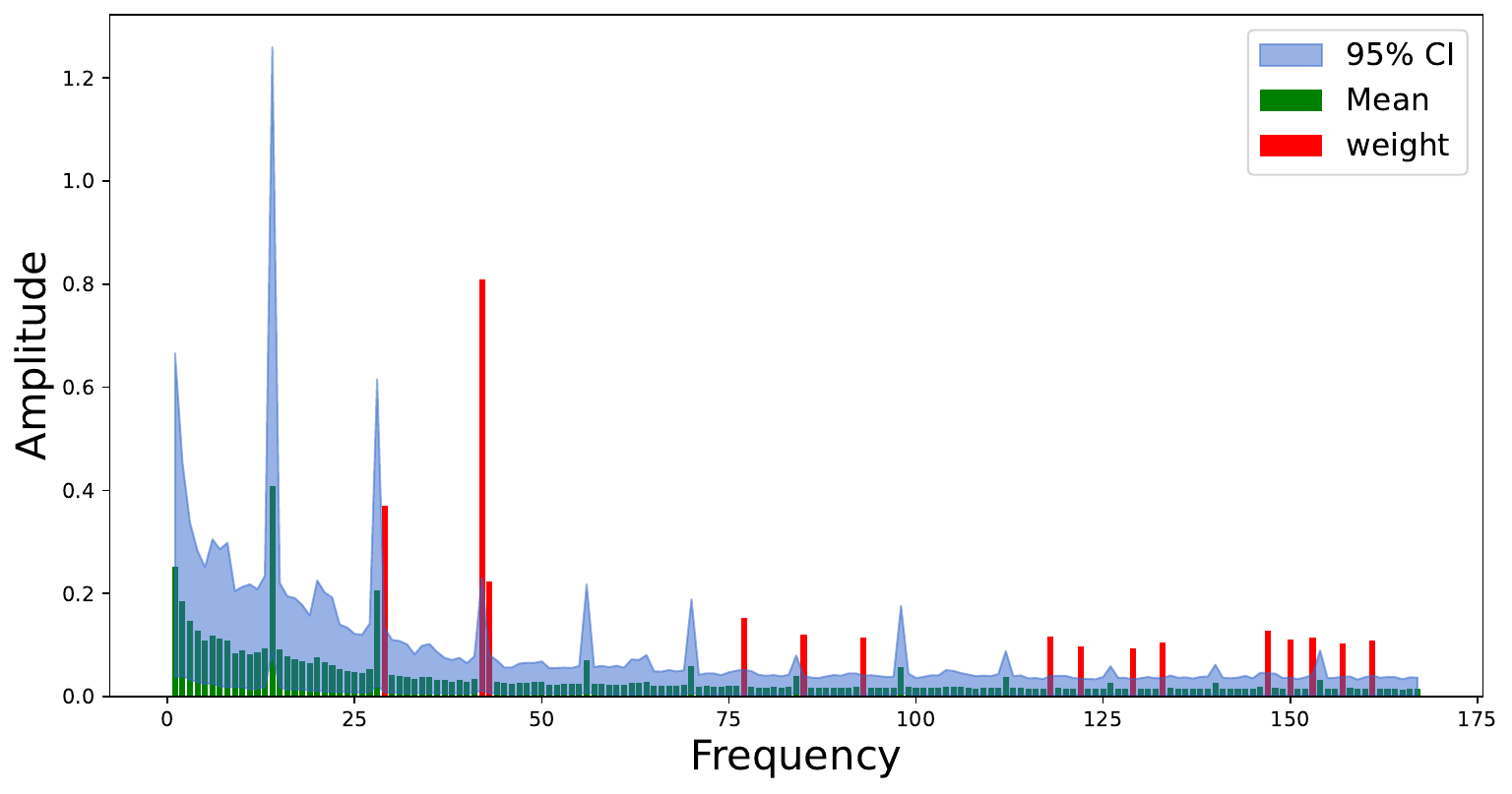}
\end{minipage}
\vspace{0.5em}
\begin{minipage}[t]{0.48\linewidth}
\centering
\includegraphics[width=0.9\textwidth]{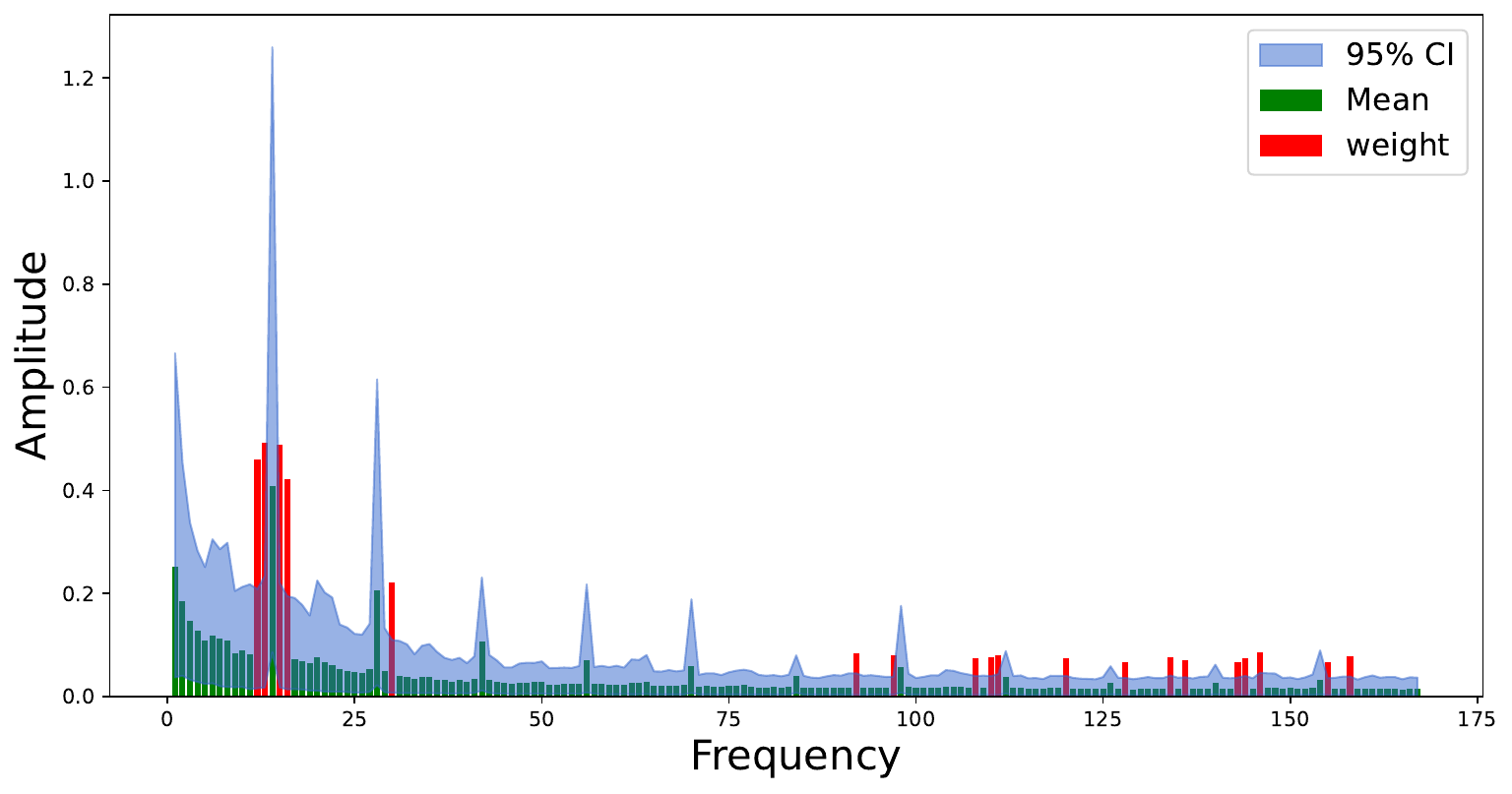}
\end{minipage}
\begin{minipage}[t]{0.48\linewidth}
\centering
\includegraphics[width=0.9\textwidth]{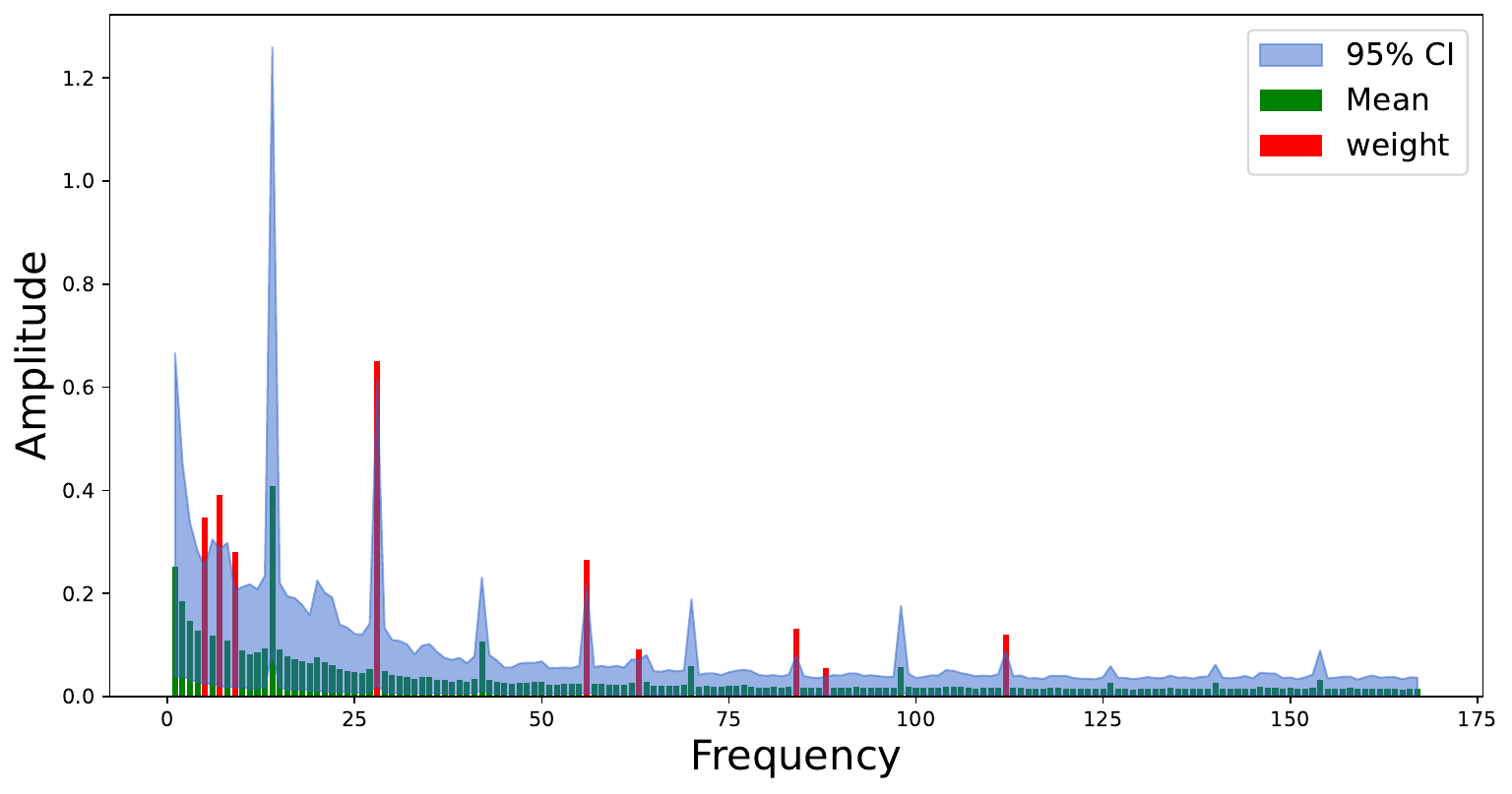}
\end{minipage}%
\vspace{0.5em}
\begin{minipage}[t]{0.48\linewidth}
\centering
\includegraphics[width=0.9\textwidth]{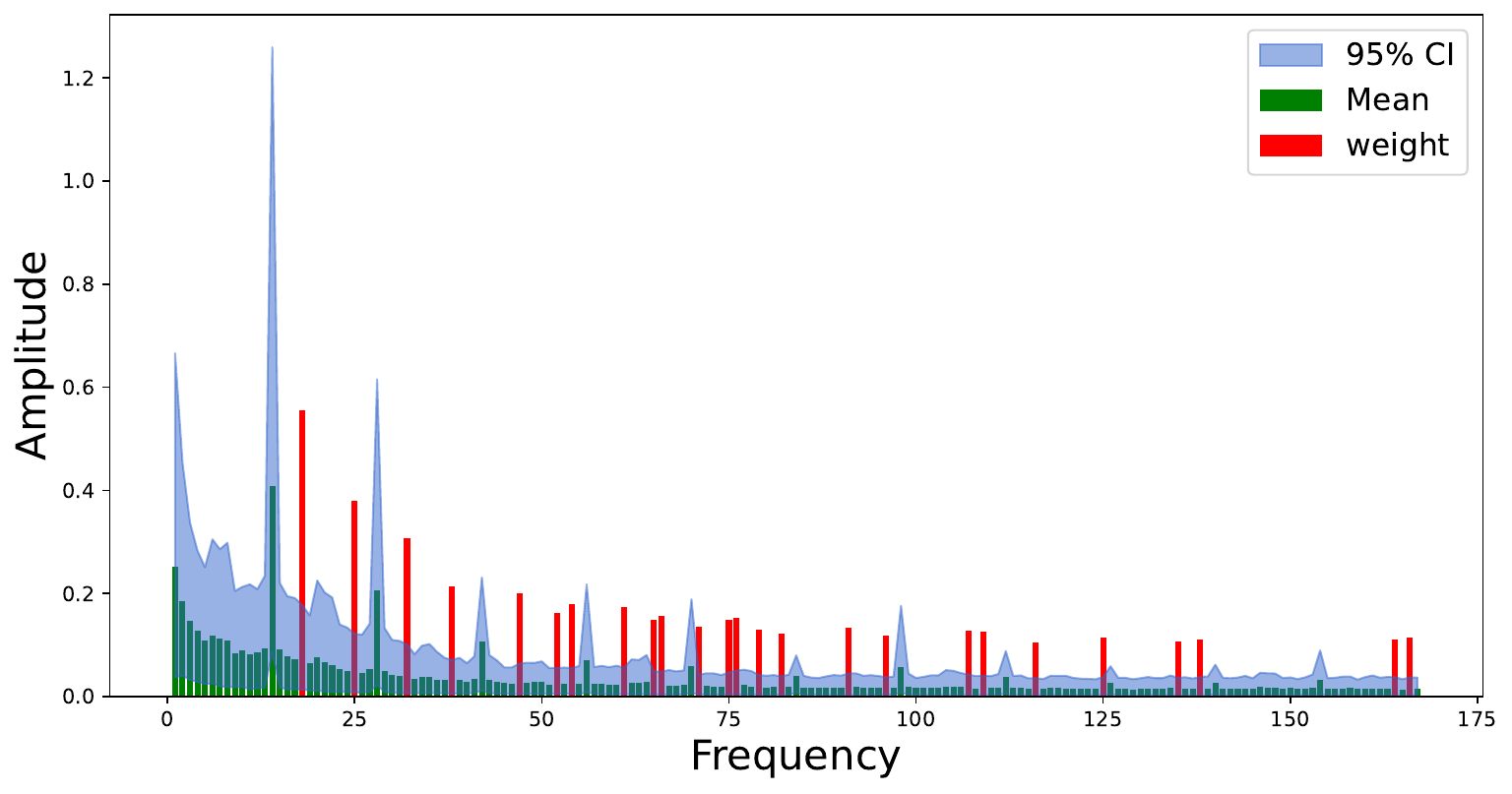}
\end{minipage}%
\begin{minipage}[t]{0.48\linewidth}
\centering
\includegraphics[width=0.9\textwidth]{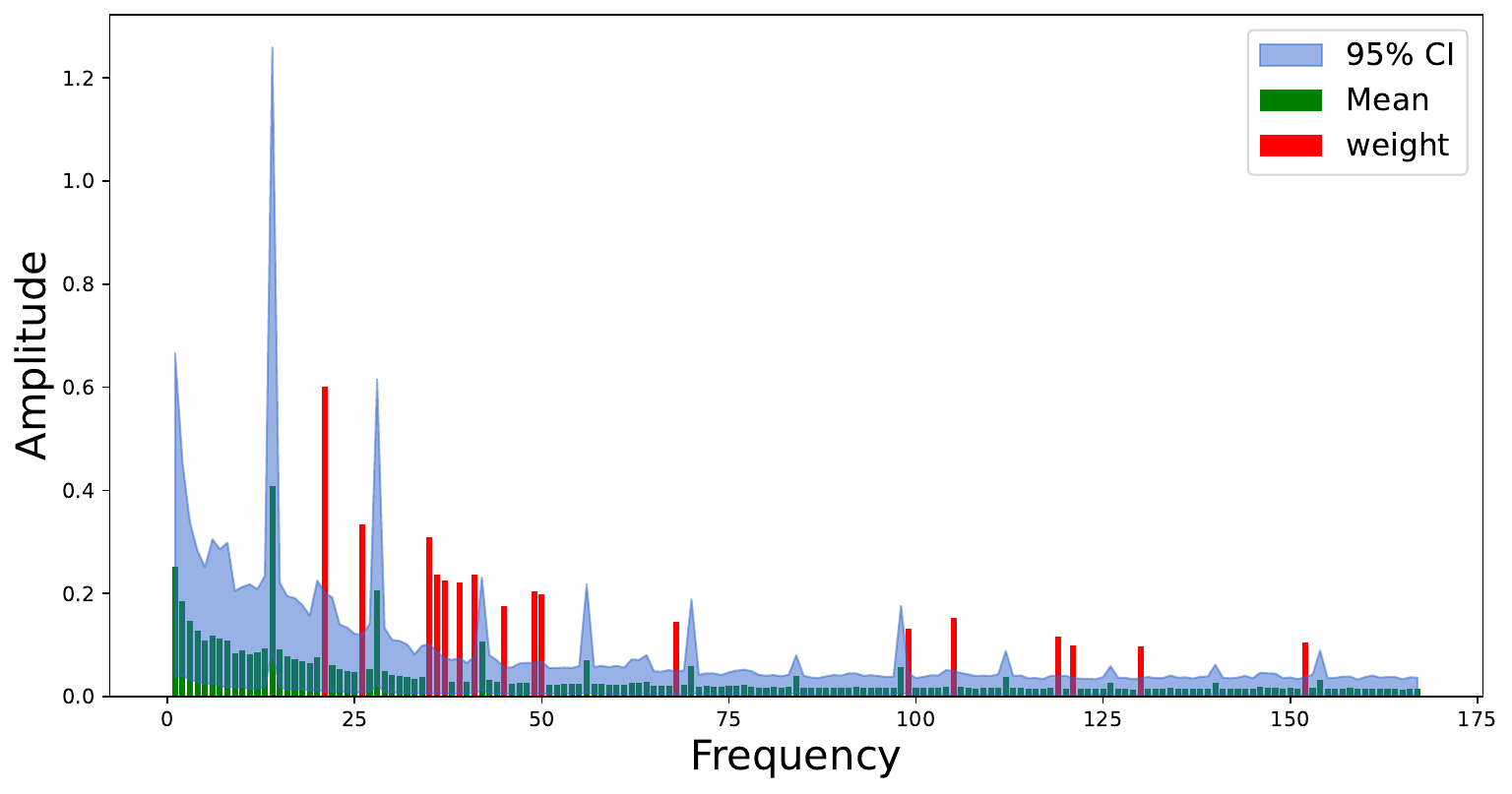}
\end{minipage}
\vspace{0.5em}
\begin{minipage}[t]{0.48\linewidth}
\centering
\includegraphics[width=0.9\textwidth]{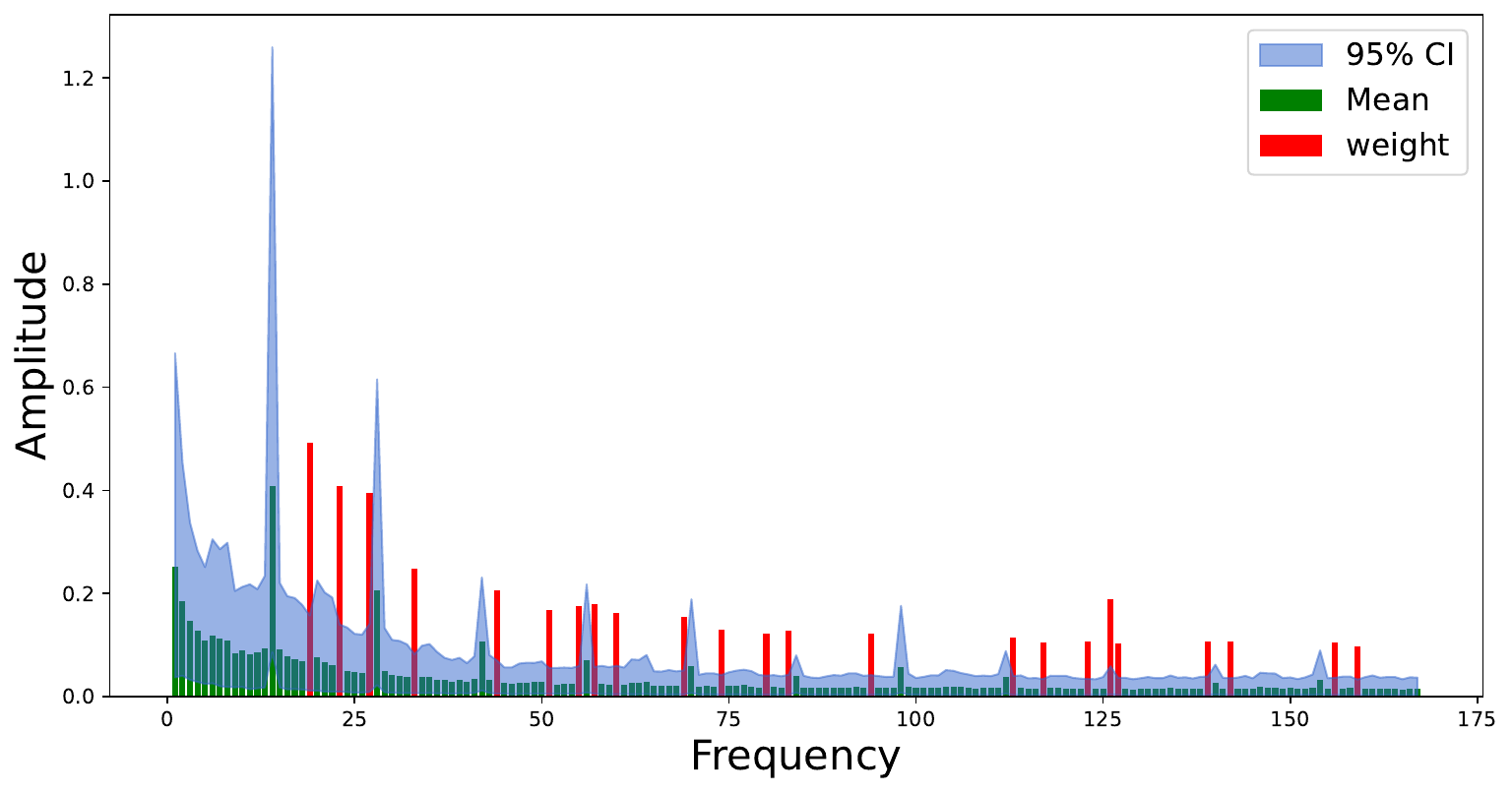}
\end{minipage}
\begin{minipage}[t]{0.48\linewidth}
\centering
\includegraphics[width=0.9\textwidth]{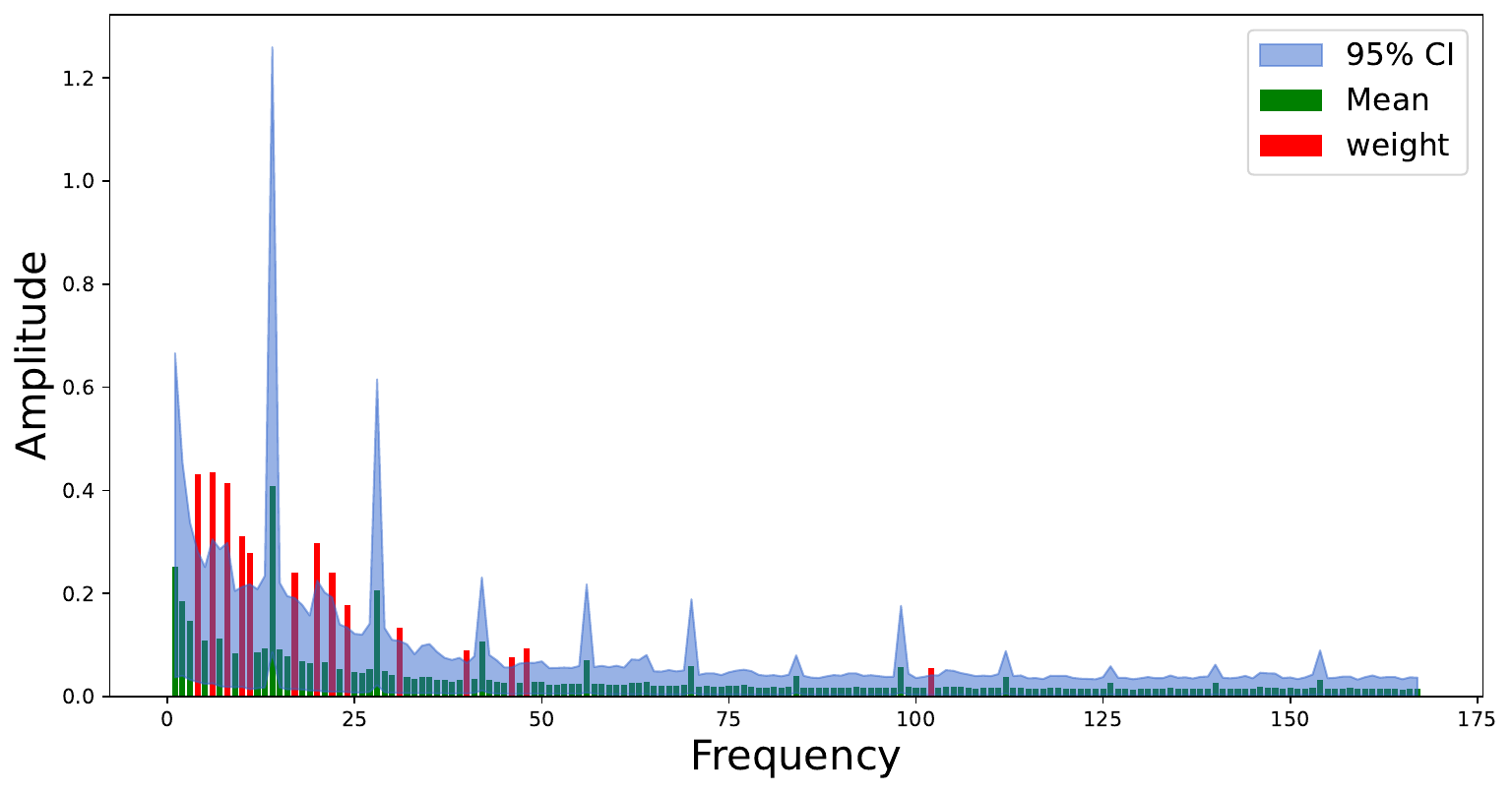}
\end{minipage}
\caption{Learned 10 Components for the ETTh1 data Compared to Its Magnitude Spectrum Distribution. The blue regions correspond to the 95\% confidence interval of magnitude spectrum, the green columns show the mean magnitude spectrum, and the red columns indicate the weights of the learned components.}
\label{components_ETTh1}
\end{figure*}

\begin{figure*}[t]
\centering
\begin{minipage}[t]{0.48\linewidth}
\centering
\includegraphics[width=0.9\textwidth]{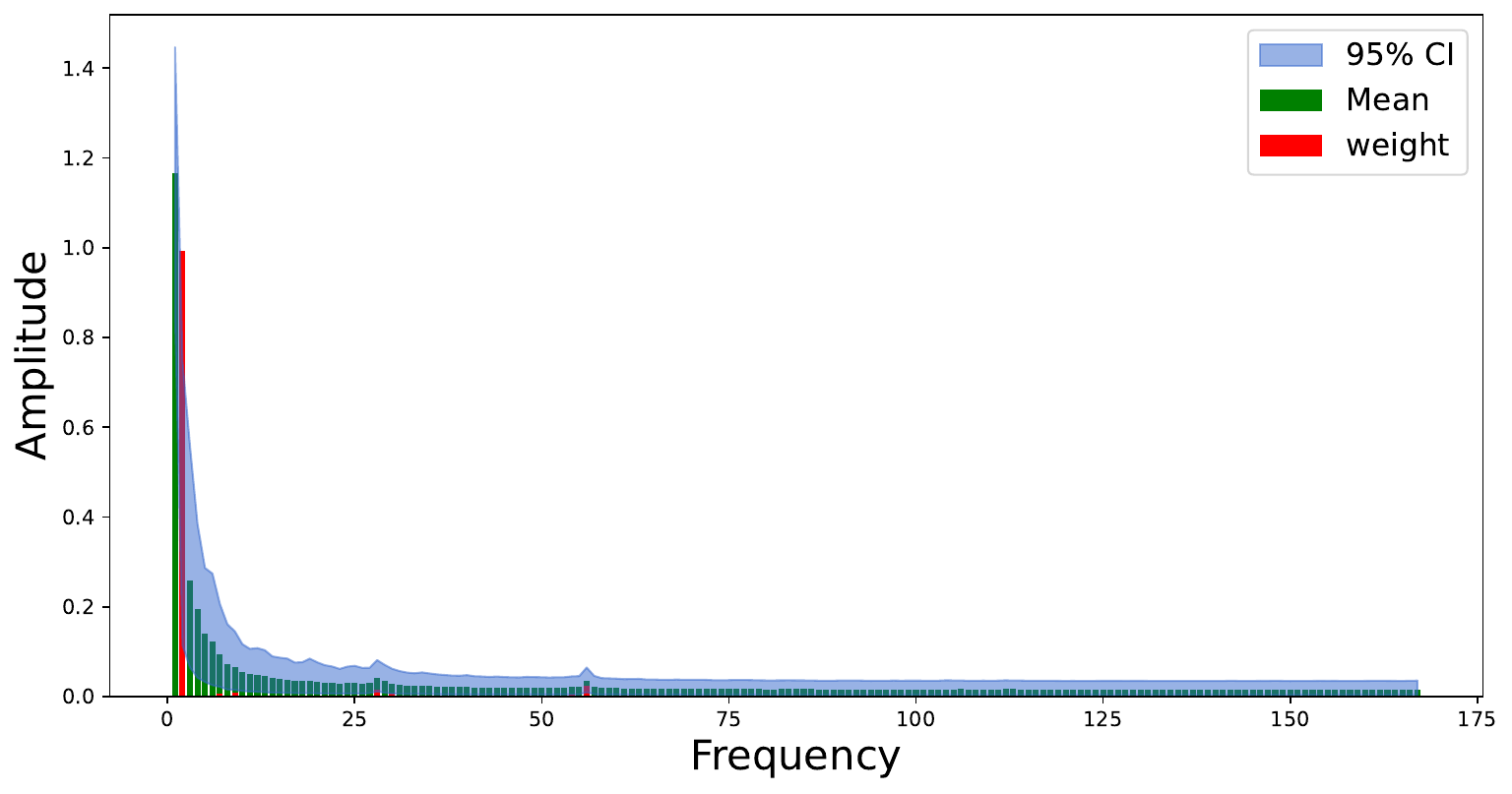}
\end{minipage}%
\begin{minipage}[t]{0.48\linewidth}
\centering
\includegraphics[width=0.9\textwidth]{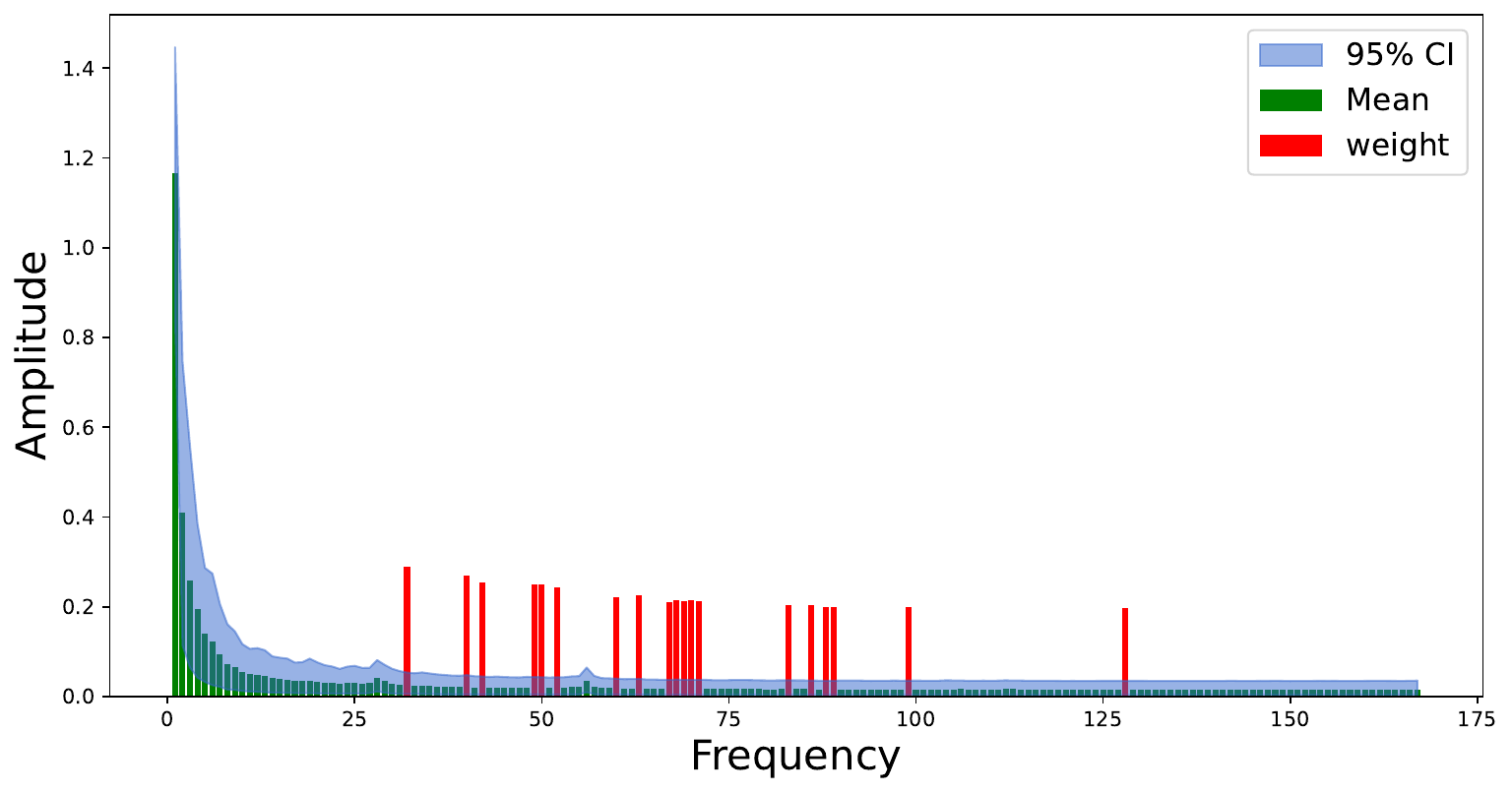}
\end{minipage}%
\vspace{0.5em}
\begin{minipage}[t]{0.48\linewidth}
\centering
\includegraphics[width=0.9\textwidth]{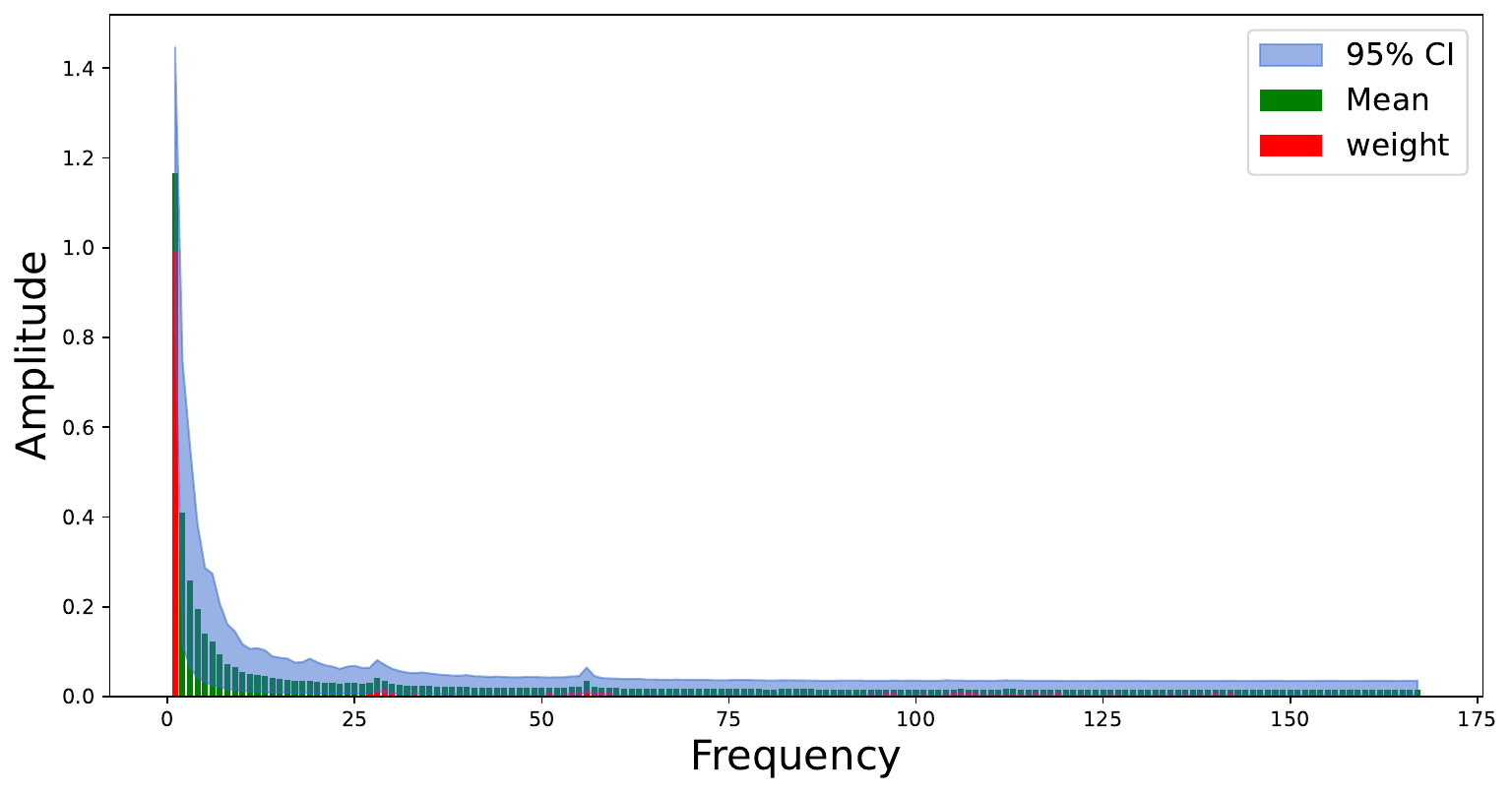}
\end{minipage}
\begin{minipage}[t]{0.48\linewidth}
\centering
\includegraphics[width=0.9\textwidth]{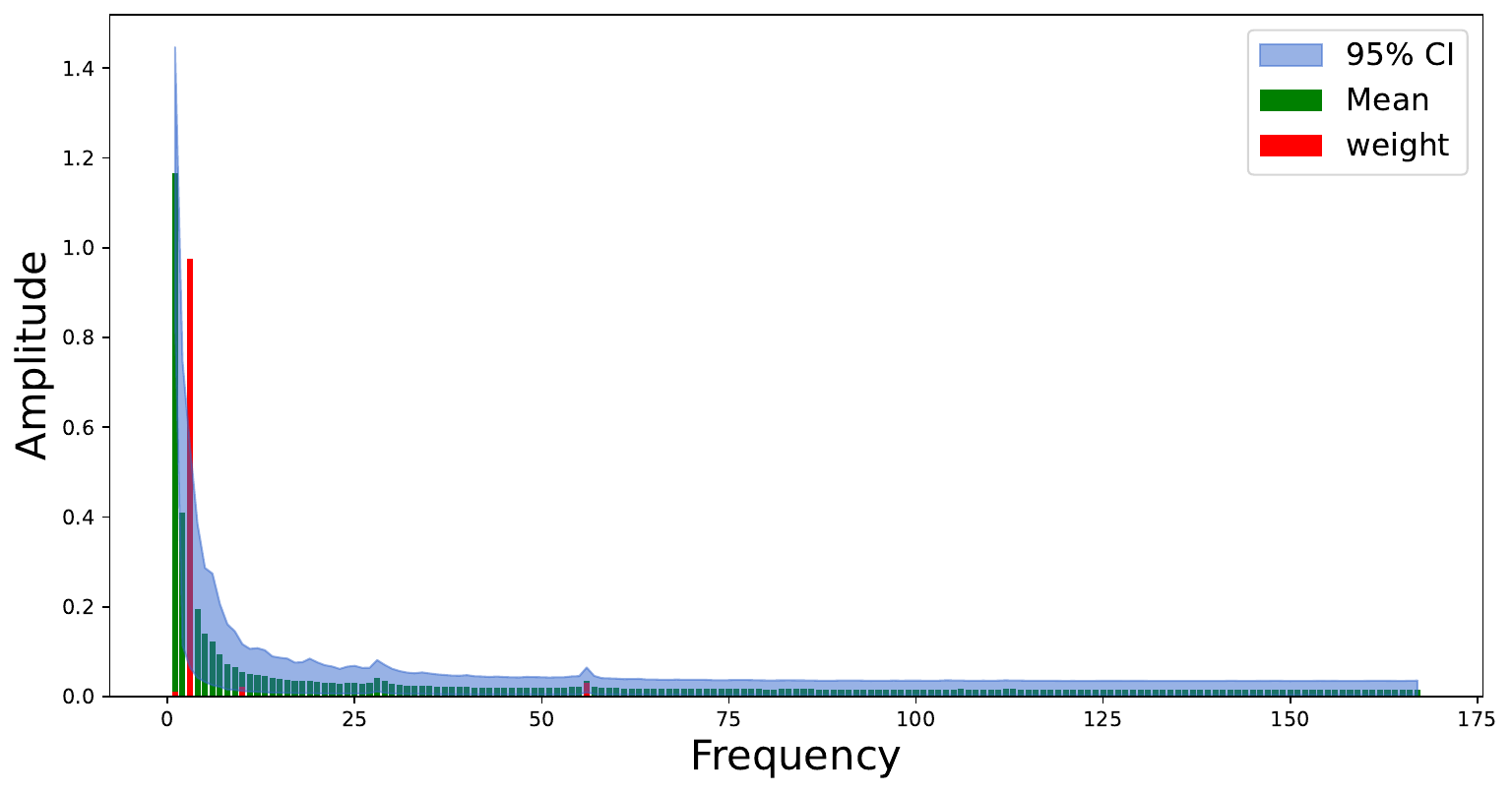}
\end{minipage}
\vspace{0.5em}
\begin{minipage}[t]{0.48\linewidth}
\centering
\includegraphics[width=0.9\textwidth]{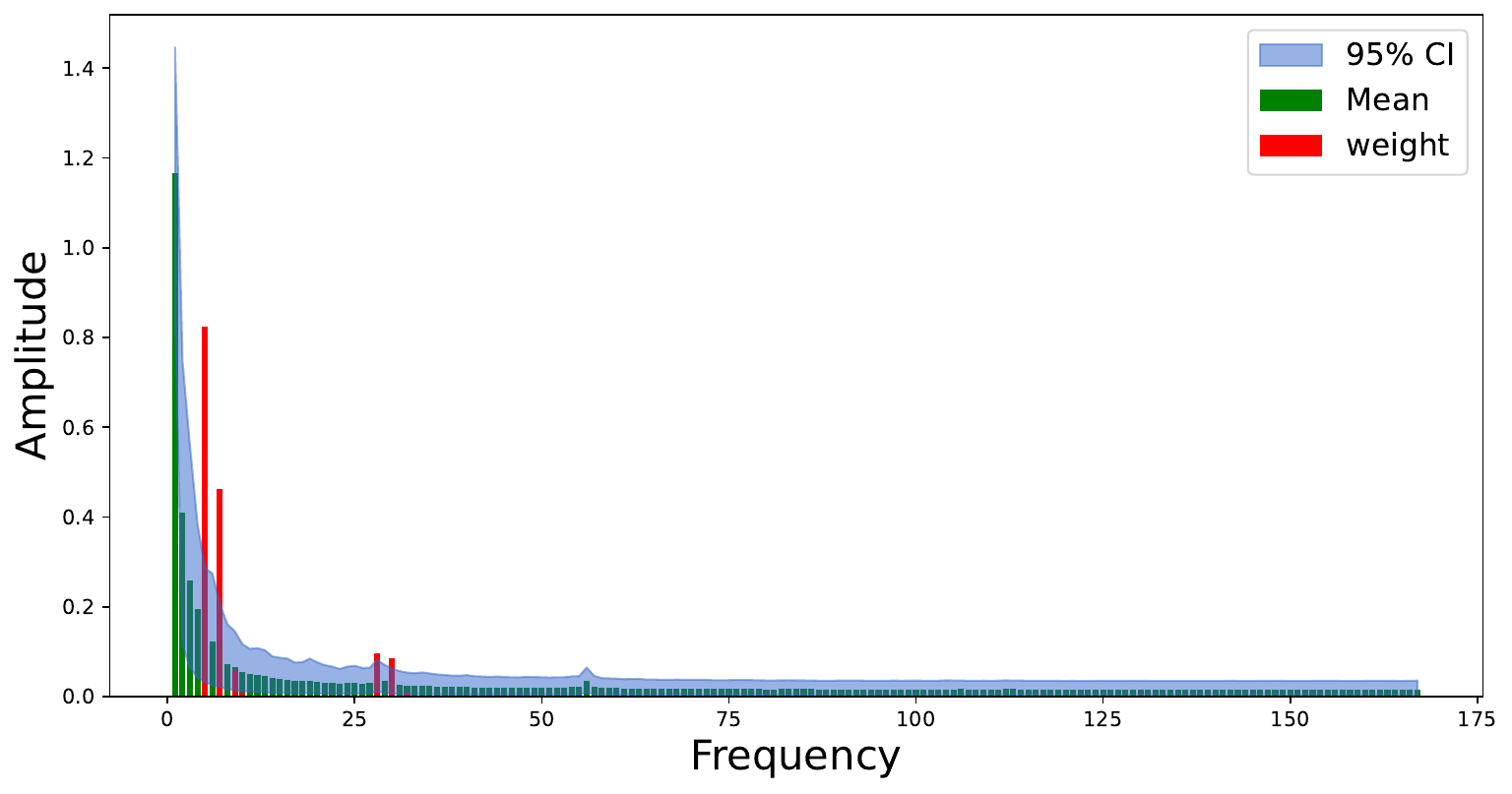}
\end{minipage}
\begin{minipage}[t]{0.48\linewidth}
\centering
\includegraphics[width=0.9\textwidth]{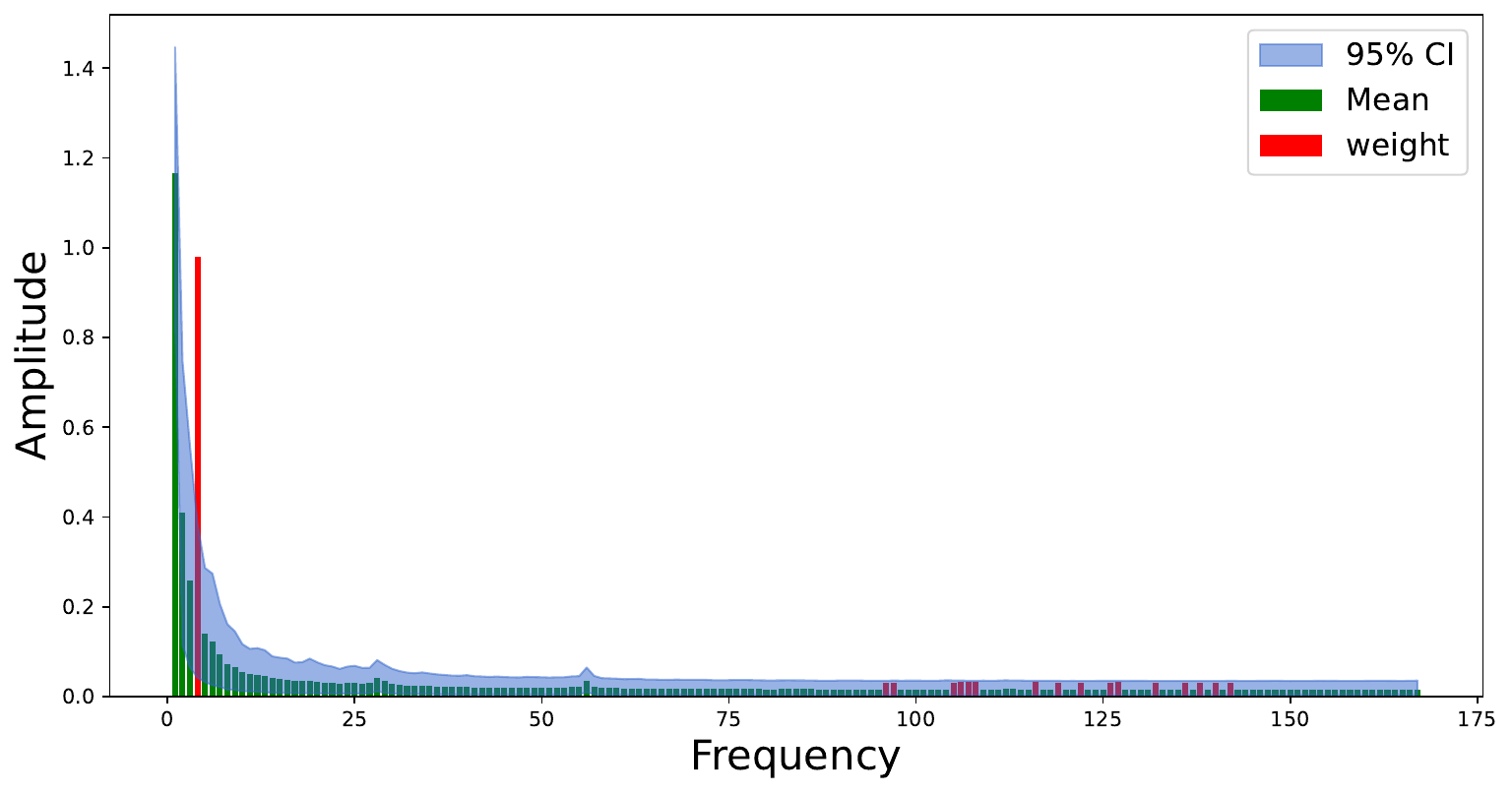}
\end{minipage}%
\vspace{0.5em}
\begin{minipage}[t]{0.48\linewidth}
\centering
\includegraphics[width=0.9\textwidth]{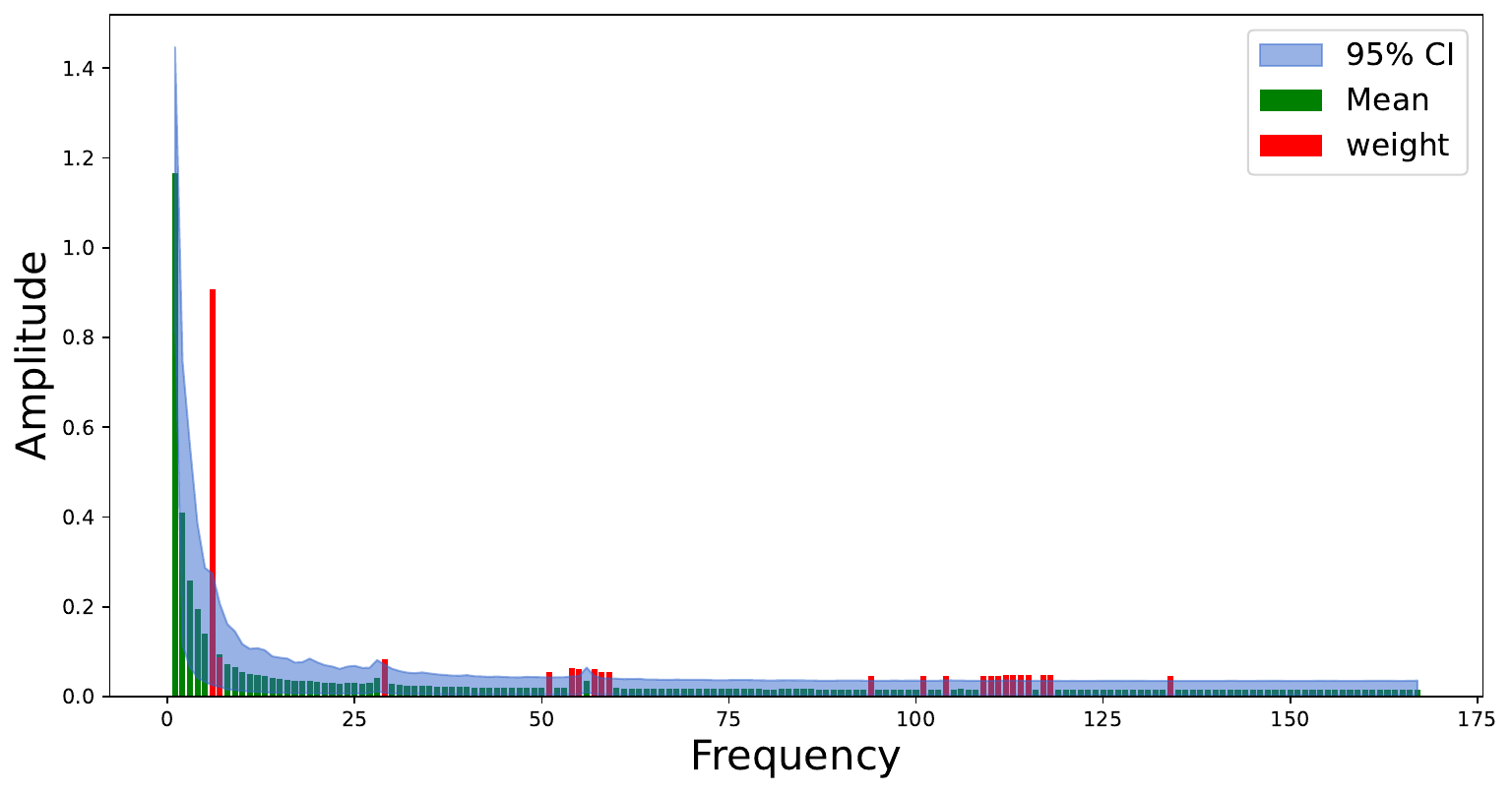}
\end{minipage}%
\begin{minipage}[t]{0.48\linewidth}
\centering
\includegraphics[width=0.9\textwidth]{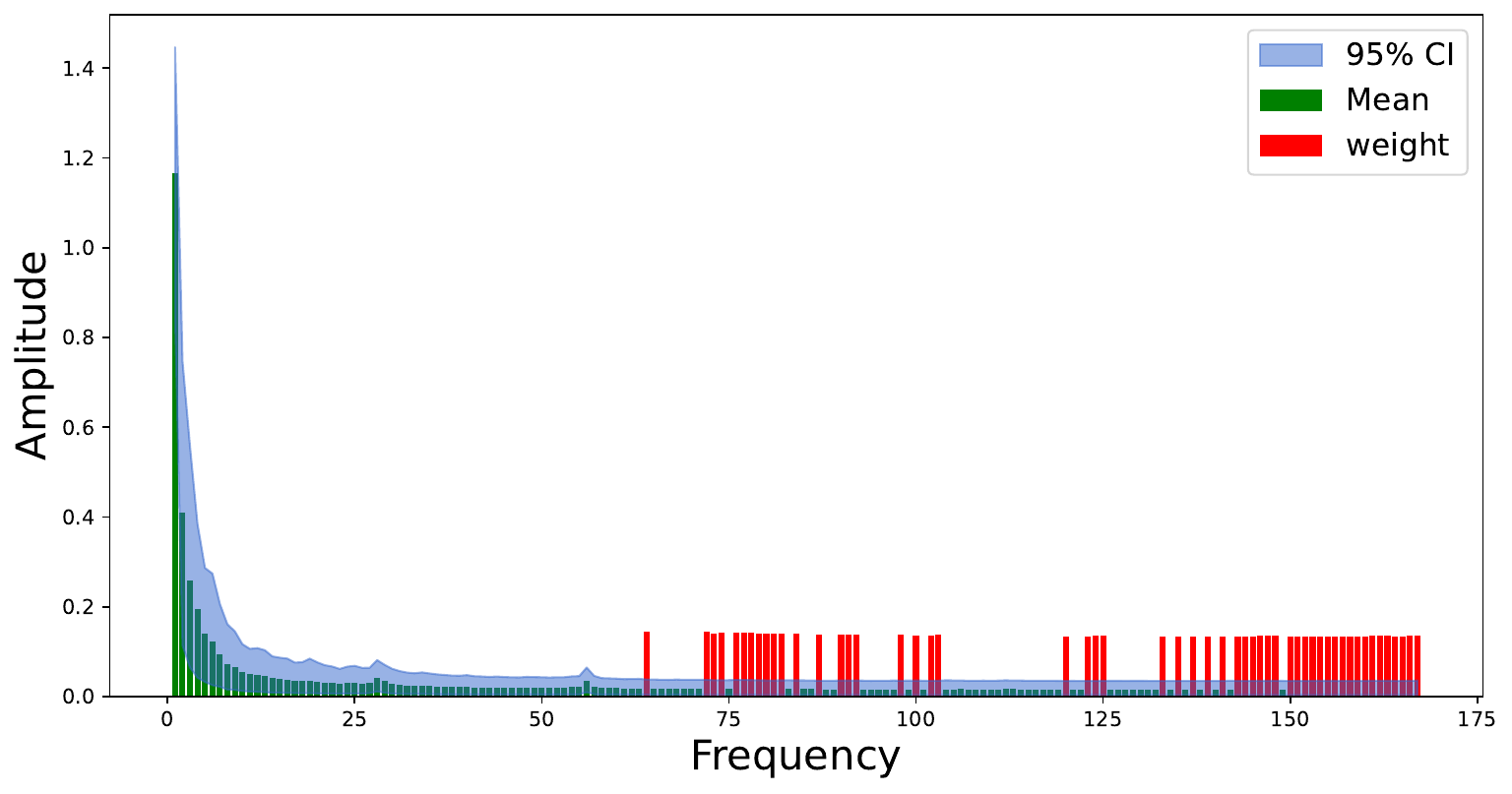}
\end{minipage}
\vspace{0.5em}
\begin{minipage}[t]{0.48\linewidth}
\centering
\includegraphics[width=0.9\textwidth]{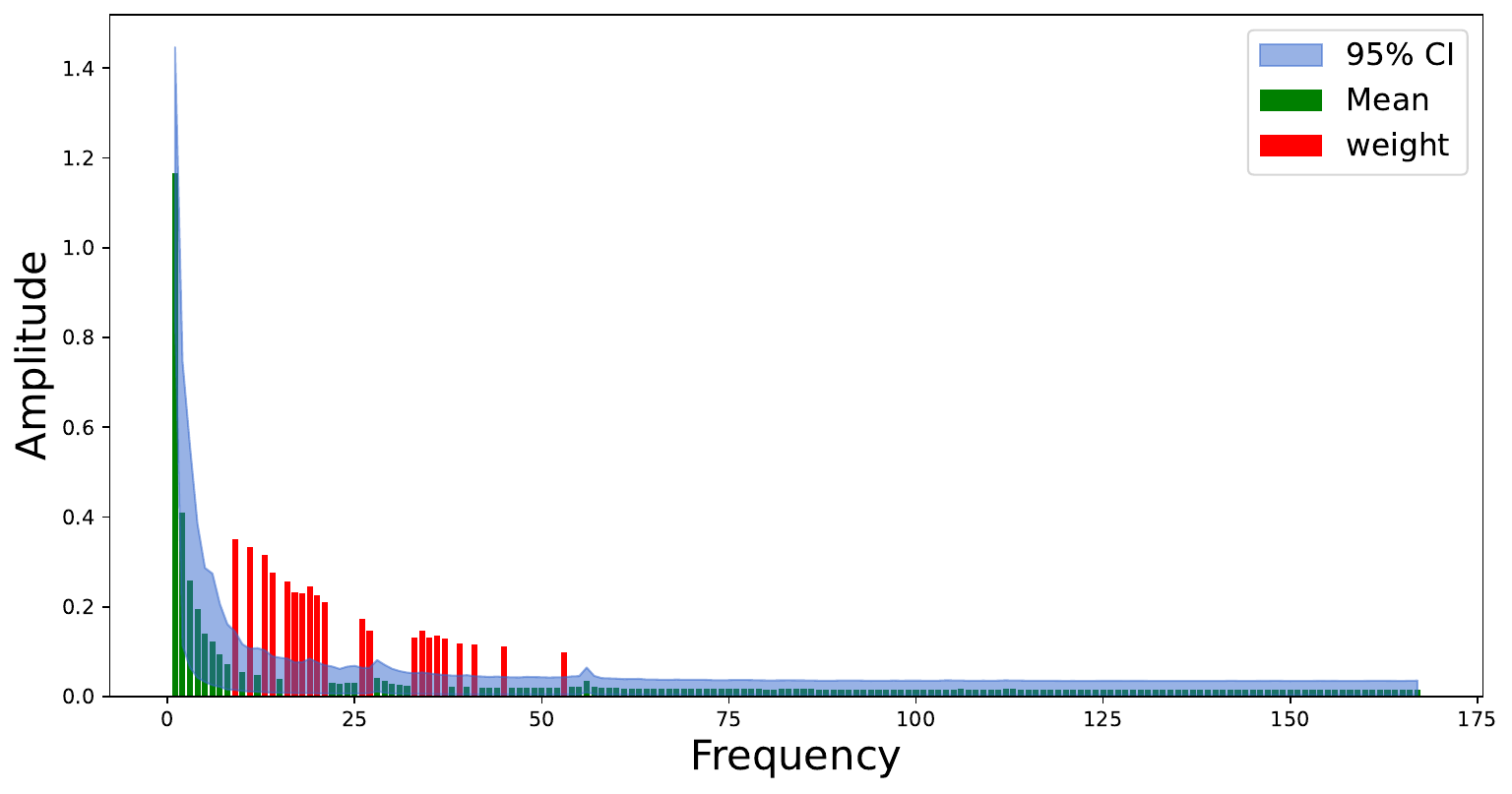}
\end{minipage}
\begin{minipage}[t]{0.48\linewidth}
\centering
\includegraphics[width=0.9\textwidth]{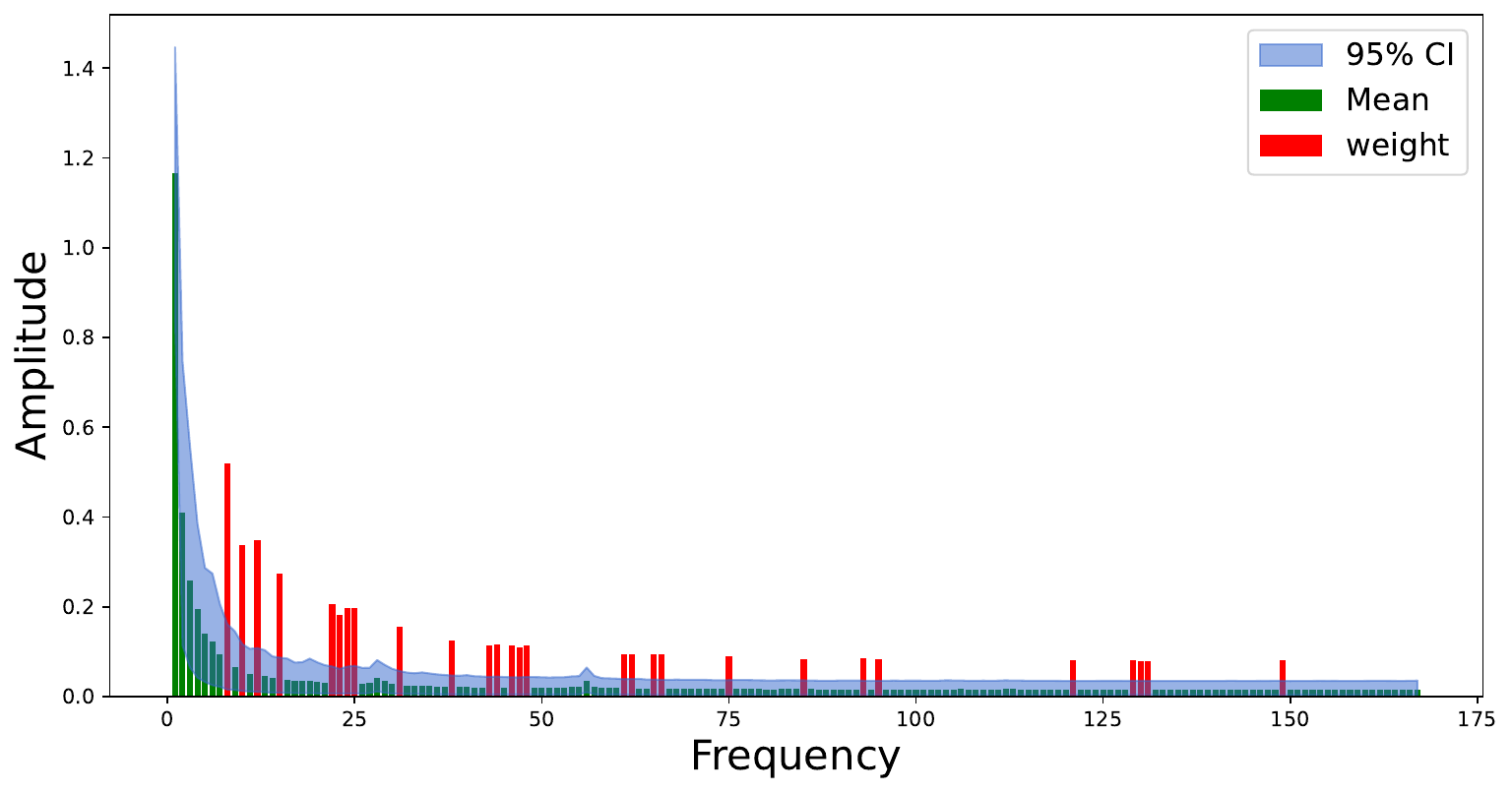}
\end{minipage}
\caption{Learned 10 Components for the PEMS03 data Compared to Its Magnitude Spectrum Distribution. The blue regions correspond to the 95\% confidence interval of magnitude spectrum, the green columns show the mean magnitude spectrum, and the red columns indicate the weights of the learned components.}
\label{components_PEMS03}
\end{figure*}

\begin{figure*}[t]
\centering
\begin{minipage}[t]{0.48\linewidth}
\centering
\includegraphics[width=0.9\textwidth]{./fig/PEMS03/weight0.pdf}
\end{minipage}%
\begin{minipage}[t]{0.48\linewidth}
\centering
\includegraphics[width=0.9\textwidth]{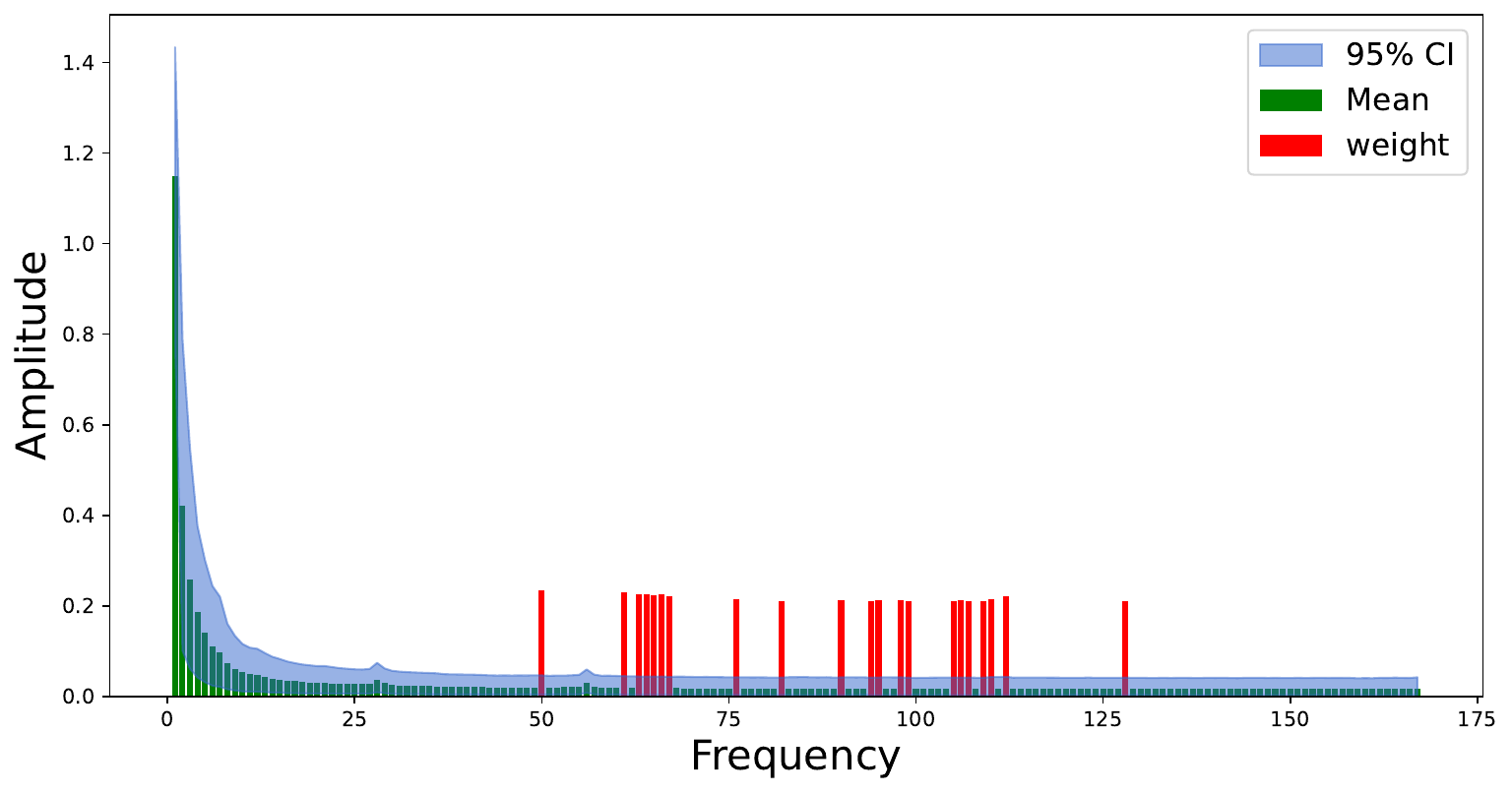}
\end{minipage}%
\vspace{0.5em}
\begin{minipage}[t]{0.48\linewidth}
\centering
\includegraphics[width=0.9\textwidth]{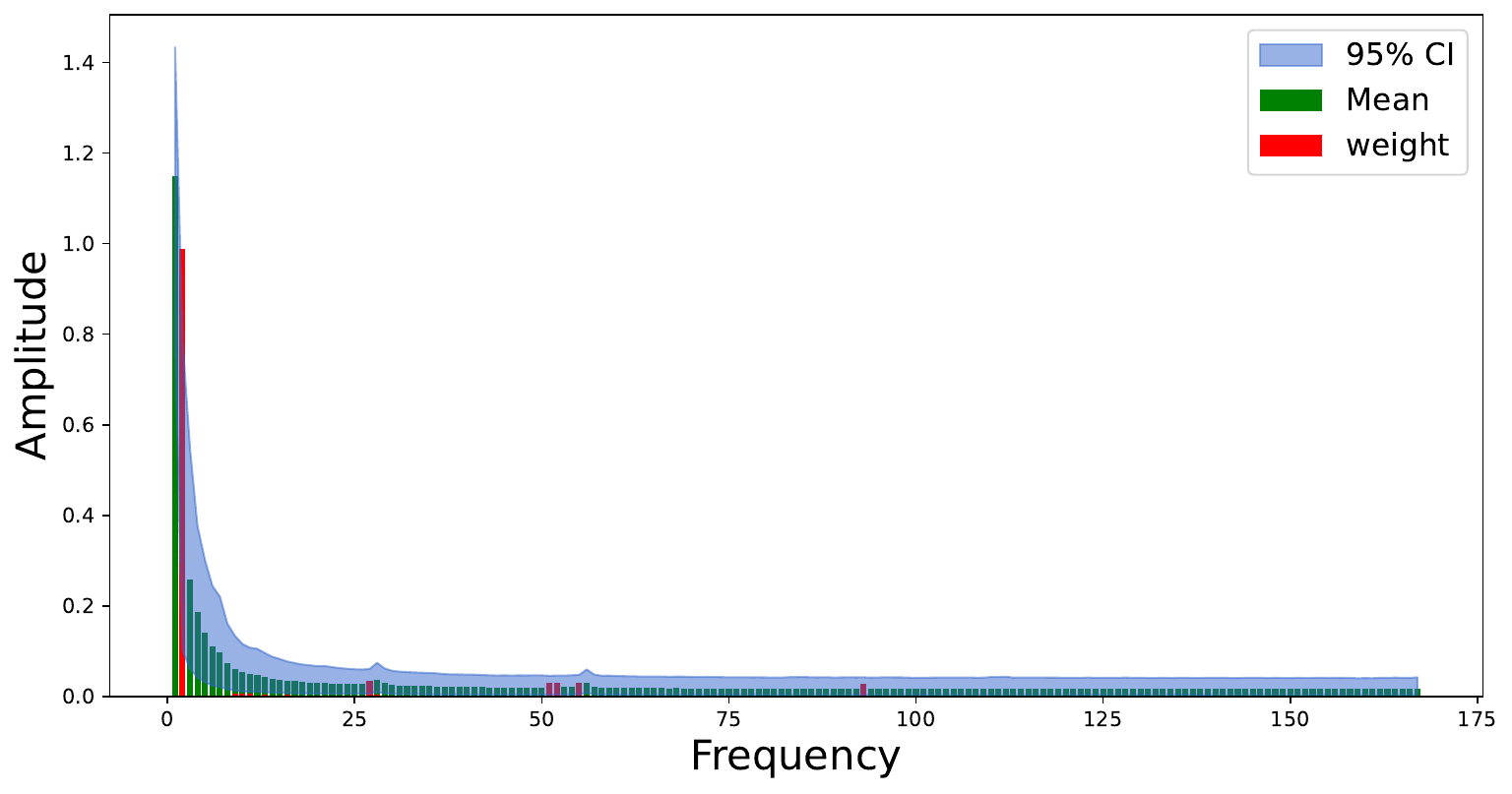}
\end{minipage}
\begin{minipage}[t]{0.48\linewidth}
\centering
\includegraphics[width=0.9\textwidth]{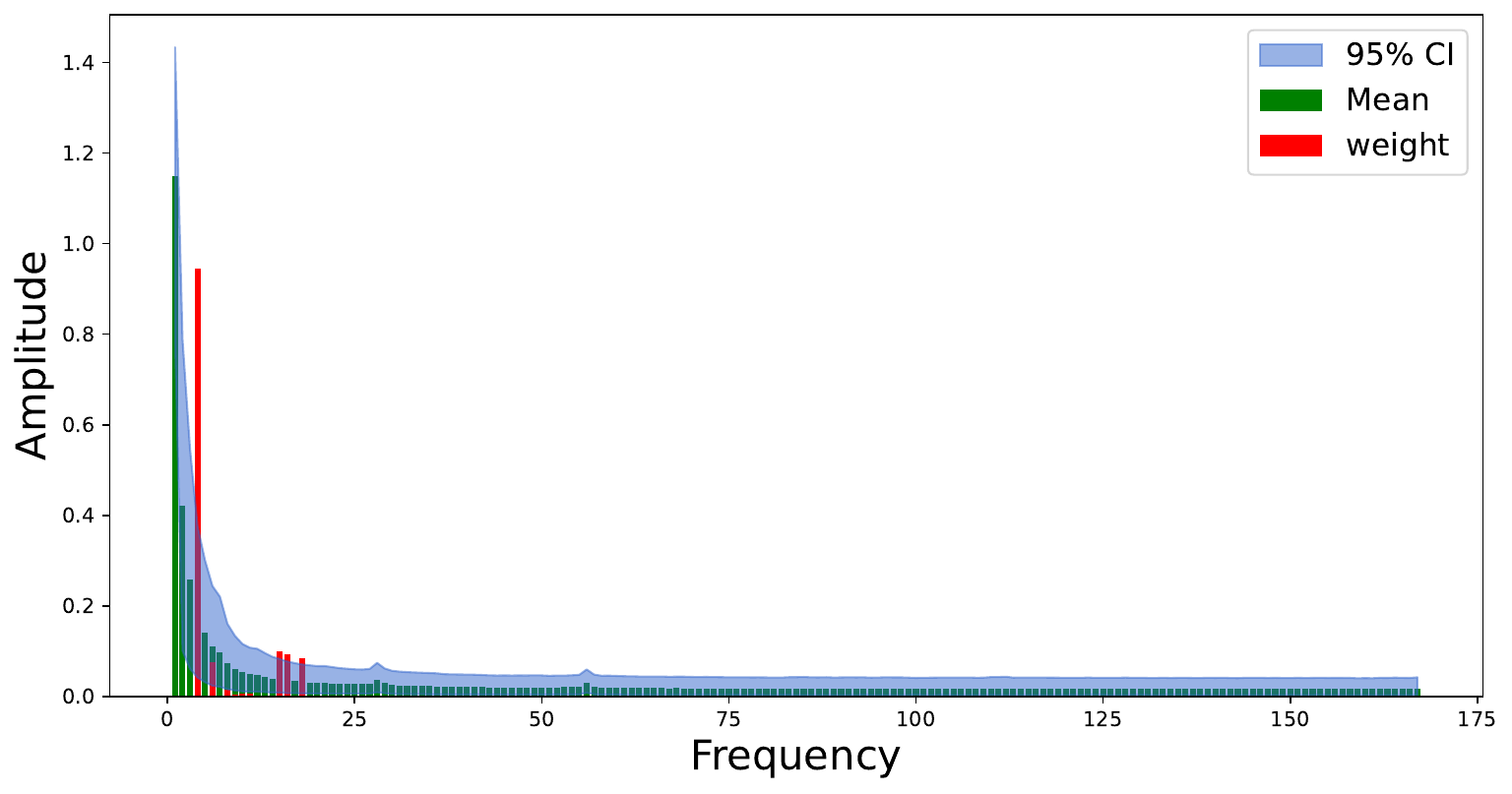}
\end{minipage}
\vspace{0.5em}
\begin{minipage}[t]{0.48\linewidth}
\centering
\includegraphics[width=0.9\textwidth]{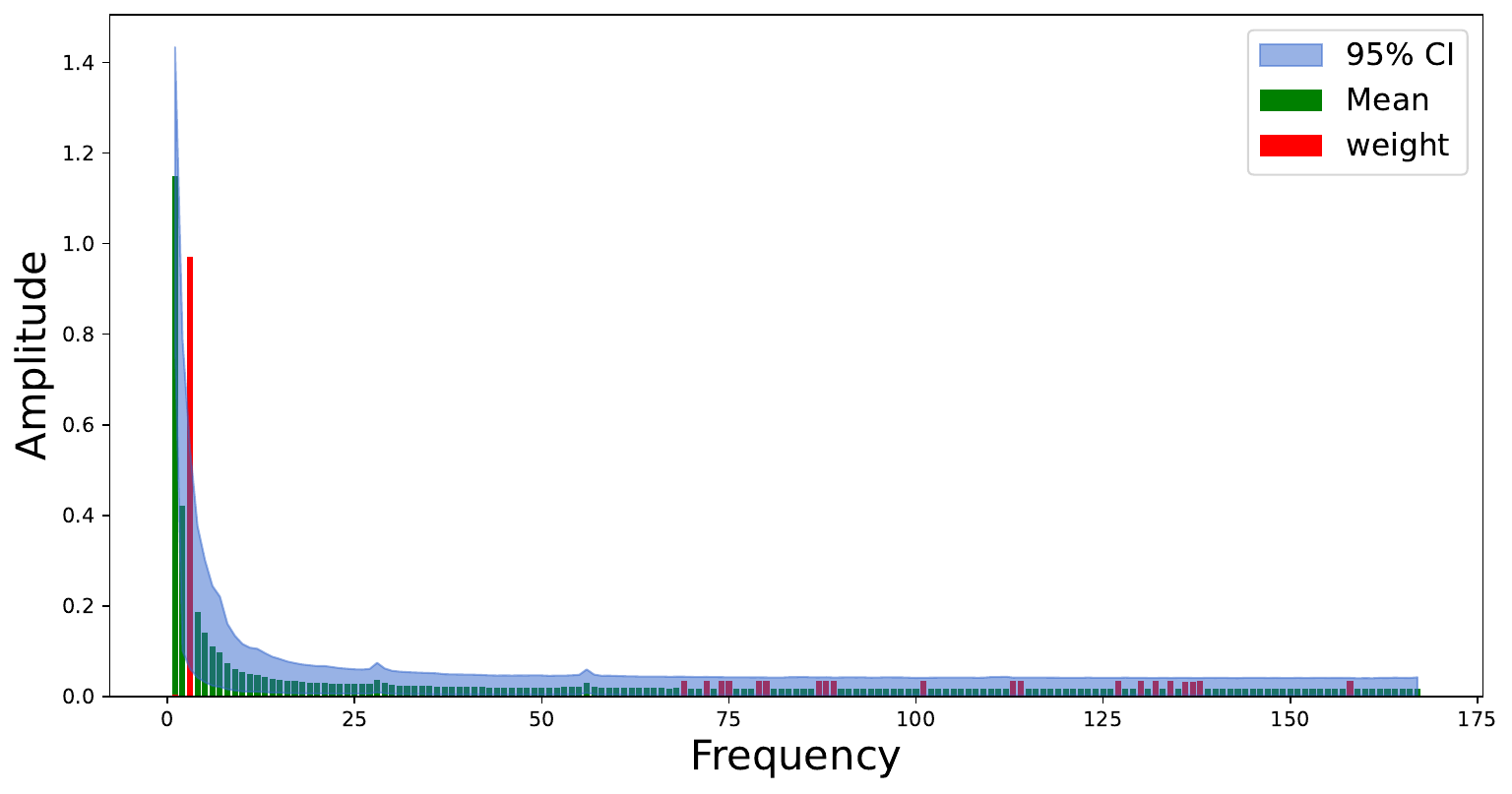}
\end{minipage}
\begin{minipage}[t]{0.48\linewidth}
\centering
\includegraphics[width=0.9\textwidth]{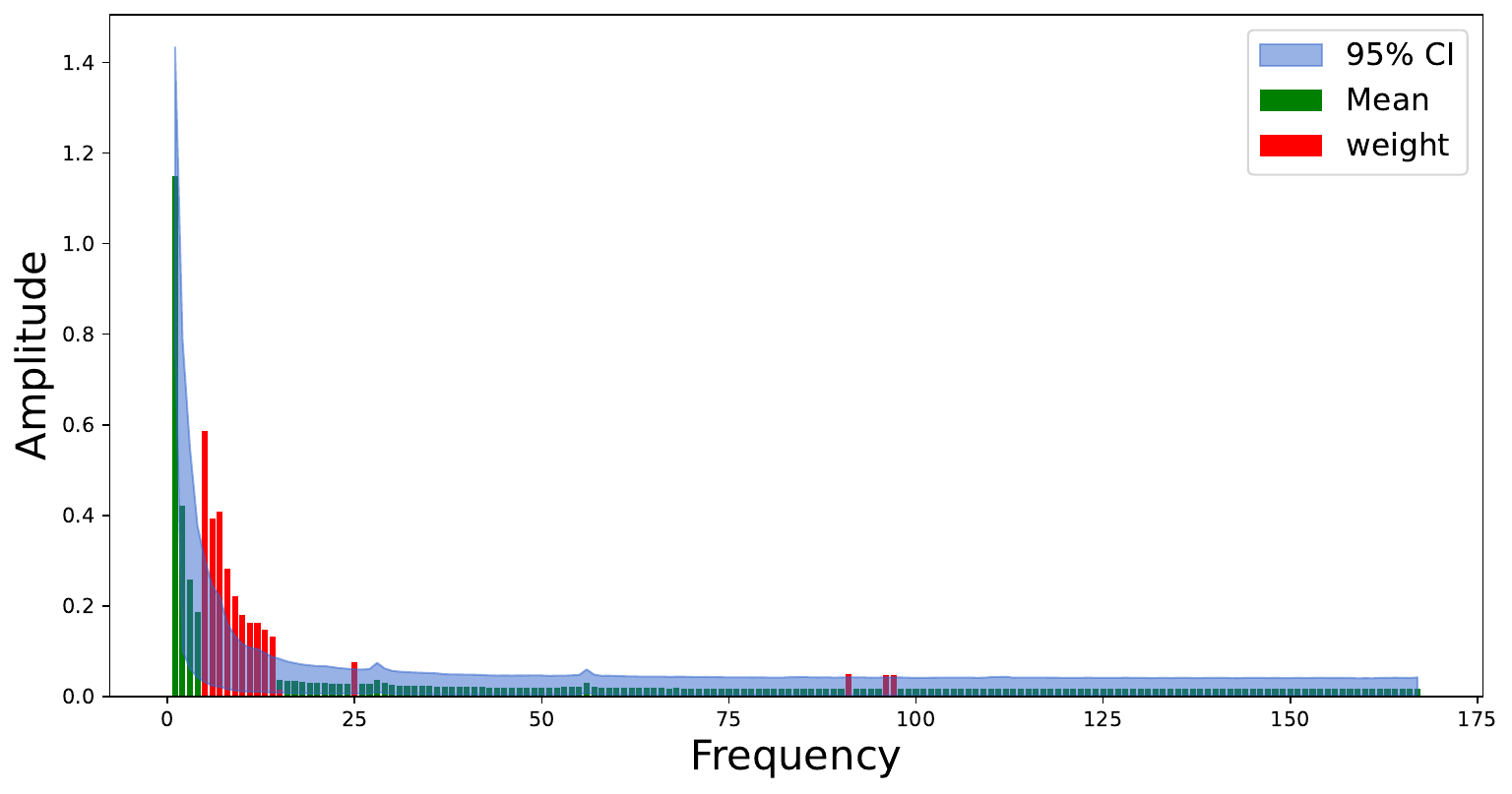}
\end{minipage}%
\vspace{0.5em}
\begin{minipage}[t]{0.48\linewidth}
\centering
\includegraphics[width=0.9\textwidth]{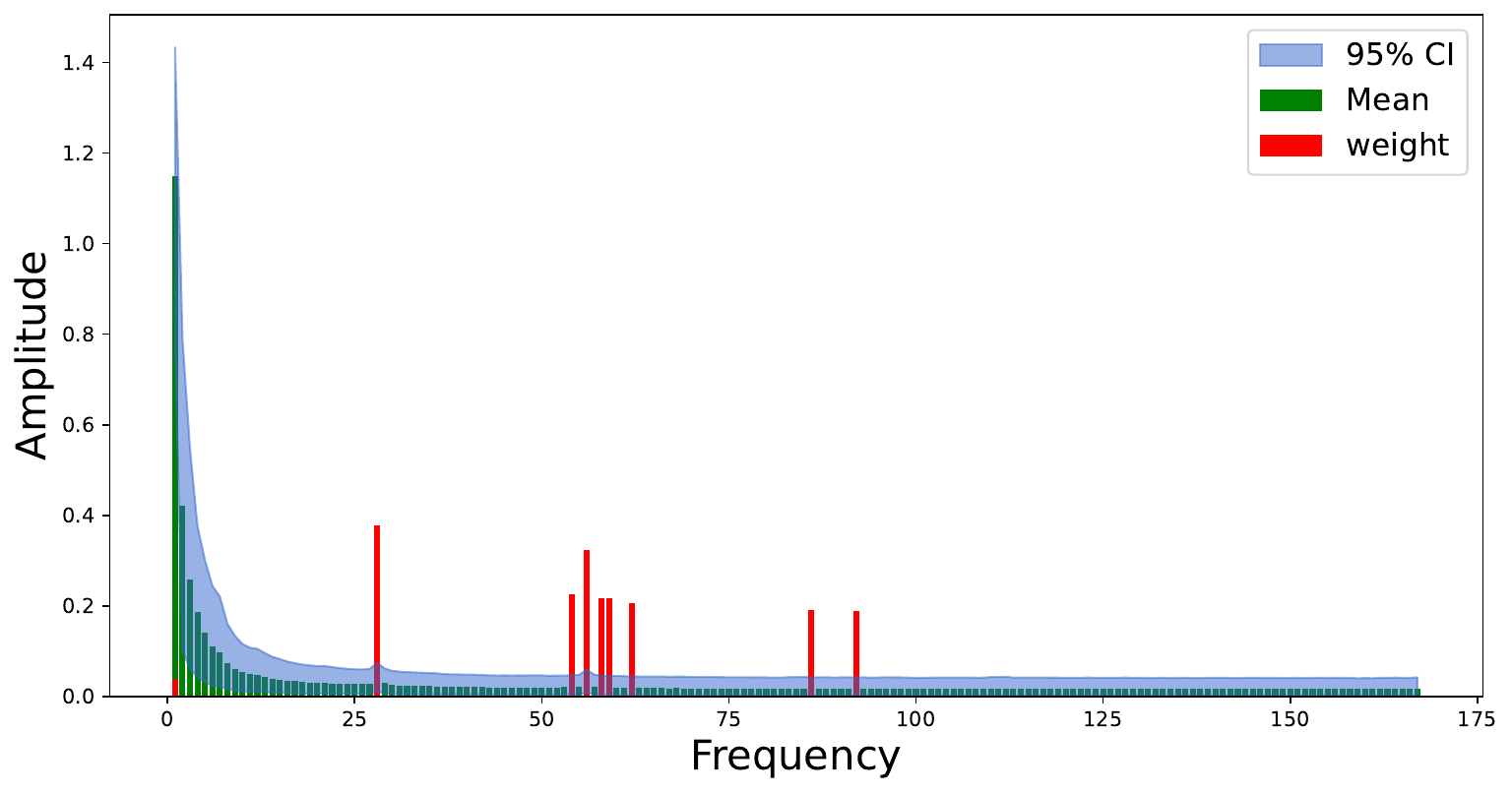}
\end{minipage}%
\begin{minipage}[t]{0.48\linewidth}
\centering
\includegraphics[width=0.9\textwidth]{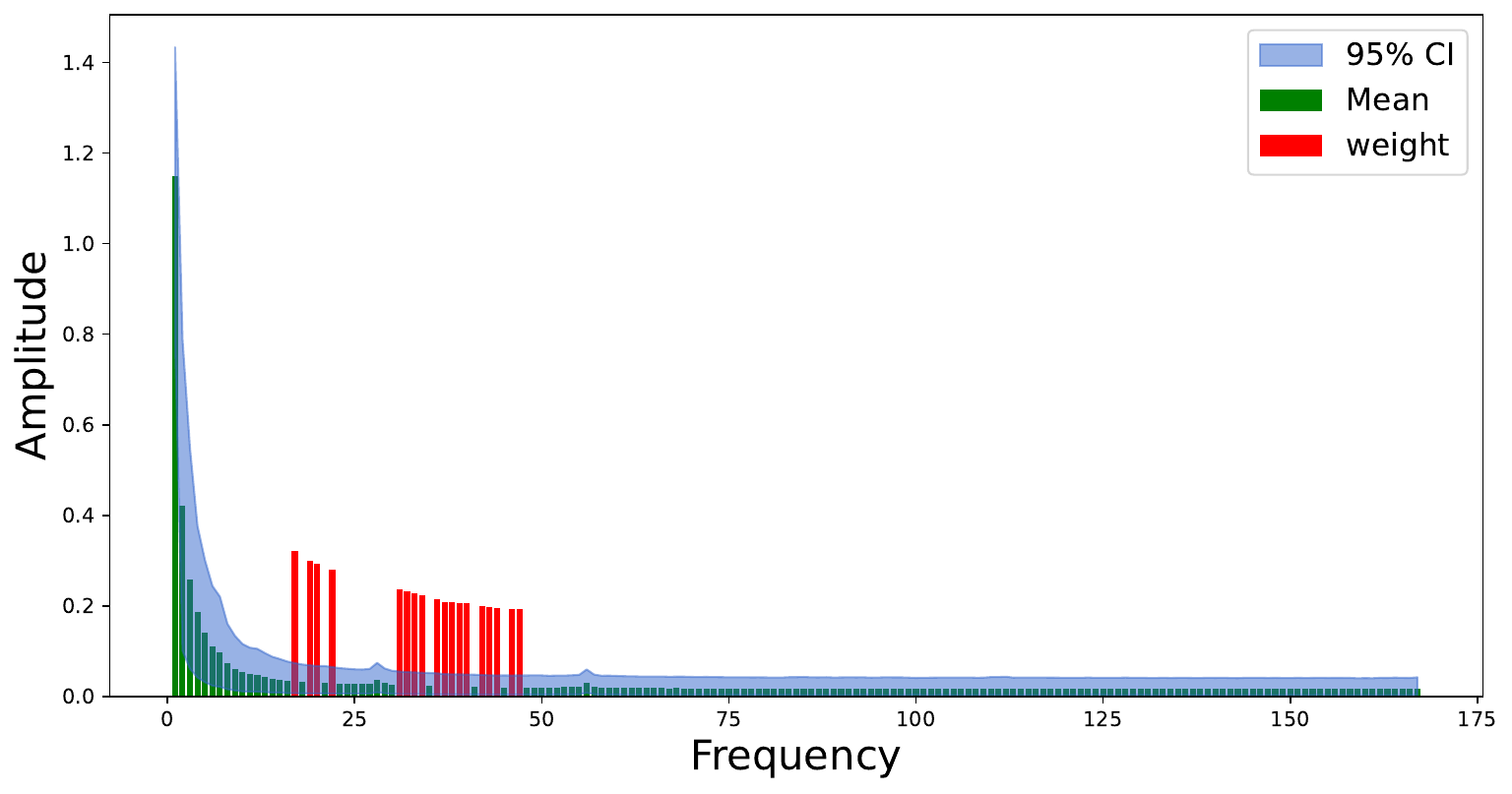}
\end{minipage}
\vspace{0.5em}
\begin{minipage}[t]{0.48\linewidth}
\centering
\includegraphics[width=0.9\textwidth]{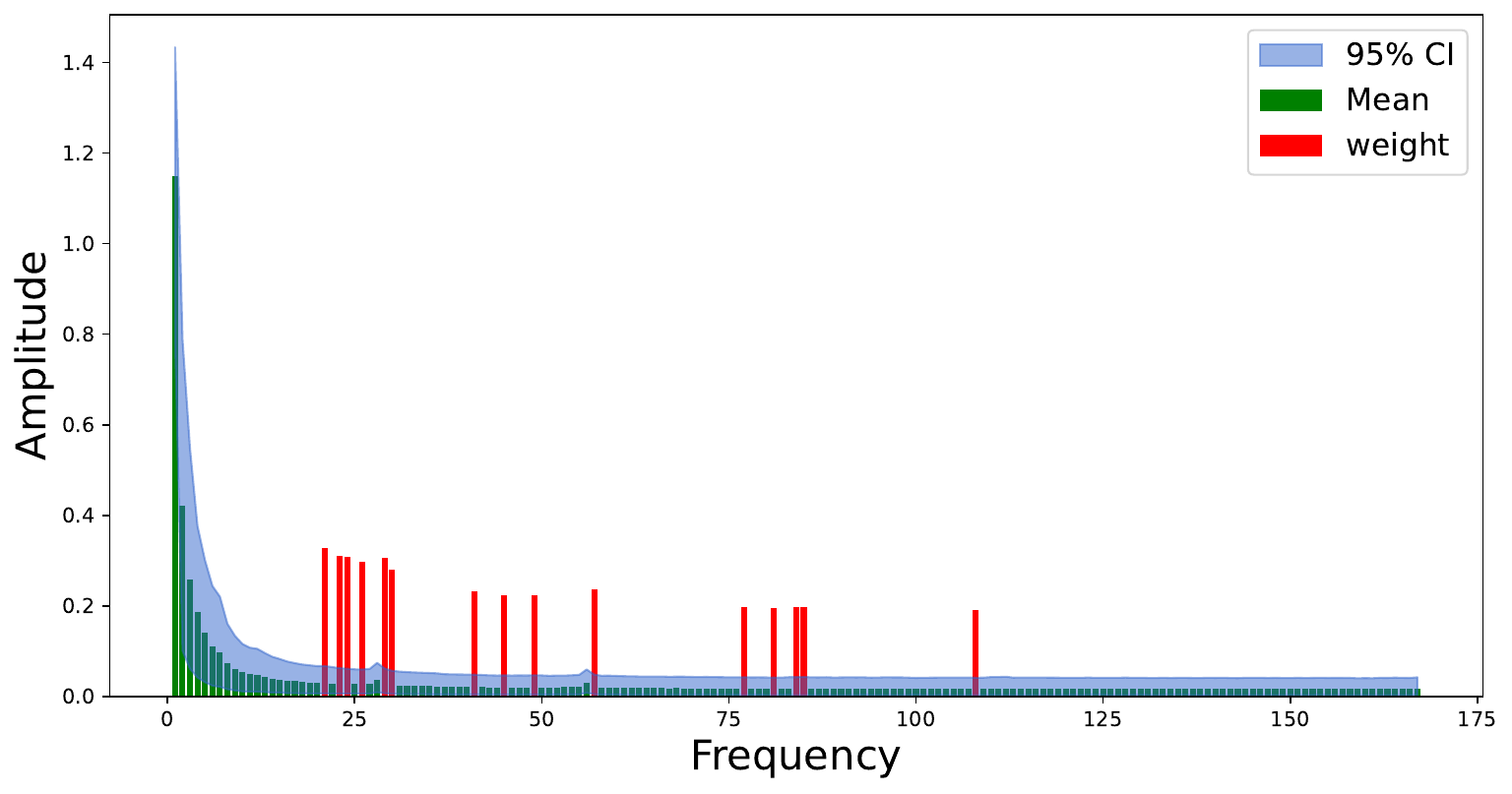}
\end{minipage}
\begin{minipage}[t]{0.48\linewidth}
\centering
\includegraphics[width=0.9\textwidth]{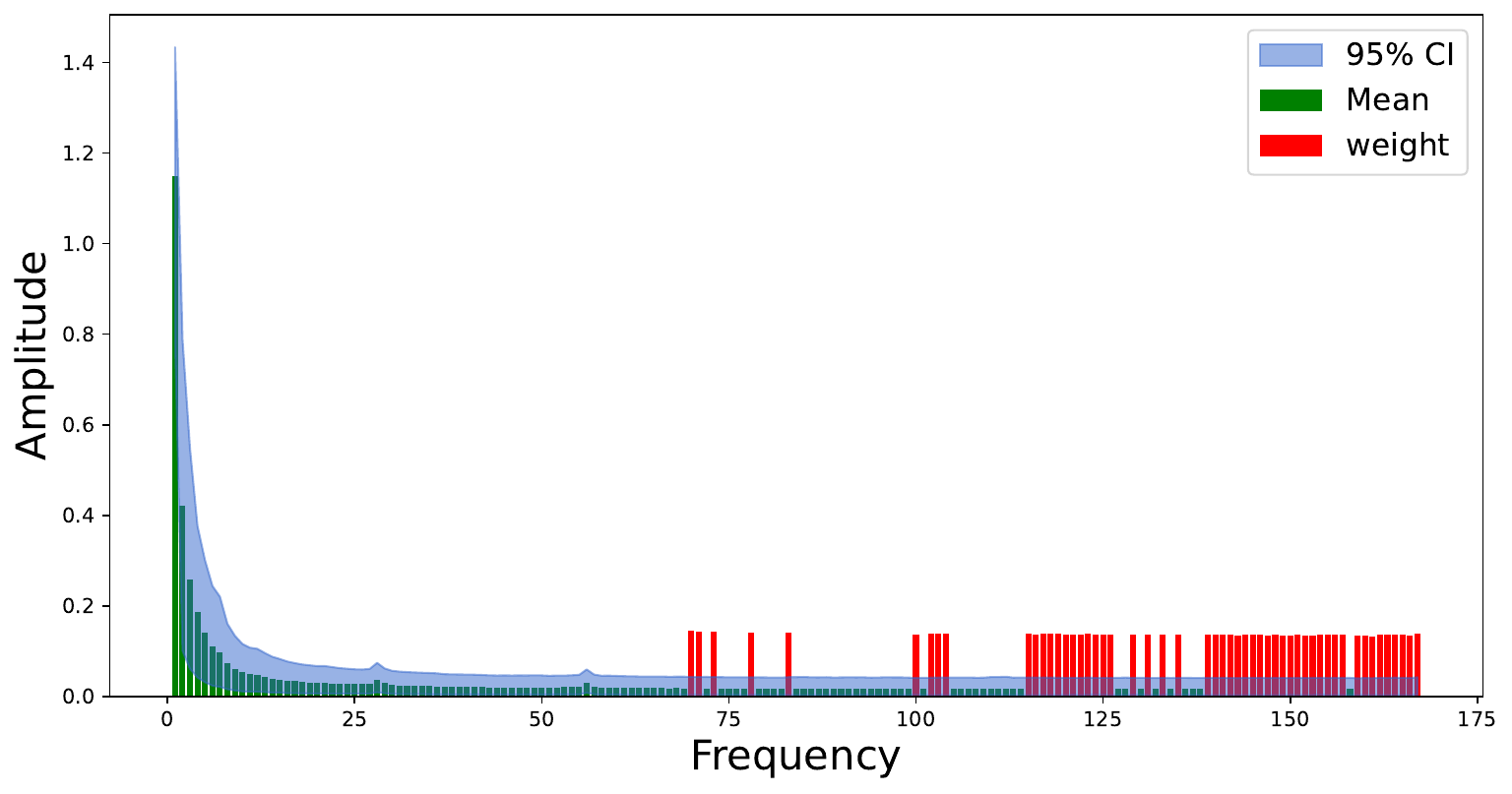}
\end{minipage}
\caption{Learned 10 Components for the PEMS04 data Compared to Its Magnitude Spectrum Distribution. The blue regions correspond to the 95\% confidence interval of magnitude spectrum, the green columns show the mean magnitude spectrum, and the red columns indicate the weights of the learned components.}
\label{components_PEMS04}
\end{figure*}

\begin{figure*}[t]
\centering
\begin{minipage}[t]{0.48\linewidth}
\centering
\includegraphics[width=0.9\textwidth]{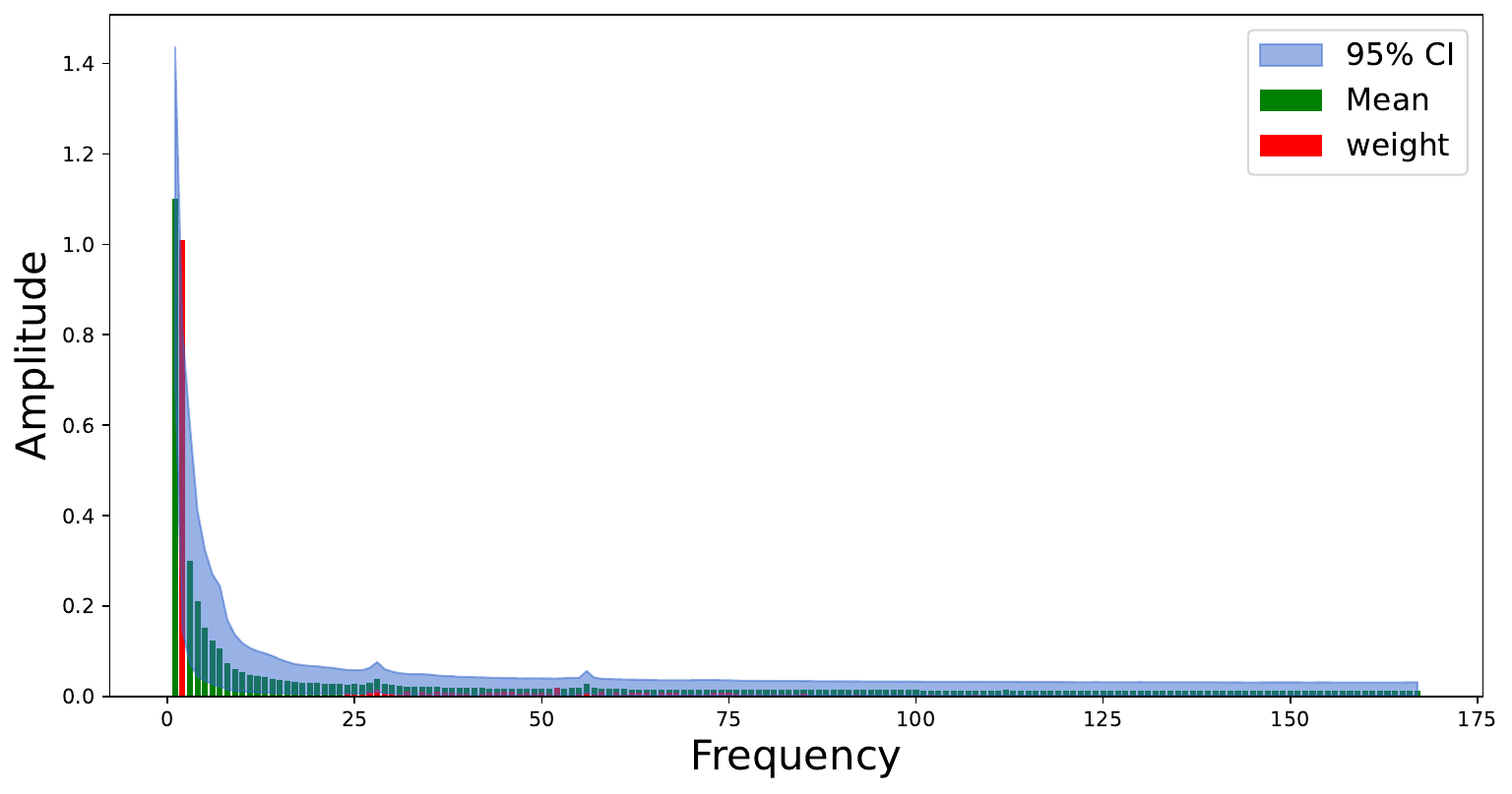}
\end{minipage}%
\begin{minipage}[t]{0.48\linewidth}
\centering
\includegraphics[width=0.9\textwidth]{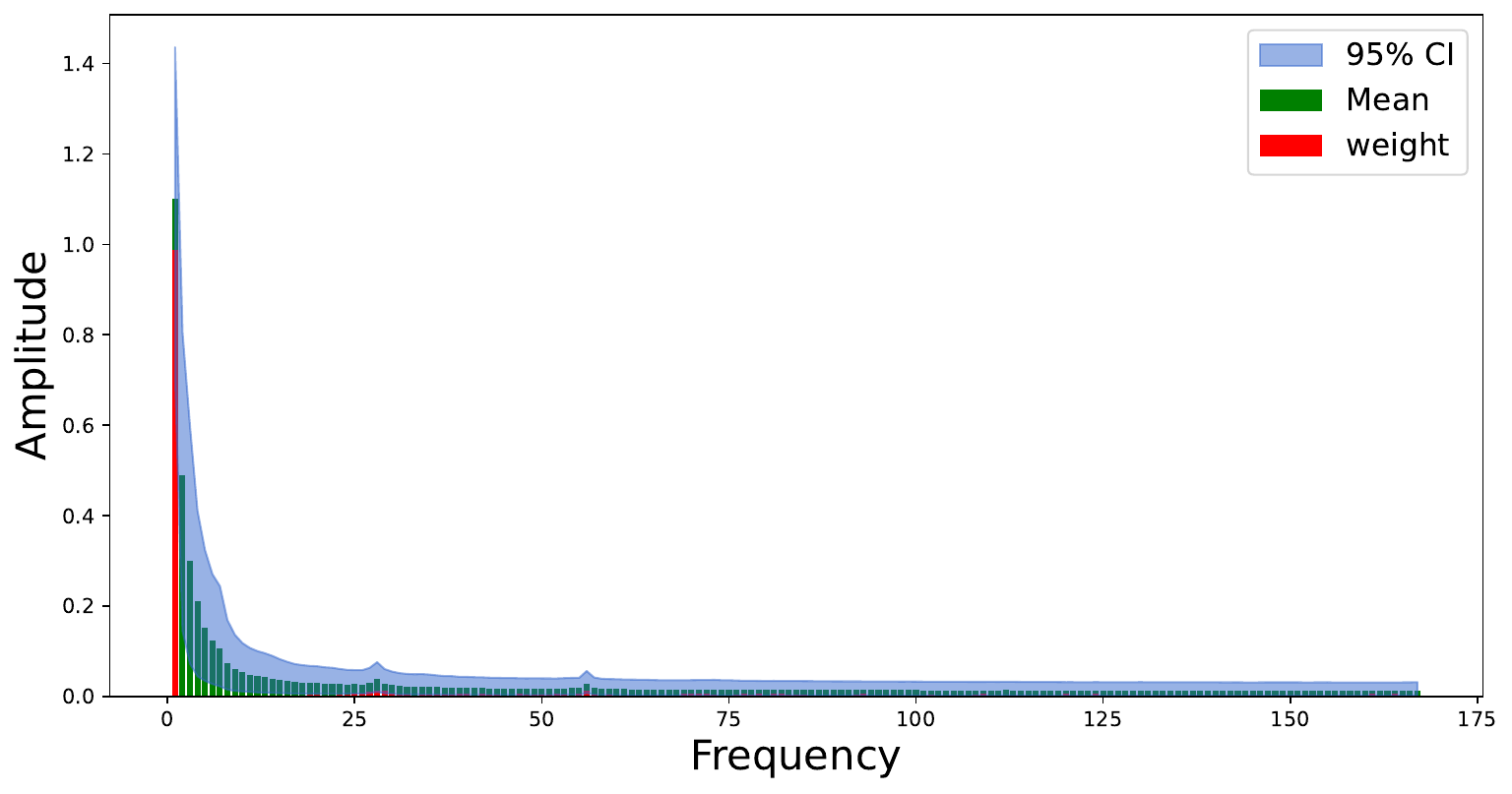}
\end{minipage}%
\vspace{0.5em}
\begin{minipage}[t]{0.48\linewidth}
\centering
\includegraphics[width=0.9\textwidth]{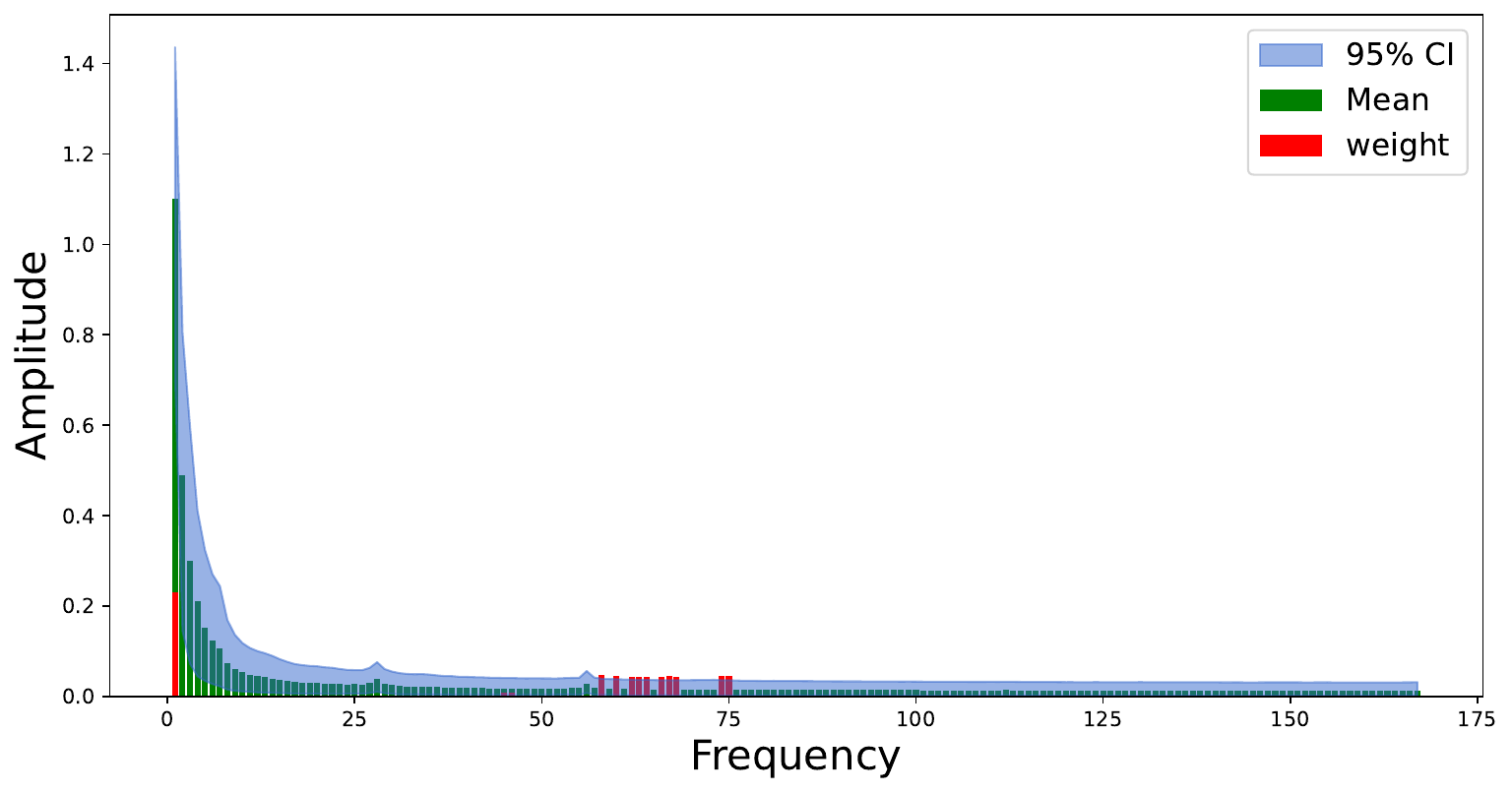}
\end{minipage}
\begin{minipage}[t]{0.48\linewidth}
\centering
\includegraphics[width=0.9\textwidth]{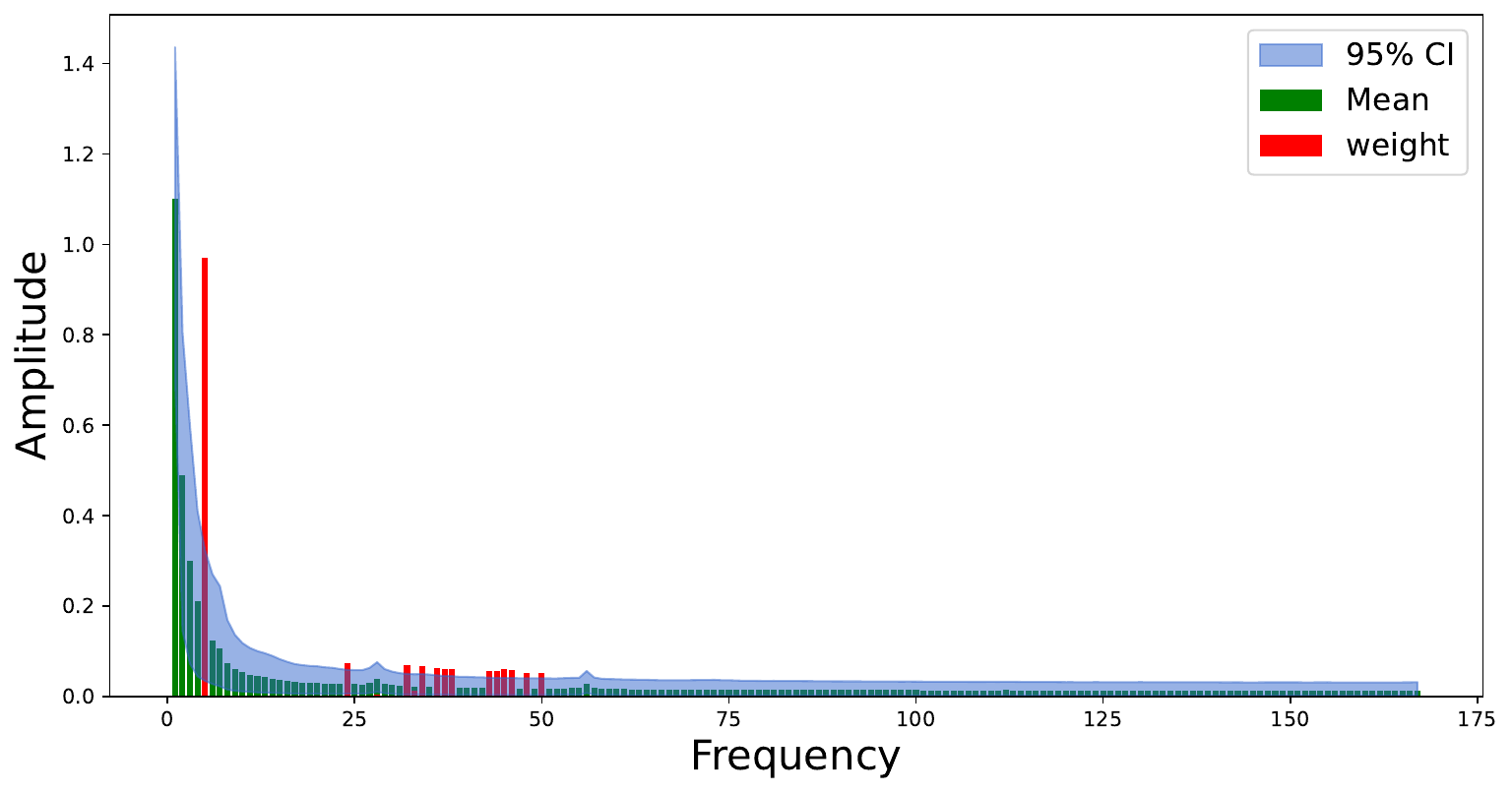}
\end{minipage}
\vspace{0.5em}
\begin{minipage}[t]{0.48\linewidth}
\centering
\includegraphics[width=0.9\textwidth]{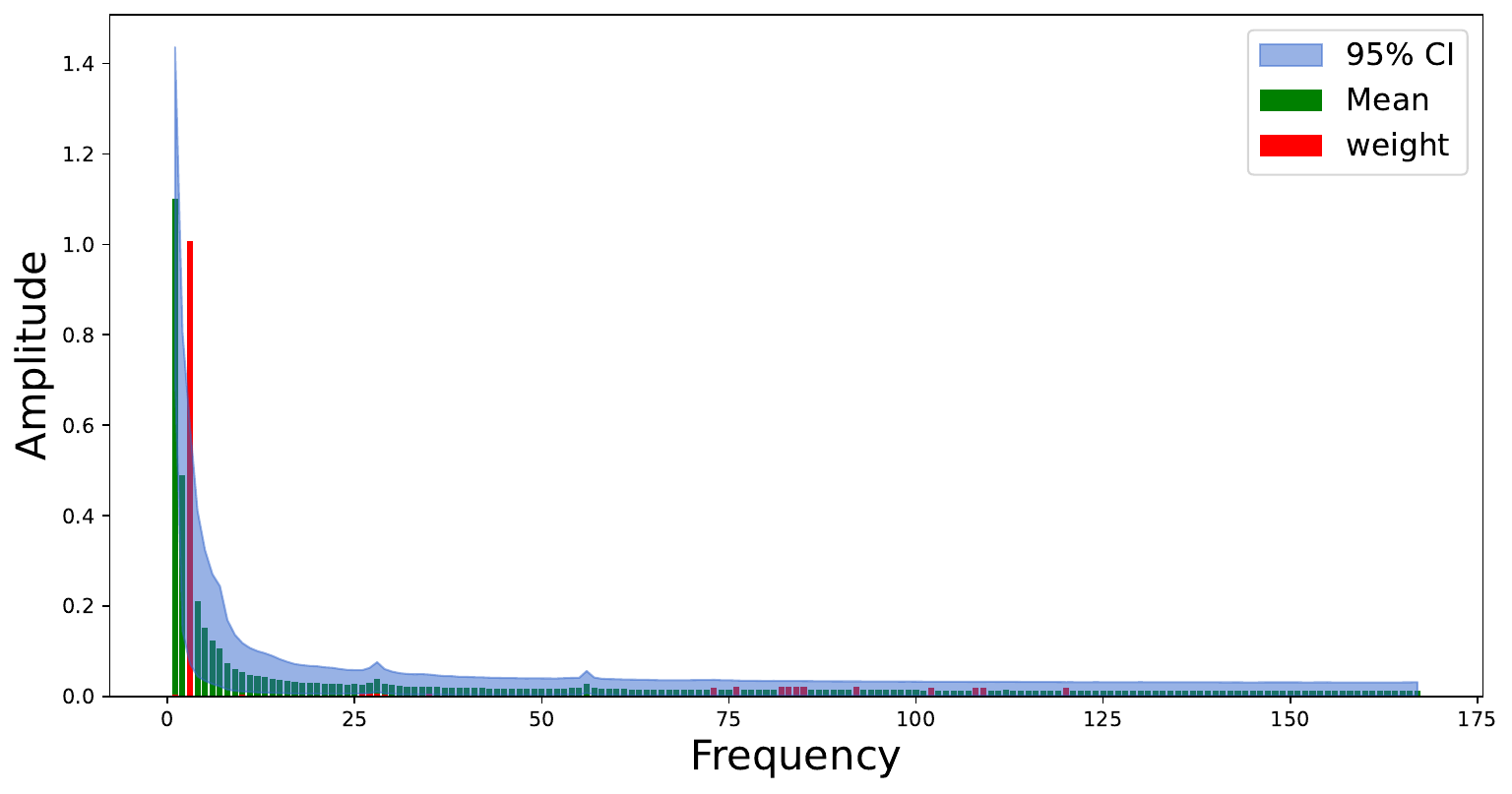}
\end{minipage}
\begin{minipage}[t]{0.48\linewidth}
\centering
\includegraphics[width=0.9\textwidth]{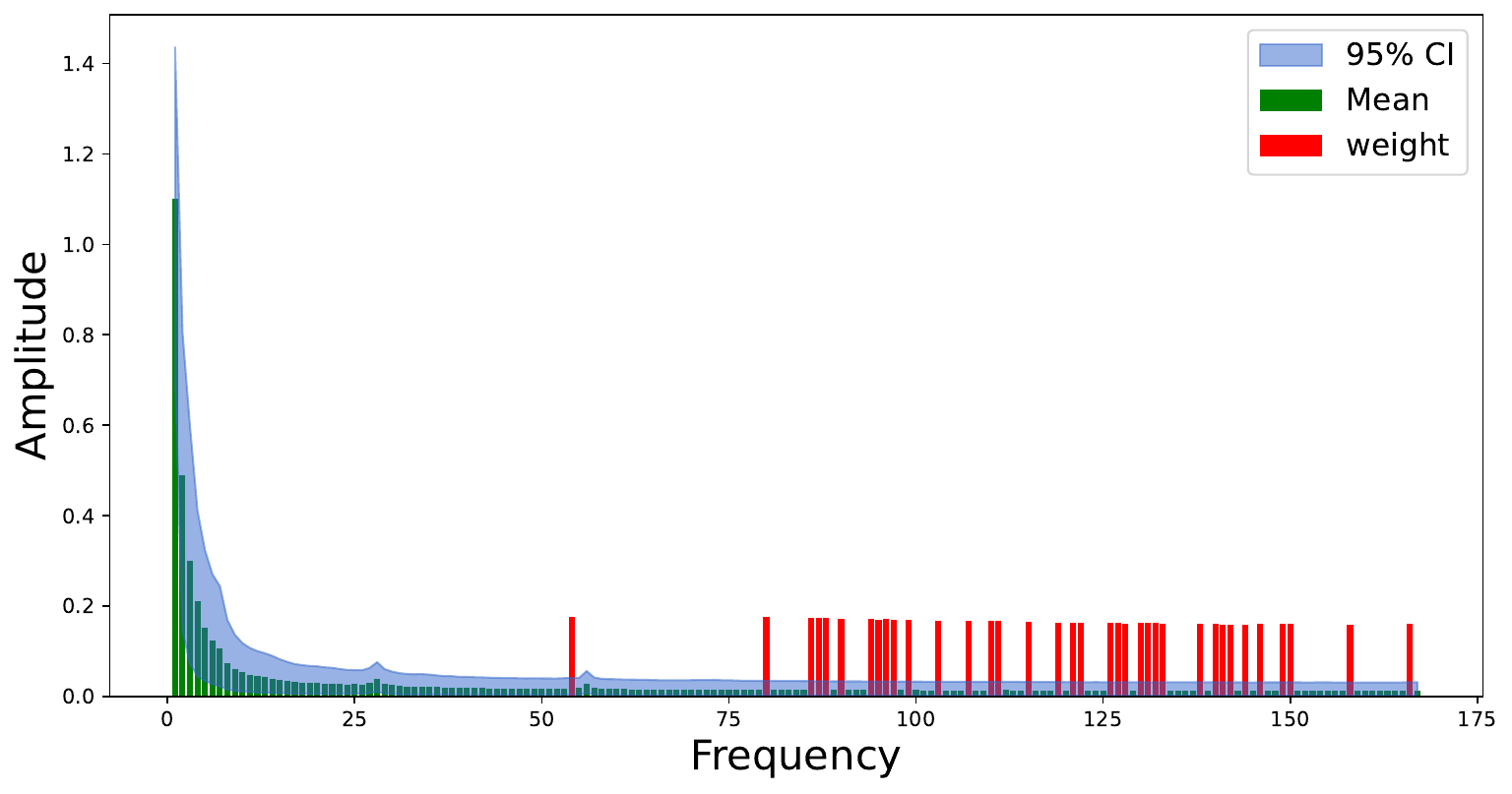}
\end{minipage}%
\vspace{0.5em}
\begin{minipage}[t]{0.48\linewidth}
\centering
\includegraphics[width=0.9\textwidth]{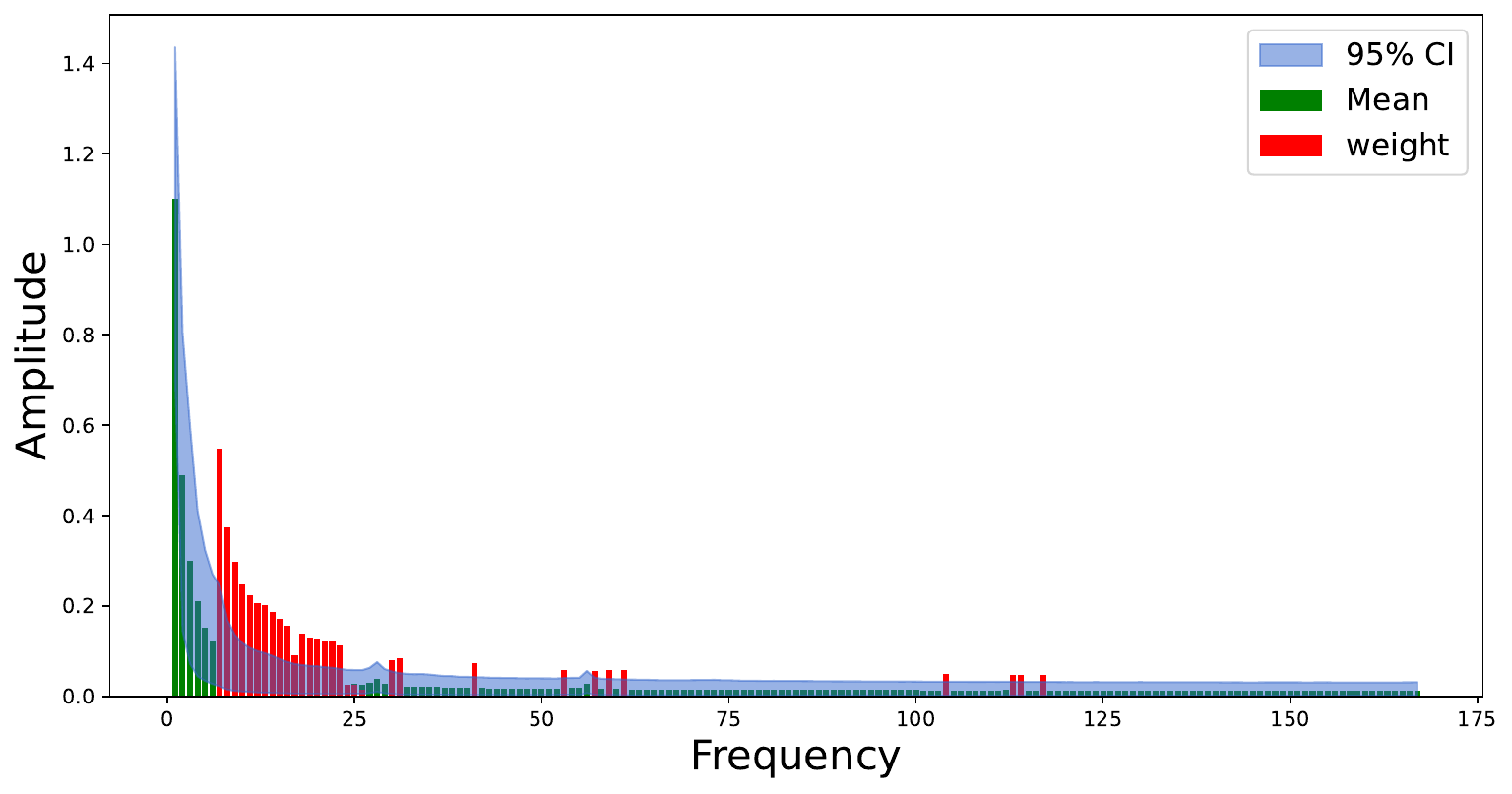}
\end{minipage}%
\begin{minipage}[t]{0.48\linewidth}
\centering
\includegraphics[width=0.9\textwidth]{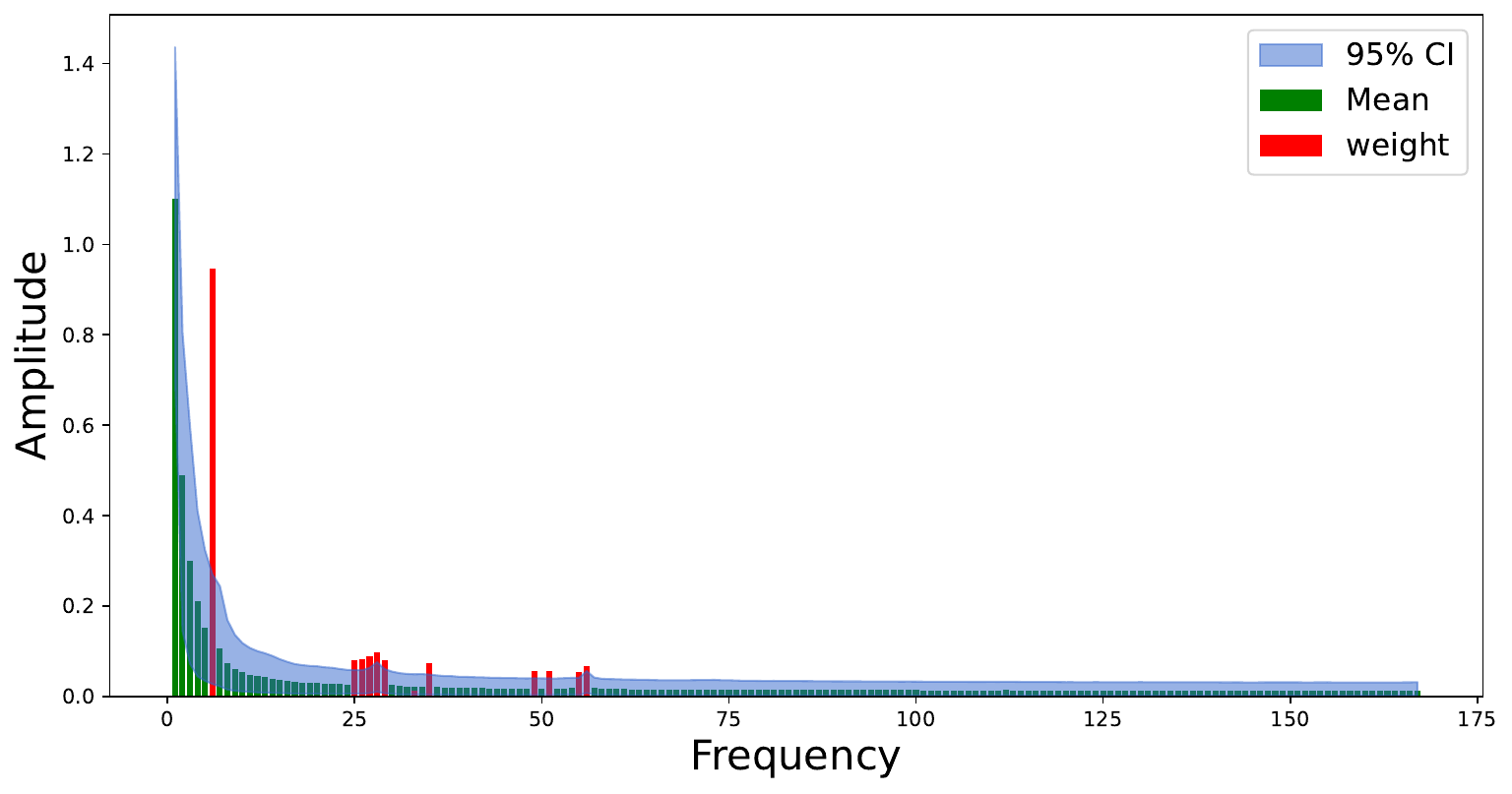}
\end{minipage}
\vspace{0.5em}
\begin{minipage}[t]{0.48\linewidth}
\centering
\includegraphics[width=0.9\textwidth]{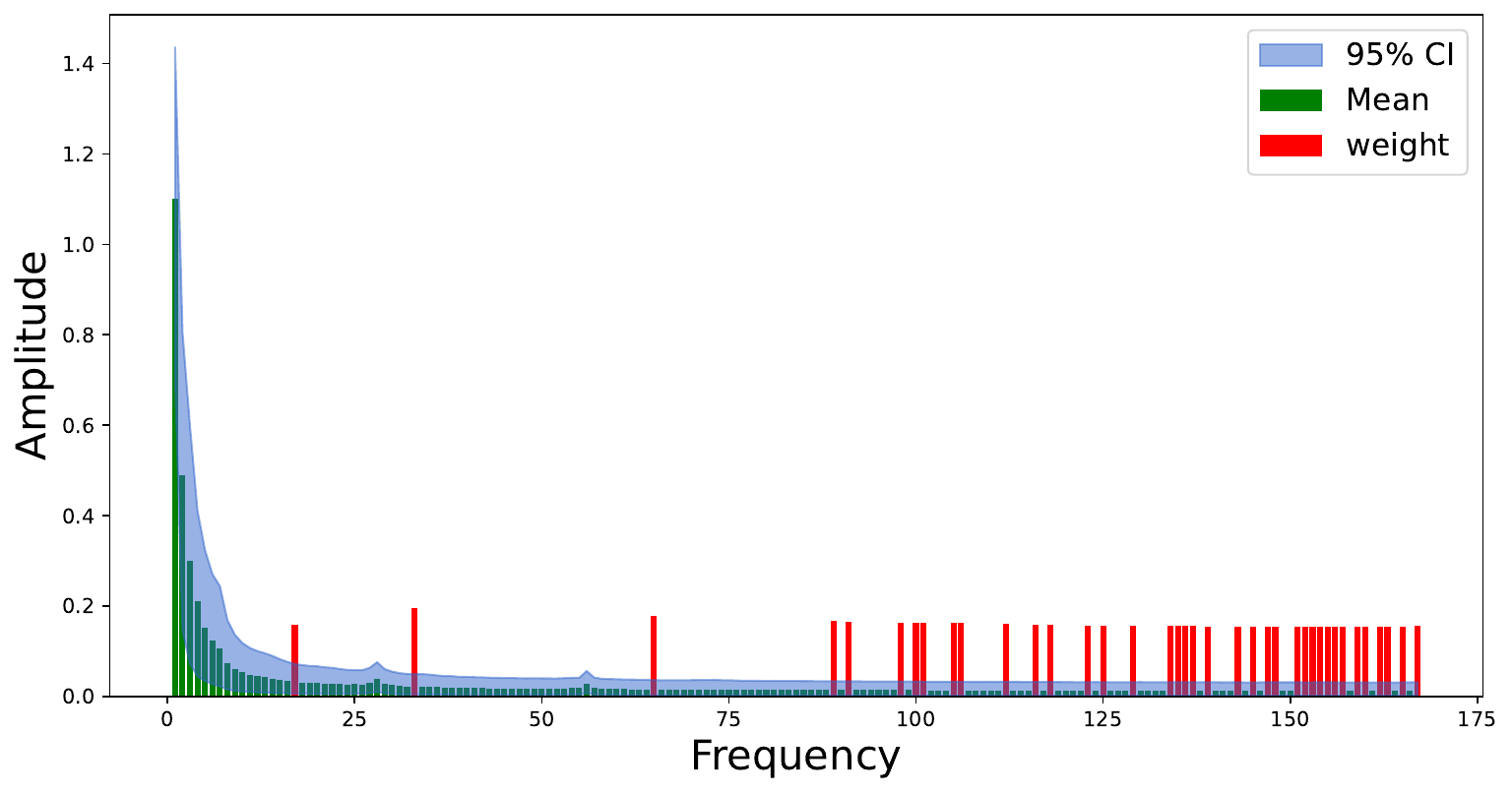}
\end{minipage}
\begin{minipage}[t]{0.48\linewidth}
\centering
\includegraphics[width=0.9\textwidth]{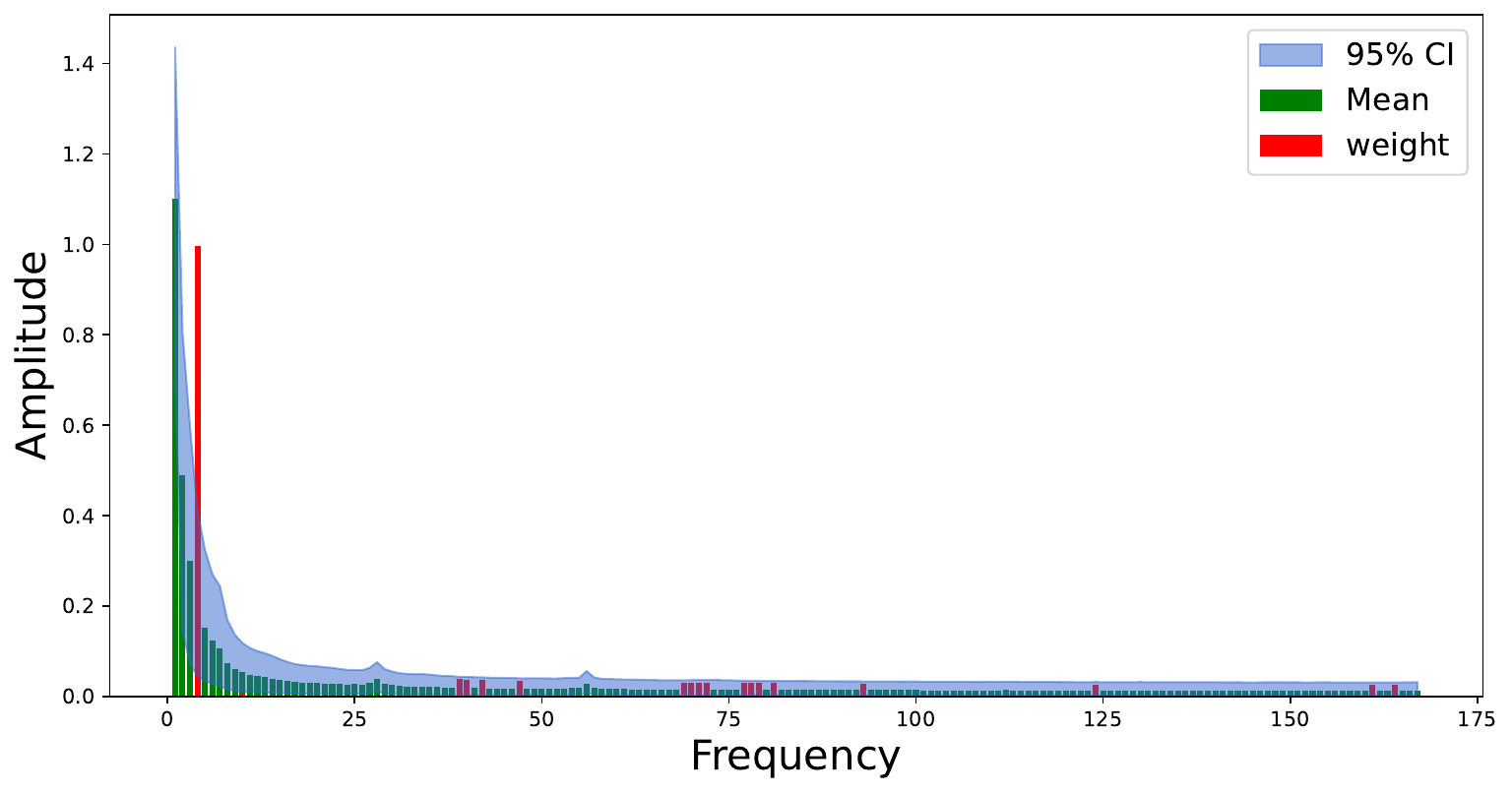}
\end{minipage}
\caption{Learned 10 Components for the PEMS07 data Compared to Its Magnitude Spectrum Distribution. The blue regions correspond to the 95\% confidence interval of magnitude spectrum, the green columns show the mean magnitude spectrum, and the red columns indicate the weights of the learned components.}
\label{components_PEMS07}
\end{figure*}

\begin{figure*}[t]
\centering
\begin{minipage}[t]{0.48\linewidth}
\centering
\includegraphics[width=0.9\textwidth]{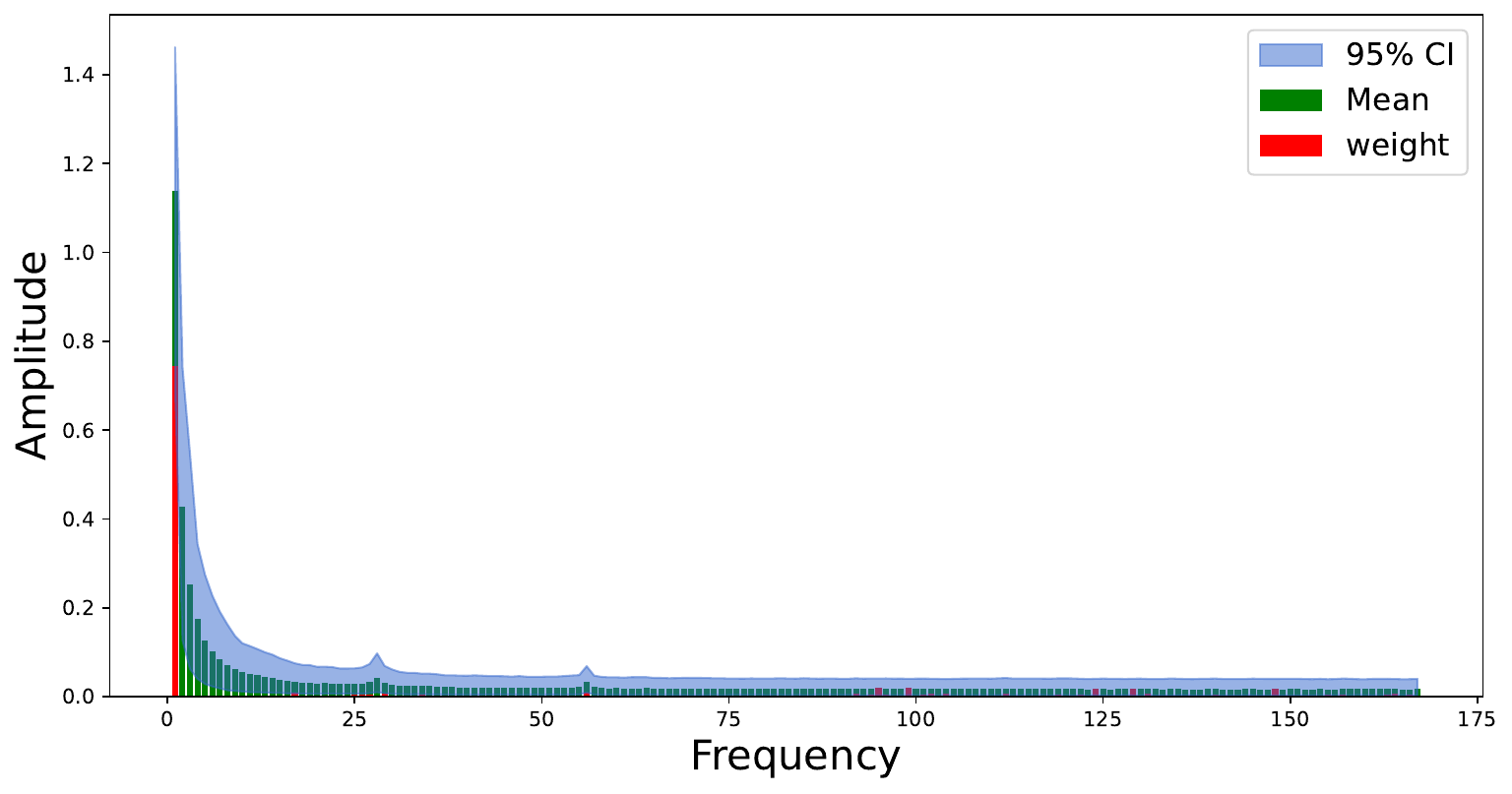}
\end{minipage}%
\begin{minipage}[t]{0.48\linewidth}
\centering
\includegraphics[width=0.9\textwidth]{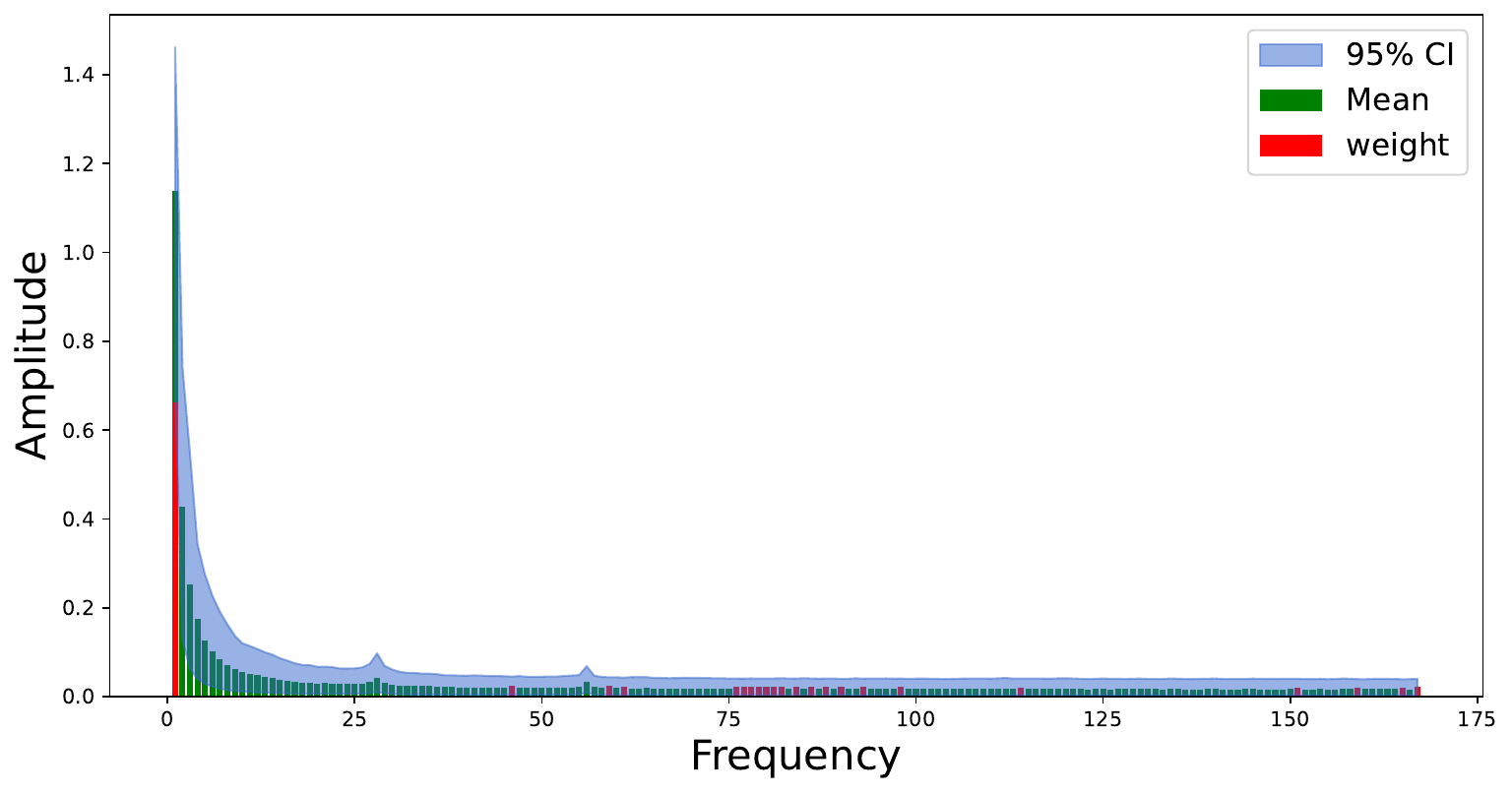}
\end{minipage}%
\vspace{0.5em}
\begin{minipage}[t]{0.48\linewidth}
\centering
\includegraphics[width=0.9\textwidth]{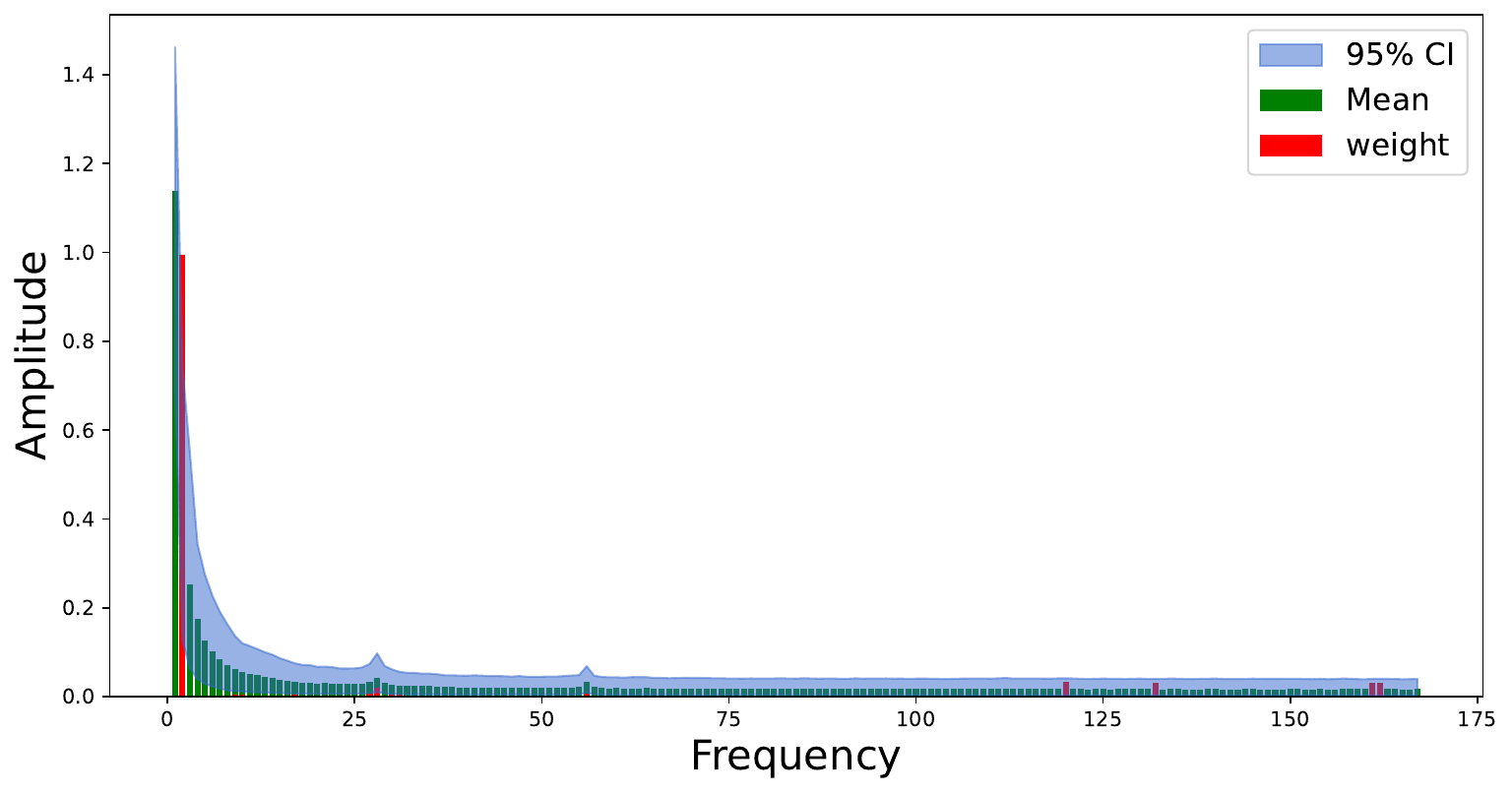}
\end{minipage}
\begin{minipage}[t]{0.48\linewidth}
\centering
\includegraphics[width=0.9\textwidth]{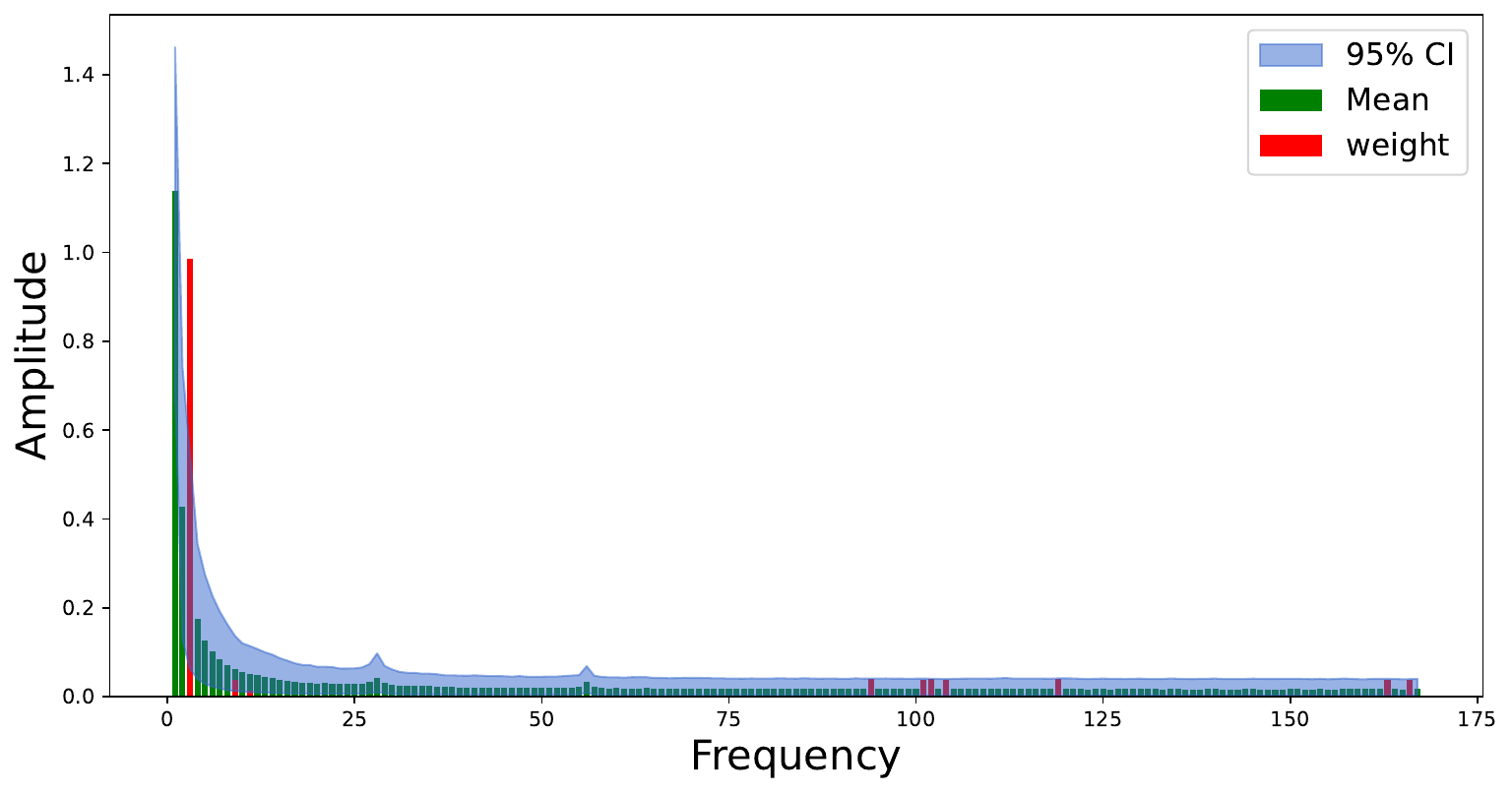}
\end{minipage}
\vspace{0.5em}
\begin{minipage}[t]{0.48\linewidth}
\centering
\includegraphics[width=0.9\textwidth]{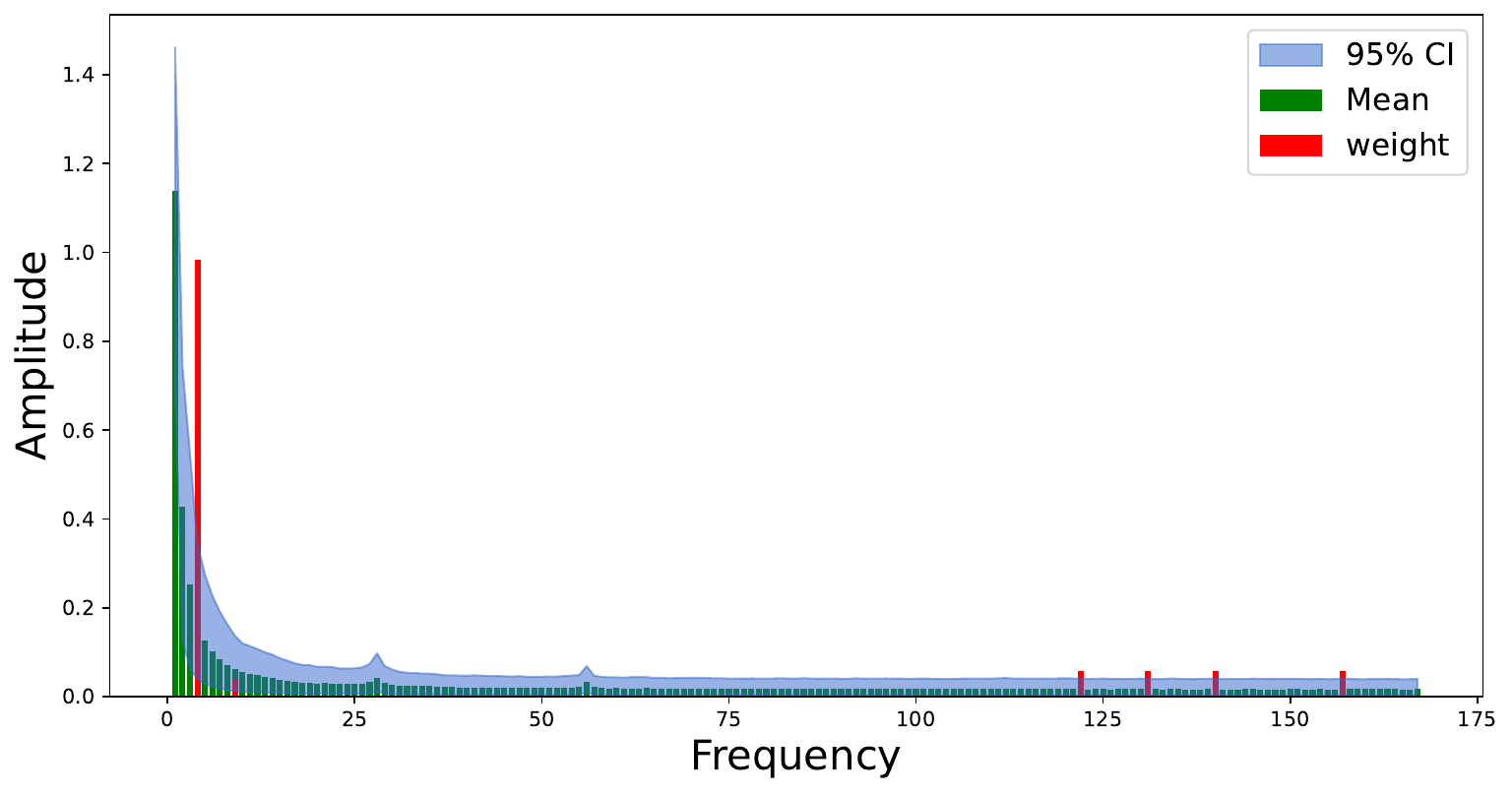}
\end{minipage}
\begin{minipage}[t]{0.48\linewidth}
\centering
\includegraphics[width=0.9\textwidth]{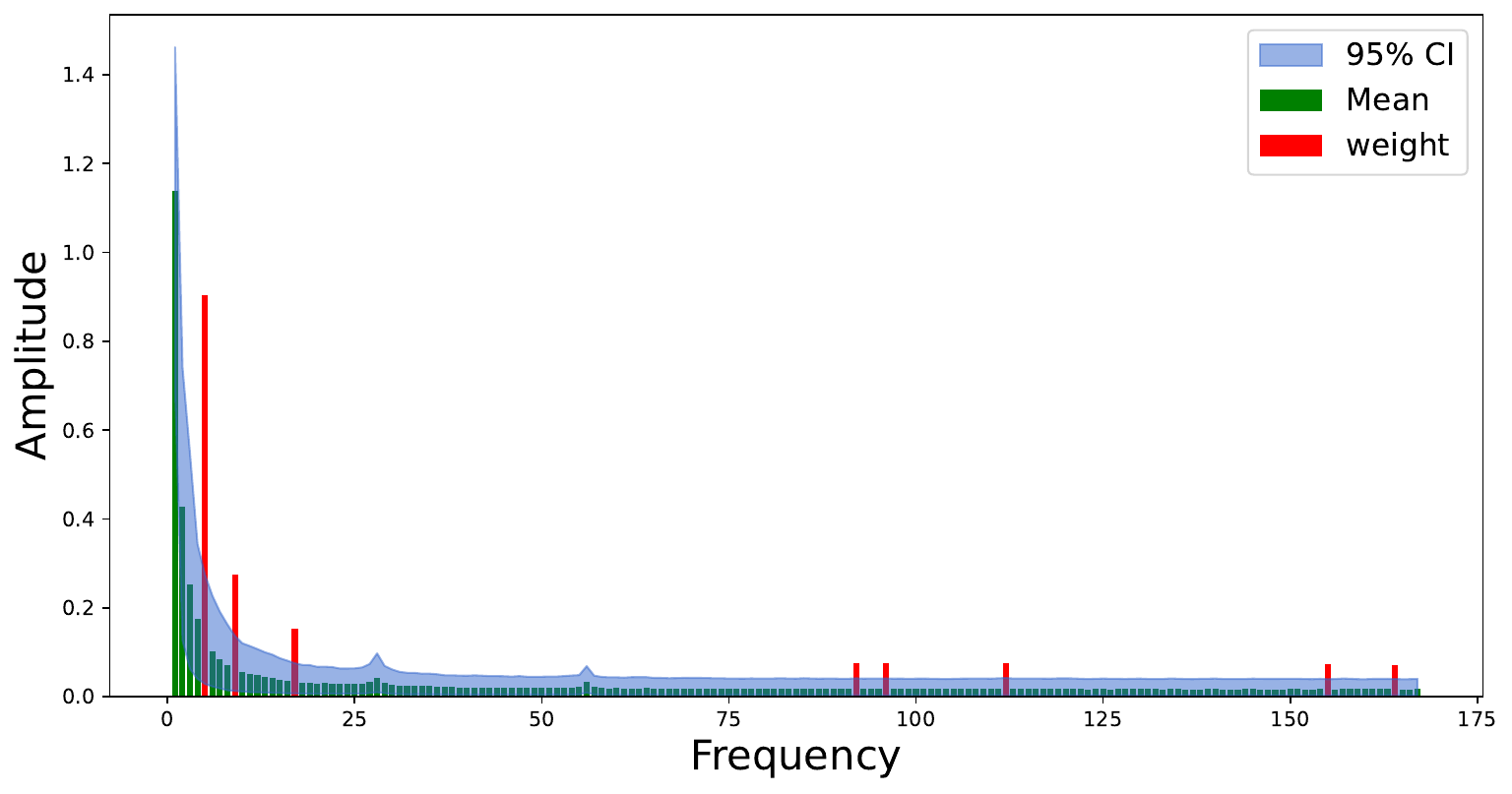}
\end{minipage}%
\vspace{0.5em}
\begin{minipage}[t]{0.48\linewidth}
\centering
\includegraphics[width=0.9\textwidth]{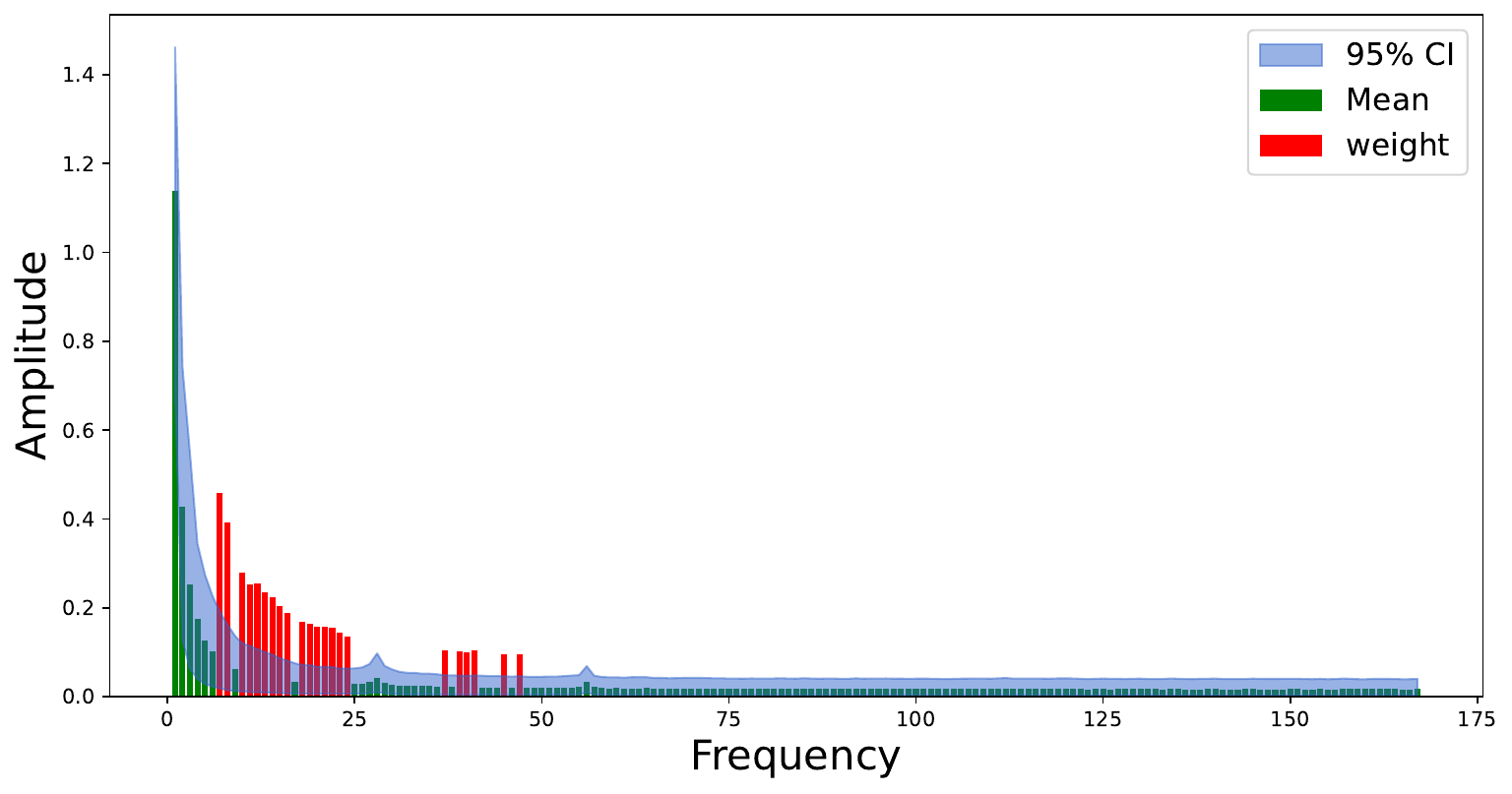}
\end{minipage}%
\begin{minipage}[t]{0.48\linewidth}
\centering
\includegraphics[width=0.9\textwidth]{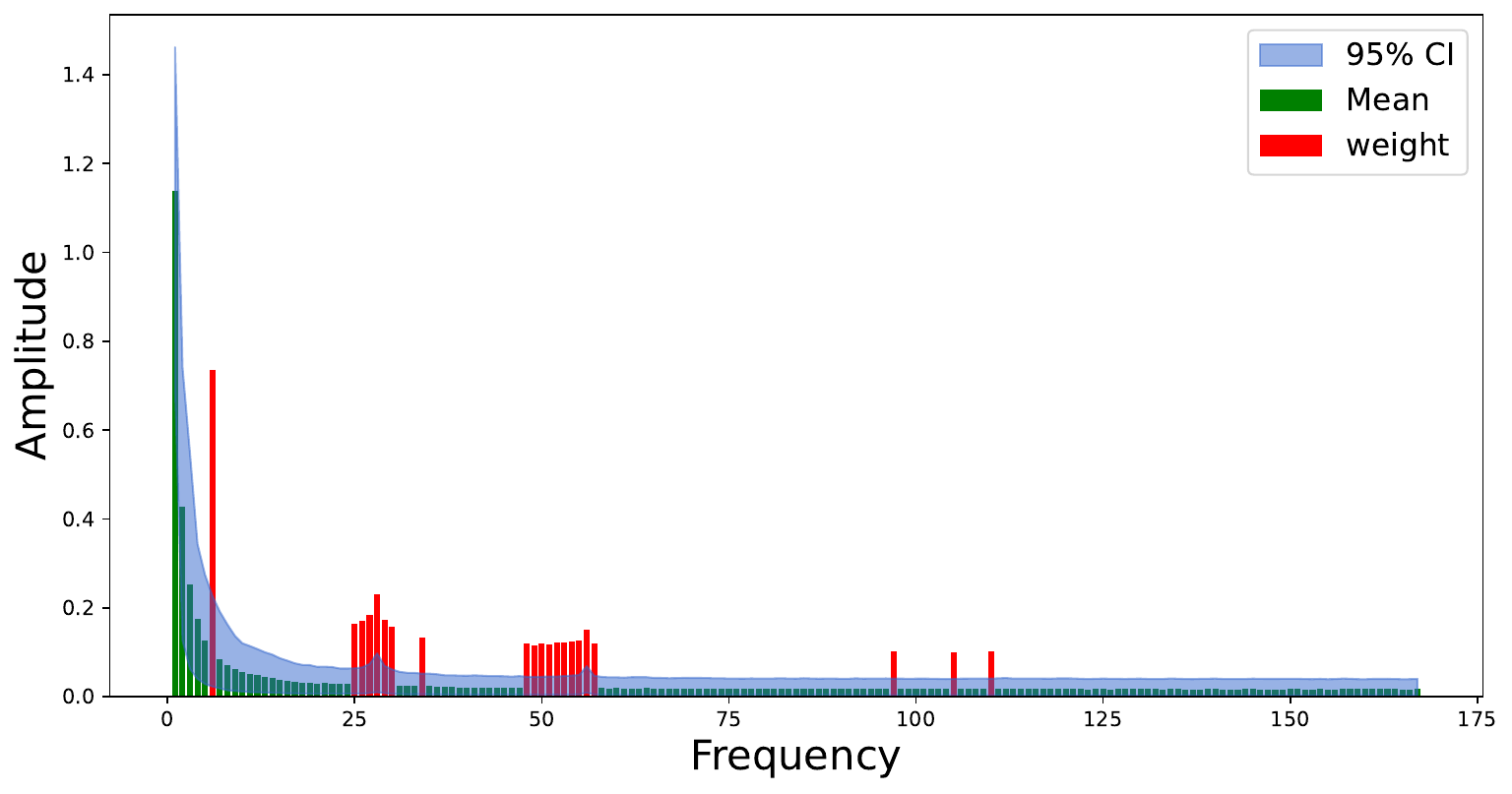}
\end{minipage}
\vspace{0.5em}
\begin{minipage}[t]{0.48\linewidth}
\centering
\includegraphics[width=0.9\textwidth]{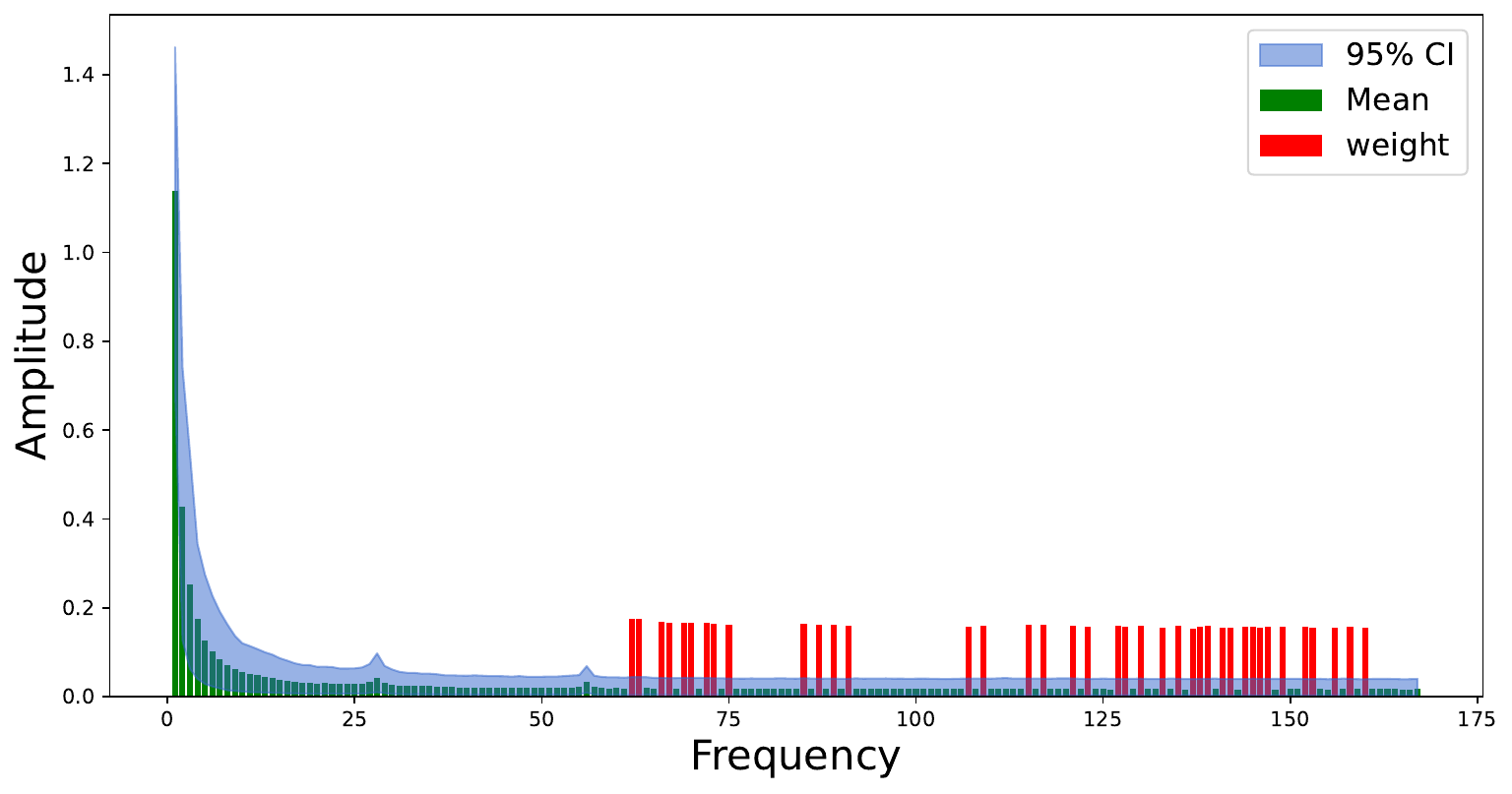}
\end{minipage}
\begin{minipage}[t]{0.48\linewidth}
\centering
\includegraphics[width=0.9\textwidth]{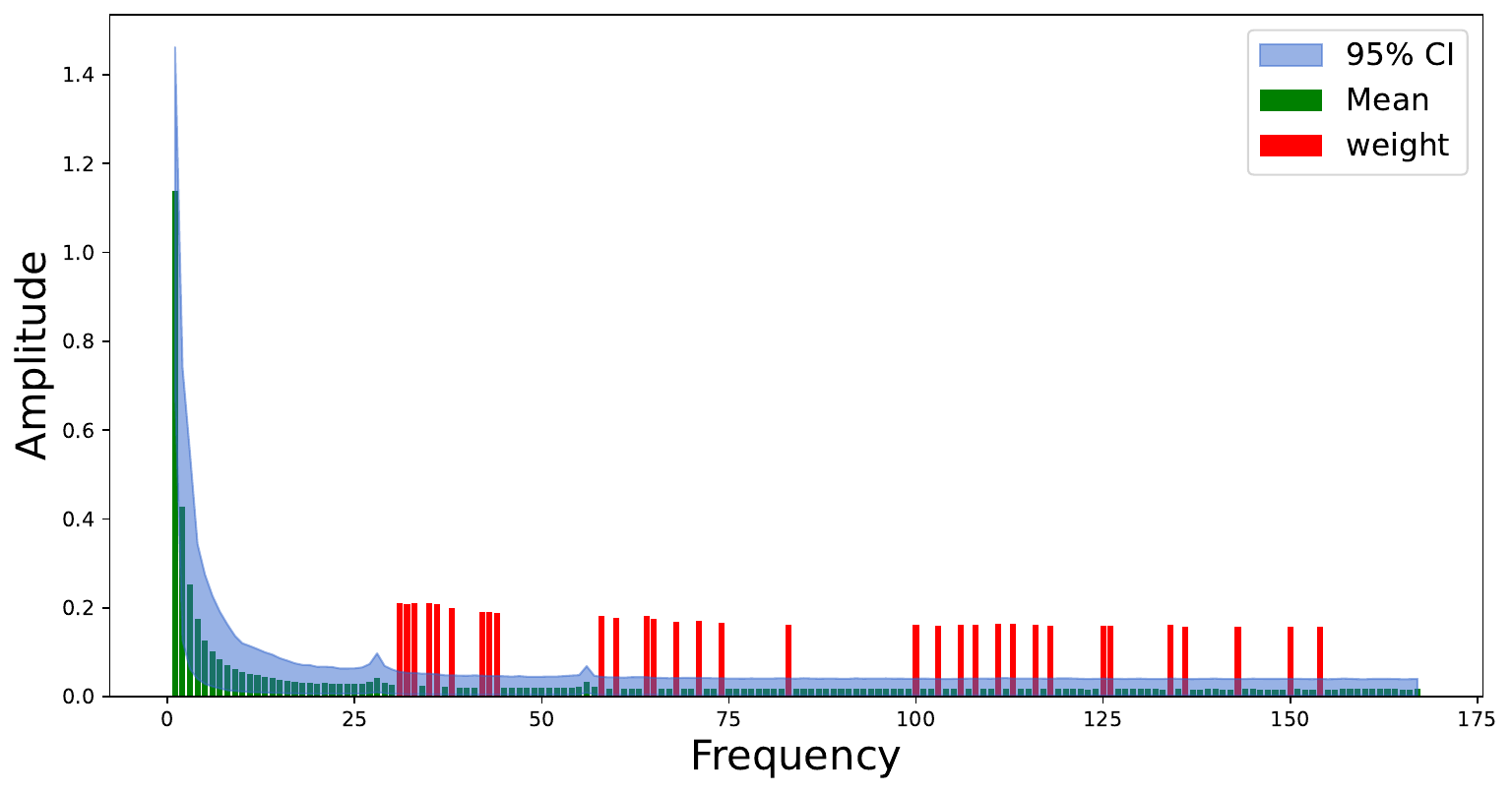}
\end{minipage}
\caption{Learned 10 Components for the PEMS08 data Compared to Its Magnitude Spectrum Distribution. The blue regions correspond to the 95\% confidence interval of magnitude spectrum, the green columns show the mean magnitude spectrum, and the red columns indicate the weights of the learned components.}
\label{components_PEMS08}
\end{figure*}

\begin{figure*}[t]
\centering
\begin{minipage}[t]{0.19\linewidth}
\centering
\includegraphics[width=\textwidth,height=0.8\textwidth]{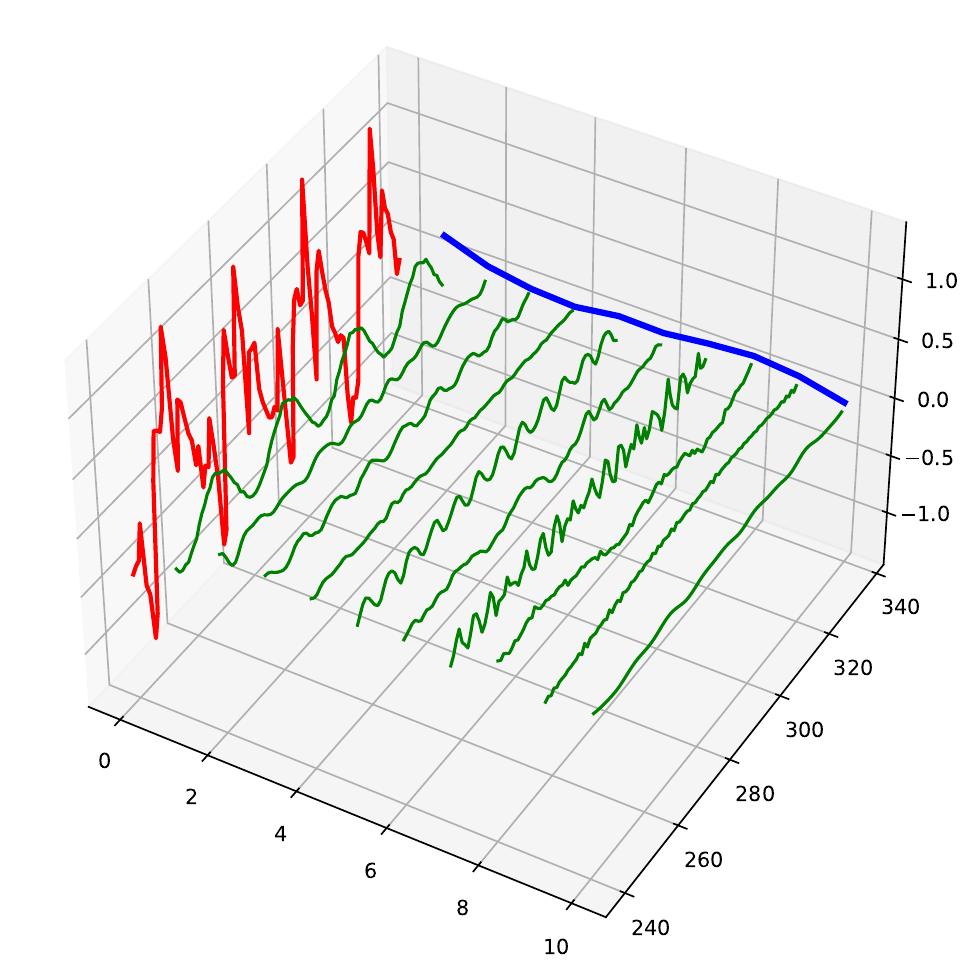}
\end{minipage}%
\begin{minipage}[t]{0.19\linewidth}
\centering
\includegraphics[width=\textwidth,height=0.8\textwidth]{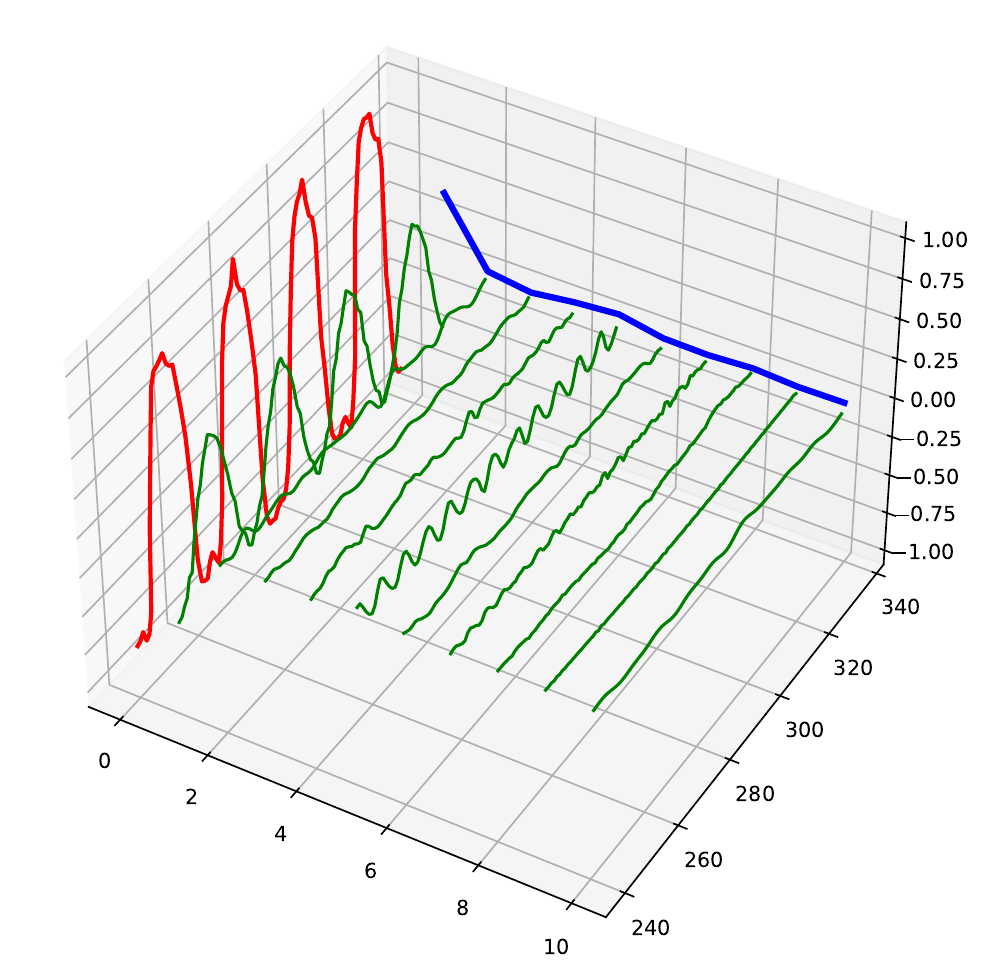}
\end{minipage}%
\begin{minipage}[t]{0.19\linewidth}
\centering
\includegraphics[width=\textwidth,height=0.8\textwidth]{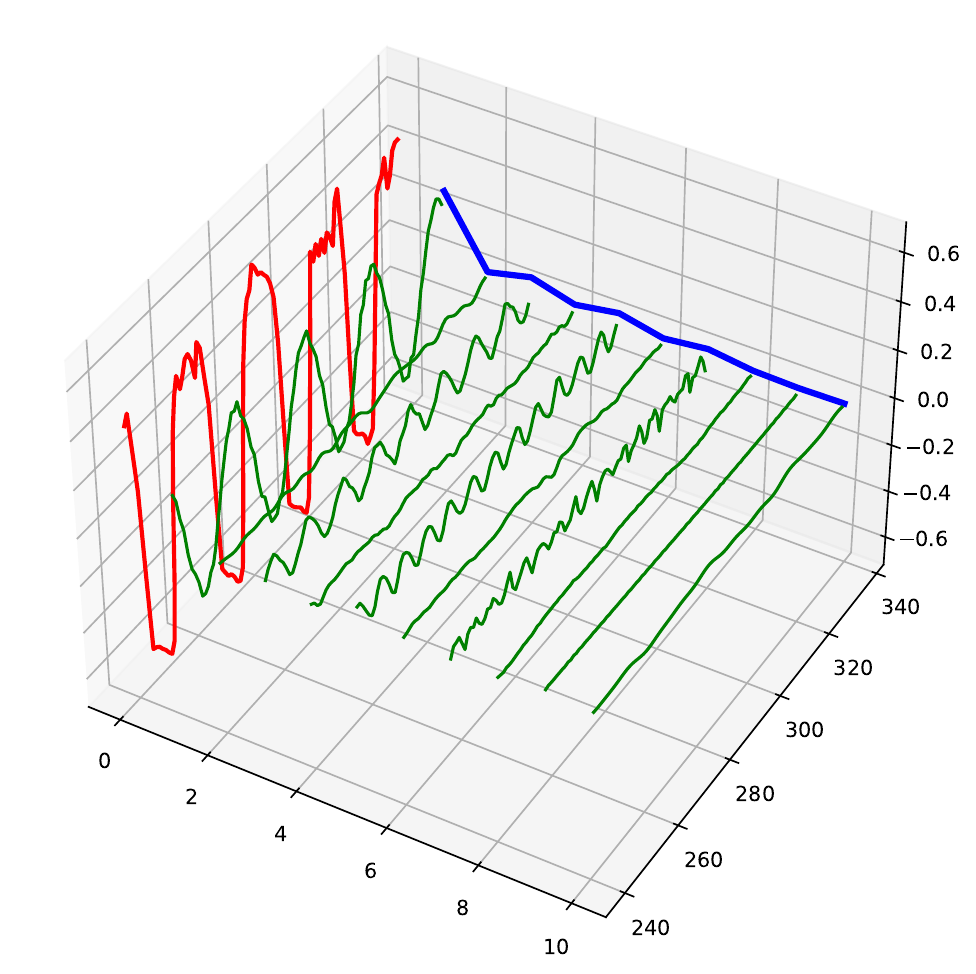}
\end{minipage}
\begin{minipage}[t]{0.19\linewidth}
\centering
\includegraphics[width=\textwidth,height=0.8\textwidth]{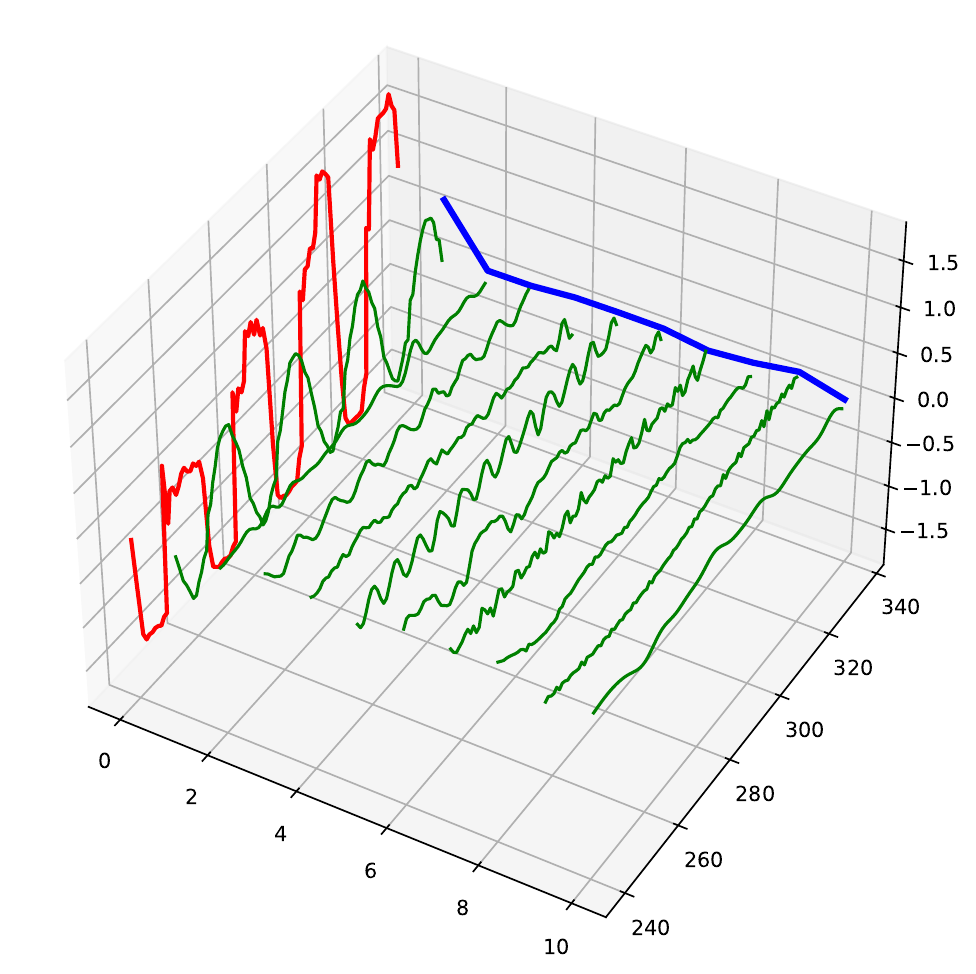}
\end{minipage}
\begin{minipage}[t]{0.19\linewidth}
\centering
\includegraphics[width=\textwidth,height=0.8\textwidth]{./fig/ECL/your_3d_plot4.pdf}
\end{minipage}
\begin{minipage}[t]{0.19\linewidth}
\centering
\includegraphics[width=\textwidth,height=0.8\textwidth]{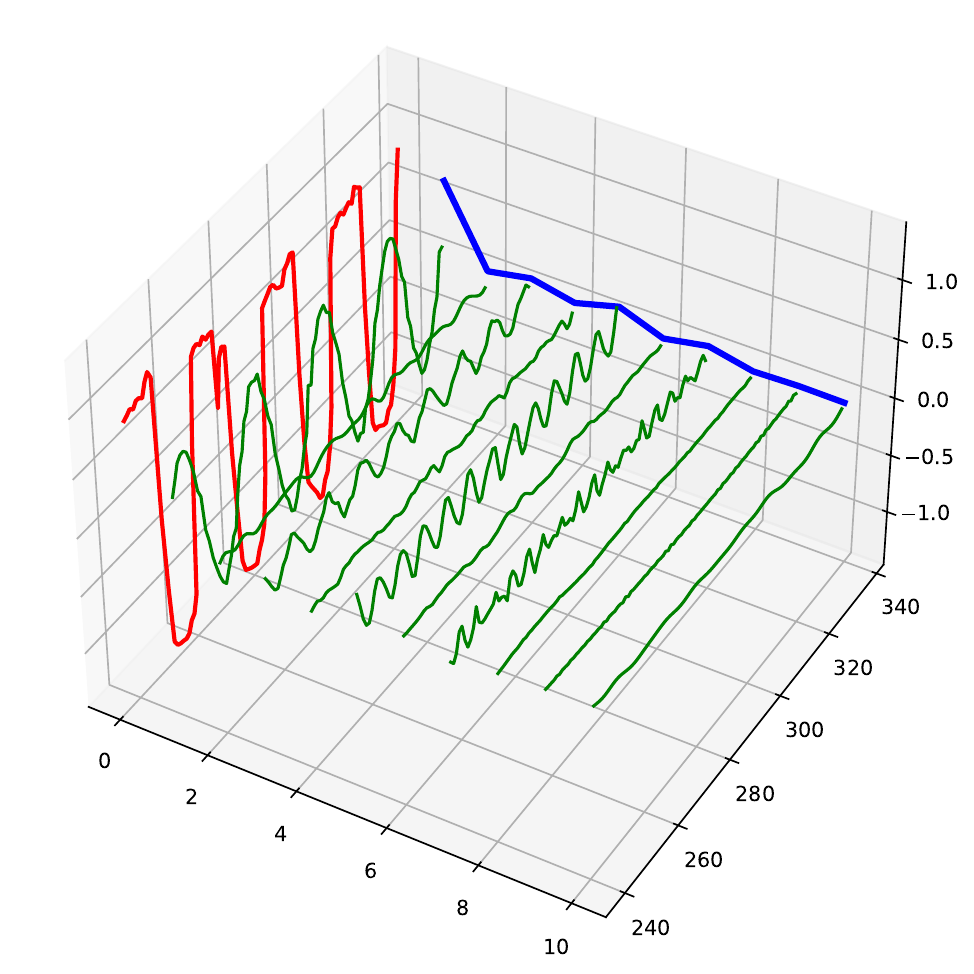}
\end{minipage}%
\begin{minipage}[t]{0.19\linewidth}
\centering
\includegraphics[width=\textwidth,height=0.8\textwidth]{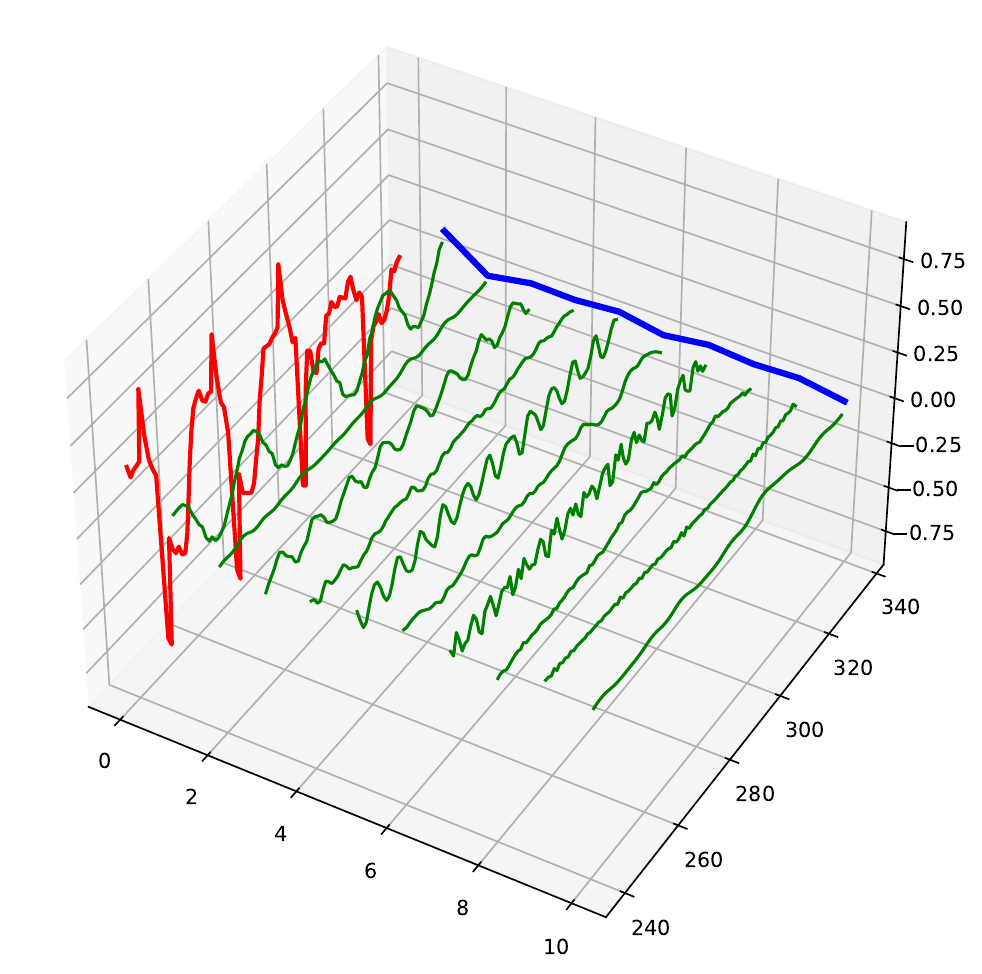}
\end{minipage}%
\begin{minipage}[t]{0.19\linewidth}
\centering
\includegraphics[width=\textwidth,height=0.8\textwidth]{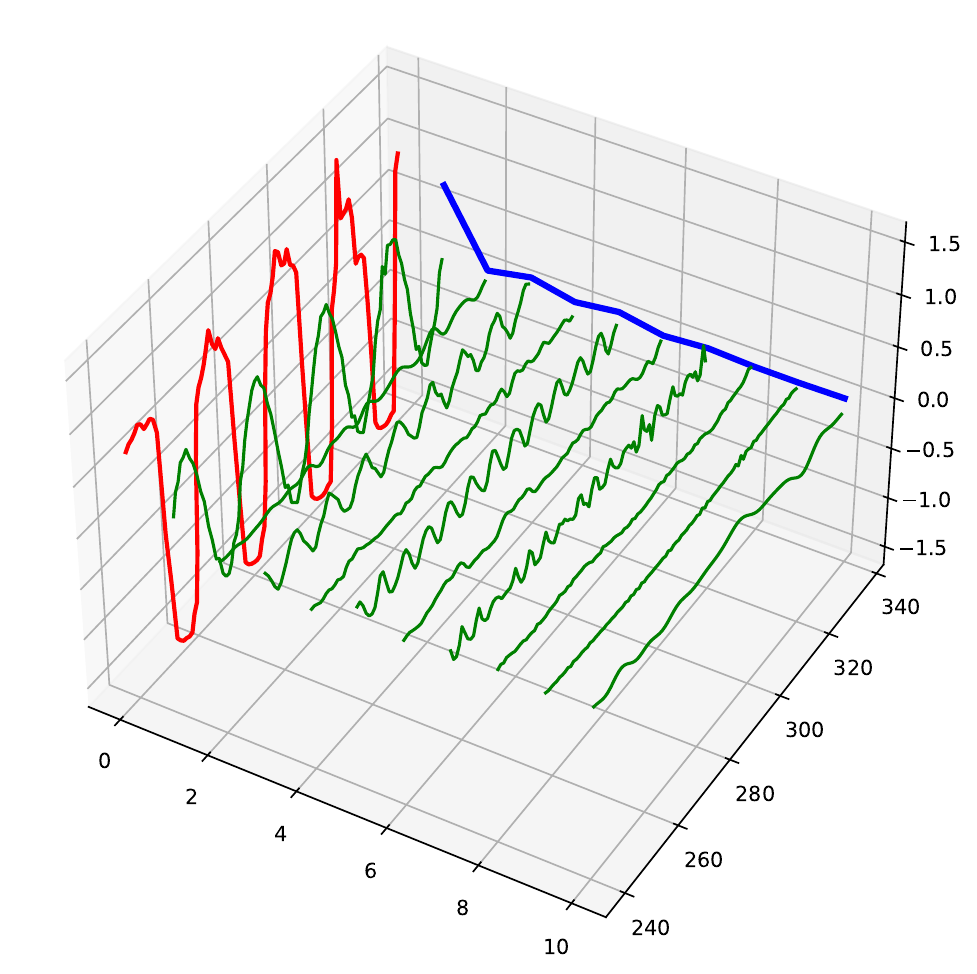}
\end{minipage}
\begin{minipage}[t]{0.19\linewidth}
\centering
\includegraphics[width=\textwidth,height=0.8\textwidth]{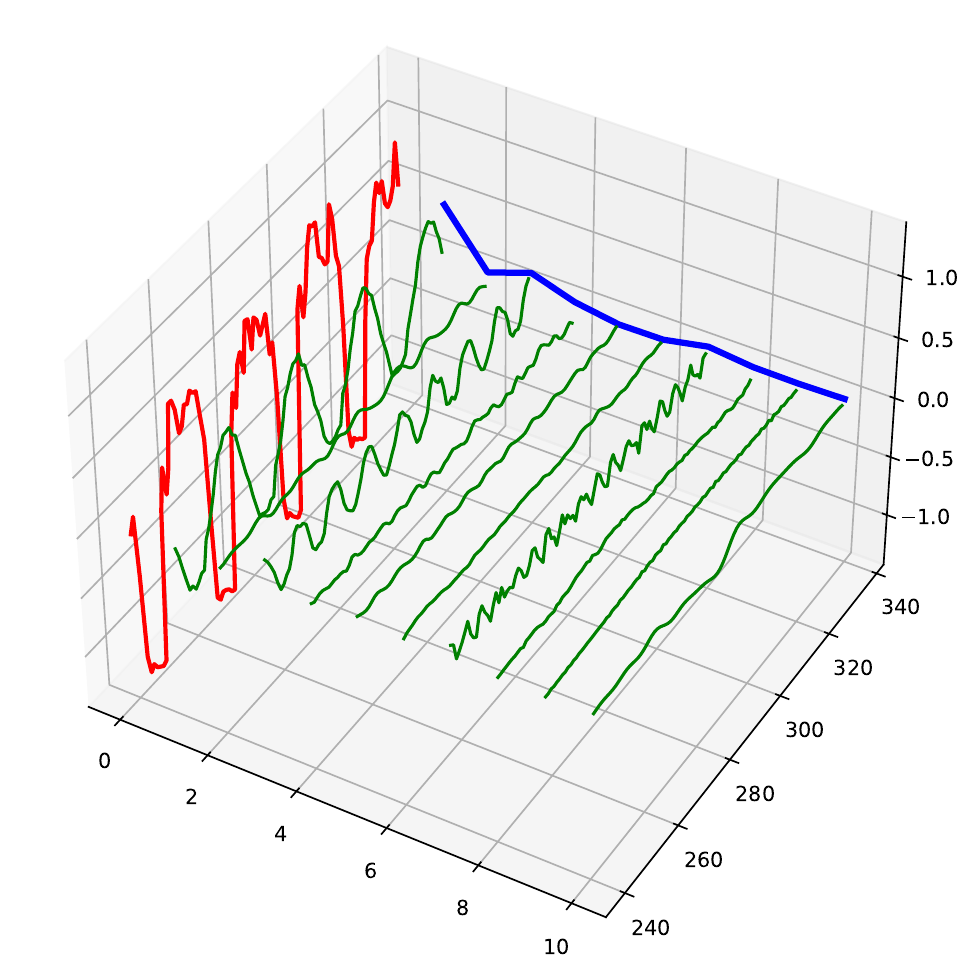}
\end{minipage}
\begin{minipage}[t]{0.19\linewidth}
\centering
\includegraphics[width=\textwidth,height=0.8\textwidth]{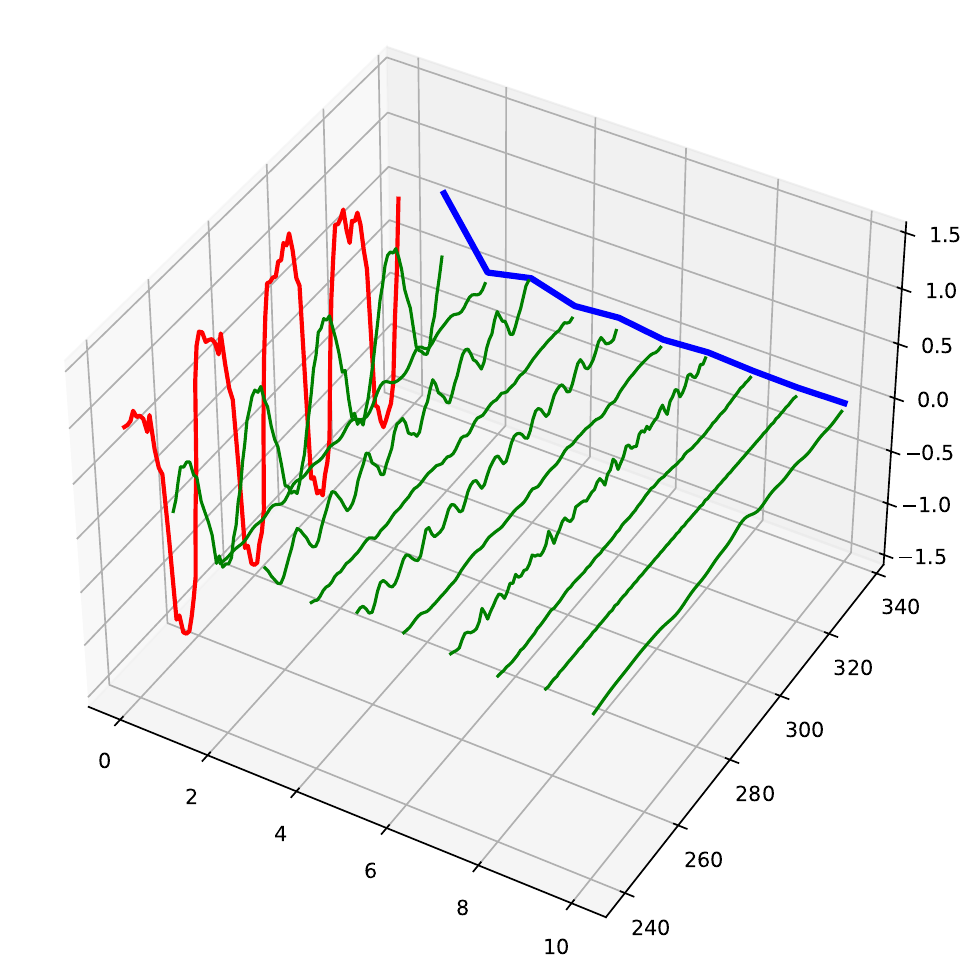}
\end{minipage}
\caption{Visualization of the MLOW Decomposition for Ten Examples on ECL.}
\label{plot1}
\end{figure*}

\begin{figure*}[t]
\centering
\begin{minipage}[t]{0.19\linewidth}
\centering
\includegraphics[width=\textwidth,height=0.8\textwidth]{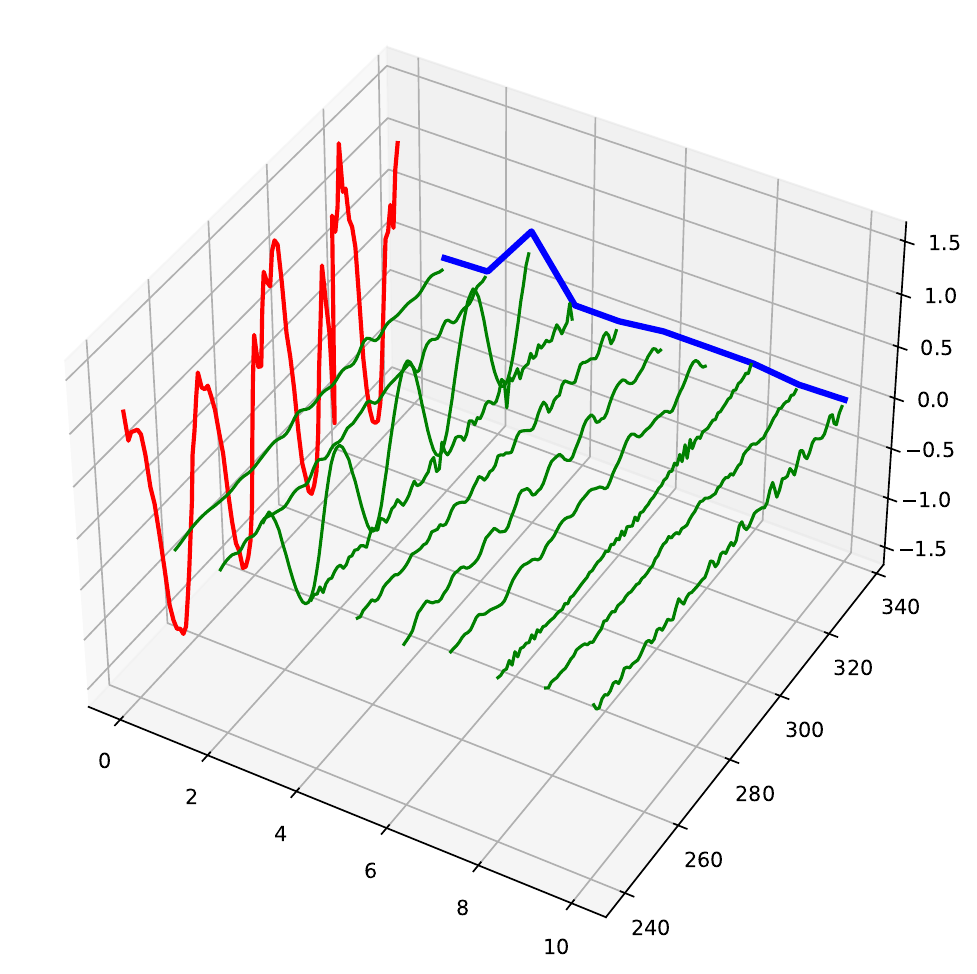}
\end{minipage}%
\begin{minipage}[t]{0.19\linewidth}
\centering
\includegraphics[width=\textwidth,height=0.8\textwidth]{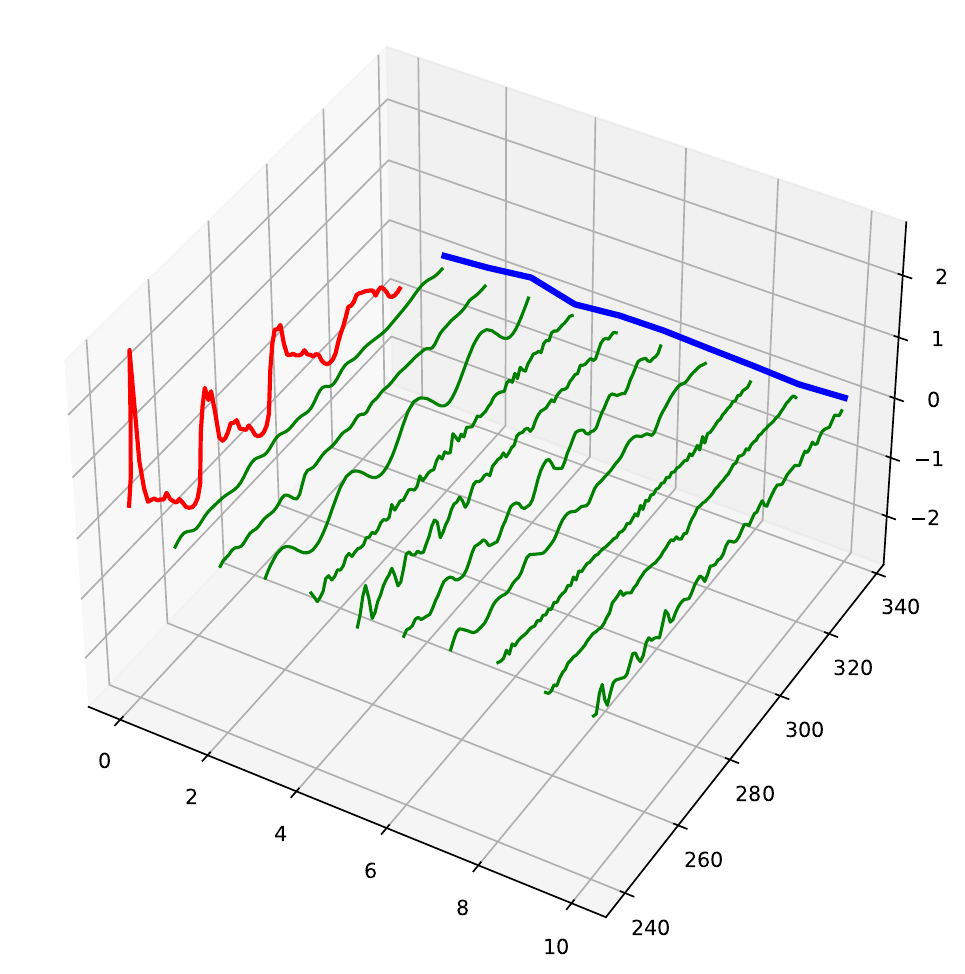}
\end{minipage}%
\begin{minipage}[t]{0.19\linewidth}
\centering
\includegraphics[width=\textwidth,height=0.8\textwidth]{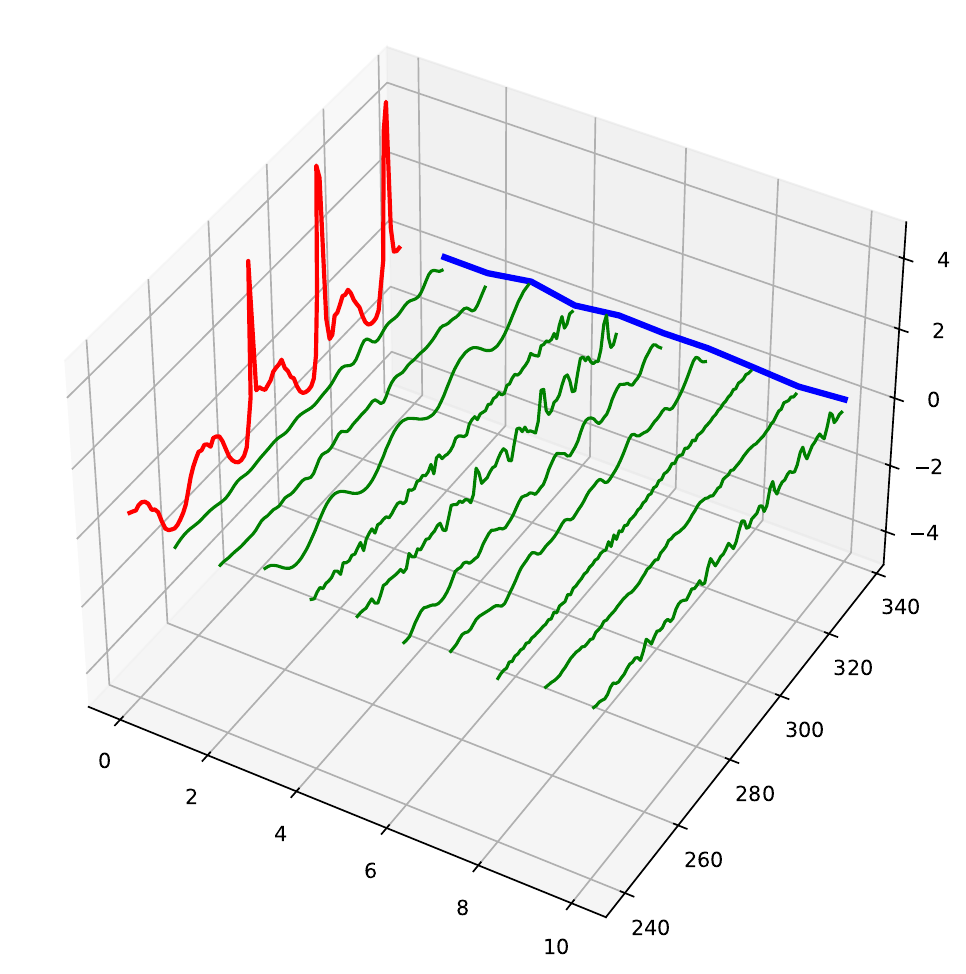}
\end{minipage}
\begin{minipage}[t]{0.19\linewidth}
\centering
\includegraphics[width=\textwidth,height=0.8\textwidth]{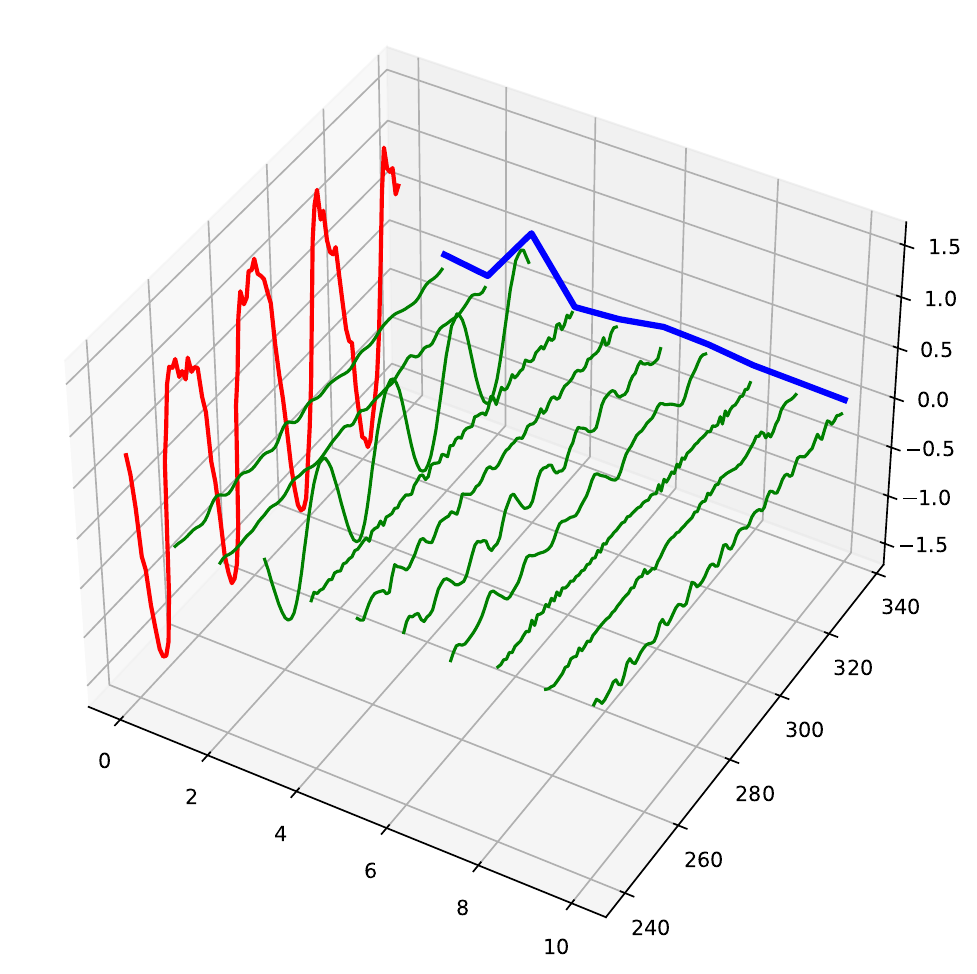}
\end{minipage}
\begin{minipage}[t]{0.19\linewidth}
\centering
\includegraphics[width=\textwidth,height=0.8\textwidth]{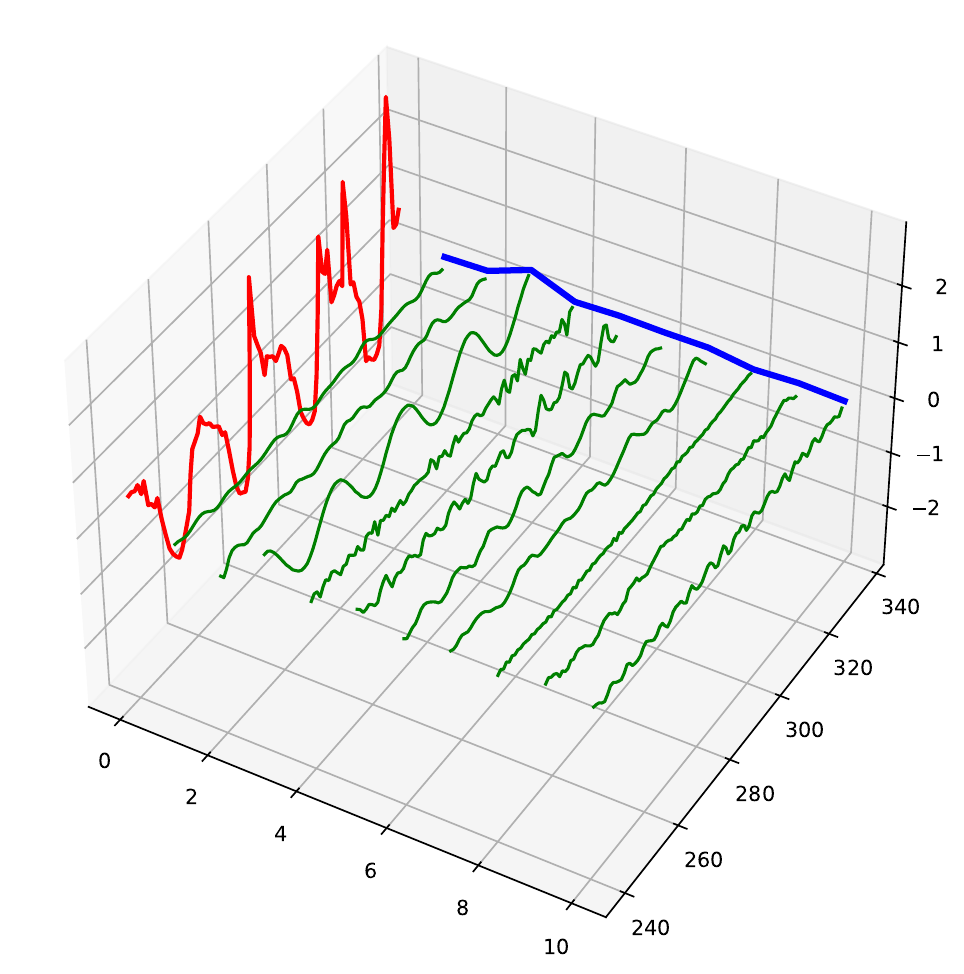}
\end{minipage}
\begin{minipage}[t]{0.19\linewidth}
\centering
\includegraphics[width=\textwidth,height=0.8\textwidth]{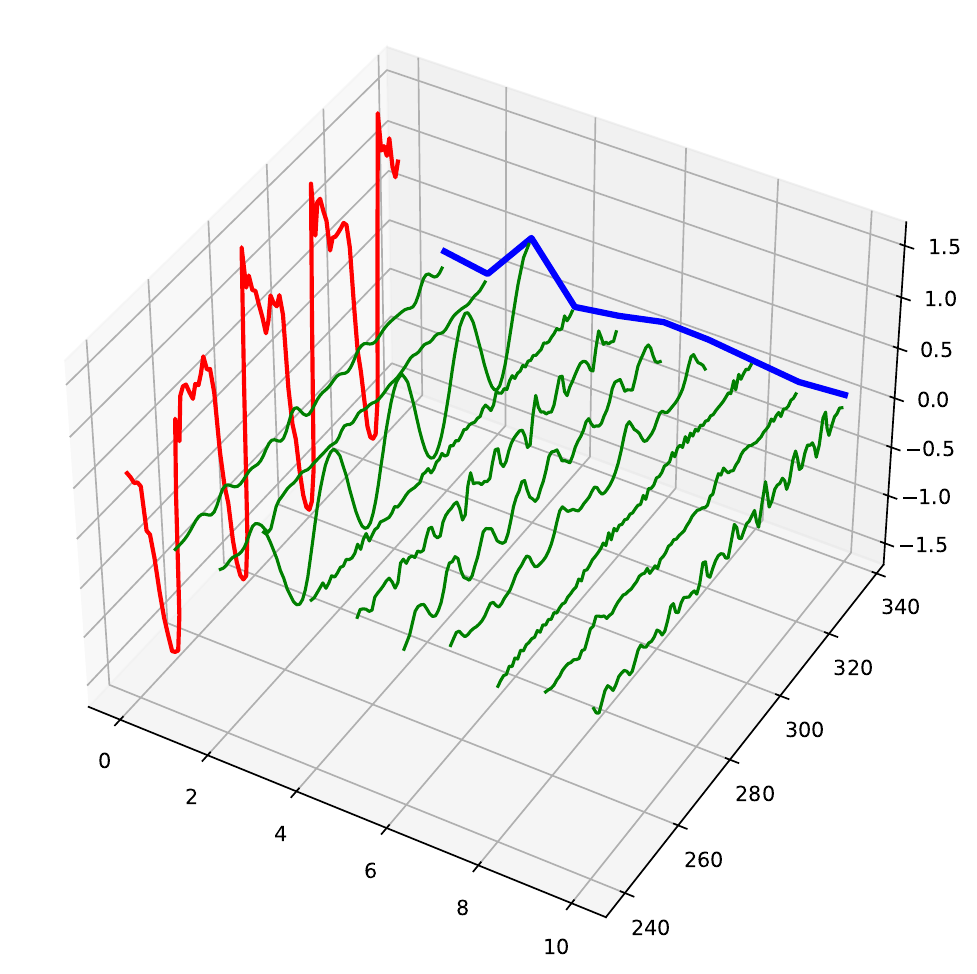}
\end{minipage}%
\begin{minipage}[t]{0.19\linewidth}
\centering
\includegraphics[width=\textwidth,height=0.8\textwidth]{./fig/Traffic/your_3d_plot6.pdf}
\end{minipage}%
\begin{minipage}[t]{0.19\linewidth}
\centering
\includegraphics[width=\textwidth,height=0.8\textwidth]{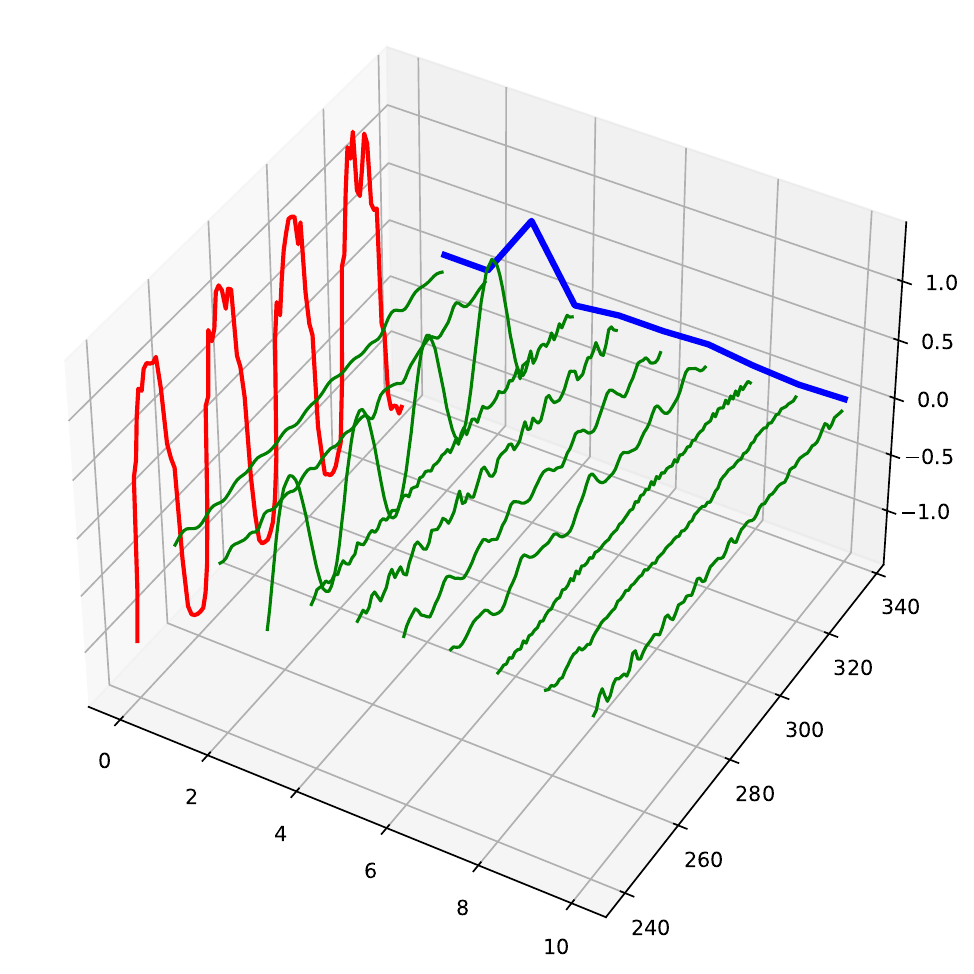}
\end{minipage}
\begin{minipage}[t]{0.19\linewidth}
\centering
\includegraphics[width=\textwidth,height=0.8\textwidth]{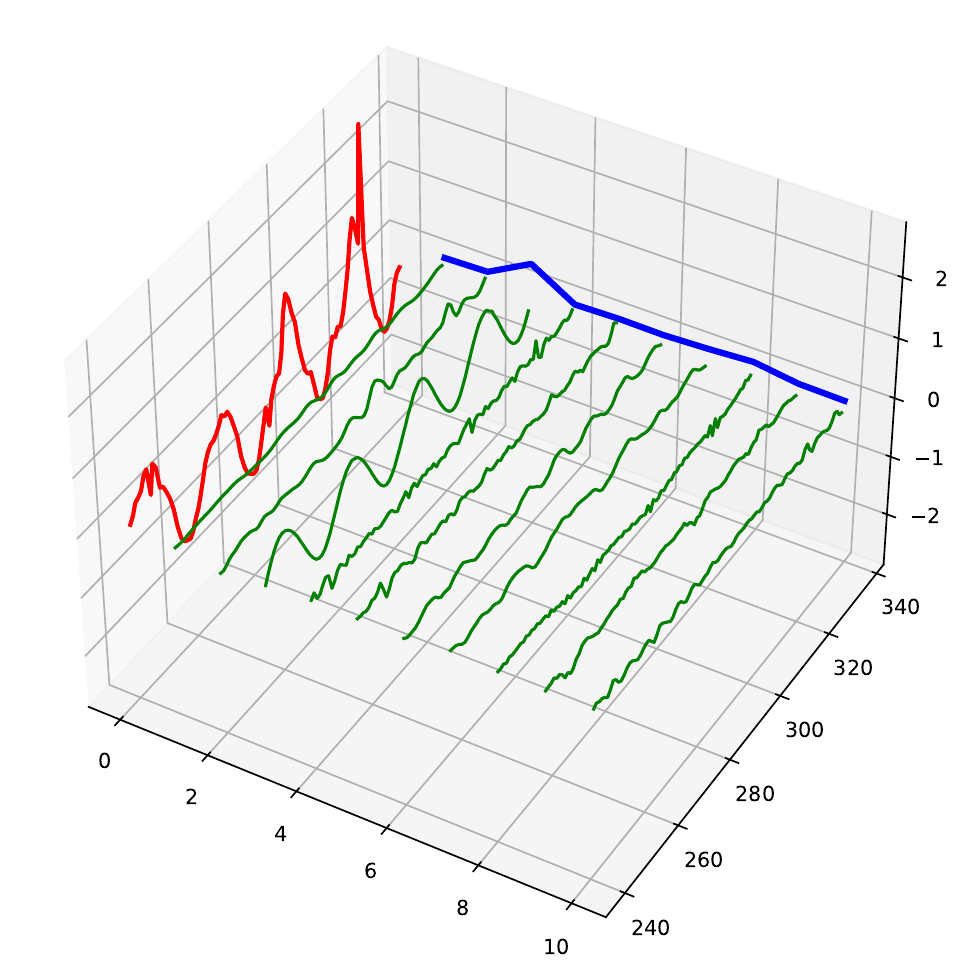}
\end{minipage}
\begin{minipage}[t]{0.19\linewidth}
\centering
\includegraphics[width=\textwidth,height=0.8\textwidth]{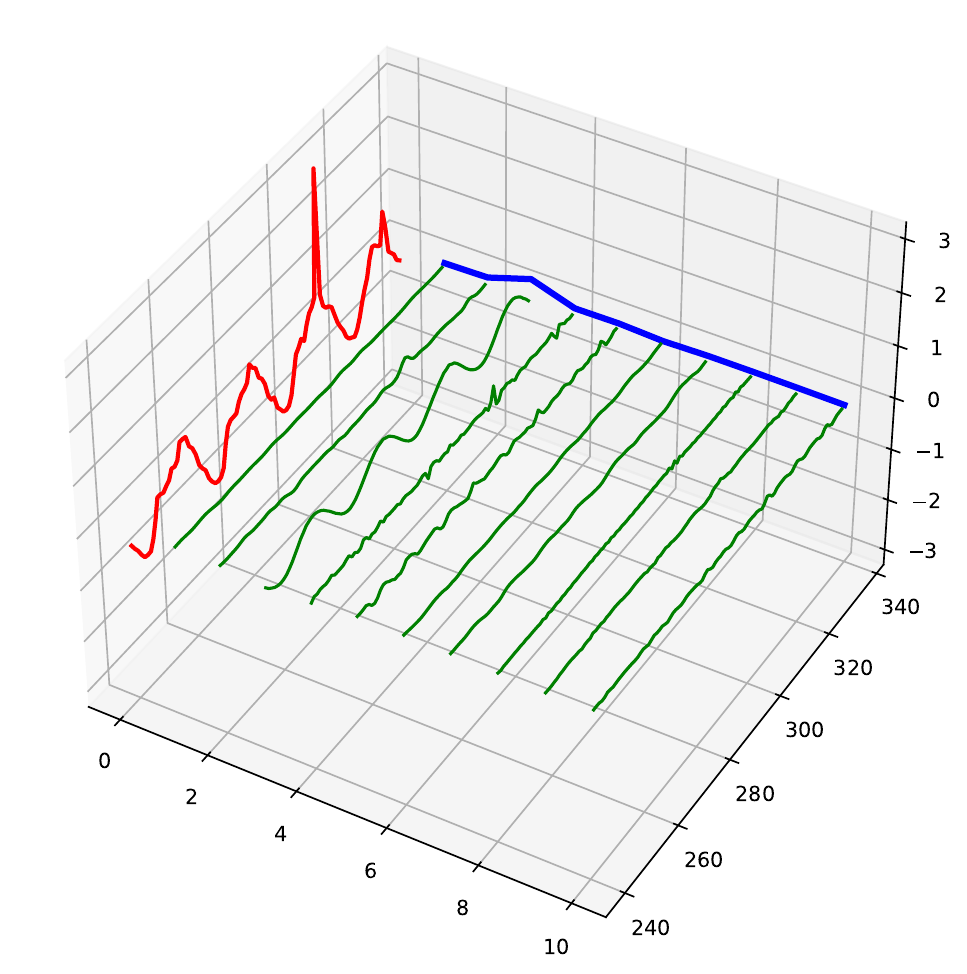}
\end{minipage}
\caption{Visualization of the MLOW Decomposition for Ten Examples on Traffic.}
\label{plot2}
\end{figure*}

\begin{figure*}[t]
\centering
\begin{minipage}[t]{0.19\linewidth}
\centering
\includegraphics[width=\textwidth,height=0.8\textwidth]{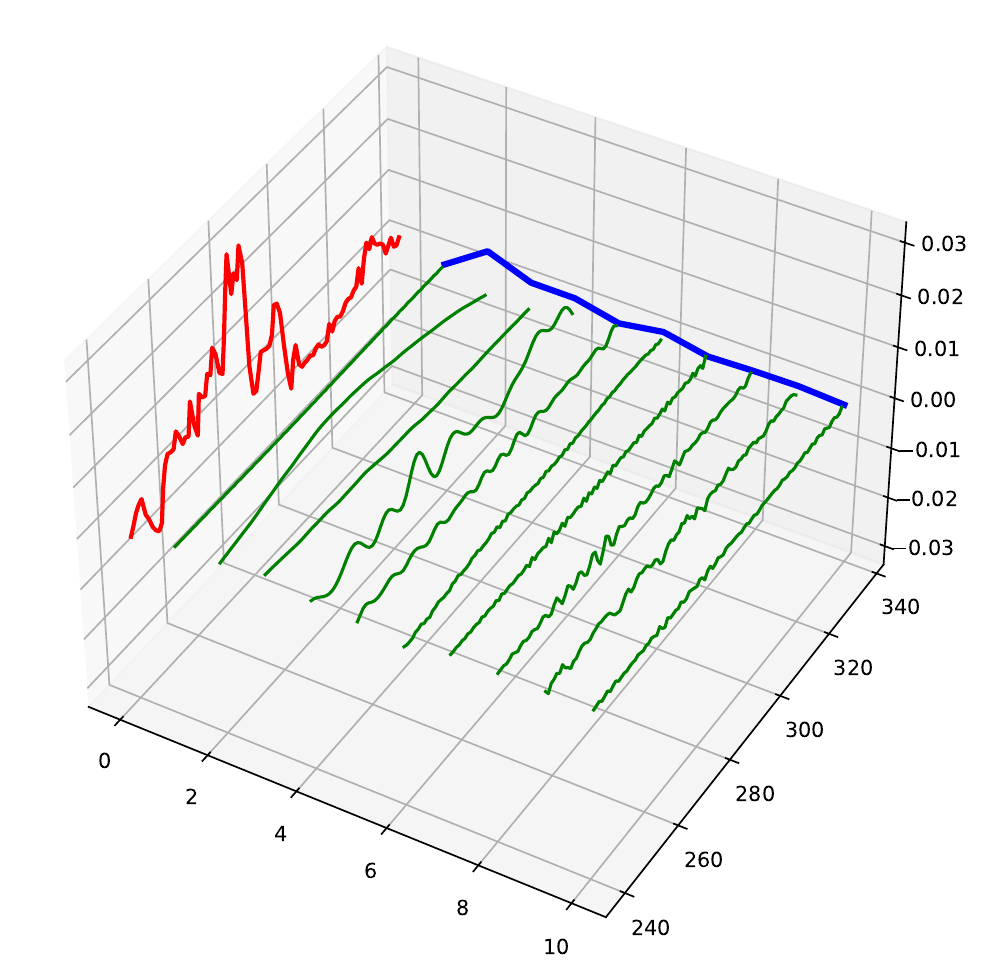}
\end{minipage}%
\begin{minipage}[t]{0.19\linewidth}
\centering
\includegraphics[width=\textwidth,height=0.8\textwidth]{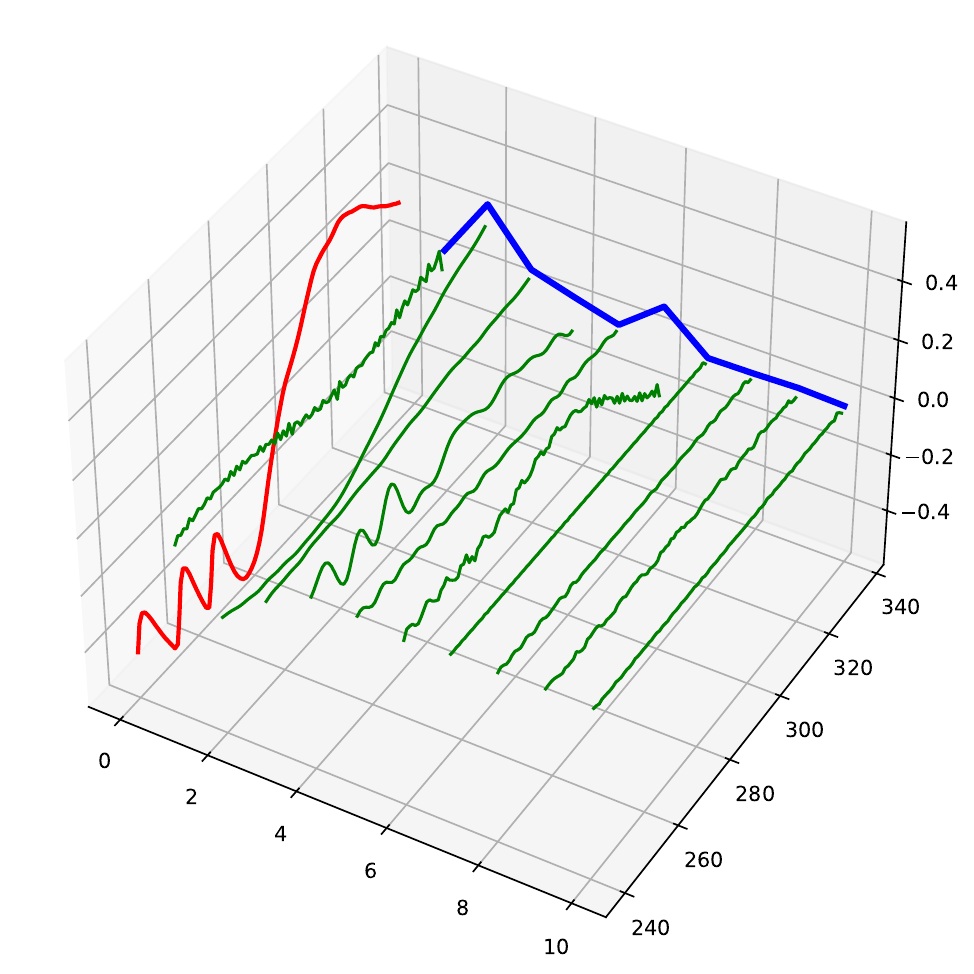}
\end{minipage}%
\begin{minipage}[t]{0.19\linewidth}
\centering
\includegraphics[width=\textwidth,height=0.8\textwidth]{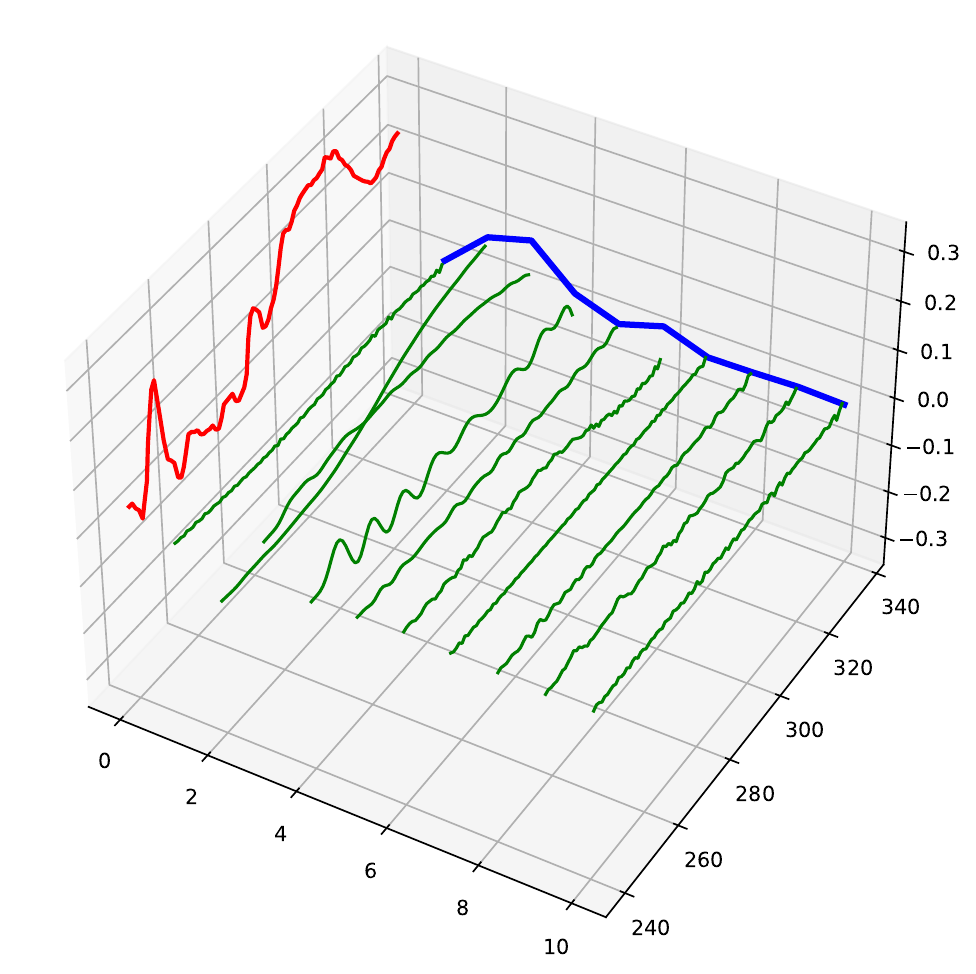}
\end{minipage}
\begin{minipage}[t]{0.19\linewidth}
\centering
\includegraphics[width=\textwidth,height=0.8\textwidth]{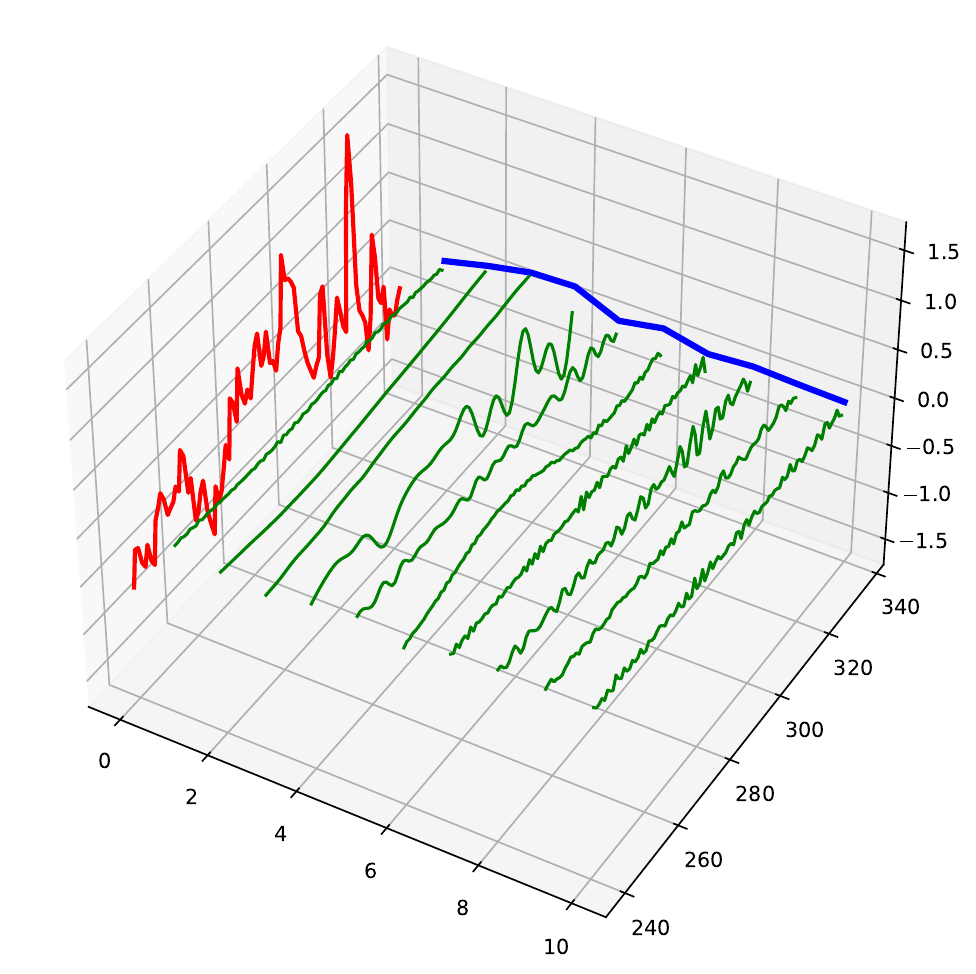}
\end{minipage}
\begin{minipage}[t]{0.19\linewidth}
\centering
\includegraphics[width=\textwidth,height=0.8\textwidth]{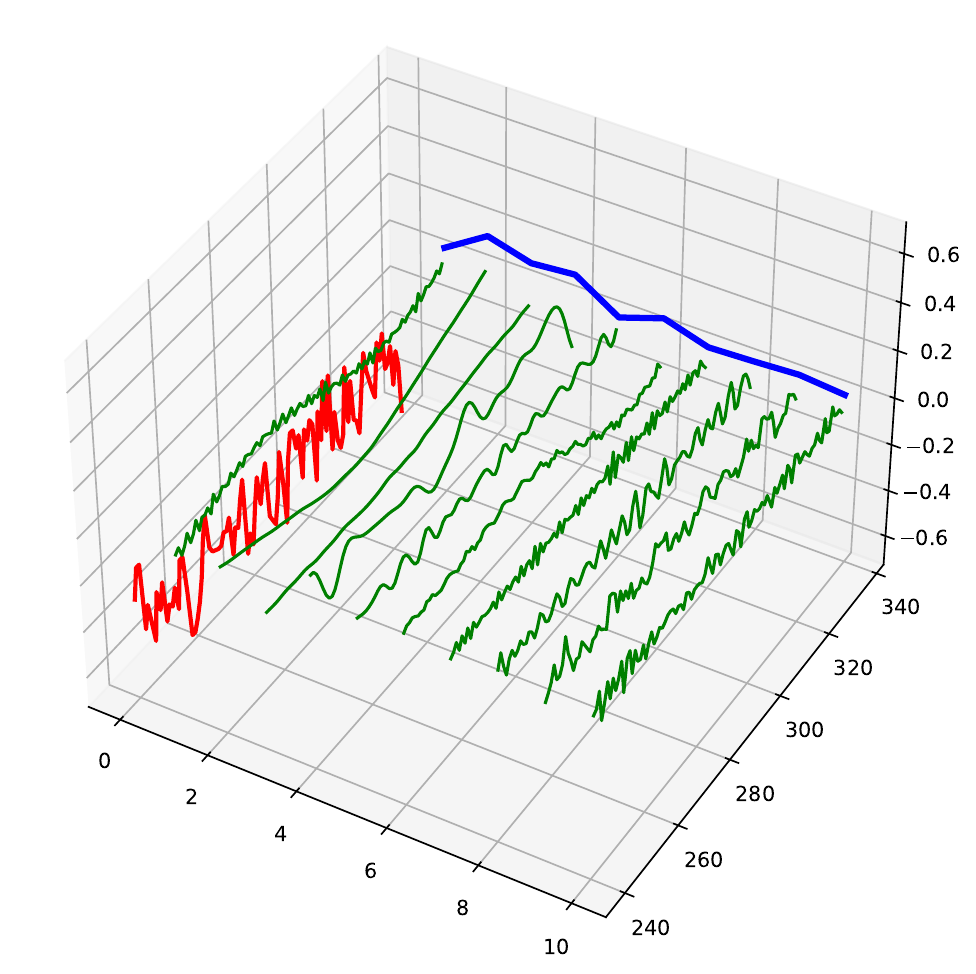}
\end{minipage}
\begin{minipage}[t]{0.19\linewidth}
\centering
\includegraphics[width=\textwidth,height=0.8\textwidth]{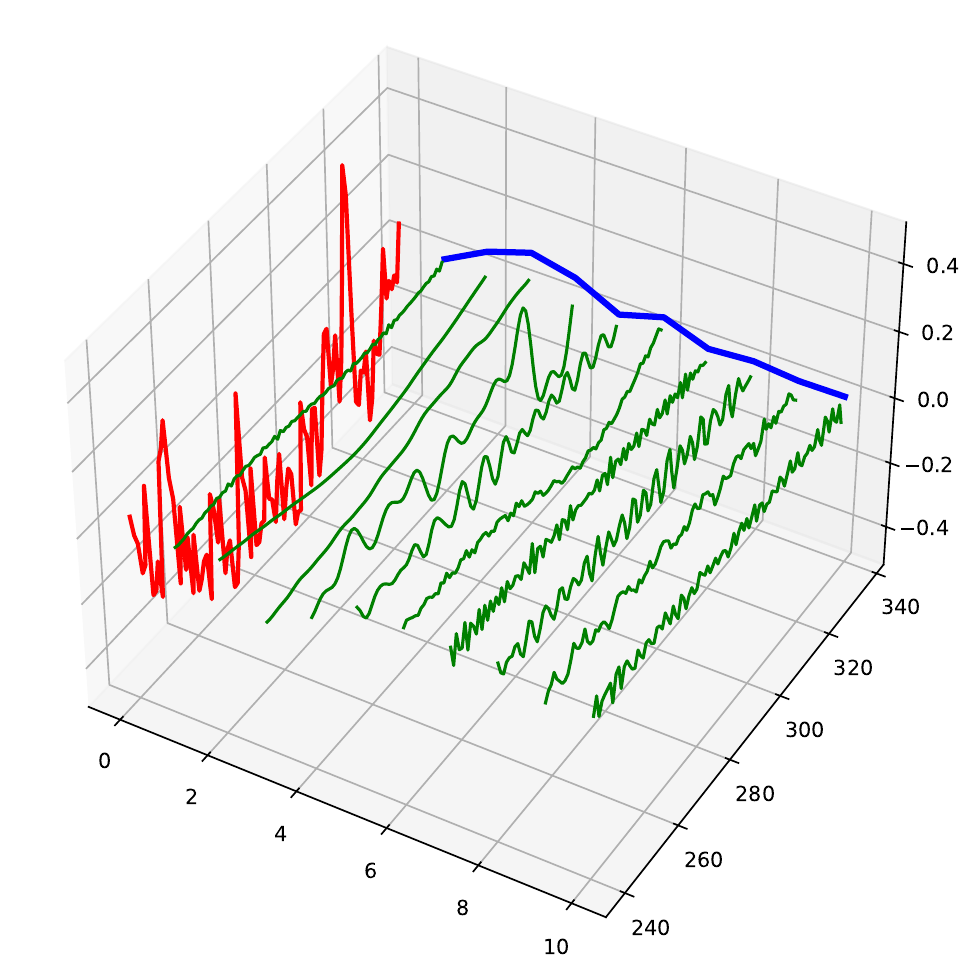}
\end{minipage}%
\begin{minipage}[t]{0.19\linewidth}
\centering
\includegraphics[width=\textwidth,height=0.8\textwidth]{./fig/Weather/your_3d_plot6.pdf}
\end{minipage}%
\begin{minipage}[t]{0.19\linewidth}
\centering
\includegraphics[width=\textwidth,height=0.8\textwidth]{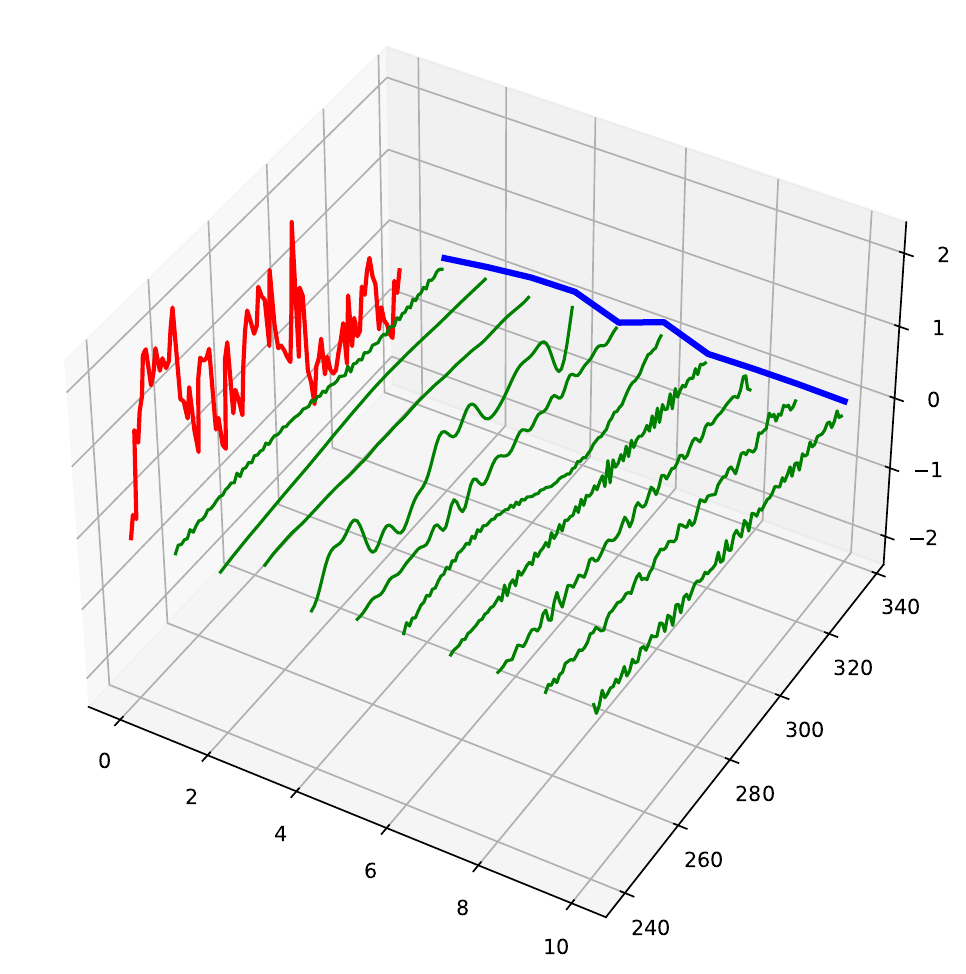}
\end{minipage}
\begin{minipage}[t]{0.19\linewidth}
\centering
\includegraphics[width=\textwidth,height=0.8\textwidth]{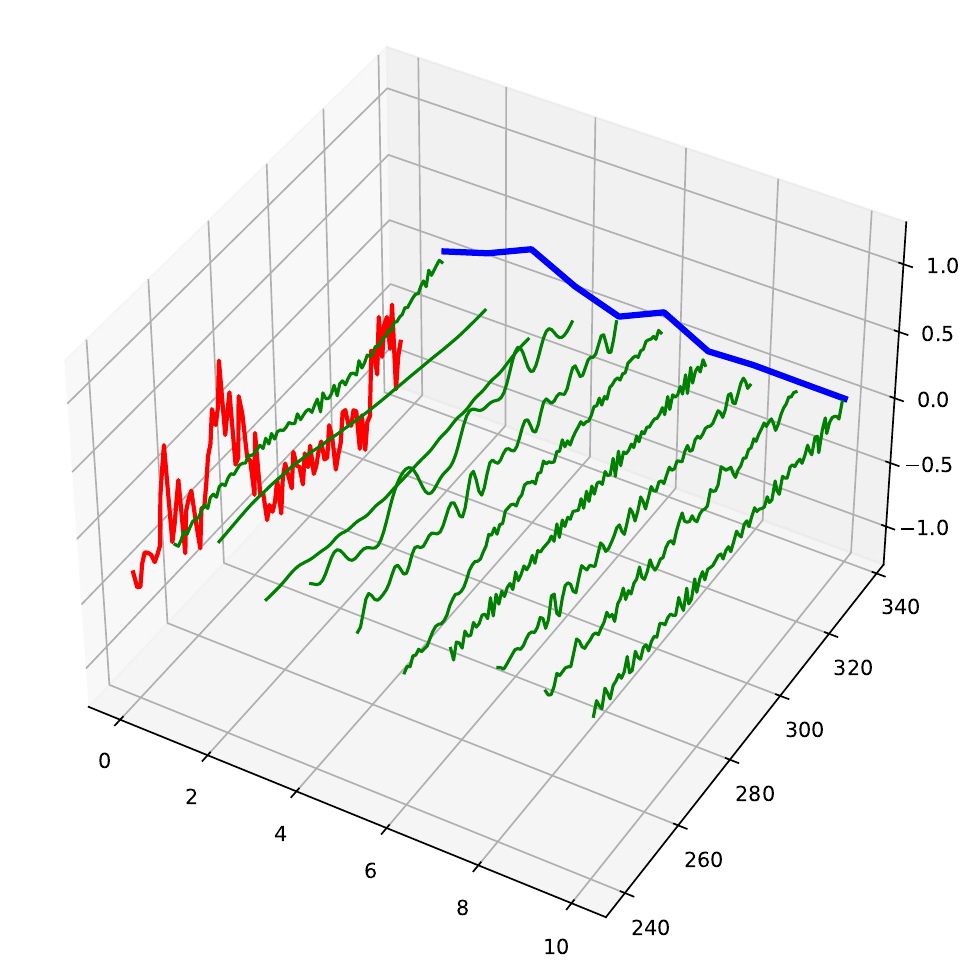}
\end{minipage}
\begin{minipage}[t]{0.19\linewidth}
\centering
\includegraphics[width=\textwidth,height=0.8\textwidth]{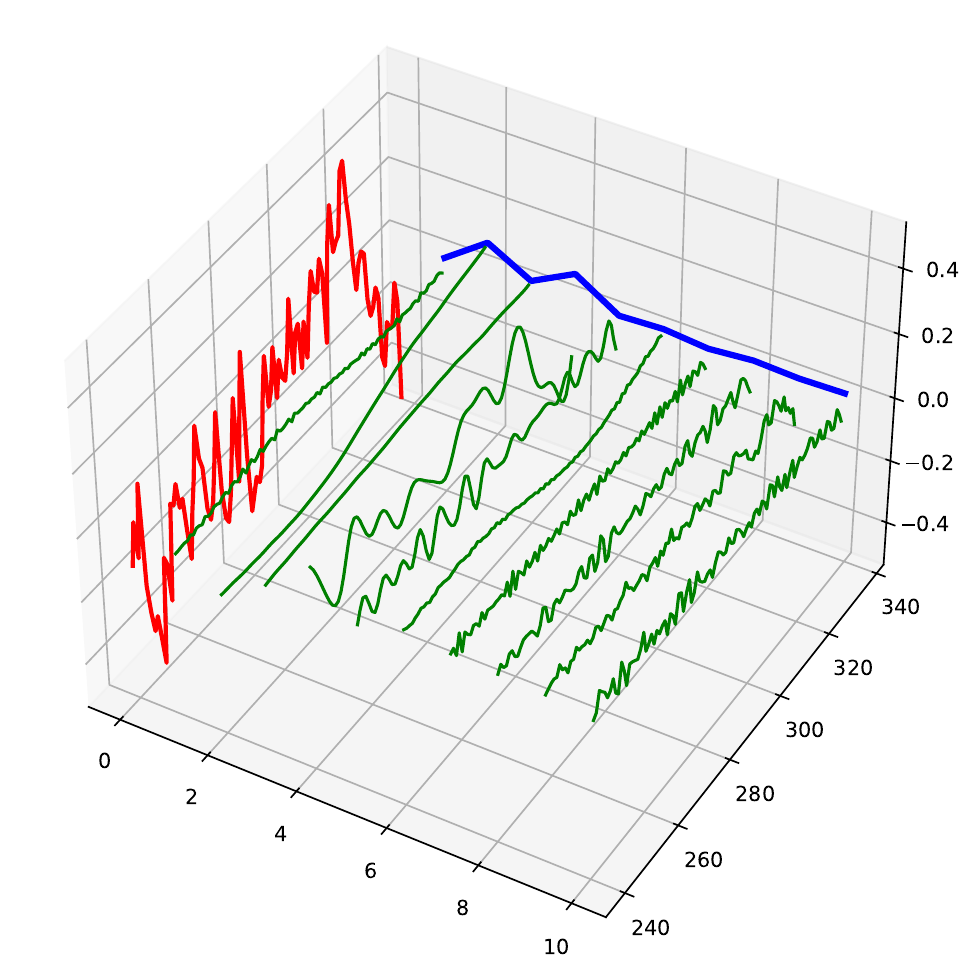}
\end{minipage}
\caption{Visualization of the MLOW Decomposition for Ten Examples on Weather.}
\label{plot3}
\end{figure*}

\begin{figure*}[t]
\centering
\begin{minipage}[t]{0.19\linewidth}
\centering
\includegraphics[width=\textwidth,height=0.8\textwidth]{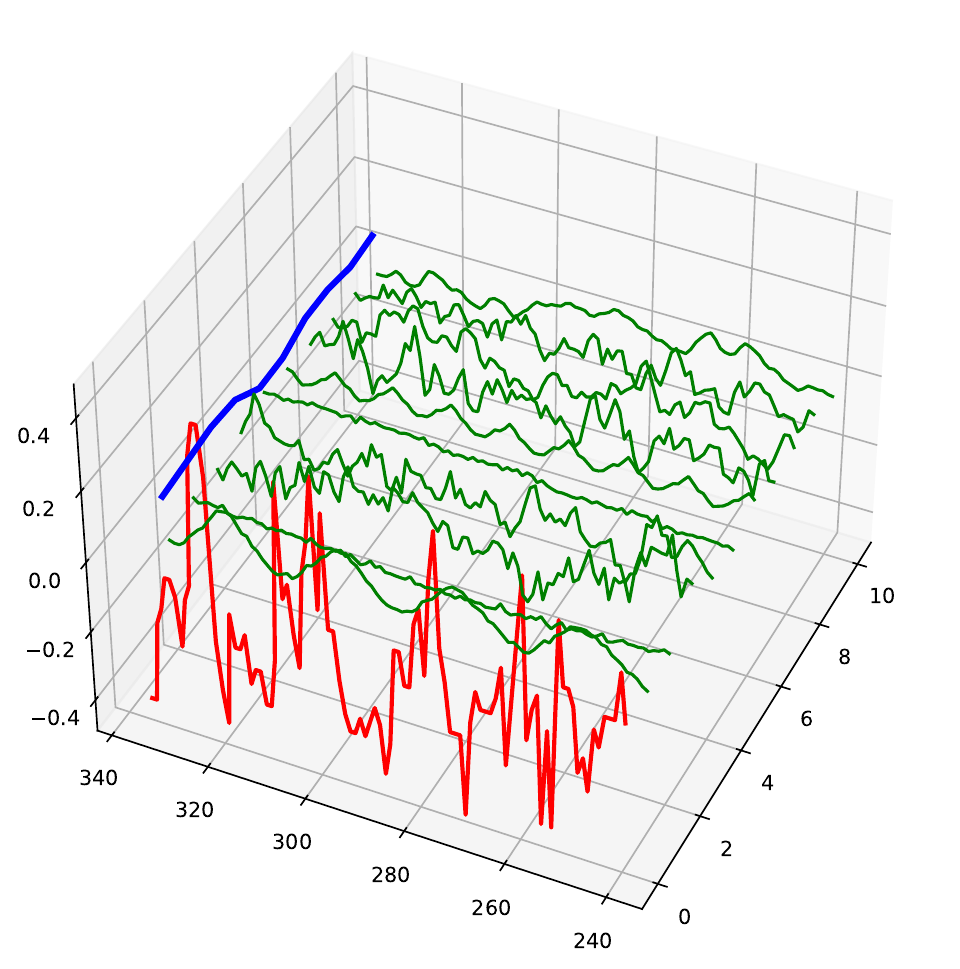}
\end{minipage}%
\begin{minipage}[t]{0.19\linewidth}
\centering
\includegraphics[width=\textwidth,height=0.8\textwidth]{./fig/ETTh1/your_3d_plot1.pdf}
\end{minipage}%
\begin{minipage}[t]{0.19\linewidth}
\centering
\includegraphics[width=\textwidth,height=0.8\textwidth]{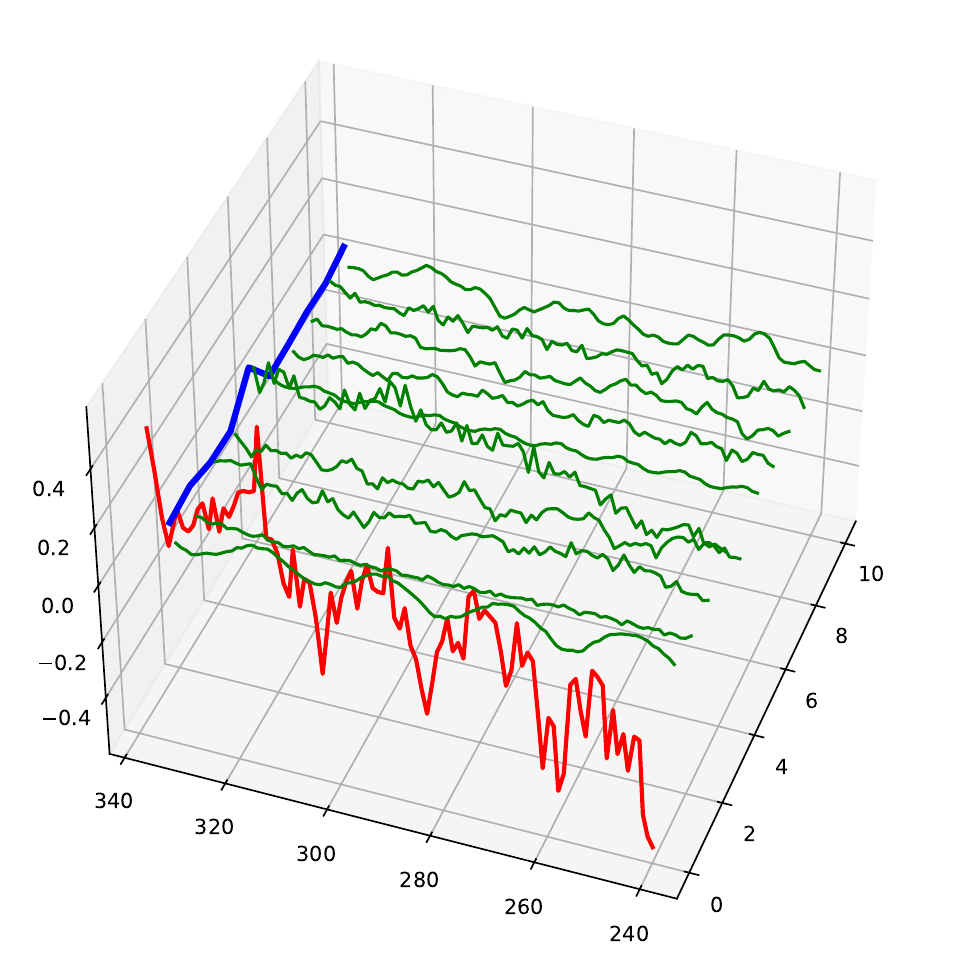}
\end{minipage}
\begin{minipage}[t]{0.19\linewidth}
\centering
\includegraphics[width=\textwidth,height=0.8\textwidth]{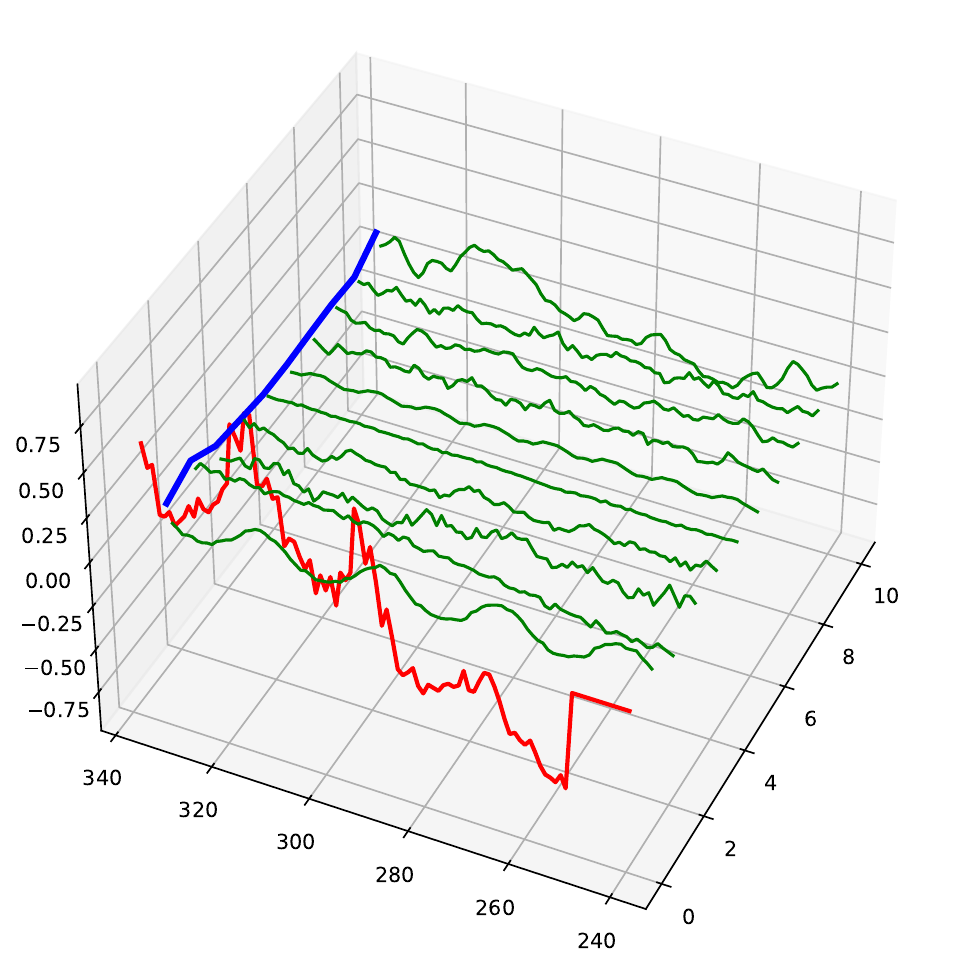}
\end{minipage}
\begin{minipage}[t]{0.19\linewidth}
\centering
\includegraphics[width=\textwidth,height=0.8\textwidth]{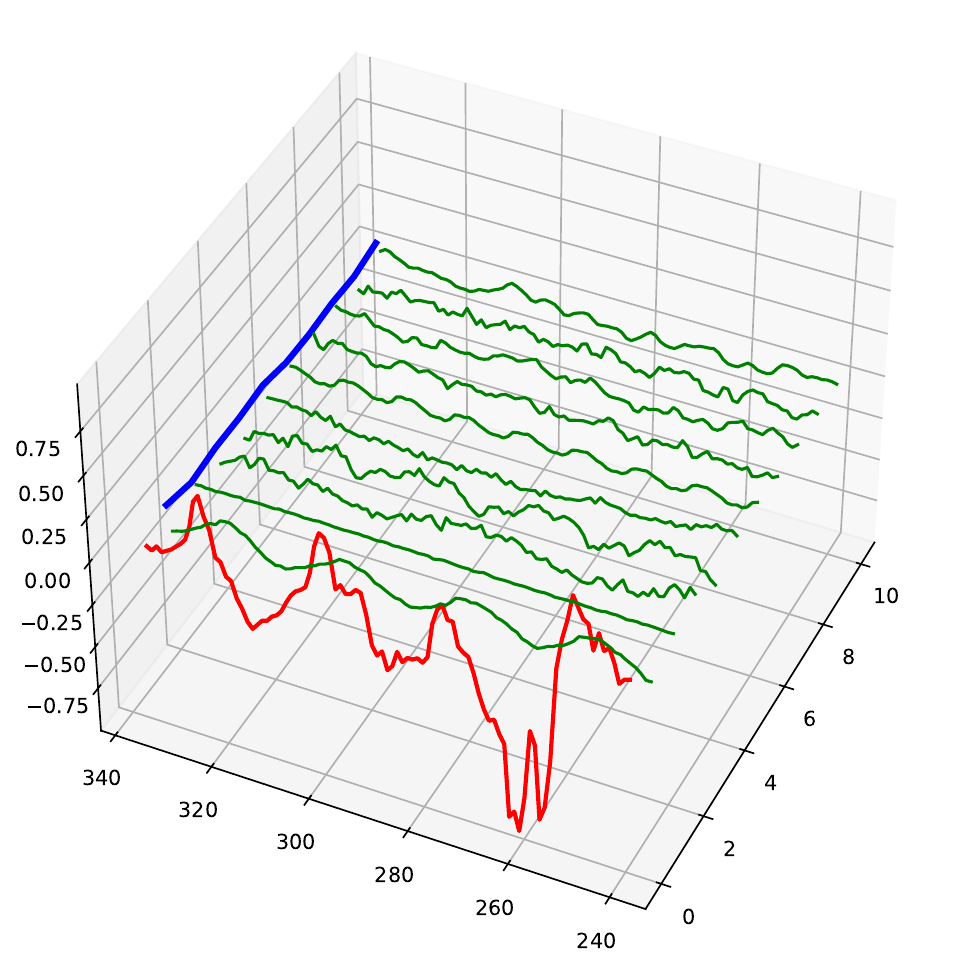}
\end{minipage}
\begin{minipage}[t]{0.19\linewidth}
\centering
\includegraphics[width=\textwidth,height=0.8\textwidth]{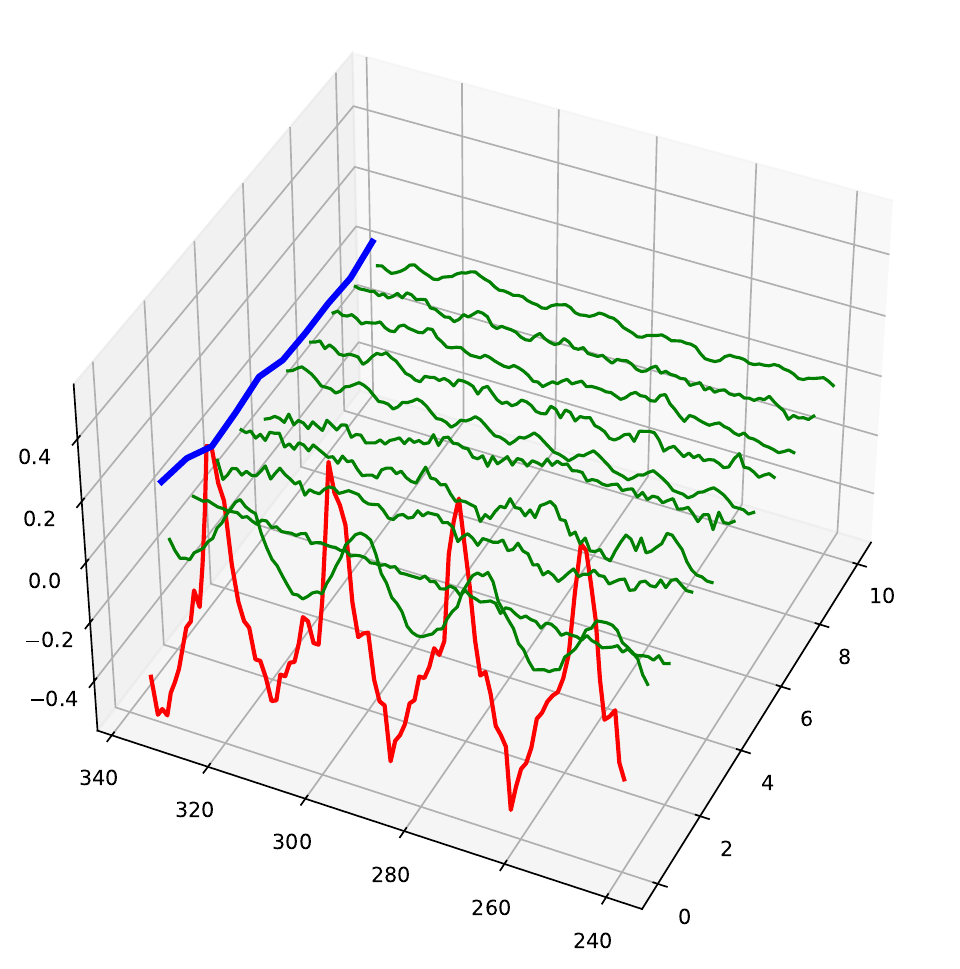}
\end{minipage}%
\begin{minipage}[t]{0.19\linewidth}
\centering
\includegraphics[width=\textwidth,height=0.8\textwidth]{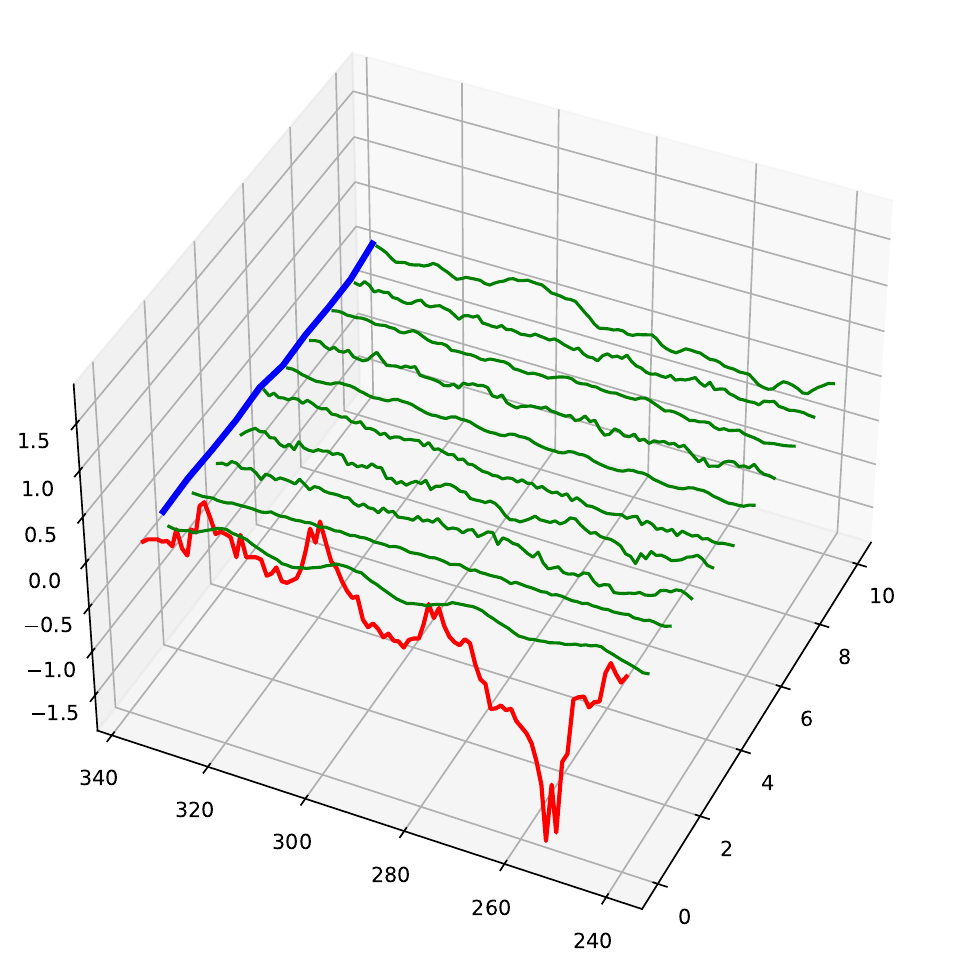}
\end{minipage}%
\begin{minipage}[t]{0.19\linewidth}
\centering
\includegraphics[width=\textwidth,height=0.8\textwidth]{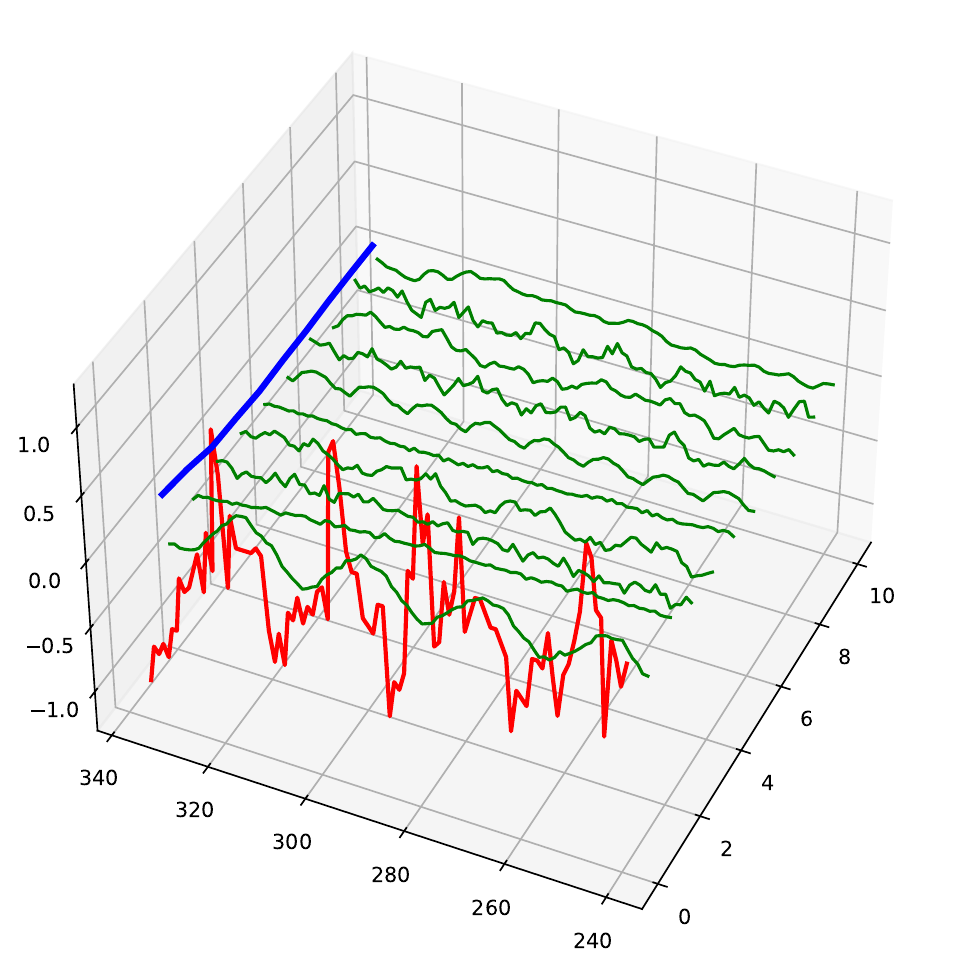}
\end{minipage}
\begin{minipage}[t]{0.19\linewidth}
\centering
\includegraphics[width=\textwidth,height=0.8\textwidth]{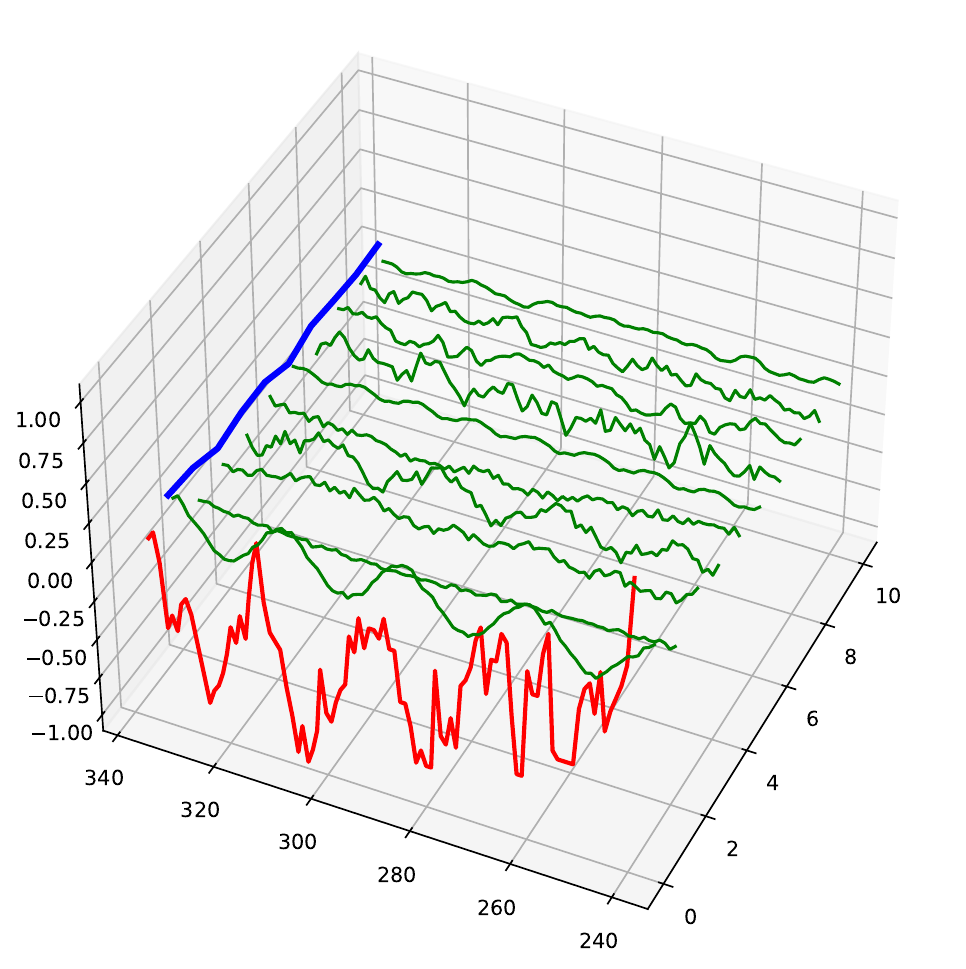}
\end{minipage}
\begin{minipage}[t]{0.19\linewidth}
\centering
\includegraphics[width=\textwidth,height=0.8\textwidth]{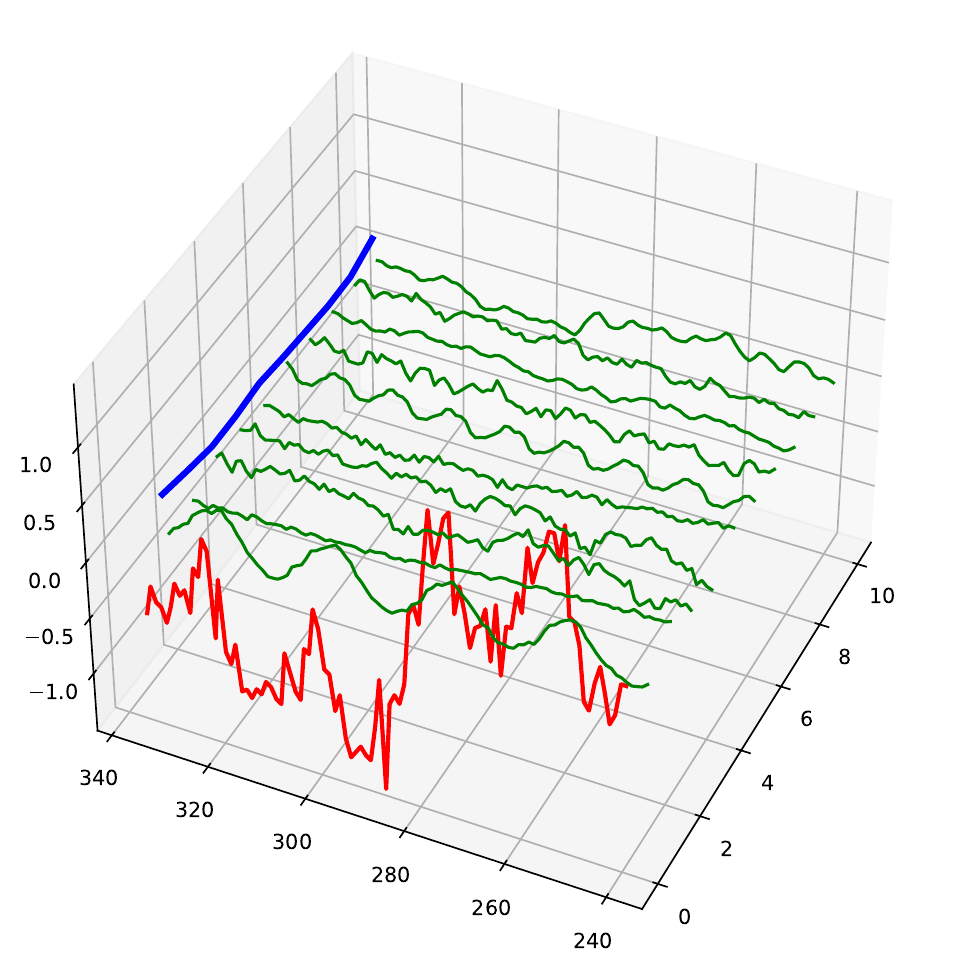}
\end{minipage}
\caption{Visualization of the MLOW Decomposition for Ten Examples on ETTh1.}
\label{plot4}
\end{figure*}

\begin{figure*}[t]
\centering
\begin{minipage}[t]{0.19\linewidth}
\centering
\includegraphics[width=\textwidth,height=0.8\textwidth]{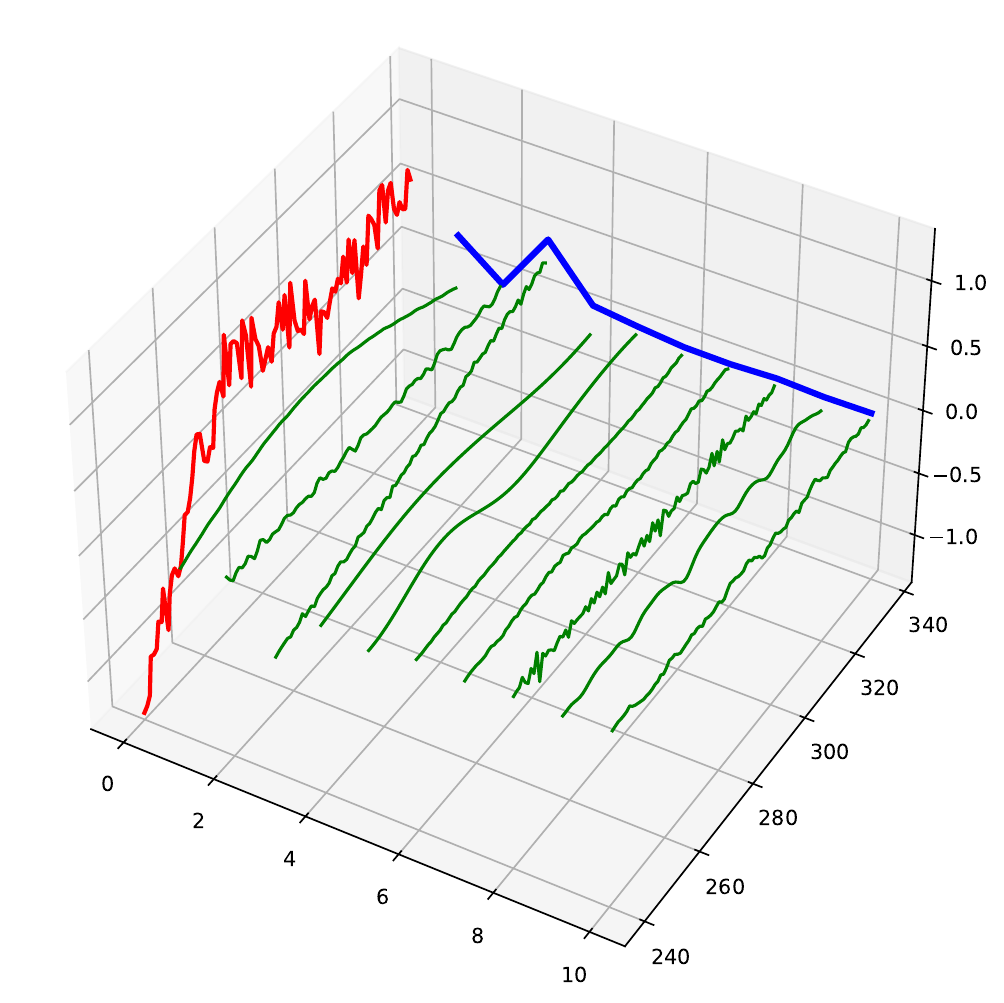}
\end{minipage}%
\begin{minipage}[t]{0.19\linewidth}
\centering
\includegraphics[width=\textwidth,height=0.8\textwidth]{./fig/PEMS03/your_3d_plot1.pdf}
\end{minipage}%
\begin{minipage}[t]{0.19\linewidth}
\centering
\includegraphics[width=\textwidth,height=0.8\textwidth]{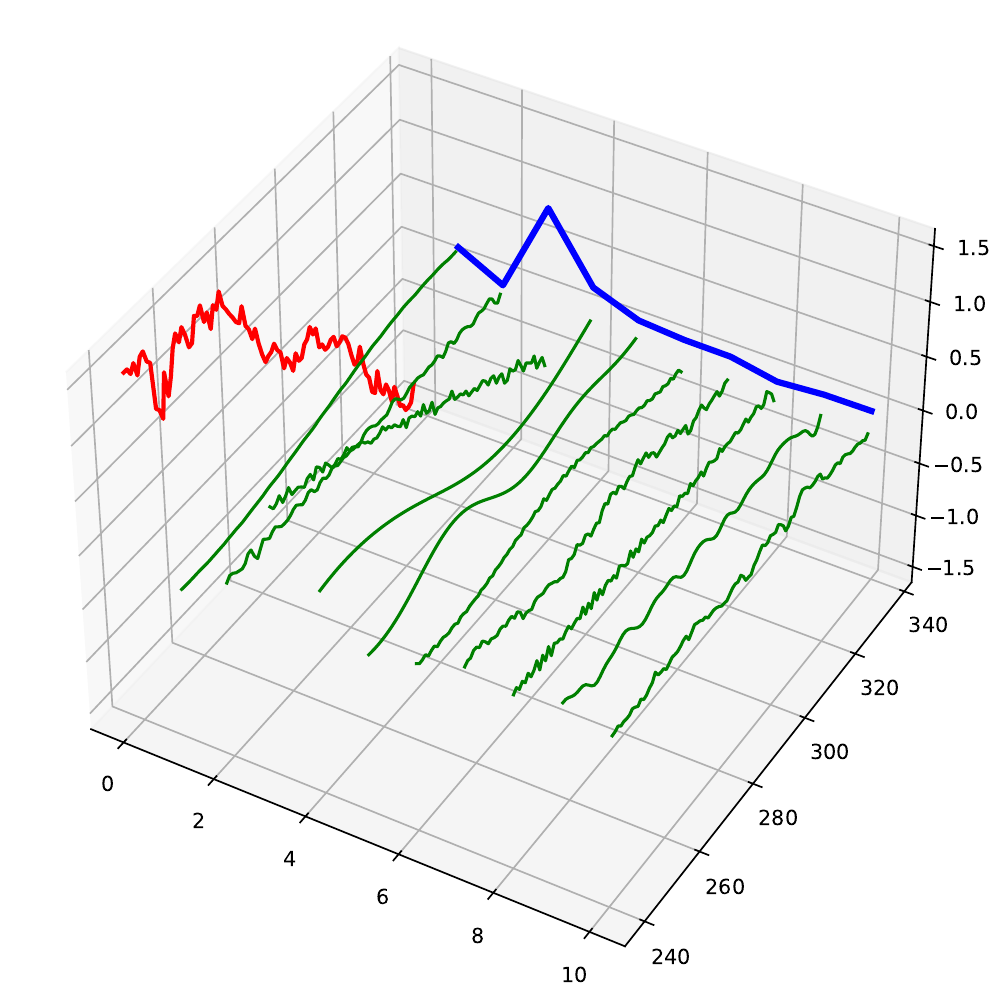}
\end{minipage}
\begin{minipage}[t]{0.19\linewidth}
\centering
\includegraphics[width=\textwidth,height=0.8\textwidth]{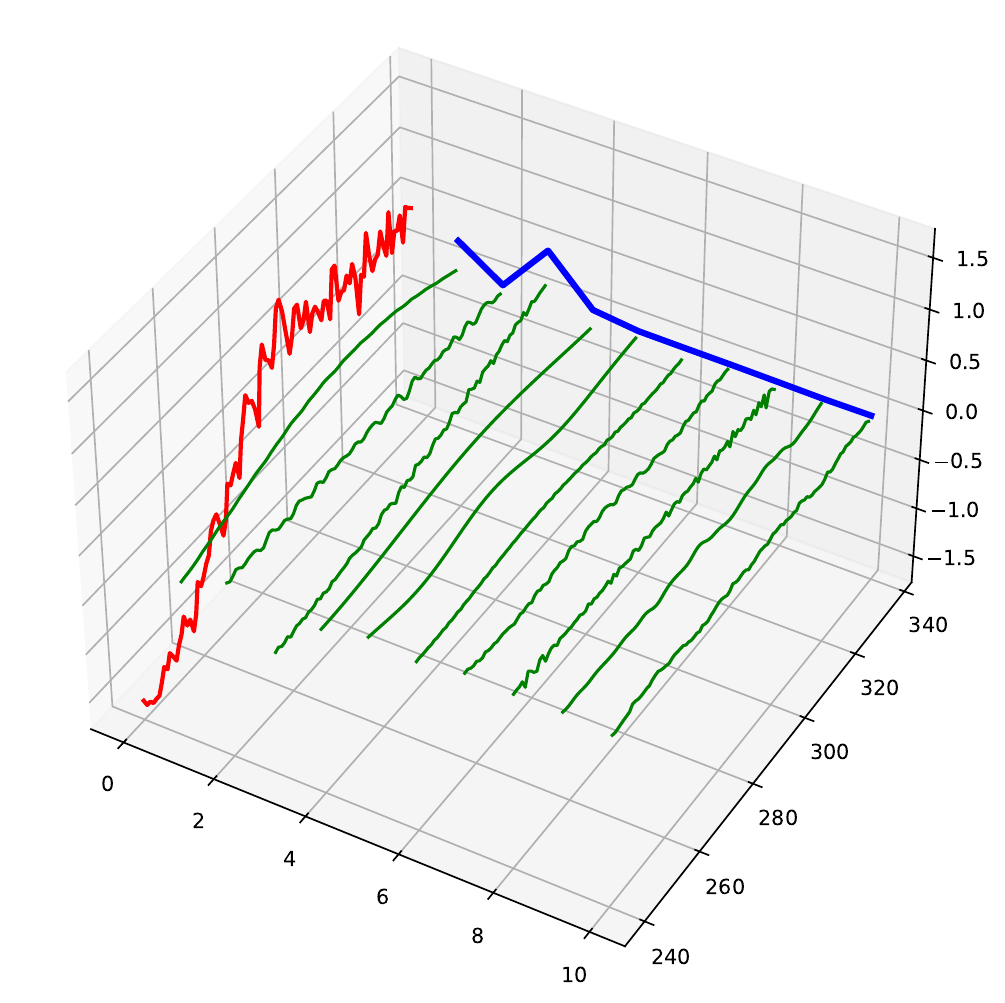}
\end{minipage}
\begin{minipage}[t]{0.19\linewidth}
\centering
\includegraphics[width=\textwidth,height=0.8\textwidth]{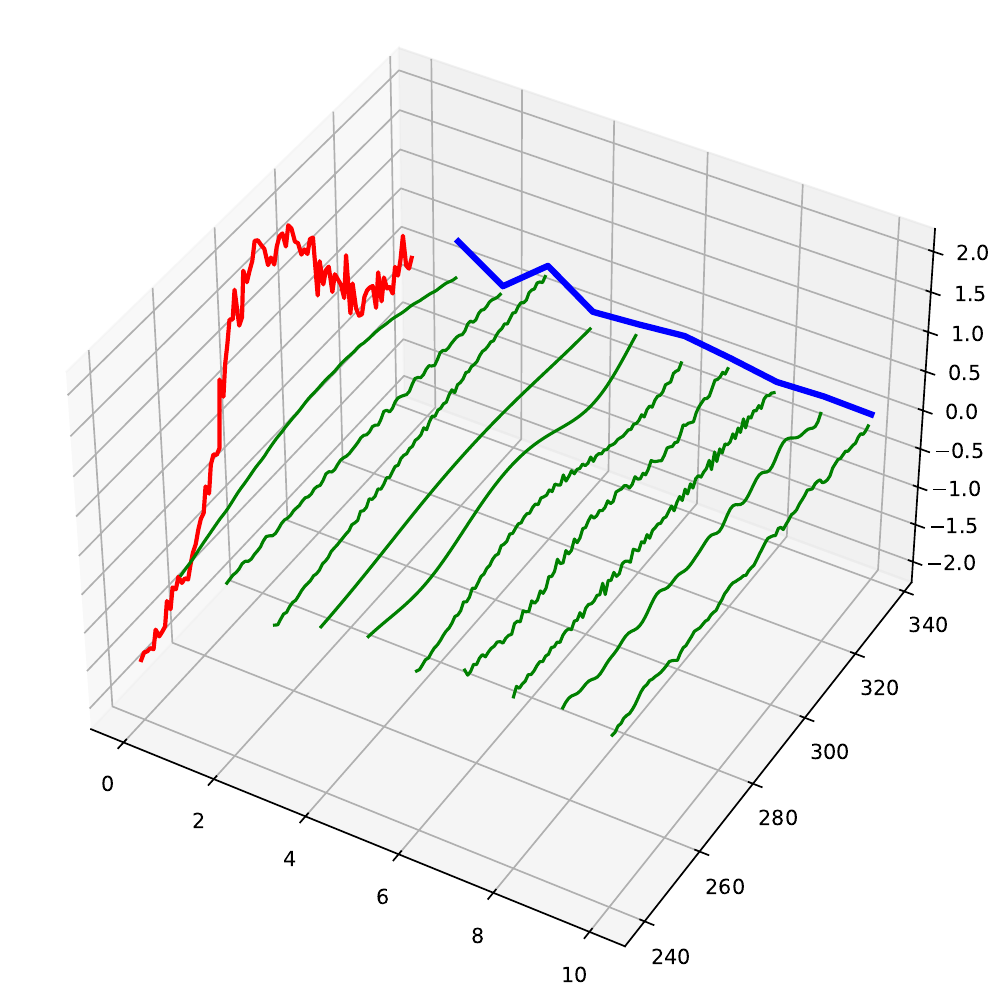}
\end{minipage}
\begin{minipage}[t]{0.19\linewidth}
\centering
\includegraphics[width=\textwidth,height=0.8\textwidth]{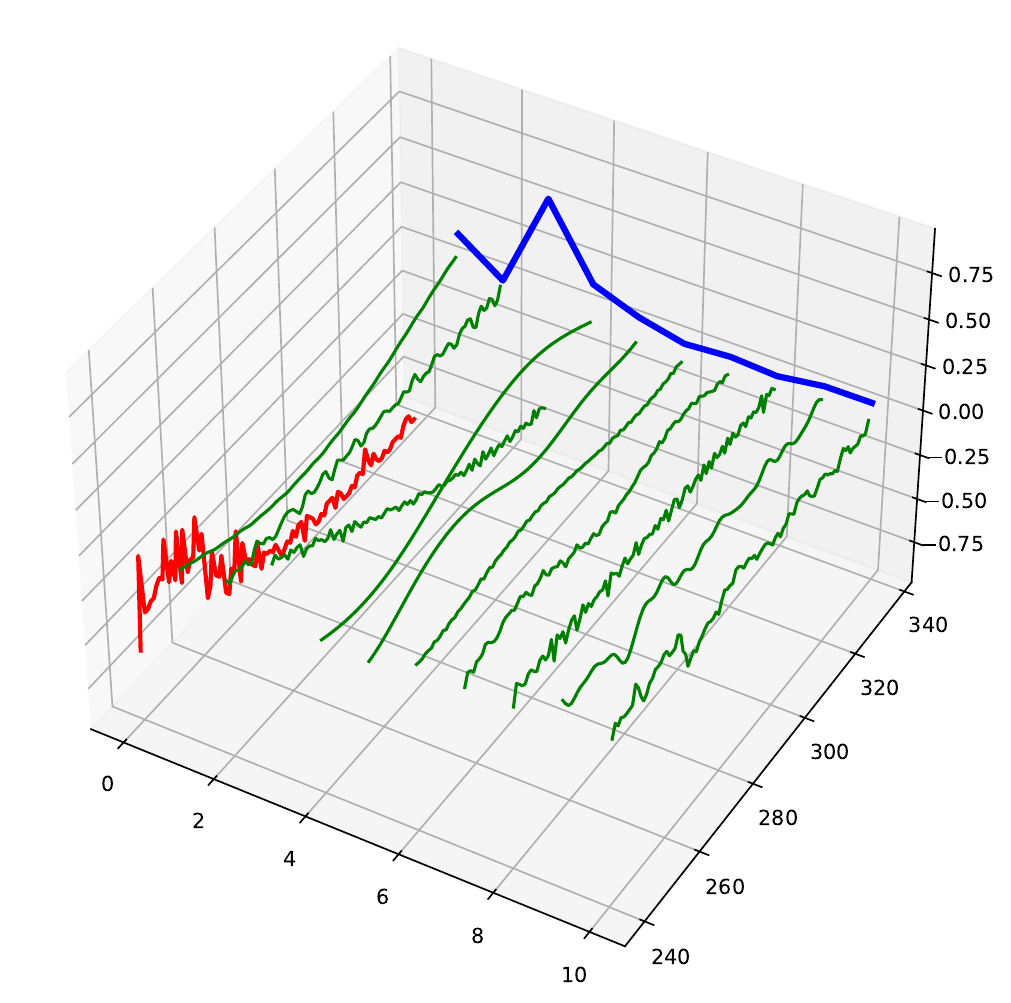}
\end{minipage}%
\begin{minipage}[t]{0.19\linewidth}
\centering
\includegraphics[width=\textwidth,height=0.8\textwidth]{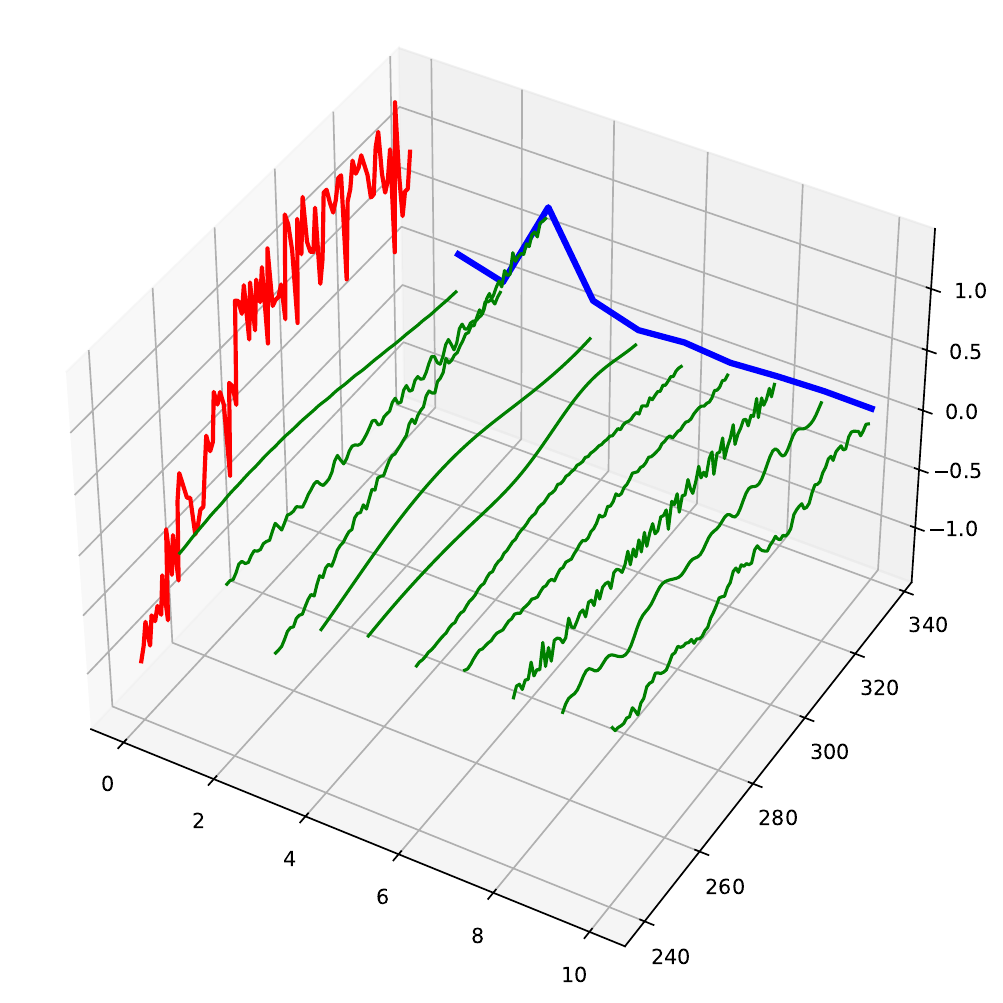}
\end{minipage}%
\begin{minipage}[t]{0.19\linewidth}
\centering
\includegraphics[width=\textwidth,height=0.8\textwidth]{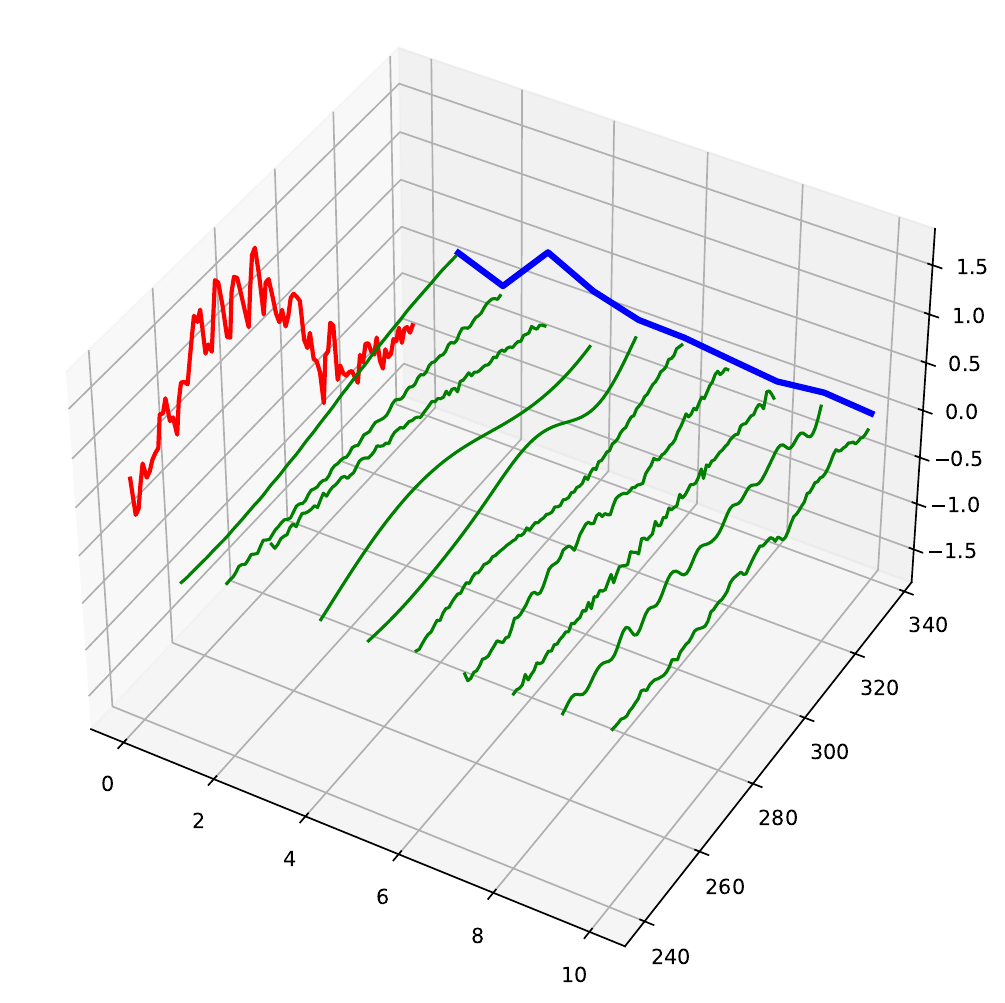}
\end{minipage}
\begin{minipage}[t]{0.19\linewidth}
\centering
\includegraphics[width=\textwidth,height=0.8\textwidth]{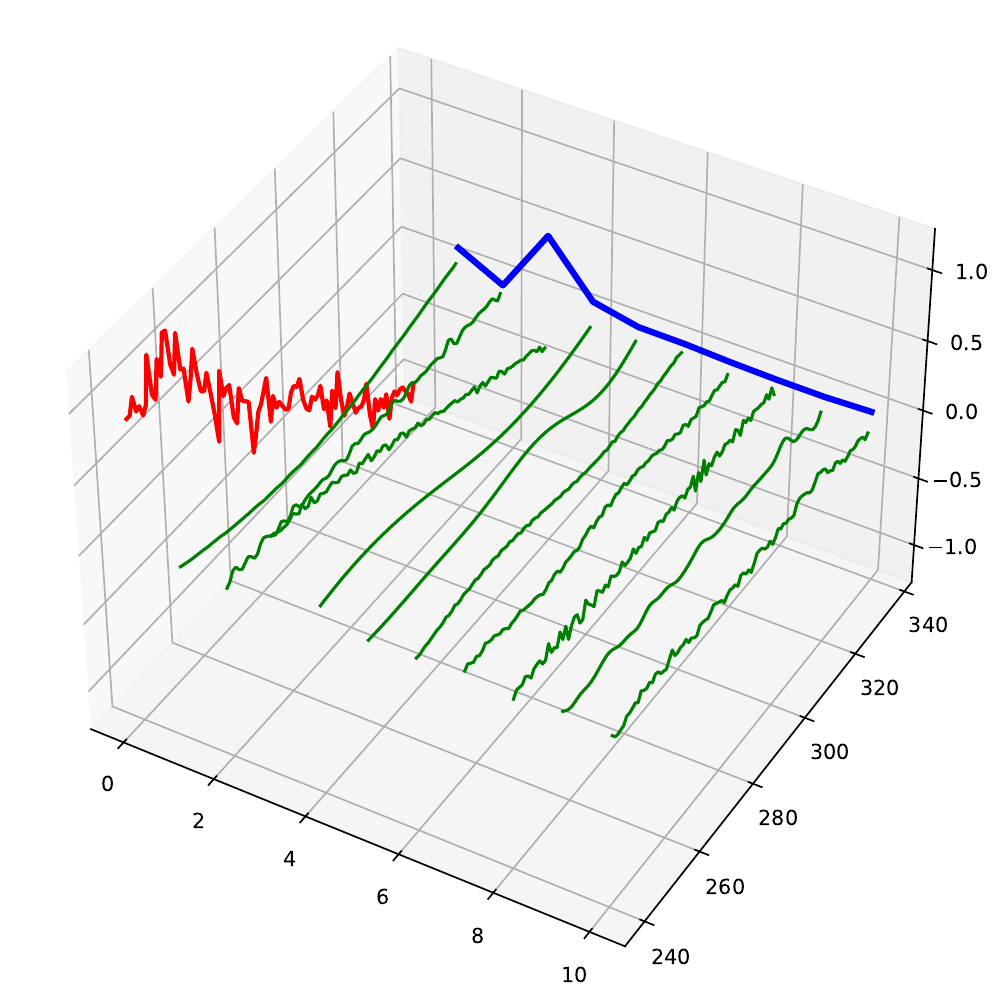}
\end{minipage}
\begin{minipage}[t]{0.19\linewidth}
\centering
\includegraphics[width=\textwidth,height=0.8\textwidth]{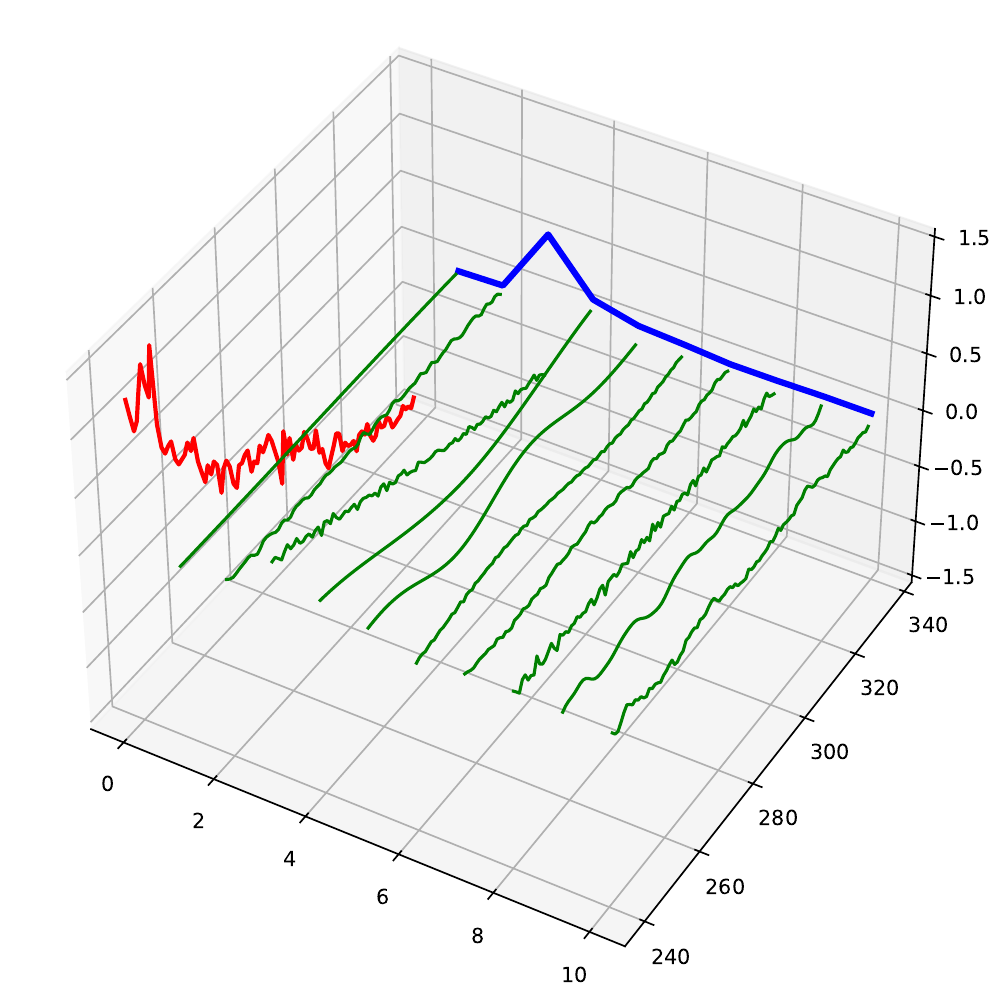}
\end{minipage}
\caption{Visualization of the MLOW Decomposition for Ten Examples on PEMS03.}
\label{plot5}
\end{figure*}

\begin{figure*}[t]
\centering
\begin{minipage}[t]{0.19\linewidth}
\centering
\includegraphics[width=\textwidth,height=0.8\textwidth]{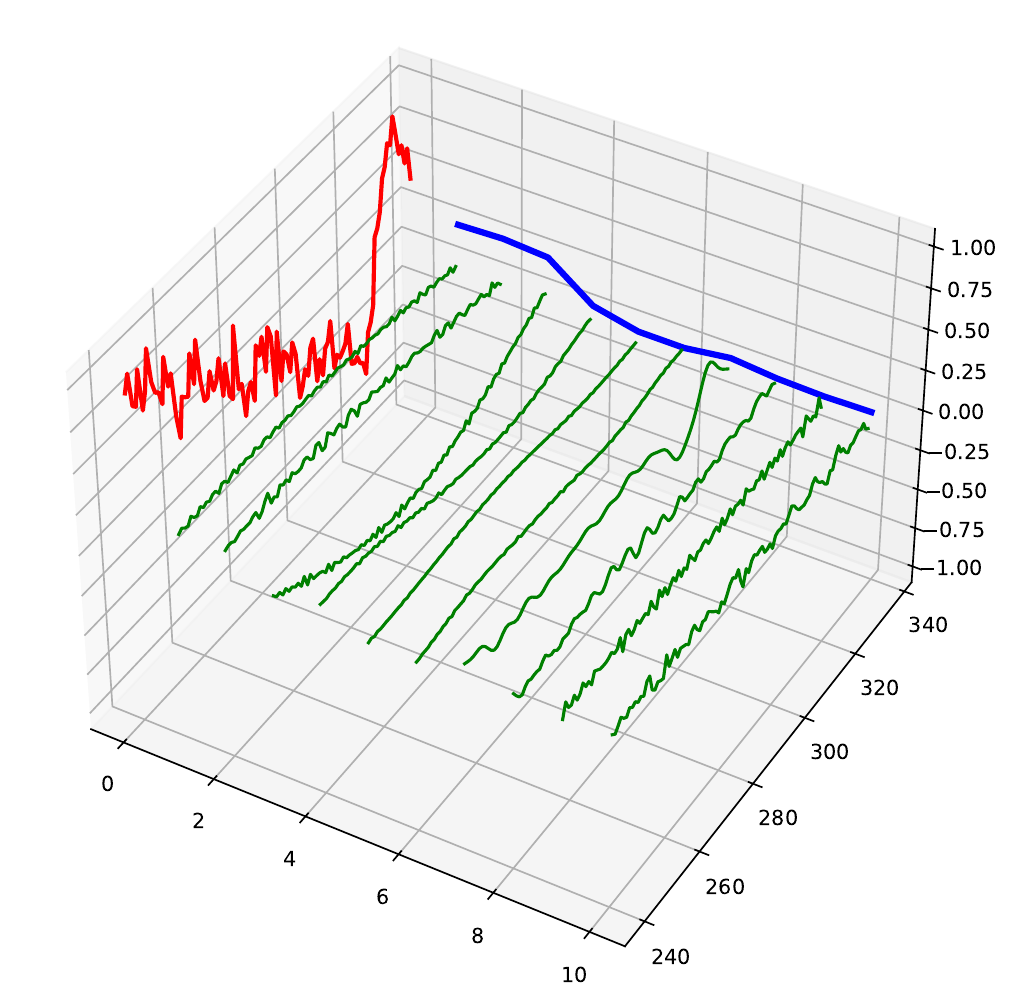}
\end{minipage}%
\begin{minipage}[t]{0.19\linewidth}
\centering
\includegraphics[width=\textwidth,height=0.8\textwidth]{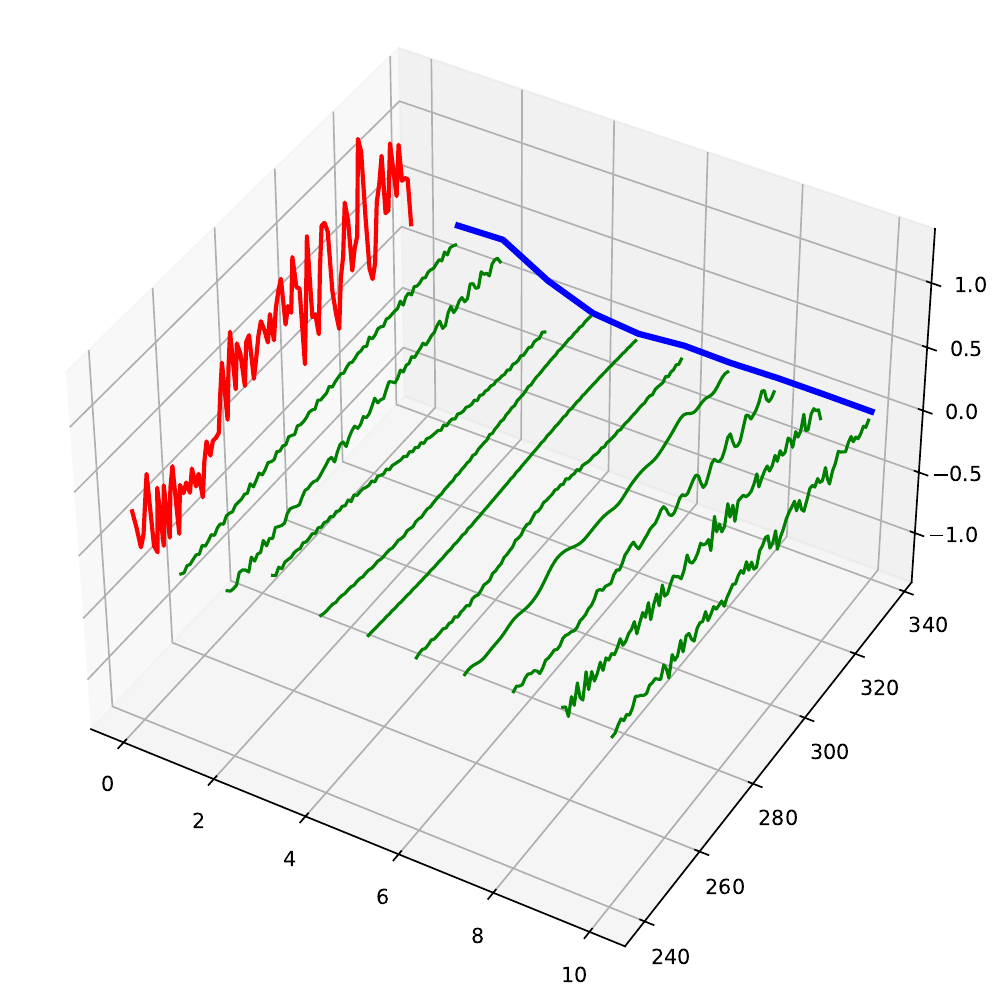}
\end{minipage}%
\begin{minipage}[t]{0.19\linewidth}
\centering
\includegraphics[width=\textwidth,height=0.8\textwidth]{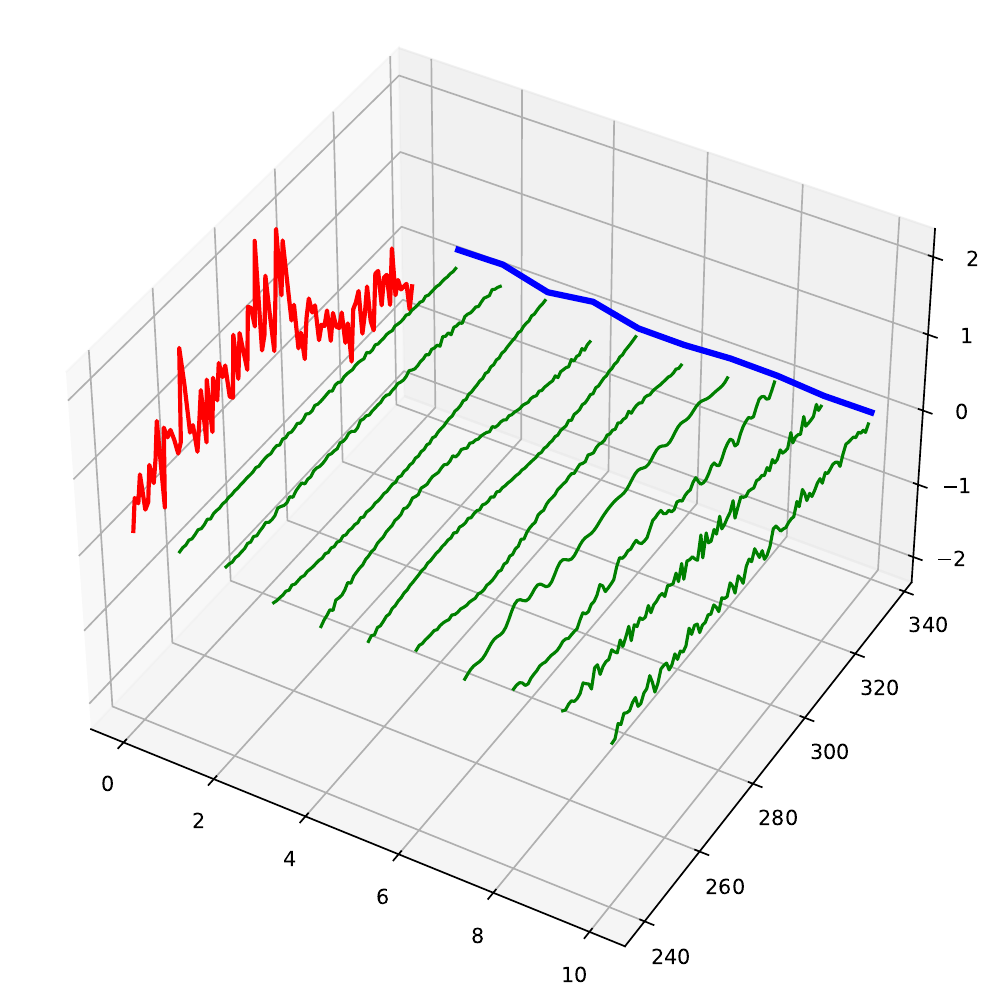}
\end{minipage}
\begin{minipage}[t]{0.19\linewidth}
\centering
\includegraphics[width=\textwidth,height=0.8\textwidth]{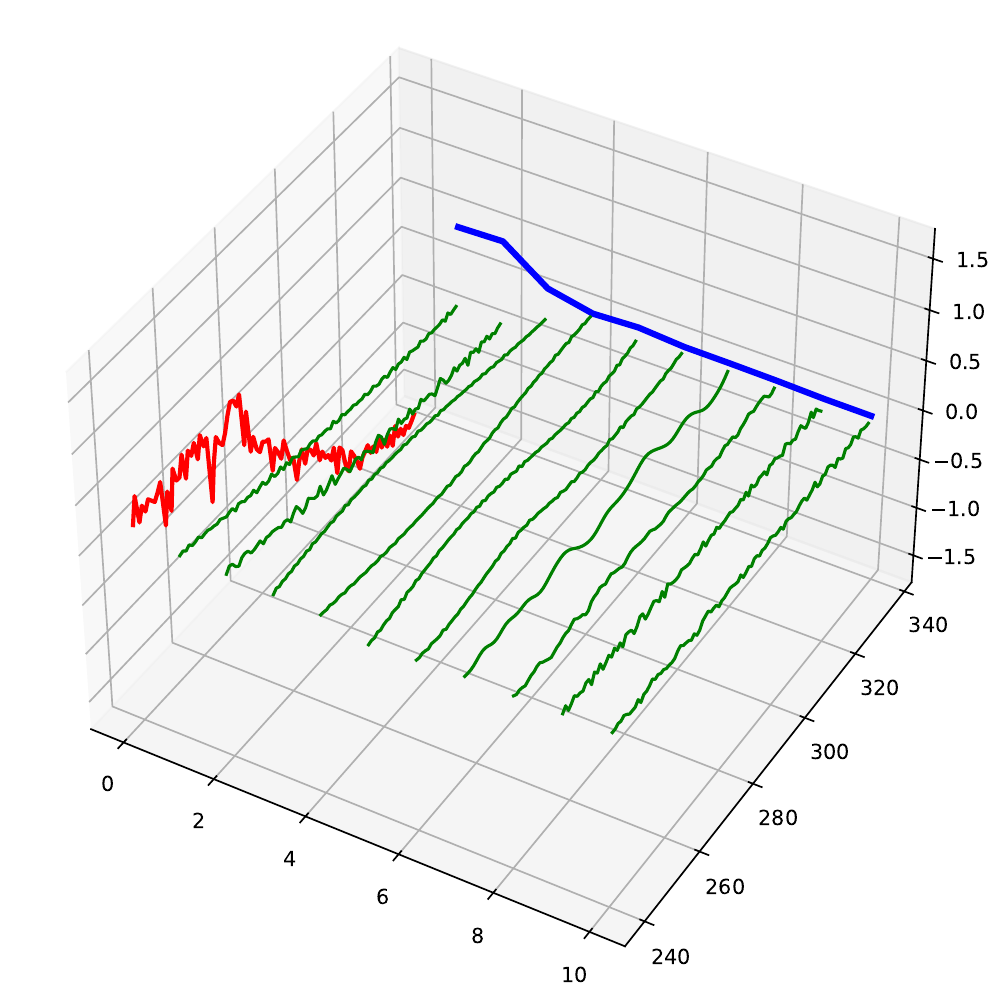}
\end{minipage}
\begin{minipage}[t]{0.19\linewidth}
\centering
\includegraphics[width=\textwidth,height=0.8\textwidth]{./fig/PEMS08/your_3d_plot4.pdf}
\end{minipage}
\begin{minipage}[t]{0.19\linewidth}
\centering
\includegraphics[width=\textwidth,height=0.8\textwidth]{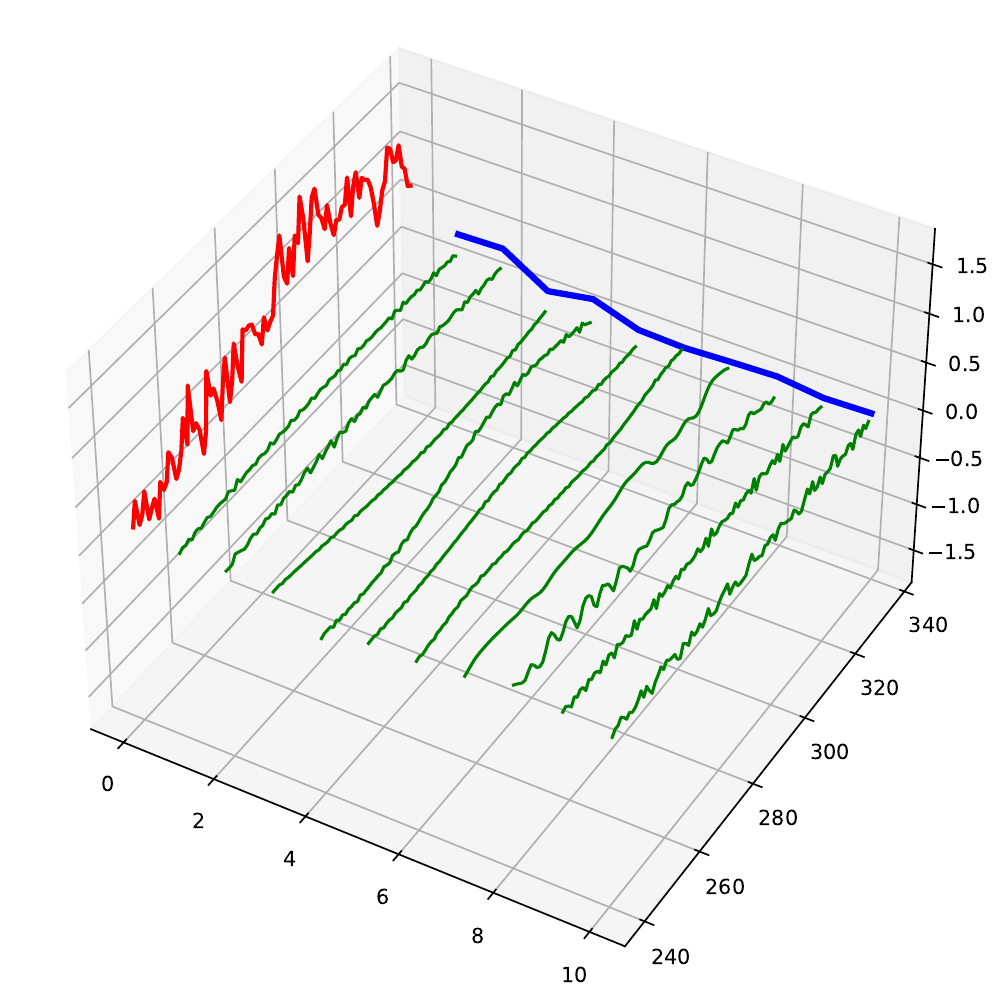}
\end{minipage}%
\begin{minipage}[t]{0.19\linewidth}
\centering
\includegraphics[width=\textwidth,height=0.8\textwidth]{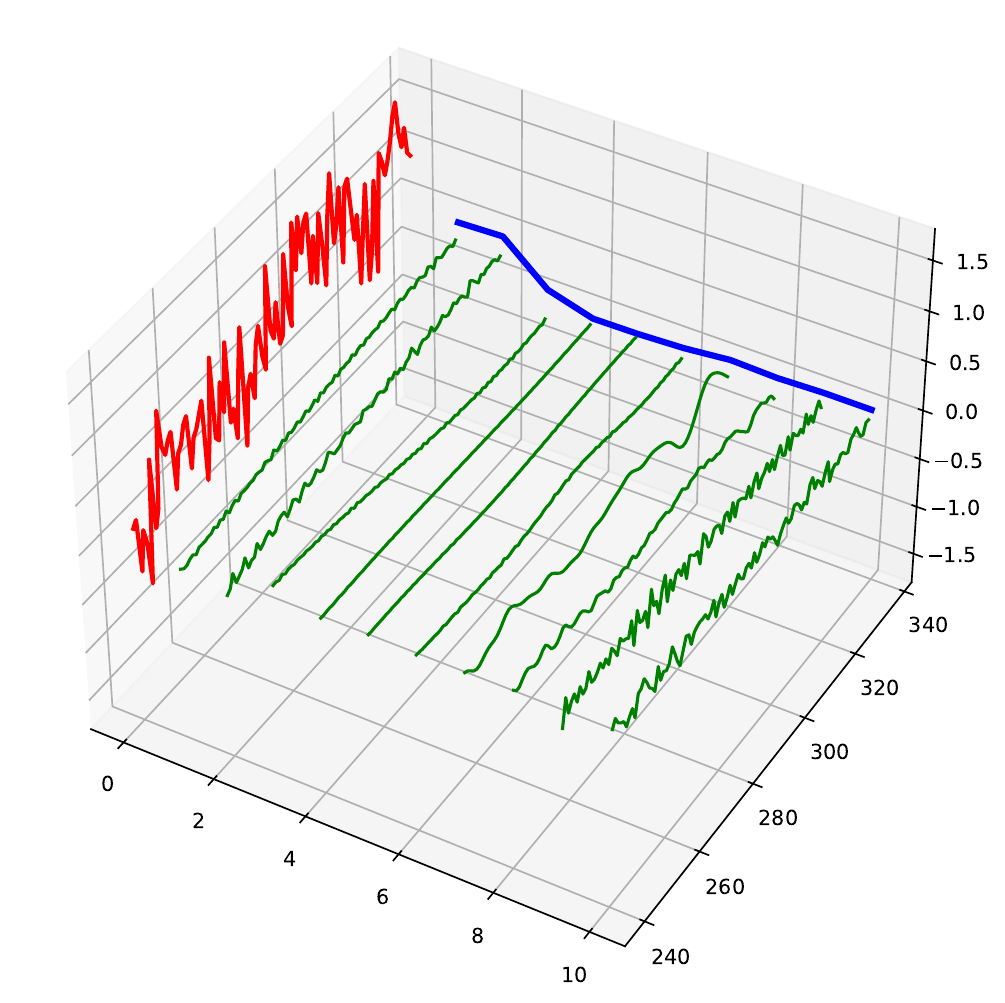}
\end{minipage}%
\begin{minipage}[t]{0.19\linewidth}
\centering
\includegraphics[width=\textwidth,height=0.8\textwidth]{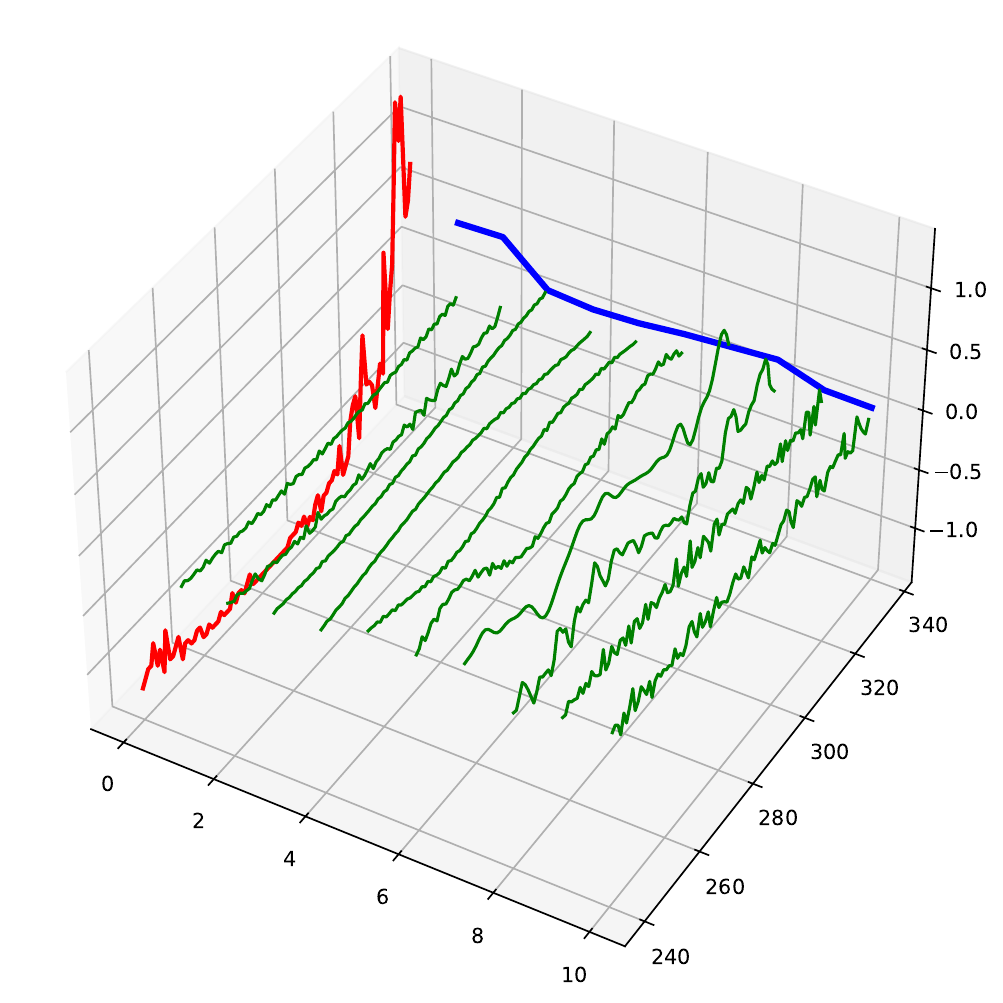}
\end{minipage}
\begin{minipage}[t]{0.19\linewidth}
\centering
\includegraphics[width=\textwidth,height=0.8\textwidth]{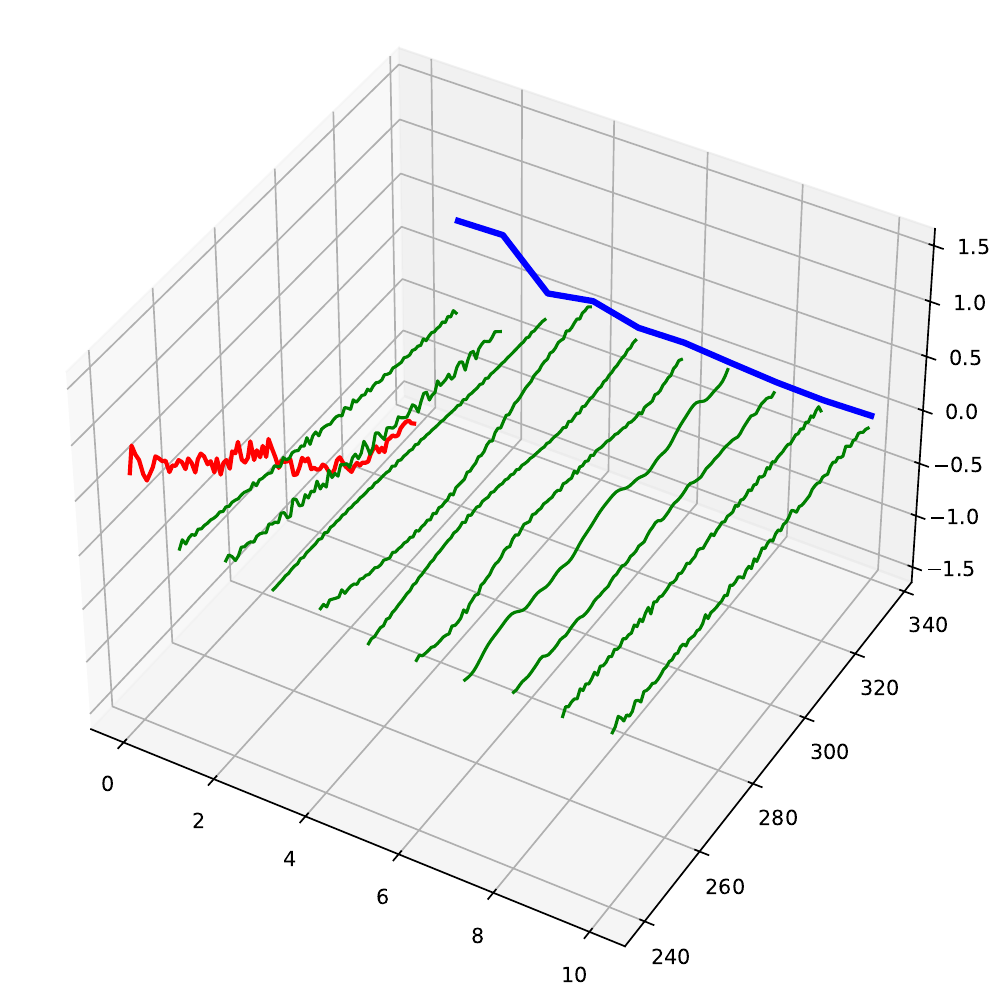}
\end{minipage}
\begin{minipage}[t]{0.19\linewidth}
\centering
\includegraphics[width=\textwidth,height=0.8\textwidth]{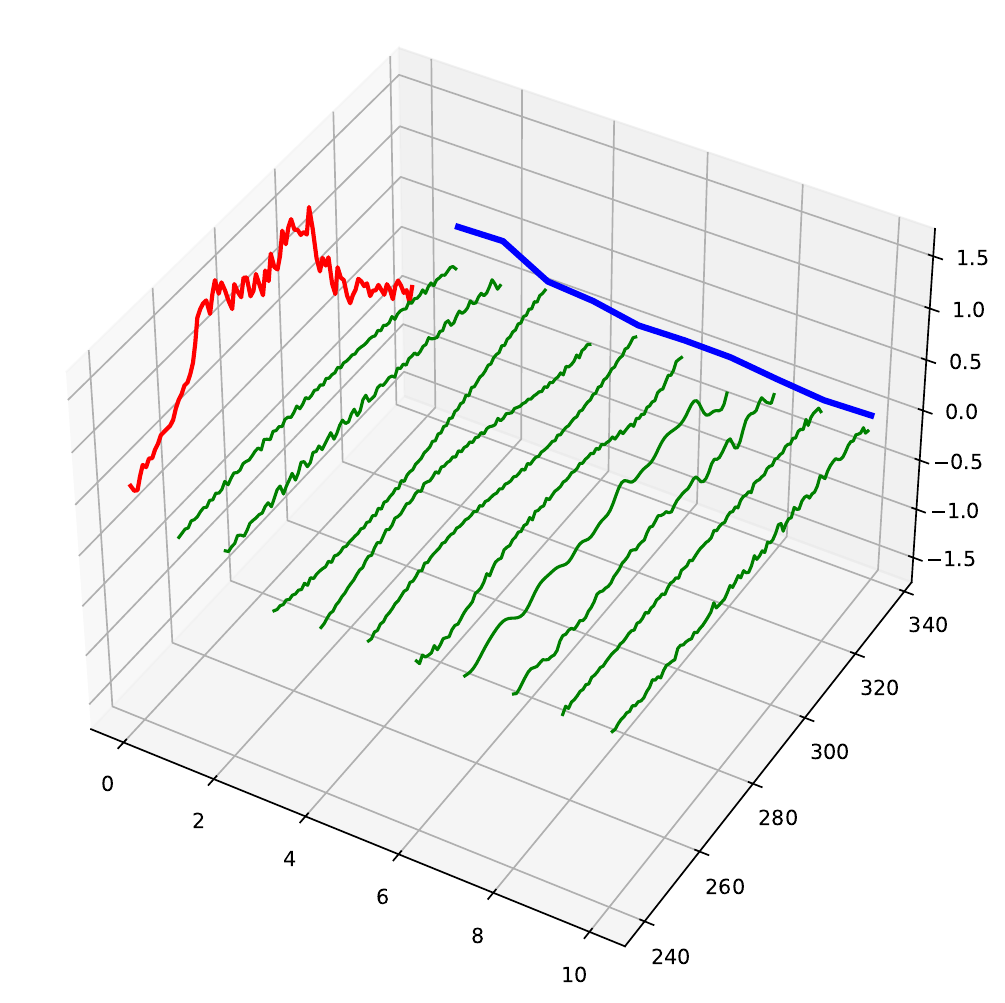}
\end{minipage}
\caption{Visualization of the MLOW Decomposition for Ten Examples on PEMS08.}
\label{plot6}
\end{figure*}

\begin{figure*}[t]
\centering
\begin{minipage}[t]{0.19\linewidth}
\centering
\includegraphics[width=\textwidth,height=0.8\textwidth]{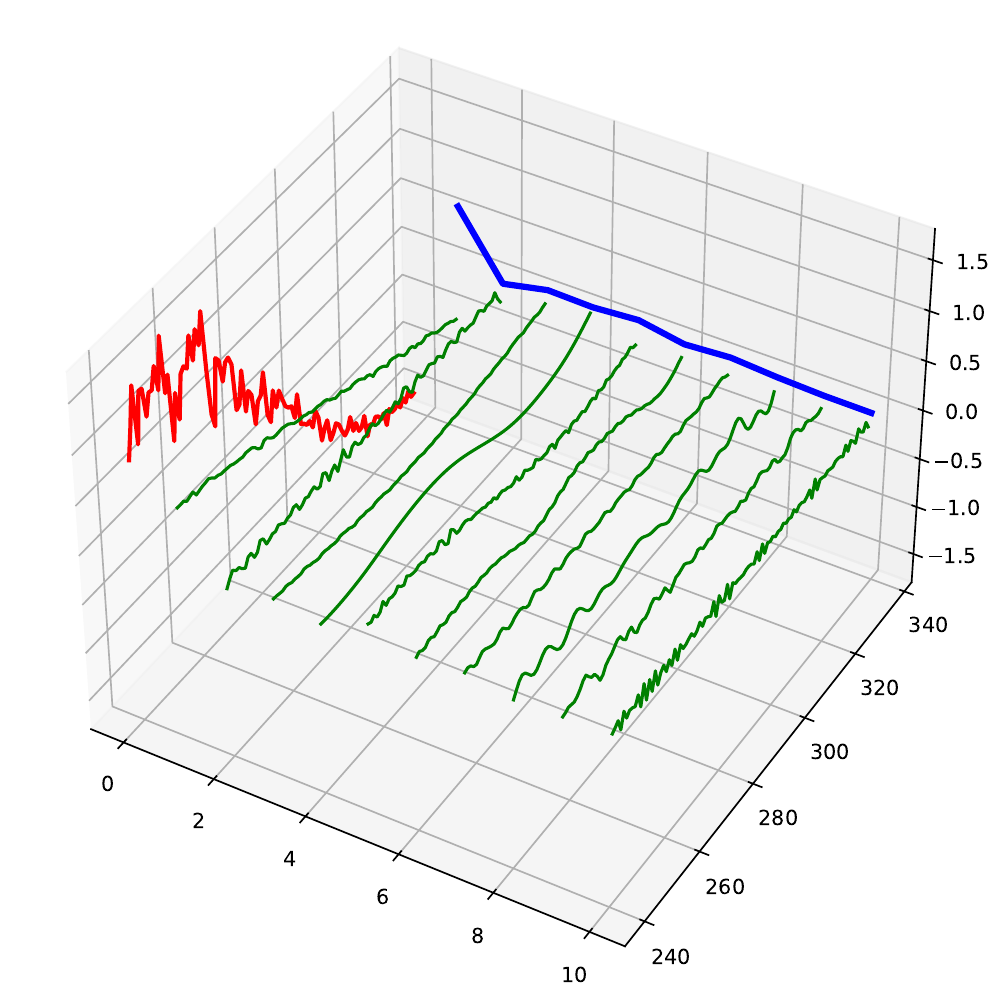}
\end{minipage}%
\begin{minipage}[t]{0.19\linewidth}
\centering
\includegraphics[width=\textwidth,height=0.8\textwidth]{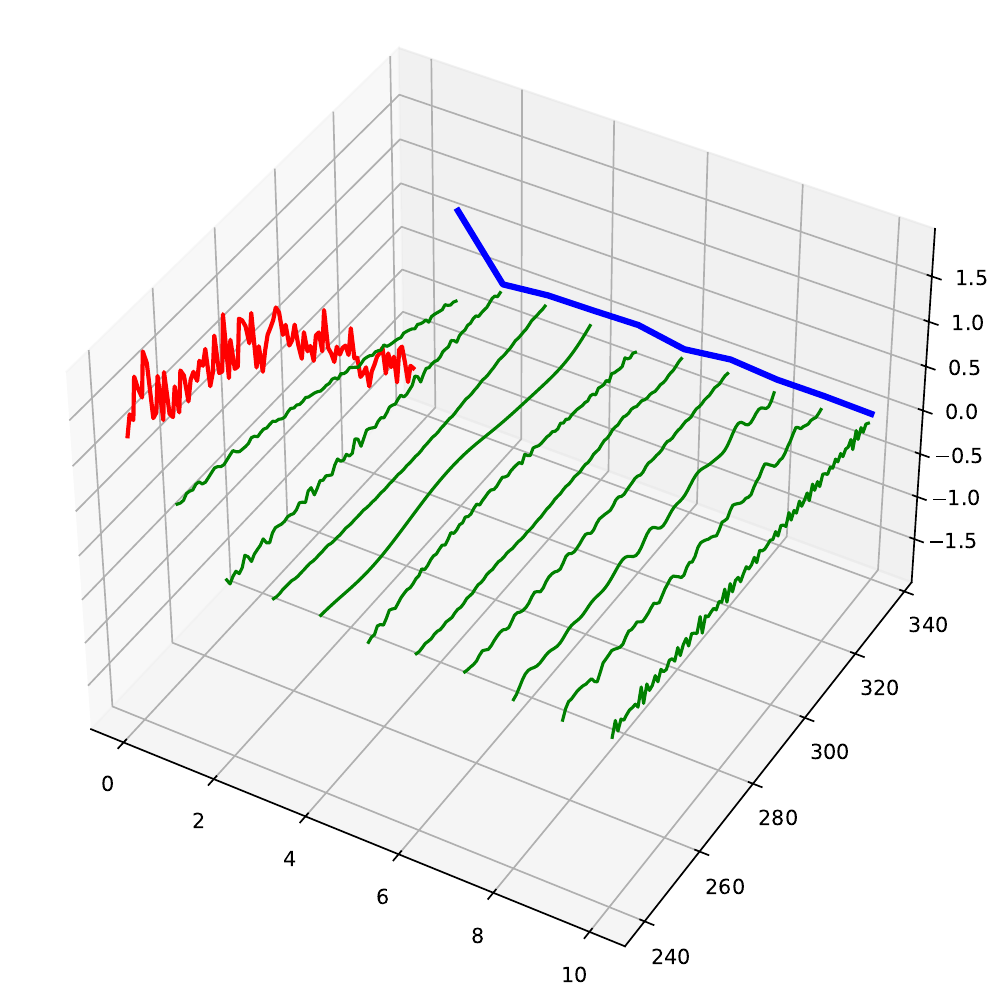}
\end{minipage}%
\begin{minipage}[t]{0.19\linewidth}
\centering
\includegraphics[width=\textwidth,height=0.8\textwidth]{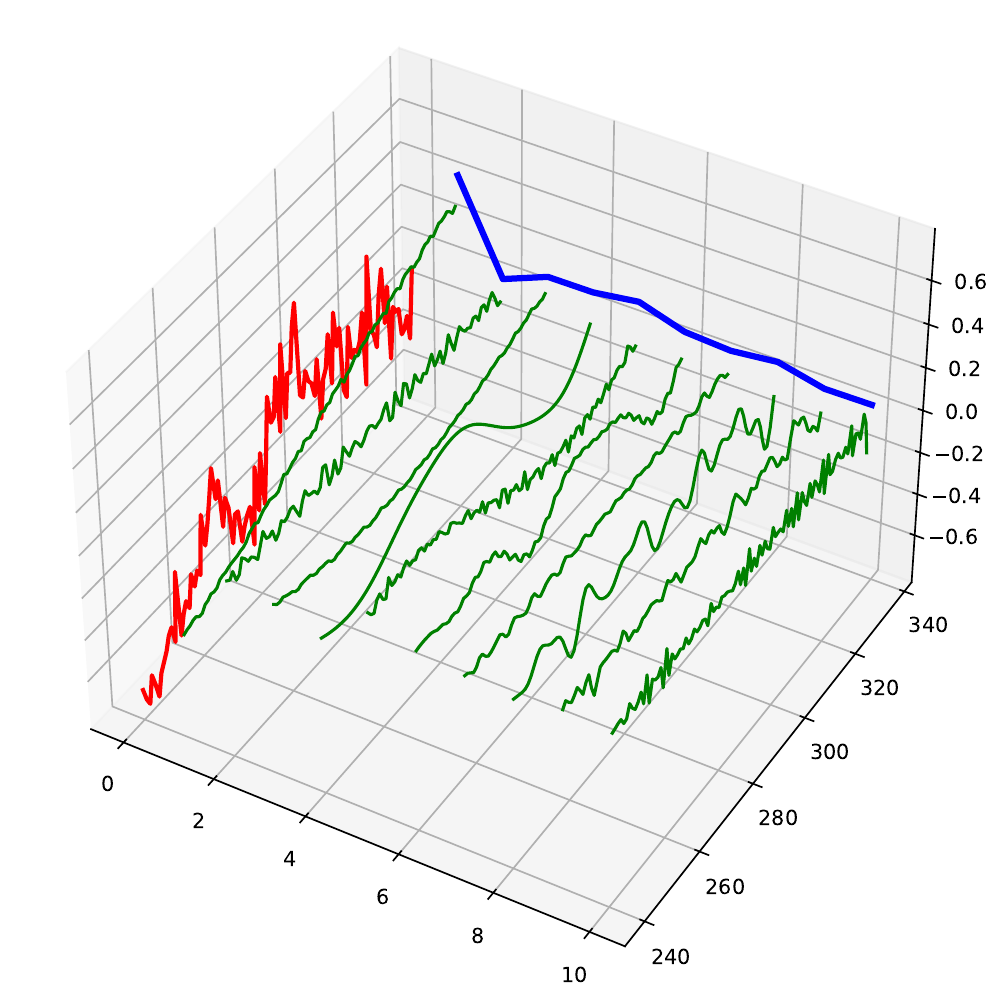}
\end{minipage}
\begin{minipage}[t]{0.19\linewidth}
\centering
\includegraphics[width=\textwidth,height=0.8\textwidth]{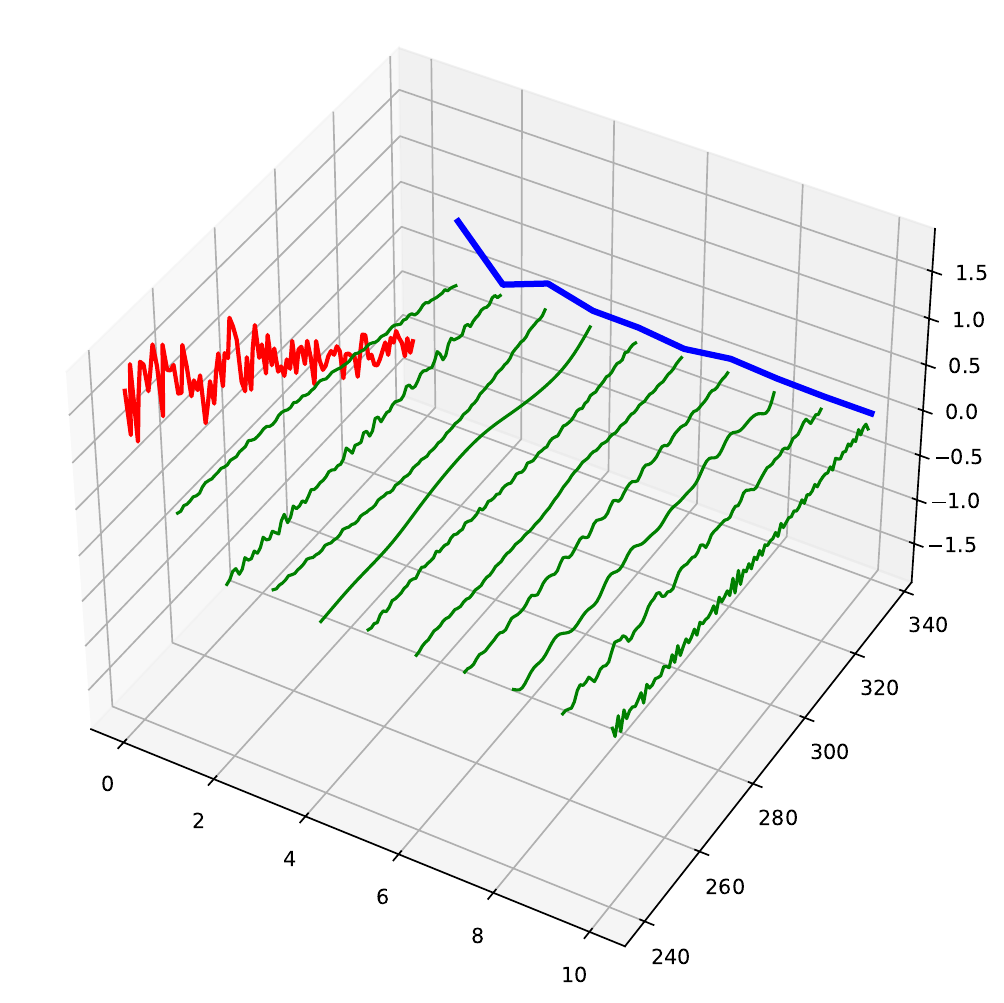}
\end{minipage}
\begin{minipage}[t]{0.19\linewidth}
\centering
\includegraphics[width=\textwidth,height=0.8\textwidth]{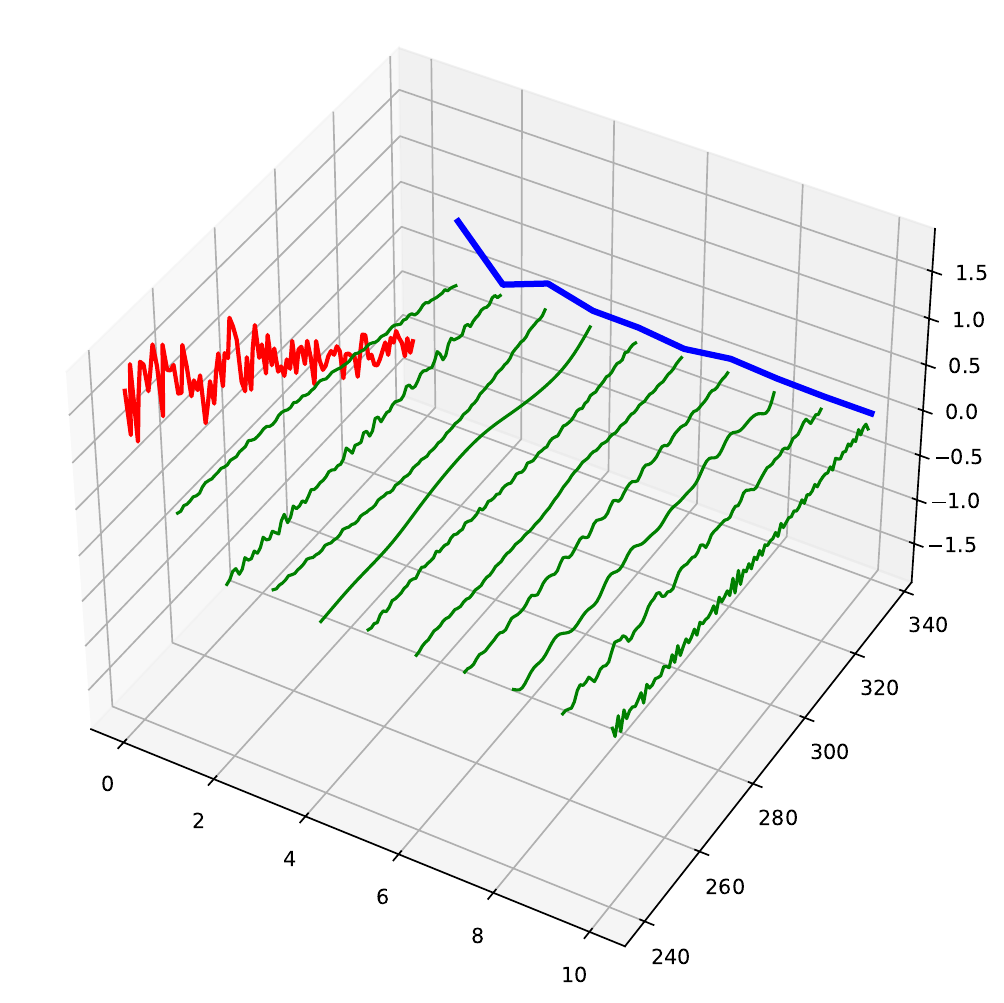}
\end{minipage}
\begin{minipage}[t]{0.19\linewidth}
\centering
\includegraphics[width=\textwidth,height=0.8\textwidth]{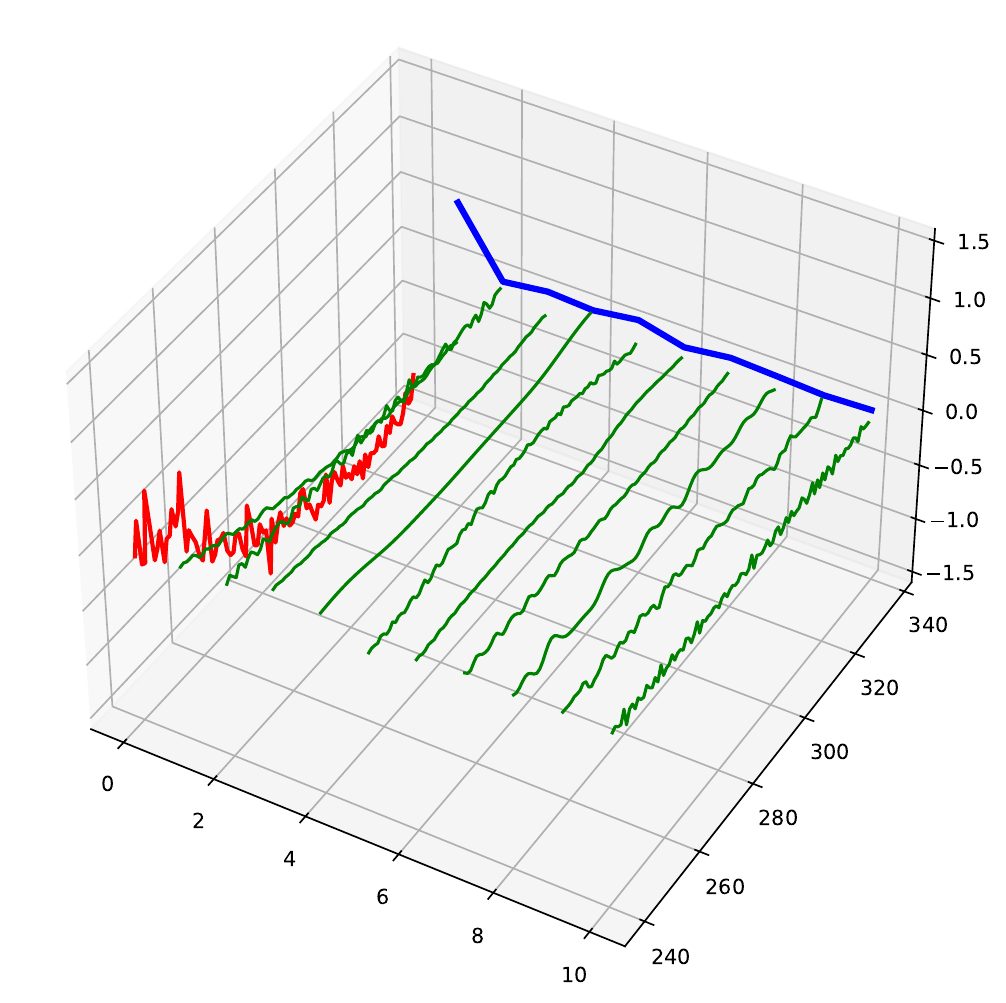}
\end{minipage}%
\begin{minipage}[t]{0.19\linewidth}
\centering
\includegraphics[width=\textwidth,height=0.8\textwidth]{./fig/PEMS04/your_3d_plot6.pdf}
\end{minipage}%
\begin{minipage}[t]{0.19\linewidth}
\centering
\includegraphics[width=\textwidth,height=0.8\textwidth]{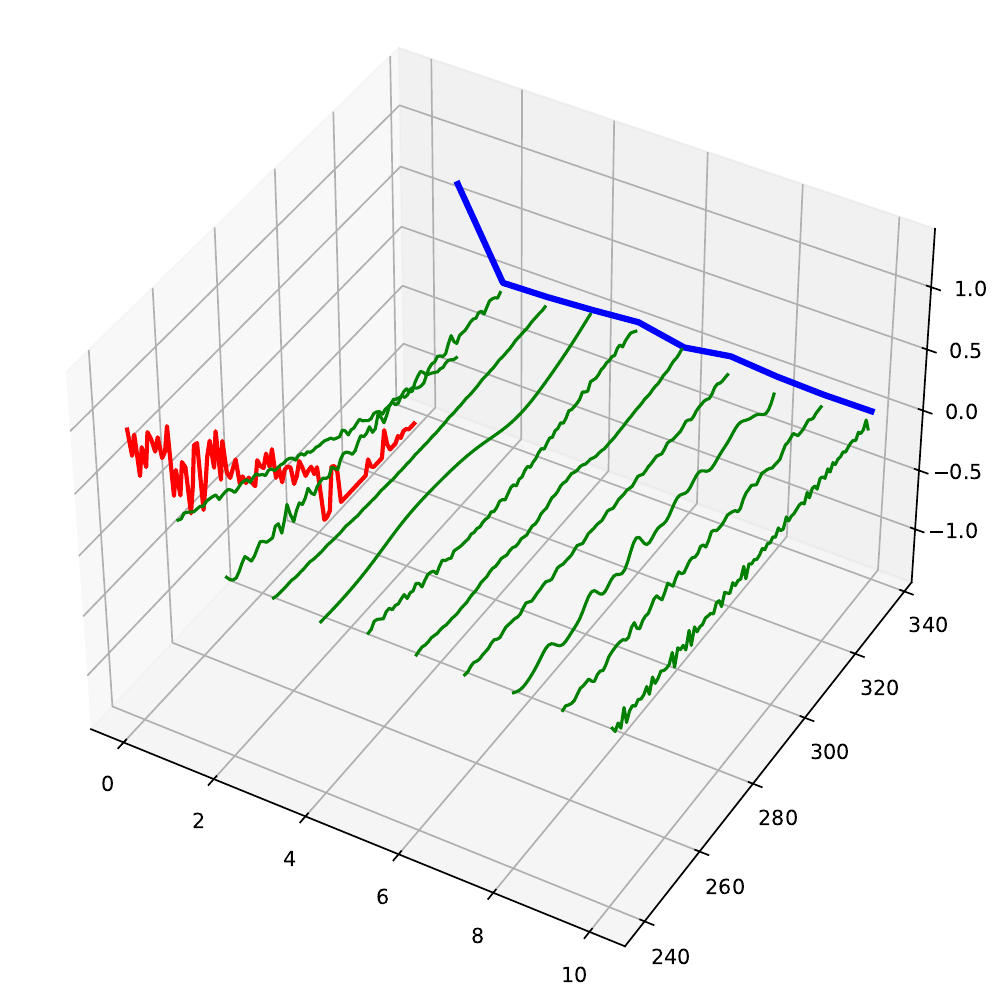}
\end{minipage}
\begin{minipage}[t]{0.19\linewidth}
\centering
\includegraphics[width=\textwidth,height=0.8\textwidth]{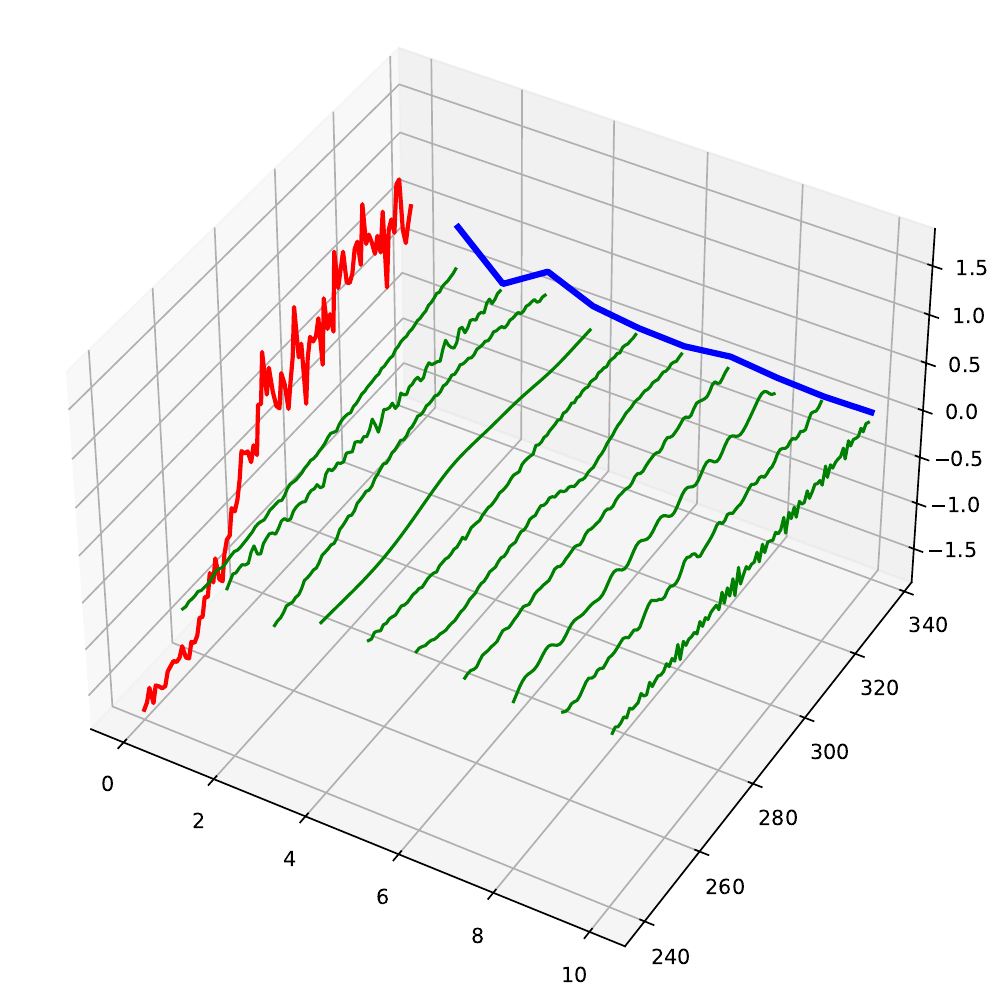}
\end{minipage}
\begin{minipage}[t]{0.19\linewidth}
\centering
\includegraphics[width=\textwidth,height=0.8\textwidth]{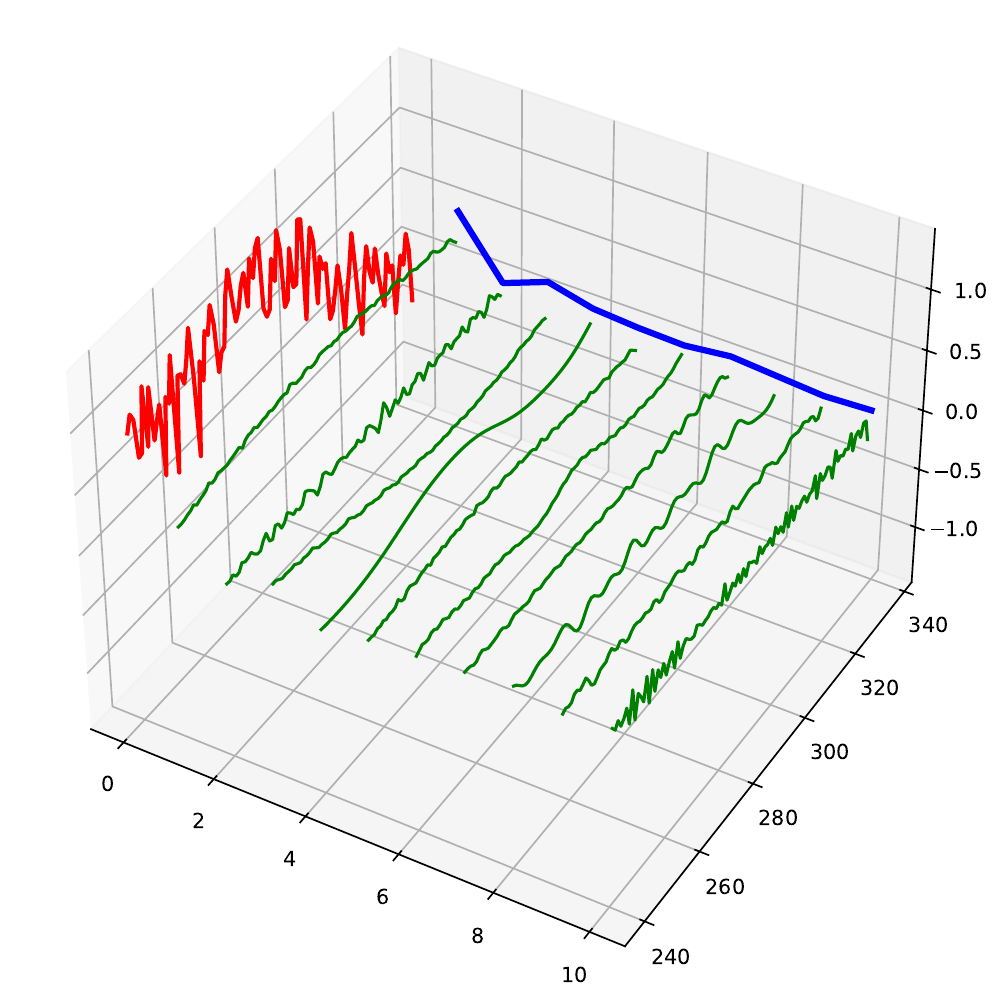}
\end{minipage}
\caption{Visualization of the MLOW Decomposition for Ten Examples on PEMS04.}
\label{plot7}
\end{figure*}

\begin{figure*}[t]
\centering
\begin{minipage}[t]{0.19\linewidth}
\centering
\includegraphics[width=\textwidth,height=0.8\textwidth]{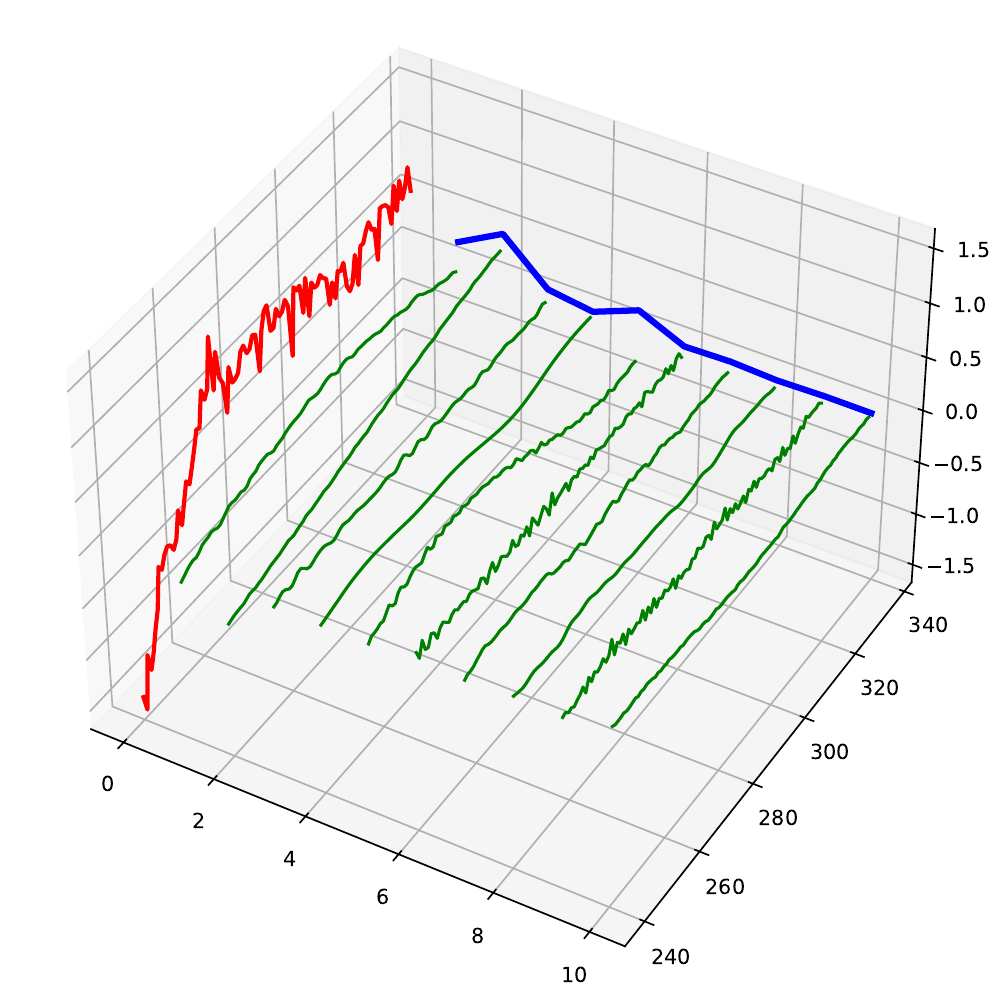}
\end{minipage}%
\begin{minipage}[t]{0.19\linewidth}
\centering
\includegraphics[width=\textwidth,height=0.8\textwidth]{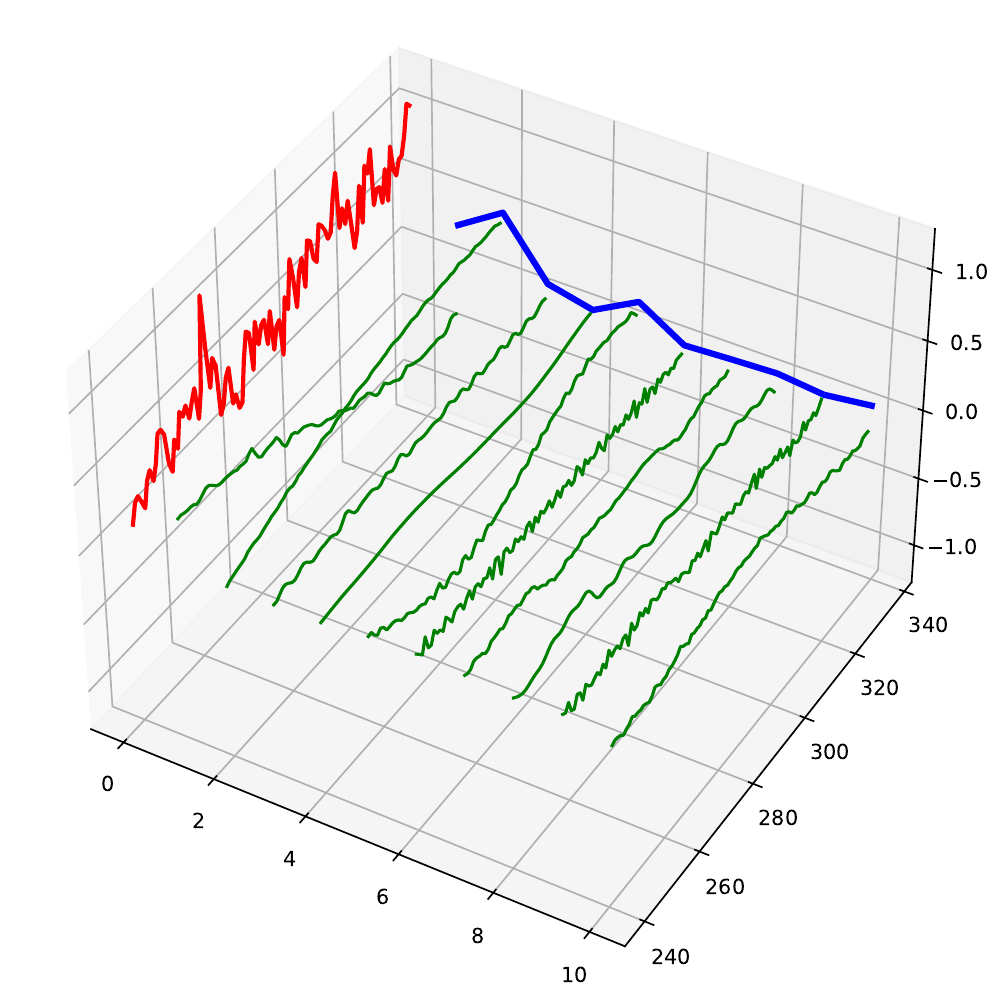}
\end{minipage}%
\begin{minipage}[t]{0.19\linewidth}
\centering
\includegraphics[width=\textwidth,height=0.8\textwidth]{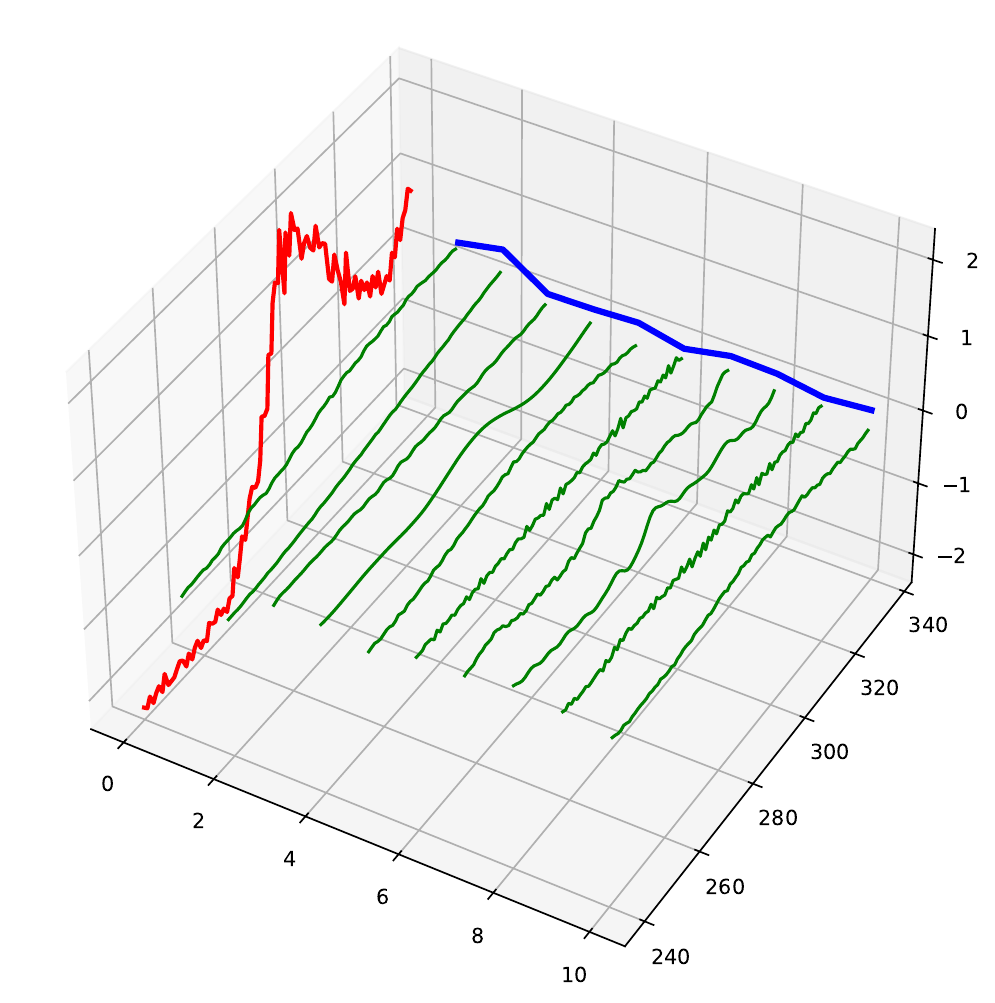}
\end{minipage}
\begin{minipage}[t]{0.19\linewidth}
\centering
\includegraphics[width=\textwidth,height=0.8\textwidth]{./fig/PEMS07/your_3d_plot3.pdf}
\end{minipage}
\begin{minipage}[t]{0.19\linewidth}
\centering
\includegraphics[width=\textwidth,height=0.8\textwidth]{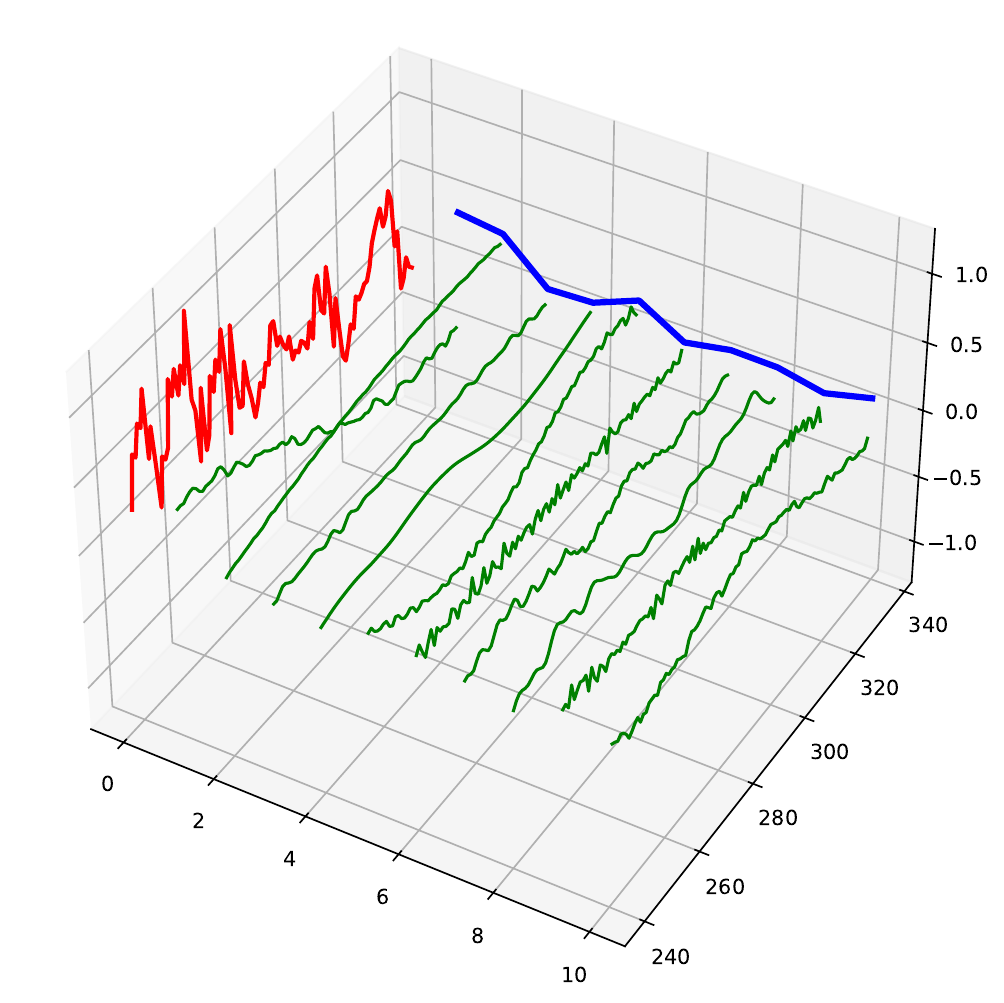}
\end{minipage}
\begin{minipage}[t]{0.19\linewidth}
\centering
\includegraphics[width=\textwidth,height=0.8\textwidth]{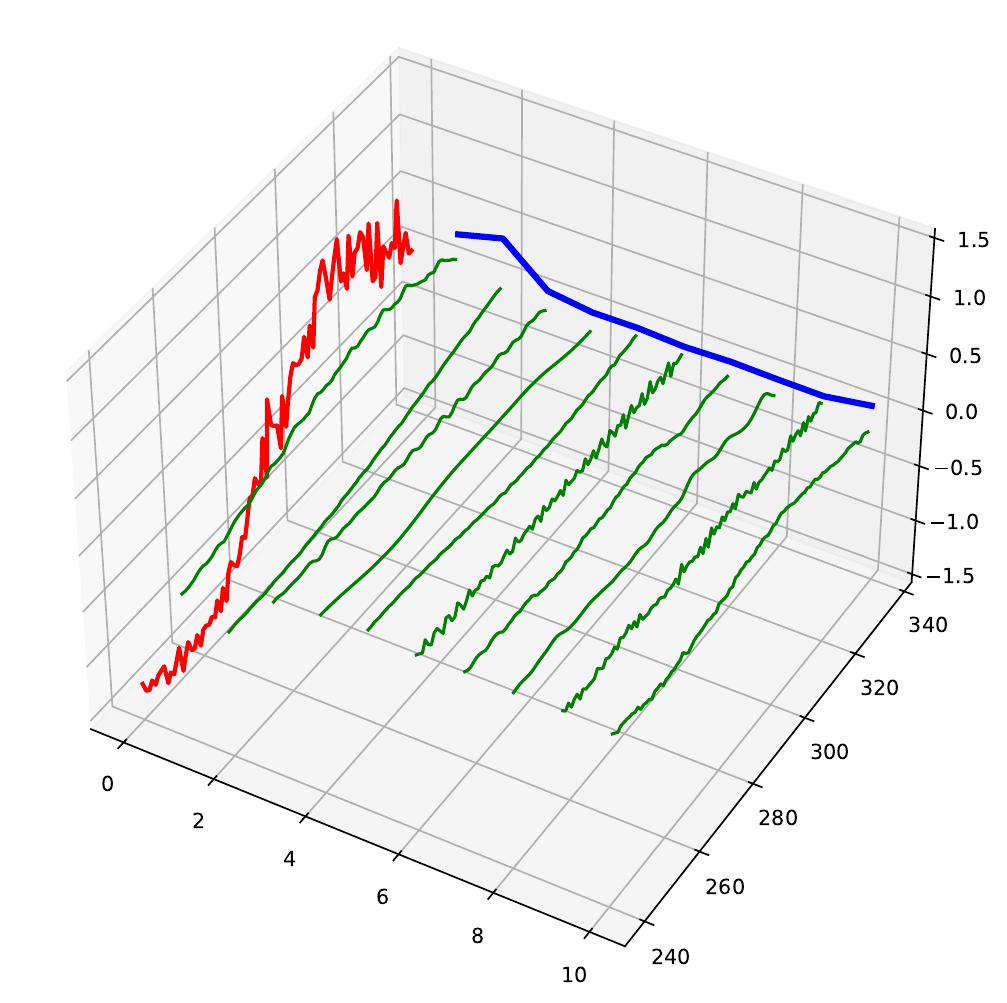}
\end{minipage}%
\begin{minipage}[t]{0.19\linewidth}
\centering
\includegraphics[width=\textwidth,height=0.8\textwidth]{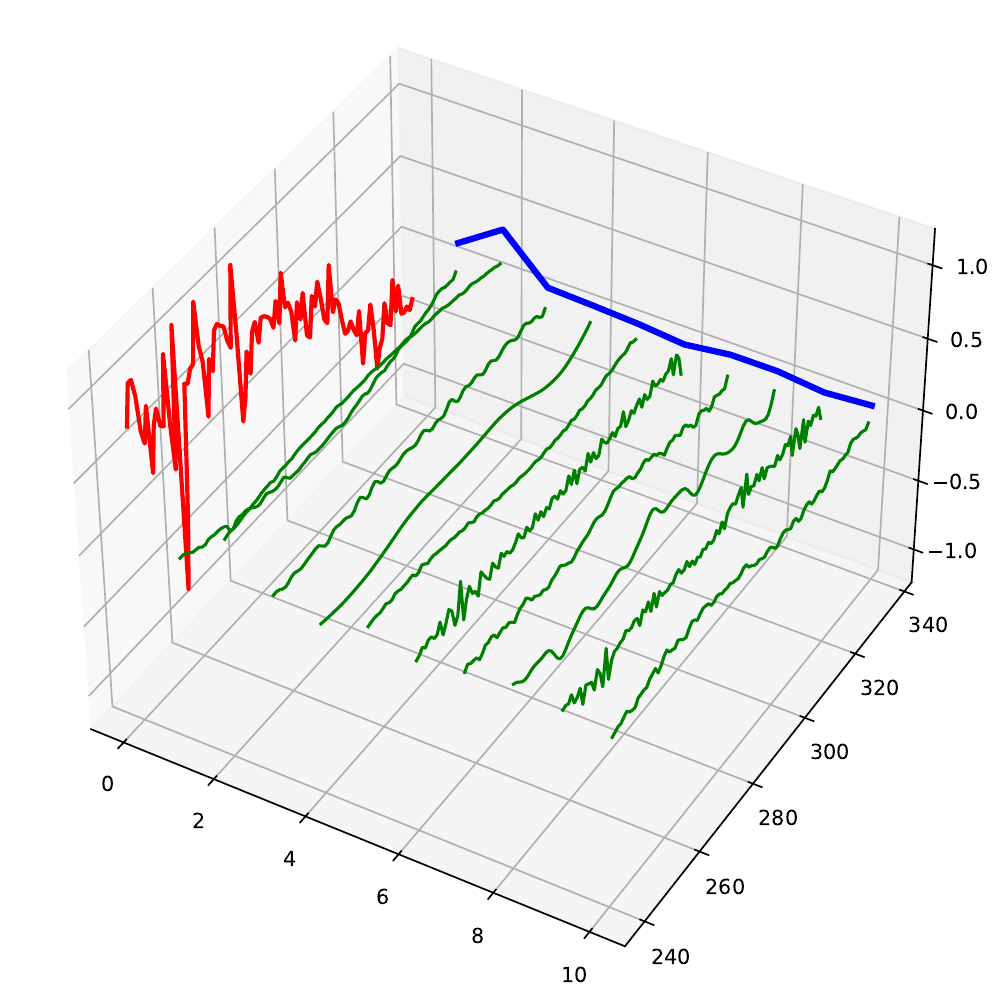}
\end{minipage}%
\begin{minipage}[t]{0.19\linewidth}
\centering
\includegraphics[width=\textwidth,height=0.8\textwidth]{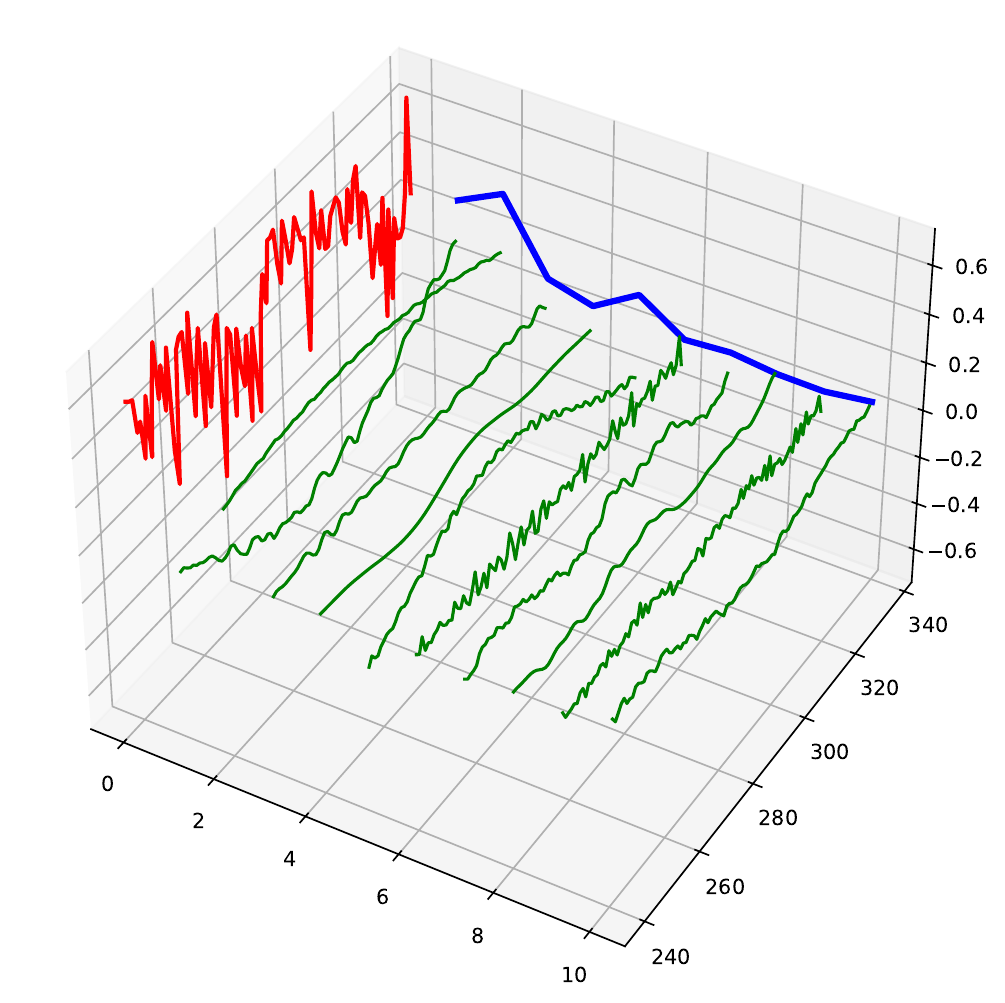}
\end{minipage}
\begin{minipage}[t]{0.19\linewidth}
\centering
\includegraphics[width=\textwidth,height=0.8\textwidth]{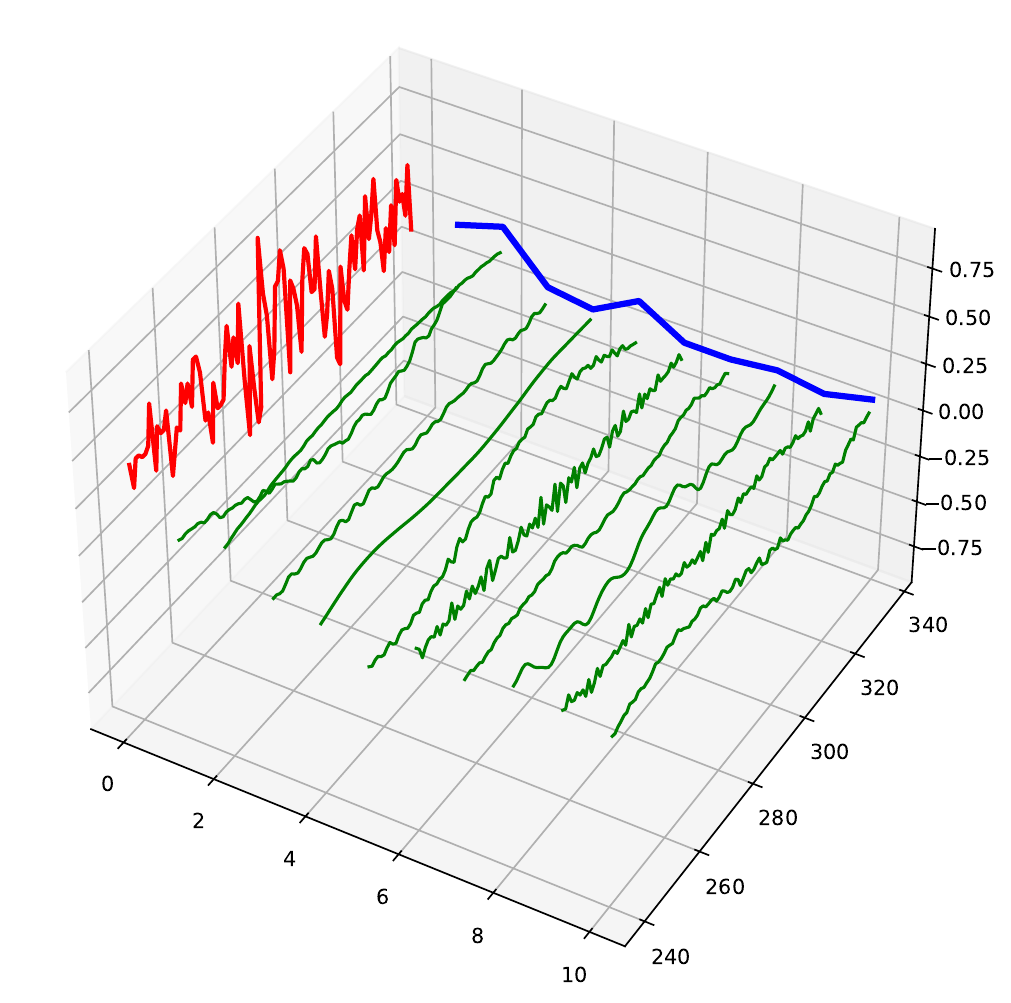}
\end{minipage}
\begin{minipage}[t]{0.19\linewidth}
\centering
\includegraphics[width=\textwidth,height=0.8\textwidth]{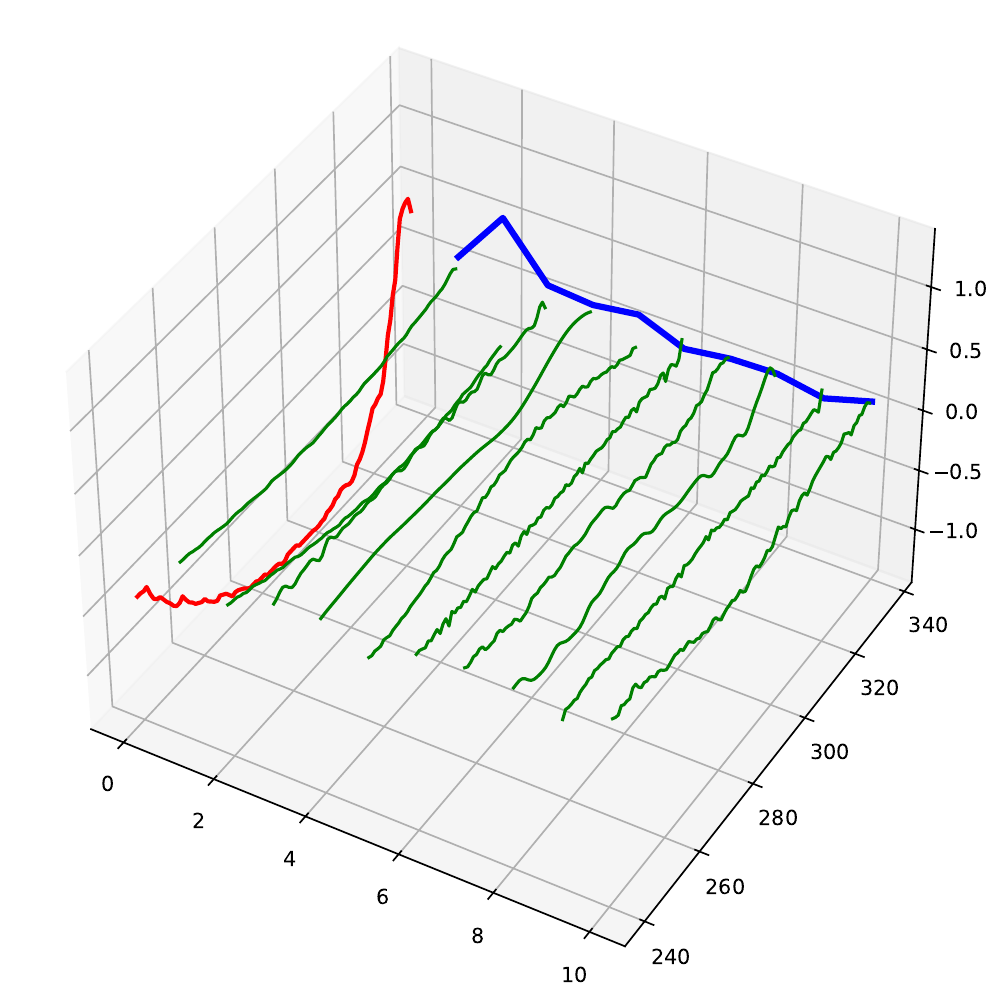}
\end{minipage}
\caption{Visualization of the MLOW Decomposition for Ten Examples on PEMS07.}
\label{plot8}
\end{figure*}

Figures \ref{components_ECL} and \ref{components_traffic} provide the visualization for the learned weights of 10 components for the ECL and Traffic data. These datasets include more seasonal effects than trending effects.  Thus, this corresponds to more patternable weights across multiple components because specific seasonal effects are more likely to be seen at multiple levels. We also find exponentially deceasing weights from low to high here. On the other hand, Figure \ref{components_Weather} and \ref{components_ETTh1} provide visualization of the learned weights of 10 components for the Weather and ETTh1 datasets. We observe a number of interesting patterns. In particular, the weather dataset exhibits many trending effects,  resulting in the learned components to show the most pronounced exponentially decaying behaviors. We further observe that multiple components exhibit similar, patternable weight structures around the peaks of the magnitude distribution in the ETTh1 dataset. Figures \ref{components_PEMS03}, \ref{components_PEMS04}, \ref{components_PEMS07}, and \ref{components_PEMS08} provide the visualization for the learned weights of 10 components for datasets PEMS03, PEMS04, PEMS07, PEMS08. These data sets include more trending effects than seasonal effects, but we also observe small peak at multiple frequency levels. Our learned weights exhibit both exponential decay patterns, such as components 9, 6, 7, and 7 for PEMS03, PEMS04, PEMS07, and PEMS08, respectively, and peak-concentrated patterns, such as components 7, 8, and 8 for PEMS04, PEMS07, and PEMS08, respectively. In conclusion, magnitude levels that tend to co-occur with high energy in observable patterns are more likely to be in the same components. Figures \ref{plot1} \ref{plot2} \ref{plot3}, \ref{plot4}, \ref{plot5}, \ref{plot6}, \ref{plot7}, and \ref{plot8} provide ten examples of MLOW decomposition for ECL, Traffic, Weather, ETTh1, PEMS03, PEMS08, PEMS04, and PEMS07, respectively, with the first five from the training sets and the last five from validation and test sets. These results demonstrate that our MLOW method can produce more interpretable decompositions by disentangling different effects and remaining robust to noises.

\begin{table}[ht]
\centering
\caption{MLOW vs. MA as Plug-and-Play Decomposition Modules on iTransformer and PatchTST under T = 96 and the Same Hyperparameter. }
\begin{adjustbox}{width=\columnwidth,center}
\begin{tabular}{cccccccc|cccccccccccccc} 
\midrule
\multicolumn{2}{c}{Methods} &\multicolumn{2}{c}{iTransformer+Ours} & \multicolumn{2}{c}{iTransformer+MA}                 & \multicolumn{2}{c}{iTransformer}                   & \multicolumn{2}{c}{PatchTST+Ours} & \multicolumn{2}{c}{PatchTST+MA}& \multicolumn{2}{c}{PatchTST} \\

\midrule
\multicolumn{2}{c}{Error} & MSE & MAE  & MSE & MAE  & MSE                    & MAE                    & MSE   & MAE                 & MSE   & MAE                   & MSE   & MAE  \\

\midrule
                     
\multirow{5}{*}{PEMS03}  & 12& \textbf{0.064} & \textbf{0.164} & 0.064  &  0.167 &  0.067 & 0.169  & \textbf{0.079} & \textbf{0.194} & 0.094 & 0.208  & 0.105 & 0.227  \\ 

                         & 24& \textbf{0.078}  & \textbf{0.179}   & 0.091  & 0.199  & 0.090  & 0.199 & \textbf{0.096} & \textbf{0.209} & 0.146  & 0.258 & 0.157& 0.271 \\ 
                        
                         & 48& \textbf{0.093} & \textbf{0.196} &  0.145  & 0.250  & 0.142 &  0.250 & \textbf{0.117}  & \textbf{0.230}& 0.220 & 0.316 & 0.234 & 0.325 & \\ 
                        
                         & 96& \textbf{0.109} &  \textbf{0.208}&  0.220 & 0.315  & 0.218  &  0.315 & \textbf{0.142}&  \textbf{0.255}  & 0.353 & 0.406 & 0.362 & 0.405 & \\ 

                         & avg& \textbf{0.086}	 & \textbf{0.186} & 	0.130 & 	0.232	 & 0.129	 &0.233	 & \textbf{0.108}	 & \textbf{0.222}	 &0.203	 &0.297	 &0.214	 &0.307
\\

\midrule
\multirow{5}{*}{PEMS04} & 12& \textbf{0.067}  & \textbf{0.163} & 0.083  & 0.183 & 0.085 & 0.187 & \textbf{0.089} & \textbf{0.213} & 0.112  & 0.234  & 0.112 & 0.227 &    \\ 

                       & 24 & \textbf{0.075} & \textbf{0.172} & 0.113 & 0.214 & 0.113  & 0.214 & \textbf{0.104}  & \textbf{0.230}  & 0.174 & 0.288  & 0.165 & 0.278 \\

                         & 48 & \textbf{0.084}  & \textbf{0.182}  & 0.171 & 0.268 & 0.170  & 0.267 & \textbf{0.132}  &  \textbf{0.254} & 0.303  & 0.376 & 0.283 & 0.359   &\\ 
                         
                         & 96& \textbf{0.096} & \textbf{0.192} & 0.262 & 0.336 & 0.262 & 0.338 & \textbf{0.156} &  \textbf{0.278} & 0.474  &  0.476 & 0.477 &  0.469 & \\

                         & avg& \textbf{0.080} &	\textbf{0.177} &	0.157	 &0.250 &	0.157 &	0.251	 &\textbf{0.120} &	\textbf{0.243}	 &0.265 &	0.343	 &0.259	 &0.333

\\

\midrule
\multirow{5}{*}{PEMS08}

                         & 12& \textbf{0.065} & \textbf{0.158} & 0.071 & 0.168 &  0.072  & 0.170 & \textbf{0.088} & \textbf{0.207} & 0.106 & 0.227 & 0.099 & 0.220  \\ 

                         & 24& \textbf{0.074} & \textbf{0.167} & 0.102 & 0.204 & 0.099 & 0.200  & \textbf{0.106} & \textbf{0.221} & 0.169 & 0.285 & 0.174  & 0.287 \\ 

                          & 48& \textbf{0.087} & \textbf{0.180} & 0.157 & 0.250 &  0.155 & 0.248   & \textbf{0.129} & \textbf{0.241} &  0.238 & 0.328 & 0.229 & 0.322  \\ 
                         
                          & 96& \textbf{0.099} & \textbf{0.187} & 0.241 & 0.312 & 0.243  & 0.311 & \textbf{0.160} & \textbf{0.260} & 0.374 & 0.413 & 0.382 &  0.406 & \\ 

                         & avg& \textbf{0.081}&	\textbf{0.173}&	0.142&	0.233&	0.142&	0.232&	\textbf{0.120}&	\textbf{0.232}&	0.221&	0.313&	0.221&	0.308

\\

\midrule
                   \multirow{5}{*}{PEMS07}  & 12& \textbf{0.050} & \textbf{0.135} & 0.064  & 0.159 & 0.065  & 0.160  &  \textbf{0.070}  & \textbf{0.190}  & 0.093 &0.217  & 0.089 & 0.206 \\ 
                   
                         & 24& \textbf{0.056} & \textbf{0.143}  & 0.092 & 0.189 & 0.092& 0.191 & \textbf{0.080} & \textbf{0.202} & 0.141& 0.259 & 0.136 & 0.254 \\ 
                        
                         & 48& \textbf{0.064} & \textbf{0.151}  & 0.140 & 0.237 & 0.139 &  0.236 & \textbf{0.096} & \textbf{0.217} & 0.253& 0.338 & 0.242 & 0.329 \\ 
                        
                         & 96& \textbf{0.071}  & \textbf{0.158} & 0.203  & 0.289 & 0.202 & 0.288 & \textbf{0.117} & \textbf{0.235} & 0.407& 0.429 & 0.393 & 0.420 \\ 

                         & avg& \textbf{0.060}&	\textbf{0.146}&	0.124&	0.218	&0.124&	0.218&	\textbf{0.090}&	\textbf{0.211}&	0.223	&0.310&	0.215&	0.302

\\

\midrule
                   \multirow{5}{*}{ECL}  & 96& \textbf{0.130} & \textbf{0.223} & 0.151  & 0.245 & 0.151  & 0.245  & \textbf{0.144} & \textbf{0.255} &  0.171 & 0.273 & 0.176 & 0.271 & \\
                   
                         & 192& \textbf{0.149} & \textbf{0.240}  &  0.167 & 0.259  & 0.167  &  0.259 & \textbf{0.161} & \textbf{0.271} & 0.182  & 0.281  & 0.188 & 0.282  \\ 
                        
                         & 336& \textbf{0.162} & \textbf{0.256}  & 0.182 & 0.276  & 0.185 & 0.276  & \textbf{0.176} & \textbf{0.284} & 0.200&0.298 &  0.203 & 0.297 &  &\\ 
                        
                         & 720& \textbf{0.181}  & \textbf{0.276} & 0.220 & 0.307 & 0.217 & 0.304 & \textbf{0.212} & \textbf{0.316} & 0.244 & 0.333& 0.254 & 0.333 &  &      \\ 

                         & avg& \textbf{0.155}&	\textbf{0.248}&	0.180&	0.271	&0.180&	0.271&	\textbf{0.173}&	\textbf{0.281}&	0.199	&0.296&	0.205&	0.295

\\

     \midrule 
                   \multirow{5}{*}{Traffic}  & 96& \textbf{0.353} & \textbf{0.229} &  0.395 & 0.266 & 0.395  & 0.268  & \textbf{0.372} & \textbf{0.278}  & 0.445 & 0.297 & 0.440 & 0.295 \\ 
                   
                         & 192& \textbf{0.380} & \textbf{0.241}  &  0.412 & 0.276 & 0.417  &  0.276 & \textbf{0.393}& \textbf{0.287} & 0.457 & 0.303 & 0.454 & 0.303 \\
                        
                         & 336& \textbf{0.402} & \textbf{0.251}  & 0.417 & 0.284 & 0.433 & 0.283  & \textbf{0.409} & \textbf{0.296} & 0.473 & 0.310  & 0.469 & 0.310 \\ 
                        
                         & 720& \textbf{0.438}  & \textbf{0.266} &   0.444 & 0.295 & 0.467 & 0.302 &  \textbf{0.446} & \textbf{0.315} & 0.508 & 0.327 & 0.508 & 0.328 \\ 
                         & avg& \textbf{0.393}	& \textbf{0.250}&	0.417&	0.280&	0.428&	0.282&	\textbf{0.405}&\textbf{	0.294}&	0.470&	0.309&	0.467&	0.309

\\

     \midrule
                     
                        \multirow{5}{*}{Weather}  & 96&  \textbf{0.155} &  \textbf{0.201} &    0.173 &  0.215 & 0.178  & 0.218 & \textbf{0.154} & \textbf{0.199}  &  0.174& 0.217  & 0.173  & 0.216 \\ 
                   
                         & 192& \textbf{0.196} & \textbf{0.242}  & 0.226 & 0.259 &  0.223  & 0.256  & \textbf{0.198} & \textbf{0.240} & 0.220  & 0.257 & 0.218  & 0.255  \\ 
                        
                         & 336&  \textbf{0.246} &  \textbf{0.280}  &  0.281 & 0.298  & 0.281  & 0.298 & \textbf{0.249} & \textbf{0.280} & 0.277 & 0.296  & 0.275 & 0.295 &  \\ 
                        
                         & 720&  \textbf{0.330} & \textbf{0.334} &   0.359& 0.350 & 0.363 & 0.359  & \textbf{0.330} & \textbf{0.334} & 0.350 & 0.343  &  0.348& 0.342    \\ 
                         & avg&\textbf{0.231}&	\textbf{0.264}&	0.259&	0.280	&0.261&	0.282&	\textbf{0.232}	&\textbf{0.263}&	0.255&	0.278&	0.253&	0.277\\

     \midrule 

                        \multirow{5}{*}{ETTh1} & 96& \textbf{0.375} & \textbf{0.404} & 0.392 &  0.409 & 0.390 & 0.407  & \textbf{0.369} & \textbf{0.398}  & 0.384 & 0.402 & 0.389 & 0.409   \\
                   
                         & 192& \textbf{0.416}  & \textbf{0.431}  &  0.443 &  0.439 & 0.443 &  0.439 & \textbf{0.413} & \textbf{0.425} & 0.427 & 0.432 & 0.430 & 0.436\\ 
                        
                         & 336& \textbf{0.439}  & \textbf{0.443}  & 0.491 & 0.464 & 0.488 & 0.462 & \textbf{0.423} & \textbf{0.432} & 0.460 & 0.451 & 0.471 & 0.454 \\ 
                        
                         & 720& \textbf{0.466} & \textbf{0.475} & 0.524 & 0.500  & 0.504 & 0.492 & \textbf{0.438} & \textbf{0.462}  & 0.474& 0.475 & 0.500 & 0.486 \\ 

                         & avg& \textbf{0.424}&	\textbf{0.438}&	0.462&	0.453&	0.456 &	0.450&	\textbf{0.410}&	\textbf{0.429}&	0.436&	0.440&	0.447&	0.446&
\\

                         
\hline
\end{tabular}
\end{adjustbox}
\label{table_full}
\end{table}

\begin{table}[ht]
\caption{MLOW + iTransformer / PatchTST vs. Other Deep TSF Methods under T = 96. }
\centering
\begin{adjustbox}{width=\columnwidth}
\begin{tabular}{cccccccccccccccccc} 
\midrule
 \multicolumn{2}{c}{\multirow{2}{*}{Methods}}  & \multicolumn{2}{c}{iTransformer+MLOW}   & \multicolumn{2}{c}{PatchTST+MLOW} & \multicolumn{2}{c}{DUET} & \multicolumn{2}{c}{CycleNet} & \multicolumn{2}{c}{SparseTSF}  & \multicolumn{2}{c}{TimeKAN}    & \multicolumn{2}{c}{TimesNet} & \multicolumn{2}{c}{TimeMixer}\\ 

 &  &  \multicolumn{2}{c}{\cite{LiuHZWWML24}}  & \multicolumn{2}{c}{\cite{nie2023time}}& \multicolumn{2}{c}{\cite{qiu2025duet}} & \multicolumn{2}{c}{\cite{lin2024cyclenet}} & \multicolumn{2}{c}{ \cite{lin2024sparsetsf}}  & \multicolumn{2}{c}{ \cite{huang2025timekan}}    & \multicolumn{2}{c}{ \cite{wu2022timesnet}} & \multicolumn{2}{c}{ \cite{TimeMixer2024}}
 
 \\

\midrule
\multicolumn{2}{c}{Error}   & MSE                    & MAE                    & MSE                    & MAE                    & MSE   & MAE        & MSE   & MAE    & MSE   & MAE   & MSE   & MAE & MSE   & MAE   & MSE   & MAE                          \\ 
\midrule
\multirow{5}{*}{PEMS03}   & 12  & \underline{0.064} & \underline{0.164} & 0.079 & 0.194 & 0.065  & 0.169 &  0.066 & 0.170  &  0.116& 0.230 & 0.099 & 0.211 & 0.083  & 0.191  & \textbf{0.063}  & \textbf{0.164}   \\ 

                         & 24 & \textbf{0.078} & \textbf{0.179} & 0.096  &0.209 & 0.089 & 0.197  & 0.092  & 0.201 &  0.173& 0.282& 0.163 &  0.268   & 0.125 & 0.234  & \underline{0.080}  & \underline{0.187}              \\ 

                         & 48  & \textbf{0.093}  & \textbf{0.196} & 0.117 & 0.230  & 0.130 & 0.242  & 0.141 & 0.245  & 0.328 & 0.388 & 0.319 & 0.388   & 0.160 & 0.270     & \underline{0.114}  & \underline{0.226}                         \\ 

                         & 96 &  \textbf{0.109}& \textbf{0.208} & \underline{0.142}  & \underline{0.255}  & 0.172 & 0.285  & 0.189 & 0.279  & 0.571 & 0.508 & 0.523 &  0.519   & 0.231  & 0.333  & 0.174 & 0.273                      \\
                          & avg & \textbf{0.086}& 	\textbf{0.186}& 	0.108	& 0.222	& 0.114	& 0.223& 	0.122	& 0.223& 	0.297	& 0.352	& 0.276	& 0.3465& 	0.149& 	0.257& 	\underline{0.107}& 	\underline{0.212}\\

\midrule
\multirow{5}{*}{PEMS08}   & 12  & \textbf{0.065} & \textbf{0.158}  & 0.088 & 0.207  & 0.068& 0.167 & 0.080 & 0.181  & 0.121 & 0.231   & 0.099 & 0.211    &0.108 & 0.209  & \underline{0.066} & \underline{0.166} \\ 

                         & 24 & \textbf{0.074} & \textbf{0.167}   & 0.106 & 0.221   & 0.085 & 0.190 & 0.112 & 0.213 & 0.180 & 0.285  & 0.161& 0.271&  0.133 & 0.219 & \underline{0.081} & \underline{0.185}         \\ 

                         & 48  & \textbf{0.087} & \textbf{0.180}   & 0.129& 0.241 & \underline{0.109} &\underline{0.221} & 0.171 & 0.260 & 0.322 & 0.383  & 0.302 & 0.381 &  0.171& 0.291 & 0.127 & 0.236                       \\ 

                         & 96 & \textbf{0.099} & \textbf{0.187} & 0.160 & \underline{0.260}  & \underline{0.156} & 0.268 &  0.240& 0.297   & 0.572 & 0.505  & 0.538& 0.523 &0.344 & 0.401  & 0.209 & 0.314                 \\ 
                       & avg & \textbf{0.081} &	\textbf{0.173} &	0.120 &	0.232 &	\underline{0.104} &	\underline{0.211}&	0.150&	0.237	 &0.298&	0.351	&0.275&	0.346&	0.189&	0.280&	0.120 &	0.225
\\
\midrule
\multirow{5}{*}{ECL}   & 96  &  \textbf{0.130} & \textbf{0.223}  & 0.145 & 0.233  & 0.146& 0.241& \underline{0.136} & \underline{0.230} & 0.197 & 0.270  & 0.179& 0.269   & 0.168  & 0.272 & 0.153 & 0.247   \\ 

                         & 192 &   \textbf{0.149} & \textbf{0.240}   &  0.163 & 0.248  & 0.163 & 0.255 & \underline{0.153} & \underline{0.245} & 0.198 & 0.273  & 0.187 & 0.280 & 0.184 &0.289  & 0.166 & 0.256  \\ 

                         & 336  &   \textbf{0.162} & \textbf{0.256}  & 0.176 &0.284& 0.175 & \underline{0.262}& \underline{0.171} & 0.265  & 0.211 & 0.291  & 0.202& 0.293        & 0.198 & 0.300  & 0.185 & 0.277                \\ 

                         & 720 &  \textbf{0.181} & \textbf{0.276}  & 0.212 & 0.316  & \underline{0.204}& \underline{0.291} & 0.211& 0.299& 0.251 & 0.321   & 0.247 & 0.328 & 0.220 & 0.320 & 0.225 & 0.310                      \\ 
                       & avg & \textbf{0.155} &	\textbf{0.248}	&0.173&	0.281	&0.172	&\underline{0.258}	& \underline{0.167}&	0.259	&0.214&	0.288&	0.203&	0.292&	0.192&	0.295&	0.182	& 0.272
\\

\midrule

    \multirow{5}{*}{Traffic}   & 96  &  \textbf{0.353} &  \textbf{0.229} & \underline{0.372}  & 0.278 & 0.407  & \underline{0.252}  & 0.457  &  0.295 &  0.575 & 0.331 &  0.560 & 0.365   & 0.593  & 0.321  & 0.462 & 0.285    \\ 

                         & 192 & \textbf{0.380} & \textbf{0.241} & \underline{0.393}  & 0.287 & 0.431  & \underline{0.262} & 0.459  & 0.297 & 0.577  & 0.332  &   0.564& 0.367 &  0.617 & 0.336  &  0.473 &  0.296              \\ 

                         & 336  & \textbf{0.402}  & \textbf{0.251}  & \underline{0.409} & 0.296  & 0.456& \underline{0.269} & 0.470 & 0.299 & 0.576 & 0.337   & 0.580 & 0.372  & 0.629  &  0.336     &  0.498 & 0.296                          \\ 

                         & 720 & \textbf{0.438} & \textbf{0.266} & \underline{0.446} & 0.315  & 0.509  & \underline{0.292}  & 0.501 & 0.314  & 0.630 & 0.356 &  0.605  &  0.384 &  0.640 & 0.350  &  0.506 & 0.313                       \\ 
                       & avg & \textbf{0.393} &	\textbf{0.250}	&\underline{0.405}&	0.294	&0.451	&\underline{0.269}&	0.471&	0.301	&0.589	&0.339&	0.577&	0.372&	0.619&	0.335&	0.484	&0.297

\\

\midrule

\end{tabular}
\end{adjustbox}
\label{others_full}
\end{table}

\begin{table}[ht]
\caption{Performance of MLOW with Different Low-Rank Methods on iTransformer and Average Inference Time for an Input Time Series on ECL.}
\centering
\begin{adjustbox}{width=\columnwidth}
\begin{tabular}{cccccccccccc} 
\midrule
 \multicolumn{2}{c}{Methods}  & \multicolumn{2}{c}{Hyperplane-NMF}  & \multicolumn{2}{c}{NMF}  & \multicolumn{2}{c}{PCA}& \multicolumn{2}{c}{Semi-NMF} \\ 

\midrule
\multicolumn{2}{c}{Error} & MSE  & MAE                                & MSE & MAE                & MSE   & MAE      & MSE  & MAE                           \\ 
\midrule
\multirow{5}{*}{PEMS08}   & 12  & \textbf{0.065} & \textbf{0.158} &  0.067 & 0.159 & 0.069 & 0.161  & 0.069 & 0.161    \\ 

                         & 24 & \textbf{0.074} & \textbf{0.167} & 0.075 & 0.168 &0.078 & 0.171  &0.080 & 0.175              \\ 

                         & 48  & \textbf{0.087} & \textbf{0.180} &0.089 & 0.182 & 0.092 & 0.185   & 0.103 & 0.199                       \\ 

                         & 96 & \textbf{0.099} & \textbf{0.187} & 0.101 & 0.191 & 0.106 & 0.194   & 0.106 & 0.203                 \\      
                          & avg  & \textbf{0.081}	 &\textbf{0.173}	 &0.083 &	0.175 &	0.086 &	0.177 &	0.089 &	0.184\\

\midrule
\multirow{5}{*}{ECL}   & 96  & \textbf{0.130} & \textbf{0.223} & 0.132 & 0.228 & 0.132 & 0.227  & 0.137 & 0.232     \\ 

                         & 192 & \textbf{0.149} & \textbf{0.240} & 0.154 & 0.247 & 0.152 & 0.244     & 0.156 & 0.248        \\ 

                         & 336  & \textbf{0.162} & \textbf{0.256} & 0.169 & 0.266 & 0.167 & 0.262  & 0.173 & 0.267                     \\ 

                         & 720 &  \textbf{0.181} & \textbf{0.276} & 0.197 & 0.294 & 0.191 & 0.289  & 0.199 & 0.295                    \\ 
                         & avg  & \textbf{0.155}&	\textbf{0.248}&	0.163&	0.258	&0.160&	0.255	&0.166&	0.260\\

\midrule
\multicolumn{2}{c}{Inference Time (s)}    & \multicolumn{2}{c}{\textbf{0.000259}}    & \multicolumn{2}{c}{0.121408}  & \multicolumn{2}{c}{0.000265} &  \multicolumn{2}{c}{0.000288} & \\ 
\midrule
\end{tabular}
\end{adjustbox}
\label{ab2}
\end{table}

\begin{table}[h]
\caption{Sensitivity Analysis for $K$ and $V$ on iTransformer.}
\centering
\begin{adjustbox}{width=\columnwidth}
\begin{tabular}{cccccccccccc} 
\midrule
\multicolumn{2}{c}{Error}   & MSE              & MAE                    & MSE                    & MAE                    & MSE   & MAE        & MSE   & MAE                                \\ 

\midrule
 \multicolumn{2}{c}{(K=168)}  & \multicolumn{2}{c}{V=5}  & \multicolumn{2}{c}{V=10}  & \multicolumn{2}{c}{V=15} & \multicolumn{2}{c}{V=20}  \\ 

\midrule
\multirow{5}{*}{PEMS08}   & 12 &  \textbf{0.065} & \textbf{0.157} & 0.065 & 0.158 & 0.065 & 0.158 &  0.065 &0.157     \\ 
                         & 24 & 0.075  & 0.167 & \textbf{0.074} & \textbf{0.167} & 0.075 & 0.169 & 0.076    &0.170               \\ 

                         & 48  & 0.088 & 0.182  & 0.087 & 0.180 & 0.088 & 0.180 & \textbf{0.087} & \textbf{0.179}                           \\ 

                             & 96 &  0.100 & 0.189 & 0.099 & \textbf{0.187} & \textbf{0.098} & 0.187 & 0.100 & 0.190                     \\ 
                             & avg & 0.082	&0.173 &	\textbf{0.081} &	\textbf{0.173} &	0.081 &	0.173	&0.082 &0.174\\

\midrule
\multirow{5}{*}{ECL} & 96  & \textbf{0.128}   & \textbf{0.221}  & 0.130 & 0.223 & 0.129 & 0.221 & 0.130 & 0.223      \\ 

                         & 192 & \textbf{0.146}  & \textbf{0.237} &  0.149 & 0.240 & 0.151 & 0.242 &0.153 & 0.244                 \\ 

                         & 336  & 0.161 & 0.253 & 0.162 & 0.256 & \textbf{0.159}  & \textbf{0.252}      & 0.162 & 0.255         \\ 

                         & 720 & \textbf{0.181}  & \textbf{0.275} &  0.181  & 0.276 & 0.183 & 0.277  &0.182  & 0.275                     \\ 
                         & avg & \textbf{0.154}	&  \textbf{0.246} &	 0.155& 0.248	&0.155	& 0.248& 0.156 & 0.251\\

\midrule
 \multicolumn{2}{c}{(K=48)} & \multicolumn{2}{c}{V=5}  & \multicolumn{2}{c}{V=10}  & \multicolumn{2}{c}{V=15} & \multicolumn{2}{c}{V=20}  \\ 
\midrule
\multirow{5}{*}{PEMS08}   & 12 & \textbf{0.065} & \textbf{0.161} & 0.066 & 0.161 & 0.066 & 0.161 & 0.065 & 0.161   \\ 
                         & 24 & 0.076 & 0.174  & 0.077 & 0.175 &0.077 &0.173  & \textbf{0.076} & \textbf{0.172}          \\ 

                         & 48  & \textbf{0.089} & \textbf{0.187} & 0.089 & 0.188 & 0.089 & 0.187 & 0.092 & 0.191                          \\ 
                         & 96 & \textbf{0.114} & \textbf{0.209} & 0.116 & 0.210 & 0.117 & 0.210 & 0.121 & 0.213                   \\ 

                          & avg & \textbf{0.086}  &	\textbf{0.182}	 &0.087	 &0.183 &	0.087	 &0.182 &	0.088	& 0.184\\

\midrule
\multirow{5}{*}{ECL} & 96 &  \textbf{0.135} & \textbf{0.226} & 0.135  & 0.227 & 0.135 & 0.226 & 0.135 & 0.227       \\ 

                         & 192 & 0.154 & 0.244 & \textbf{0.153}  & \textbf{0.242} & 0.154  & 0.244  &  0.155 & 0.245 &           \\ 

                         & 336  & 0.167 & 0.259 & \textbf{0.166} & \textbf{0.258} & 0.168 & 0.261 & 0.166  &  0.258      \\ 

                         & 720 & 0.189 & 0.282 & \textbf{0.187}  & \textbf{0.279} & 0.190 & 0.282 &  0.193 &  0.284                 \\ 
                         & avg & 0.161 & 0.252&	\textbf{0.160} & \textbf{0.251} & 0.161  & 0.253& 0.162 & 0.253	\\

\midrule
\end{tabular}
\end{adjustbox}
\label{ab4}
\end{table}

\begin{table}[ht]
\caption{Data Granularity}
\centering
\begin{adjustbox}{width=\columnwidth}
\begin{tabular}{c|l|c|c|c|c|c}
\toprule
Tasks & Dataset & Dim & Series Length & Dataset Size & Frequency  &\scalebox{0.8}{Information} \\
\toprule
 & ETTh1 & 7 & \scalebox{0.8}{\{96, 192, 336, 720\}} & (8545, 2881, 2881) & Hourly  &\scalebox{0.8}{Temperature} \\
\cmidrule{2-7}
Long-term & Electricity & 321 & \scalebox{0.8}{\{96, 192, 336, 720\}} & (18317, 2633, 5261) & Hourly & \scalebox{0.8}{Electricity} \\
\cmidrule{2-7}
Forecasting & Traffic & 862 & \scalebox{0.8}{\{96, 192, 336, 720\}} & (12185, 1757, 3509) & Hourly& \scalebox{0.8}{Transportation} \\
\cmidrule{2-7}
 & Weather & 21 & \scalebox{0.8}{\{96, 192, 336, 720\}} & (36792, 5271, 10540)  &10 mins   &\scalebox{0.8}{Weather} \\
\midrule
& PEMS03 & 358 & \scalebox{0.8}{\{12, 24, 48, 96\}} & (18185,2568,5135) & 5 mins  & \scalebox{0.8}{Transportation}\\
\cmidrule{2-7}
Short-term& PEMS04 & 307 &  \scalebox{0.8}{\{12, 24, 48, 96\}} & (11859,1688,3375) & 5 mins  & \scalebox{0.8}{Transportation}\\
\cmidrule{2-7}
Forecasting& PEMS07 & 883 &  \scalebox{0.8}{\{12, 24, 48, 96\}} & (19722,2811,5622) & 5 mins  & \scalebox{0.8}{Transportation}\\
\cmidrule{2-7} & PEMS08 & 170 &  \scalebox{0.8}{\{12, 24, 48, 96\}} & (12434,1774,3548) & 5 mins  & \scalebox{0.8}{Transportation}\\
\bottomrule
\end{tabular}
\end{adjustbox}
\label{data_detail}
\end{table}

\begin{table}[ht]
\caption{iTransformer+MLOW Training Details}
\centering
\begin{adjustbox}{width=\columnwidth}
\begin{tabular}{ccccc|ccccccc}
\toprule
 & \multicolumn{4}{c}{iTransformer Hyperparameter} & \multicolumn{5}{c}{Training Process} \\
\midrule   
Hyperparameters& Dropout & $e_{Layers}$ & $d_{model}$ & $d_{ff}$  & LR & Loss & lradj & Batch Size & Epochs & Patience\\
\midrule   
ETTh1 & 0.15 & 3 & 256 & 256 & $10^{-4}$ & MAE & TST& 128 & 100  & 10\\
\midrule
Weather &  0.15 & 3 & 256 & 256 & $10^{-4}$ & MAE & TST & 128 & 100 & 10\\
\midrule
Electricity &  0.15 & 3 & 512 & 512 & $10^{-4}$ & MAE & TST & 16 & 100 & 10\\
\midrule
Traffic &  0.15 & 5 &512& 512  & $10^{-4}$ & MAE & TST & 16 & 100 & 10\\
\midrule
PEMS03 &  0.15 & 3 & 512& 512 & $4*10^{-4}$ & MAE & TST& 64 & 150 & 20 \\
\midrule
PEMS04 &  0.15 & 3 & 512 & 512& $4*10^{-4}$ & MAE& TST & 64 & 150 & 20\\
\midrule
PEMS07 &  0.15 & 3 & 512 & 512& $4*10^{-4}$ & MAE& TST & 64 & 150 & 20 \\
\midrule
PEMS08 &  0.15 & 3 & 512& 512 & $4*10^{-4}$ & MAE& TST & 64 & 150 & 20 \\
\bottomrule
\label{detail1}
\end{tabular}
\end{adjustbox}
\end{table}

\begin{table}[ht]
\caption{PatchTST+MLOW Training Details}
\centering
\begin{adjustbox}{width=\columnwidth}
\begin{tabular}{ccccccc|cccccc}
\toprule
 & \multicolumn{6}{c}{PatchTST Hyperparameter} & \multicolumn{6}{c}{Training Process} \\
\midrule
Hyperparameters & Dropout & $e_{Layers}$ & $d_{model}$  & $d_{ff}$ & Patch Length & Stride & LR & lradj & Loss &Batch Size & Epochs & Patience\\
\midrule
ETTh1 & 0.15 & 3 & 64 & 128 & 8 &8 & $10^{-4}$ & MAE & TST & 128 & 100  & 10\\
\midrule
Weather &  0.15 & 3 & 128 & 256 & 8 &8& $10^{-4}$ & MAE & TST & 128 & 100 & 10\\
\midrule
Electricity &  0.15 & 3 & 128 & 256 &  8 &8&$10^{-4}$ & MSE & TST & 16 & 100 & 10\\
\midrule
Traffic &  0.15 & 5 &128& 256  & 8 &8& $10^{-4}$ & MSE & TST& 16 & 100 & 10\\
\midrule
PEMS03 &  0.15 & 3 & 128& 256 &  8 &8&$10^{-4}$ & MAE& TST & 64 & 150 & 20 \\
\midrule
PEMS04 &  0.15 & 3 & 128 & 256& 8 &8& $10^{-4}$ & MAE& TST& 64 & 150 & 20\\
\midrule
PEMS07 &  0.15 & 3 & 128 & 256& 8 &8& $10^{-4}$ & MAE& TST & 64 & 150 & 20 \\
\midrule
PEMS08 &  0.15 & 3 & 128& 256 & 8 &8& $10^{-4}$ & MAE & TST& 64 & 150 & 20 \\
\bottomrule
\label{detail2}
\end{tabular}
\end{adjustbox}
\end{table}

\end{document}